\documentclass[twoside]{report}

\usepackage[a4paper,width=150mm,top=25mm,bottom=25mm,bindingoffset=10.5mm]{geometry}

\usepackage[all]{nowidow}

\usepackage{hyperref}

\usepackage{arabtex}
\usepackage{utf8}
\setcode{utf8}

\usepackage[most]{tcolorbox}
\usepackage{ragged2e}
\usepackage{titlesec}
\titleformat{\chapter}[display]   
{\normalfont\huge\bfseries\RaggedRight}{\chaptertitlename\ \thechapter}{20pt}{\Huge}   
\titlespacing*{\chapter}{0pt}{-50pt}{35pt}

\titleformat{\section}[block]
{\Large\bfseries}%
{\thesection}
{5mm}
{}
\titleformat{\subsection}[block]
{\large\bfseries}%
{\thesubsection}
{5mm}
{}

\titlespacing{\section}{0pt}{3ex}{1.5ex}
\titlespacing{\subsection}{0pt}{3ex}{1ex}

\hypersetup{
    colorlinks=true,
    linkcolor=black,
    filecolor=black,
    urlcolor=black,
    citecolor=black,
}

\usepackage{fancyhdr}
\usepackage{thesis}
\usepackage{times}
\usepackage{url}
\usepackage{latexsym}
\usepackage{natbib}
\usepackage{layout}
\usepackage{float}
\usepackage{booktabs, multirow}
\usepackage[]{graphicx}
\usepackage{subfig}
\usepackage{hyphenat}
\usepackage{subfiles}
\usepackage{multicol}
\usepackage{enumerate}

\newcommand*{\myfont}{\fontfamily{lmss}\selectfont}
\DeclareTextFontCommand{\textmyfont}{\myfont}

\usepackage{setspace}

\begin{document}
\setlength\topmargin{0.2in}

\title{
    \bf \LARGE Language Modelling Approaches to \\ Adaptive Machine Translation \\
    \bigbreak \bigbreak
    \centering\tiny{-----------------------------------------------------------------------} \\
    \large PhD Thesis \\
    \centering\tiny{-----------------------------------------------------------------------}
    \vspace{-8ex}
}

\author{
    \centering
    \name{\bf \large \textsc{Yasmin Moslem}} \\
    \vspace{3ex}
    \addr{\normalsize School of Computing \\
     Faculty of Engineering and Computing \\ 
     Dublin City University} \\
    \vspace{1ex}
    \begin{figure}[H]
    \centering
    \includegraphics[width=0.14\textwidth]{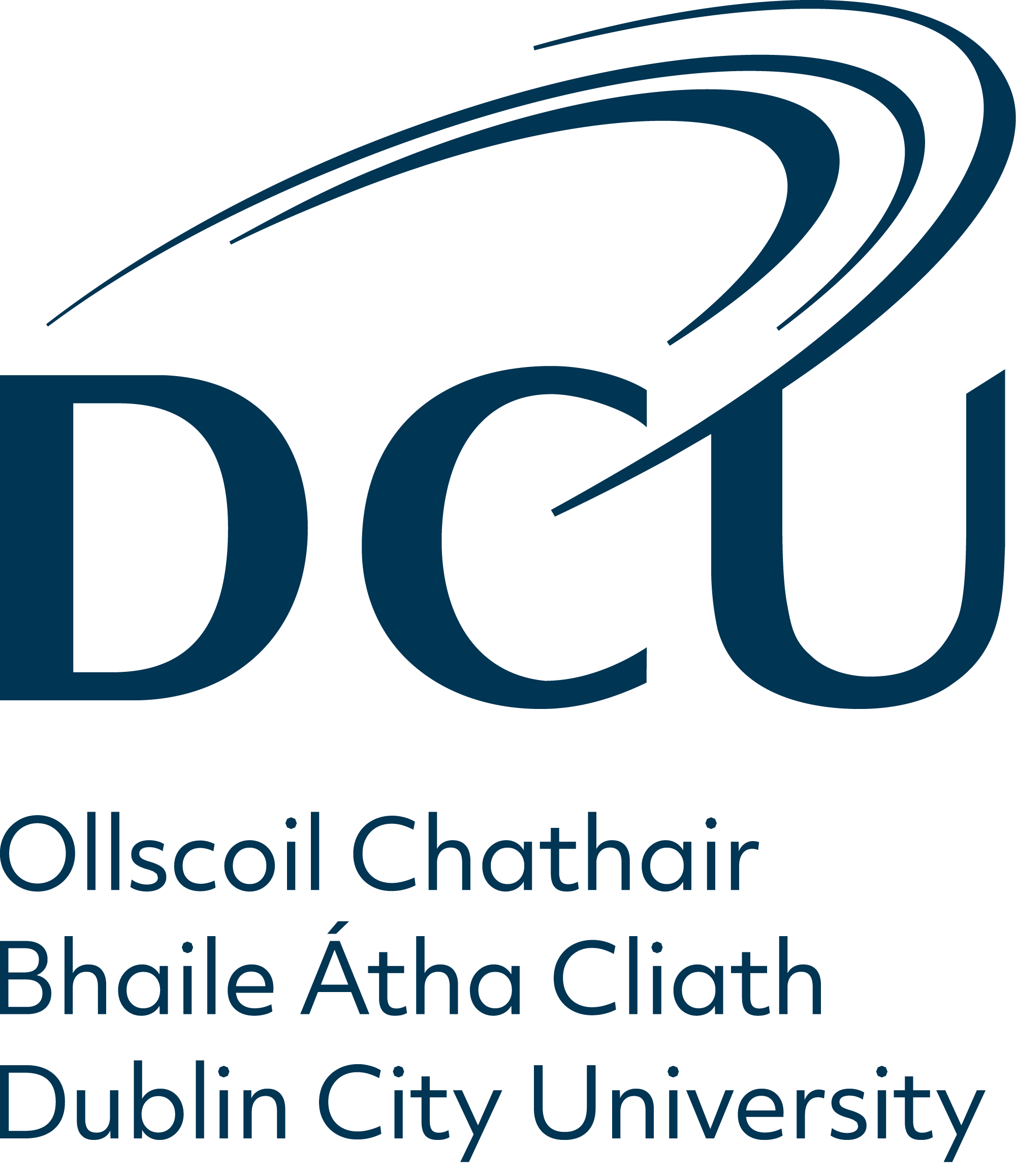}
    \end{figure}
    \bigbreak \bigbreak \bigbreak
    \vspace{7ex}
    \name{\textbf{Supervised by:} \\
    \vspace{1ex}
    Prof Andy Way \\
    {\small School of Computing, Dublin City University} \\
    \vspace{1ex}
    Dr Rejwanul Haque \\
    {\small Department of Computing, South East Technological University} \\
    \vspace{1ex}
    Prof John D. Kelleher \\
    {\small Hamilton Institute, Maynooth University} \\
    \vspace{10ex}
    {\small Ireland, Dublin \\
    January 2024
    }
    } \\
}

\maketitle

\newpage
\pagestyle{empty}

\setlength\topmargin{-0.7in}

\pagenumbering{gobble}


\section*{Declaration}

I hereby certify that this material which I now submit for assessment on the programme of study leading to the award of PhD is entirely my own work, and that I have exercised reasonable care to ensure that the work is original, and does not to the best of my knowledge breach any law of copyright, and has not been taken from the work of others, save and to the extent that such work has been cited and acknowledged within the text of my work.

\vspace{15pt}

\begin{multicols}{2}

Signed: Yasmin Moslem (Candidate)

ID No.: 19215697

Date: 1 January 2024

\begin{figure}[H]
\includegraphics[angle=3,width=0.25\textwidth]{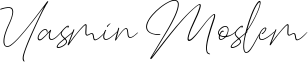}
\end{figure}

\end{multicols}

\newpage
\section*{Dedication}

\vspace{20pt}
\begin{large}
\centering

\< إلى أحبتي: >

{\Large\< ماما * بابا * منار >}

\< بوركتم >

\end{large}

\newpage
\section*{Acknowledgements}

This work would not have been possible without the generous support of the \mbox{Science} Foundation Ireland (SFI) Centre for Research Training in Digitally-Enhanced Reality (d-real) under Grant No. 18/CRT/6224, the ADAPT Centre for Digital Content Technology under SFI's Grant No. \mbox{13/RC/2106\_P2}, and Microsoft Research.

I am truly grateful to my supervisors, Prof Andy Way, Dr Rejwanul Haque, and Prof John Kelleher, for their invaluable guidance throughout my PhD journey. Their insightful feedback and constant encouragement played a significant role in my academic achievement and were a source of support in challenging times. I sincerely appreciate their contributions and wish them good health and further success.

Moreover, I would like to extend my sincere thanks to Julie Locquet, Senior \mbox{Linguist}; Philippe \mbox{Locquet}, Senior Linguist and Academic Program Manager at Wordfast; and Dr~Muhammed Yaman Muhaisen, Ophthalmologist and Linguist, for conducting the linguistic evaluation of the translation tasks.

Finally, special thanks go to colleagues and alumni of Irish universities as well as colleagues in the language technology industry for generously offering advice and inspiration.

\newpage
\section*{Publications}
\begin{enumerate}
    \item \textbf{Yasmin Moslem}, Rejwanul Haque, John D. Kelleher, and Andy Way. 2023. \href{https://aclanthology.org/2023.eamt-1.22/}{\textit{Adaptive Machine Translation with Large Language Models}}. In Proceedings of the 24th Annual Conference of the European Association for Machine Translation, pages 227–237, Tampere, Finland. European Association for Machine Translation.
    
    \item \textbf{Yasmin Moslem}, Gianfranco Romani, Mahdi Molaei, John D. Kelleher, Rejwanul Haque, and Andy Way. 2023. \href{https://aclanthology.org/2023.wmt-1.82/}{\textit{Domain Terminology Integration into Machine Translation: Leveraging Large Language Models}}. In Proceedings of the Eighth Conference on Machine Translation, pages 902–911, Singapore. Association for Computational Linguistics.
    
    \item \textbf{Yasmin Moslem}, Rejwanul Haque, and Andy Way. 2023. \href{https://arxiv.org/abs/2312.12740}{\textit{Fine-tuning Large Language Models for Adaptive Machine Translation}}. arXiv preprint arXiv:2312.12740 [cs.CL].
    
    \item \textbf{Yasmin Moslem}, Rejwanul Haque, John D. Kelleher, and Andy Way. 2022. \href{https://aclanthology.org/2022.amta-research.2/}{\textit{Domain-Specific Text Generation for Machine Translation}}. In Proceedings of the 15th biennial conference of the Association for Machine Translation in the Americas (Volume 1: Research Track), pages 14–30, Orlando, USA. Association for Machine Translation in the Americas.
    
    \item \textbf{Yasmin Moslem}, Rejwanul Haque, and Andy Way. 2022. \href{https://aclanthology.org/2022.wmt-1.119/}{\textit{Translation Word-Level Auto-Completion: What Can We Achieve Out of the Box?}} In Proceedings of the Seventh Conference on Machine Translation (WMT), pages 1176–1181, Abu Dhabi, United Arab Emirates. Association for Computational Linguistics.
    
    \item Alp \"Oktem, Rodolfo Zevallos, \textbf{Yasmin Moslem}, G\"une\c{s} \"Ozt\"urk, and Karen \c{S}arhon. 2022. \href{https://aclanthology.org/2022.eurali-1.18/}{\textit{Preparing an endangered language for the digital age: The Case of Judeo-Spanish}}. In Proceedings of the Workshop on Resources and Technologies for Indigenous, Endangered and Lesser-resourced Languages in Eurasia within the 13th Language Resources and Evaluation Conference, pages 105–110, Marseille, France, June 2022. European Language Resources Association.
    
    \item \textbf{Yasmin Moslem}, Rejwanul Haque, and Andy Way. 2020. \href{https://aclanthology.org/2020.nlptea-1.2/}{\textit{Arabisc: Context-Sensitive Neural Spelling Checker}}. In Proceedings of the 6th Workshop on Natural Language Processing Techniques for Educational Applications, pages 11–19, Suzhou, China. Association for Computational Linguistics.
   
    \item Rejwanul Haque, \textbf{Yasmin Moslem}, and Andy Way. 2020. \href{https://aclanthology.org/2020.icon-adapmt.4/}{\textit{Terminology-Aware Sentence Mining for NMT Domain Adaptation: ADAPT’s Submission to the Adap-MT 2020 English-to-Hindi AI Translation Shared Task}}. In Proceedings of the 17th International Conference on Natural Language Processing (ICON): Adap-MT 2020 Shared Task, pages 17–23, Patna, India. NLP Association of India (NLPAI).

    \item Rejwanul Haque, \textbf{Yasmin Moslem}, and Andy Way. 2020. \href{https://aclanthology.org/2020.ngt-1.17/}{\textit{The ADAPT System Description for the STAPLE 2020 English-to-Portuguese Translation Task}}. In Proceedings of the Fourth Workshop on Neural Generation and Translation, pages 144–152, Online. Association for Computational \mbox{Linguistics}.

\end{enumerate}

\tableofcontents


\onehalfspacing

\newpage

\thispagestyle{plain}

\begin{center}

\begin{Large}
Language Modelling Approaches to Adaptive Machine Translation
\end{Large}

\begin{large}
\textsc{Yasmin Moslem}
\end{large}

\end{center}

\vspace{5ex}

\begin{abstract}

\smallskip
\nohyphens{
\large
Consistency is a key requirement of high-quality translation. It is especially important to adhere to pre-approved terminology and adapt to corrected \mbox{translations} in domain-specific projects.~Machine translation (MT) has achieved significant progress in the area of domain adaptation. However, in-domain data scarcity is common in translation settings, due to the lack of specialised datasets and terminology, or inconsistency and inaccuracy of available in-domain translations. In such scenarios where there is insufficient in-domain data to fine-tune MT models, producing translations that are consistent with the relevant context is challenging. While real-time adaptation can make use of smaller amounts of in-domain data to improve the translation on the fly, it remains challenging due to supported context limitations and efficiency constraints. Large language models (LLMs) have recently shown interesting capabilities of in-context learning, where they learn to replicate certain input-output text generation patterns, without further fine-tuning. Such capabilities have opened new horizons for domain-specific data augmentation and real-time adaptive MT. This work attempts to address two main relevant questions:~1) in scenarios involving human interaction and continuous feedback, can we employ language models to improve the quality of adaptive MT at inference time? and 2) in the absence of sufficient in-domain data, can we use pre-trained large-scale language models to improve the process of MT domain adaptation?
}
\end{abstract}


\newpage
\pagestyle{fancy}

\pagenumbering{arabic}
\setcounter{page}{1}

\begin{large}
\chapter{Introduction}

Neural Machine Translation (NMT) is capable of producing high-quality translations in terms of fluency and adequacy. The emergence of the Transformer architecture \citep{Bahdanau2015-JointlyLearn,Vaswani2017-attention} has revolutionised the field of Natural Language Processing (NLP) in general and NMT in particular, and paved the way for many subsequent breakthroughs in the field. Nevertheless, NMT still faces some challenges when it comes to translation of out-of-domain texts \citep{Koehn2017-qj}. \mbox{Domain} adaptation of MT systems using in-domain parallel texts has been an active area of research to handle this situation. Several research works on domain adaptation assume the availability of in-domain data. However, in-domain data scarcity is common in translation settings, due to the lack of specialised datasets and terminology, or inconsistency and inaccuracy of available in-domain translations \citep{Axelrod2011-Data,Haddow2012-Data}.

Recent advances in language modelling techniques in general and large-scale \mbox{language} models (LLMs) in particular have shown significant potential in improving a wide range of NLP tasks. Inspired by this idea, this research aims to answer two major \textbf{Research Questions (RQ)}:

\begin{itemize}
\setlength{\itemindent}{0.6em}
    \item[\bf RQ1] In scenarios involving human interaction and continuous feedback, can we employ language models to improve the quality of adaptive MT at inference time? In the subsequent sections, I will be referring to this question as ``\textbf{Adaptive and \mbox{Interactive} MT}''.
    \item[\bf RQ2] In the absence of sufficient in-domain data, can we use pre-trained LLMs to improve the process of NMT domain adaptation? In the following sections, I will be referring to this question as ``\textbf{Domain-specific Text Generation for MT}''.
\end{itemize}


\newpage
\begin{figure}[H]
\includegraphics[width=1\textwidth]{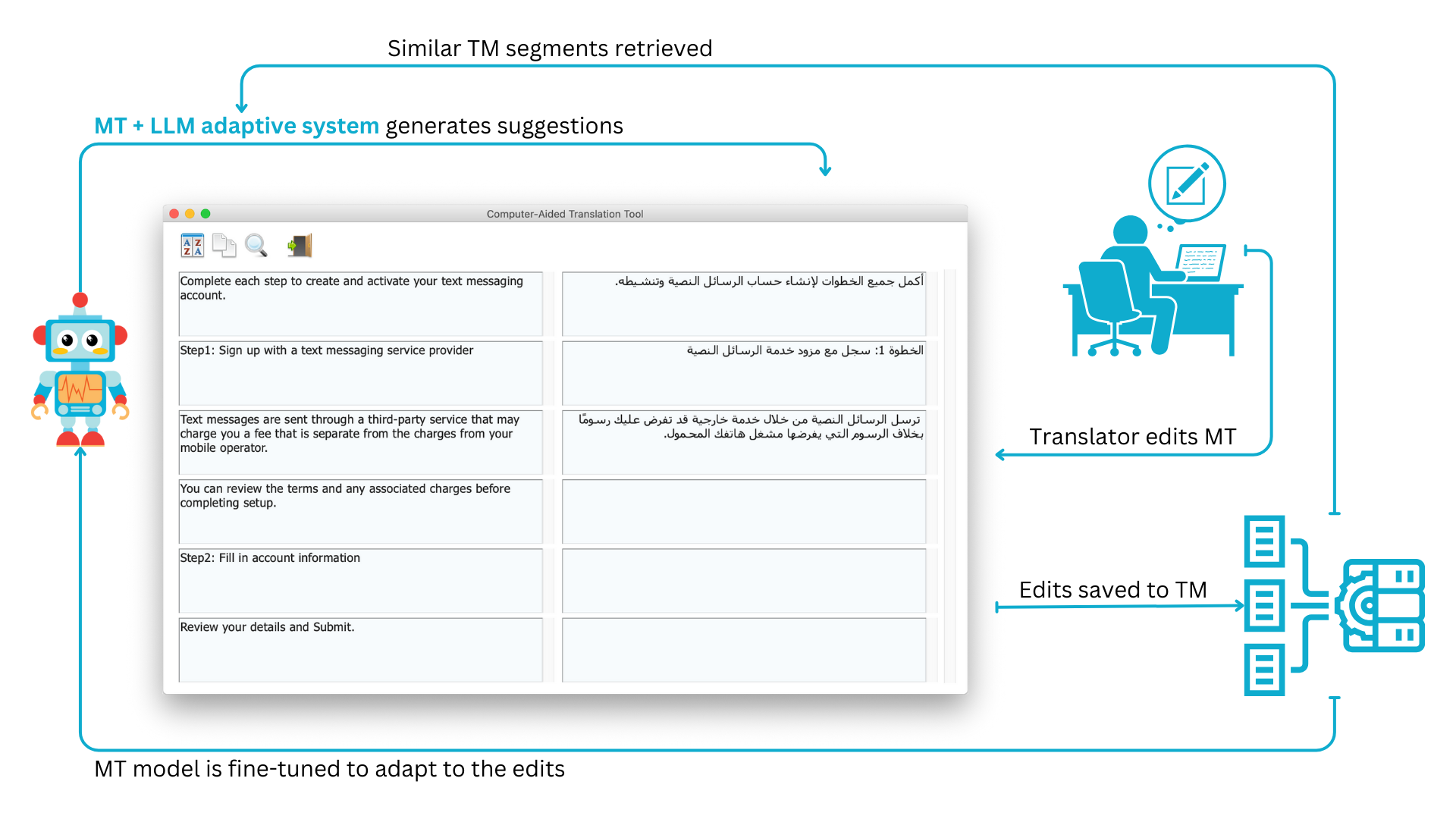}
\caption{Translation environment, involving adaptive MT and human interaction}
\label{fig:cat}
\end{figure}


\section{Background and Motivation}

Initially, as translators transitioned to using computers and text editors, they manually maintained previous translations and terminology in regular electronic documents. The introduction of computer-aided translation tools has significantly transformed traditional approaches, changing the landscape of translation workflows.

Figure \ref{fig:cat} illustrates a typical workflow in translation environments nowadays, where the process starts with an MT suggestion that a translator revises to achieve the required quality. Approved translations are then saved into a translation memory (TM). After some time, bilingual text segments stored in the TM are used to fine-tune the MT model to adapt to the required domain and style. The fine-tuned MT model is then used to generate translation suggestions. Ideally, the model should be able to make use of similar translation pairs retrieved from the TM at inference time. It should also be able to interact with the translator by providing them with suggestions and adapting to their edits during translation. However, such real-time adaptivity and interactivity features are not currently implemented in the majority of MT systems that translators have been using, which highlights the challenges of transitioning research in this area into real-world production workflows.

Throughout my career in the translation technology industry, I have heard from several translators about situations where they edit a term or expression, yet subsequent MT-ed segments continue to repeat the same mistakes. Despite the wide research conducted on real-time adaptive MT and related fields such as context-aware MT and document-level MT, practical application of these techniques in real-world tools remains limited. Some of these approaches were proposed with architectures before the Transformer model\footnote{Given that LSTM models generally have been superseded by Transformer models \citep{Vaswani2017-attention}, performance gains achieved via methods originally proposed with LSTM might not be noticeable. This can also be due to the complexity of the Transformer architecture or using variations of tokenisation and data preparation methods \citep{Popovic2023-Reproducibility}.} or verified only through medium-sized experiments, which sometimes means that such methods cannot achieve the same performance gains when implemented on a large scale. Others suffer from lack of efficiency, which makes deploying them in production challenging for several language service providers \citep{Meng2022-kNN-MT-Fast,Martins2023-kNN-MT,Treviso2023-Efficient-NLP}. In addition, even the details of the limited state-of-the-art commercial applications of adaptive MT are not necessarily shared in publicly accessible publications.

The emergence of the in-context learning capability of LLMs has opened new avenues for diverse NLP tasks, including MT. In-context learning refers to the ability of a model to refine its output based on the context provided within the input itself, without the need for fine-tuning on specific scenarios \citep{Brown2020-GPT-3}. This capability is especially useful in retrieval-augmented generation, a technique that integrates external knowledge sources to enhance the responses of the model. For adaptive MT, this means that LLMs can dynamically adapt translations to various sources of context, such as fuzzy matches or terminology, leading to more accurate translations.

\enlargethispage{0.4\baselineskip}

\section{Research Questions}

This section provides an overview of the aforementioned research questions, while the following chapters elaborate on experimental setups and results. It is worth noting that the two questions overlap in several aspects, which make it possible to use any of the employed approaches to address both of them.

\subsection{\scshape RQ1. Adaptive and Interactive MT}

Adaptive MT utilises user feedback to improve translation quality over time, particularly in domain-specific scenarios where baseline MT systems may lack relevant data. Incorporating user feedback into the translation process, especially at inference time, poses challenges.

In my research, I investigated two forms of utilising language modelling to improve MT real-time adaptivity and interactivity. This can be outlined into the following sub-questions:
\begin{itemize}
\setlength{\itemindent}{1.2em}
    \item [RQ1.a] Can we utilise the autoregressive property of NMT models, i.e. their ability to decode the target sentence word by word according to the translation history, to generate relevant autosuggestions? So instead of providing the most probable word, the system can interact with typed sequences at inference time to generate more accurate translations. In my research \citep{Moslem2022-WLAC}, I employed random sampling techniques to generate diverse alternatives, which led to improving the ability of an MT system in the scenario where the user types a few characters, and the system is expected to predict and auto-complete the correct word, given the current context (cf. Section \ref{sec:q-autosuggest}).
    
    \item [RQ1.b] Can we improve adaptive MT by employing the in-context learning capability of LLMs? This involves learning from similar translations (fuzzy matches) found in approved TMs. In my research \citep{Moslem2023-AdaptiveMT}, I explored leveraging LLMs to boost adaptive MT at inference time. I concluded that LLMs can be leveraged for adapting new translations to match the terminology and style of pre-approved fuzzy matches, post-editing translations from encoder-decoder MT systems, and performing terminology-constrained MT (cf. Section \ref{sec:q-adaptive-mt}).
    
    \item [RQ1.c] Can we reinforce MT adherence to terminology through prompting LLMs to use pre-approved terms in translations? In my work \citep{Moslem2023-AdaptiveMT}, I proposed adding relevant term pairs during translation with an LLM to enhance real-time terminology-constrained MT. Similarly, in my research \citep{Moslem2023-Terminology}, I used LLMs for terminology-constrained automatic post-editing, where an LLM is instructed to incorporate missing terms in translations originally generated by an MT model  (cf. Section \ref{sec:q-term-mt}).
\end{itemize}

\subsubsection{RQ1.a Translation autosuggestions and autocompletion}
\label{sec:q-autosuggest}


Looking at the NMT decoder as an autoregressive language model, that can predict the next word depending on previously generated words, several researchers studied the capabilities that this perspective can bring to the quality of NMT in general, and domain adaptation in particular. Research in this direction involves: a) iterative prediction–correction approaches, b) information retrieval from training datasets, and c) employing external language models at inference time.

In a user survey I conducted \citep{Moslem2022-WLAC}, participants indicated that suggesting translation alternatives can be a source of inspiration. Moreover, it can be easier or faster than typing, and it can limit their need to refer to external resources. When a high-quality baseline MT model is employed, MT auto-completion can yield higher quality translation \citep{Green2014-zv}.

The WMT’s Word-Level AutoCompletion (WLAC) shared task\footnote{\url{https://statmt.org/wmt22/word-autocompletion.html}} addresses a more specific scenario, where the user types a few characters, and the NMT system predicts and auto-completes the correct word, given the current context. In 2022, I made submissions for Chinese-to-English, English-to-Chinese, German-to-English, and English-to-German language directions. I employed random sampling to generate diverse alternatives, and achieved excellent results (1st and 2nd places in the shared task) based on both automatic and human evaluation. Random sampling is a decoding mode that randomly samples tokens from the model output distribution. To obtain diverse generations from the MT model, I rely on randomness in the decoding method, in particular through \mbox{top-K} sampling that samples the next word from the top-K most probable choices \citep{Fan2018-zm, Holtzman2018-rf, Radford2019-GPT-2}, instead of aiming to decode text that maximises likelihood. More details on this topic can be found in Chapter \ref{chapter:autosuggest}.

\subsubsection{RQ1.b Adaptive MT with LLMs}
\label{sec:q-adaptive-mt}

Real-time translation that can adapt to changes in the context and terminology remains a challenging task. Autoregressive decoder-only LLMs such as BLOOM \citep{BLOOM2022}, \mbox{Falcon} \citep{Penedo2023-Falcon}, \mbox{GPT-3} \citep{Brown2020-GPT-3}, and \mbox{GPT-4} \citep{OpenAI2023-GPT-4} are trained to predict the next word given the previous context. In-context learning allows these models to adapt their output to adhere to the terminology and style used in previously approved translation pairs without further training.

In my research \citep{Moslem2023-AdaptiveMT}, I explored employing the in-context learning feature of LLMs to enhance real-time adaptive MT. The findings illustrated in the paper show that LLMs can effectively adapt to specific domains and terminologies at translation time, outperforming strong encoder-decoder MT systems, especially for high-resource languages. In this paper, I employed an embedding-based similarity approach for retrieving similar translation pairs (fuzzy matches) from a TM instead of providing random examples. The performance further improves as more fuzzy matches are added to the context.

One significant advantage of in-context learning is its capacity for real-time customisation. This advantage can give MT systems the capability to learn from previous interactions and adapt their output to align with the user's preferred terminology and style at inference time \citep{Moslem2023-AdaptiveMT}. Although in-context learning can improve adaptive MT without any special fine-tuning, LLMs can be fine-tuned to further enhance their in-context learning ability at translation time \citep{Moslem2023-AdaptiveMT-LLM-Finetuning}. Overall, my research demonstrates the practical implications of using in-context learning of LLMs for real-time adaptive MT, offering opportunities to improve translation quality and efficiency in diverse language pairs and domains at inference time.

\subsubsection{RQ1.c Terminology-constrained MT with LLMs}
\label{sec:q-term-mt}

Terminology-constrained MT is the scenario where domain-specific terminology can be \mbox{enforced} in a multi-domain NMT model at translation time. The approach should be capable of handling unseen terminology while retaining NMT’s ability to produce fluent output sequences \citep{Hokamp2017-ConstrainedDecoding, Dinu2019-TerminologyConstraintsTraining, Exel2020-TerminologyConstrainedMT}.

In my work \citep{Moslem2023-AdaptiveMT}, I investigated terminology-constrained MT with LLMs. Simply put, I added lists of relevant terms while prompting an LLM for translation. In addition to automatic evaluation, human evaluation for terminology-constrained MT with LLMs was conducted by professional linguists, who evaluated the adherence of the model to required terms and its impact on the overall translation quality. The evaluation compared diverse scenarios, such as zero-shot translation, zero-shot with glossary terms, two-shot translation with fuzzy matches, and two-shot translation with both fuzzy matches and glossary terms, to assess the usage of provided terms in the translations. The evaluators found that for Arabic, French, and Spanish, terminology-constrained MT was more successful at incorporating the provided glossary terms into the translations.

Similarly, in my research \citep{Moslem2023-Terminology}, LLMs were used for terminology-constrained automatic post-editing to insert missing terms into translations generated by an encoder-decoder MT system. In other words, if an MT model does not produce a translation that includes the terms provided by the organisers, the translation is fed into the LLM, prompting it to incorporate these terms while retaining the rest of the translation. While the approach can be used independently with a generic MT model, in this paper, it was used after fine-tuning an MT model on synthetic in-domain data to assess the extra gains that can be achieved from this step. The whole process almost doubled the use of terms across three language pairs, compared to the translations generated by a baseline generic model.

\subsection{\scshape RQ2. Domain-specific Text Generation for MT}

In-domain data scarcity is common in translation settings, which makes it challenging to fine-tune MT models, and produce translations that are consistent with the relevant context \citep{Axelrod2011-Data,Haddow2012-Data,Moslem2022-MT-LM}. Purely synthetic data may be beneficial for domain adaptation when even monolingual in-domain natural data is unavailable, either due to lack of resources or an extremely narrow target domain. Synthetic sentences may be produced by a template or an external model, in conjunction with forward-translation or back-translation \citep{Saunders2022-Survey}.

Recently, there has been a considerable advancement in training LLMs, not only for English, but also for diverse languages. My research investigates the feasibility of domain-specific text generation using LLMs and seeks to answer the following sub-questions:

\begin{itemize}
\setlength{\itemindent}{1.2em}
    \item [RQ2.a] Can we improve the quality of MT domain adaptation by using \textit{a small in-domain dataset} to generate huge amounts of synthetic in-domain data? This includes two main scenarios: (a) where there is only a small bilingual dataset available, and (b) where only the monolingual source text to be translated is available. Given a pre-trained language model for the target language, I fed existing in-domain target sentences as language model prompts to generate additional examples, then back-translated the new synthetic target sentences to obtain the equivalent source sentences. Finally, I used the resulting synthetic bilingual data for domain adaptation of the baseline NMT system. This topic was covered in my paper, \textit{Domain-Specific Text Generation for Machine \mbox{Translation}} \citep{Moslem2022-MT-LM}, which won a Best Presentation Award at AMTA 2022. The work proposes a novel approach to domain adaptation, that makes use of huge amounts of synthetic in-domain data to fine-tune baseline NMT models, and demonstrates significant improvements in translation of in-domain texts in both scenarios (cf. Section \ref{sec:q-insufficient-data}).
    
    \item [RQ2.b] Can we improve the quality of MT domain adaptation by using \textit{approved terminology} to generate huge amounts of synthetic bilingual in-domain data? In my WMT 2023 paper \citep{Moslem2023-Terminology}, I built on the approach described in the first sub-question (RQ2.a). Instead of generating the target only with an LLM and using back-translation to generate the source, I employed an LLM to generate both the source and target sides of the synthetic data, instructing it to incorporate a term pair in the generated translation pair (cf. Section \ref{sec:q-term-generation}).
\end{itemize}

\subsubsection{RQ2.a Insufficient in-domain data}
\label{sec:q-insufficient-data}

In \citet{Moslem2022-MT-LM}, I investigated utilising large-scale autoregressive (GPT-like) language models \citep{Radford2019-GPT-2,Brown2020-GPT-3}, pre-trained to predict the next word in a sequence. When there is a small in-domain dataset, that is insufficient to fine-tune a baseline MT system, I can use each target sentence as a prompt to generate text. Interestingly, the generated text simulates the domain and linguistic characteristics of the authentic in-domain data. Combining the idea of in-domain text generation with back-translation, I was able to generate huge amounts of synthetic bilingual in-domain data. Finally, I fine-tuned the baseline MT model, on a mix of synthetic in-domain data and generic data. More details on the approach can be found in Chapter \ref{chapter:generation}.

If there is no in-domain dataset at all, we can first forward-translate the source text to be translated, or a portion of it, using the baseline MT model. Then, the same aforementioned steps are applied \citep{Moslem2022-MT-LM}.

In the case that we have a list of pre-approved terms, we can use each target term or term pair to generate synthetic translation pairs that include the term \citep{Moslem2023-Terminology}. Then, we can use the bilingual term-based synthetic data to fine-tune an MT model. The process is elaborated in the next section.

\subsubsection{RQ2.c Terminology-aware text generation}
\label{sec:q-term-generation}

In \citet{Haque2020-Terminology}, I investigated using terminology for data mining of a target-side large monolingual dataset, to extract sentences similar to those in an in-domain test set. Then, back-translation is employed to generate the source. Finally, mixed fine-tuning \citep{Chu2017-mixed-fine-tuning} is applied to the MT baseline to train an in-domain model. This terminology can be either pre-approved and provided by the client or mined with a terminology extraction tool such as KeyBERT\footnote{\url{https://github.com/MaartenGr/KeyBERT}} \citep{Grootendorst2020-KeyBERT}, or TM2TB\footnote{\url{https://github.com/luismond/tm2tb}} \citep{Mondragon2021-TM2TB}. Even more similar terms can be generated using sense2vec\footnote{\url{https://github.com/explosion/sense2vec}} \citep{Trask2015-sense2vec}. Nowadays, LLMs can even be used for bilingual terminology extraction, as I demonstrated in previous work \citep{Moslem2023-Terminology}, which is addressed in Chapter~\ref{chapter:adaptive-MT} of this thesis. We can also use already identified terms from tasks such as the WMT biomedical shared task, ClinSpEn\footnote{\url{https://doi.org/10.5281/zenodo.6497350}} \citep{Neves2022-ClinSpEn}.

There are multiple approaches to employing terminology and LLMs to generate more \mbox{in-domain} sentences:

\vspace{-1ex}

\paragraph{Term-based generation:}

Each term can be used as a prompt to generate in-domain sentences. Alternatively, as in \citet{Haque2020-Terminology}, we can extract sentences that include the terms, and then generate text using these sentences as prompts \citep{Moslem2022-MT-LM}.

\paragraph{Term-constrained generation:}

\citet{Lin2020-CommonGen} and \citet{Zhang2020-POINTER} investigated using lexical constraints while generating texts using language models. They introduced CommonGen models, which are fine-tuned versions of BLOOM, GPT-2, and T5.
Unlike prefix-constrained generation, CommonGen generates text that includes the term at any position, i.e. not necessarily at the beginning. This can help generate diverse sentence patterns.

\paragraph{Bilingual few-shot in-context learning:}

During unsupervised pre-training, a language model develops a broad set of pattern recognition abilities. It then uses these abilities at inference time to rapidly adapt to or recognise the desired task. In their experiments, \citet{Brown2020-GPT-3}, use the term ``in-context learning" to describe the inner loop of this process.
In-context learning is a scenario where a pre-trained language model at inference time learns to replicate certain input-output text generation patterns, without further fine-tuning. They showed that autoregressive large language models such as GPT-3 can perform well in diverse tasks, through zero-shot, one-shot, and few-shot in-context learning without weight updates. Zero-shot generation can be used to create sentences from terms with strong LLMs out of the box. For some other LLMs, one-shot or few-shot in-context learning can achieve better results than zero-shot generation. The approach can be applied to a number of generation cases, including a) term-to-term generation, b) source-to-target-sentence generation, c) term-to-target-sentence generation, and d) term-to-bilingual-sentence generation.

In the paper describing our submission to the WMT 2023 Terminology Shared Task \citep{Moslem2023-Terminology}, ChatGPT was used to generate bilingual sentence pairs, based on the terms provided by the organisers. So, given a target term, the model was asked to generate multiple translation pairs, including both the source (e.g. German) and the target (e.g. English). This approach can be particularly useful in distilling bilingual terminology-based knowledge from an LLM, to a specialised encoder-decoder MT model, which can improve both the quality and efficiency, and reduce computing costs at inference time. After fine-tuning an encoder-decoder MT model with this terminology-based bilingual synthetic data, the model adherence to the pre-approved terminology was improved (cf. Chapter~\ref{chapter:terminology}).

\newpage
\section{Publications}

The following publications are directly related to the research questions, and they have been incorporated as chapters within this thesis:

\begin{itemize}
    \item \textit{\textbf{Adaptive \mbox{Machine} Translation with Large Language Models}} {\small \citep{Moslem2023-AdaptiveMT}} \\
    {\small \textbf{Yasmin Moslem}, Rejwanul Haque, John Kelleher, and Andy Way. 2023.  In Proceedings of the 24th Annual Conference of the European Association for Machine Translation (Research: Technical), pages 227–237, Tampere, Finland. Association for Machine Translation in the Americas.}\footnote{This paper is already well-cited in public literature.}
    
    \item \textit{\textbf{Domain Terminology Integration into Machine Translation: Leveraging Large Language Models}} {\small \citep{Moslem2023-Terminology}} \\
    {\small \textbf{Yasmin Moslem}, Gianfranco Romani, Mahdi Molaei, Rejwanul Haque, John Kelleher, and Andy Way. 2023. In Proceedings of the Eighth Conference on Machine Translation, Sentosa, Singapore. Association for Computational Linguistics.}

    \item \textit{\textbf{Fine-tuning Large Language Models for Adaptive Machine Translation}} {\small \citep{Moslem2023-AdaptiveMT-LLM-Finetuning}} \\
    {\small \textbf{Yasmin Moslem}, Rejwanul Haque, and Andy Way. 2023. arXiv preprint arXiv:2312.12740 [cs.CL].}
    
    \item \textit{\textbf{Domain-Specific Text Generation for Machine Translation}} {\small \citep{Moslem2022-MT-LM}} \\
    {\small \textbf{Yasmin Moslem}, Rejwanul Haque, John Kelleher, and Andy Way. 2022. In Proceedings of the 15th biennial conference of the Association for Machine Translation in the Americas (Volume 1: Research Track), pages 14–30, Orlando, USA. Association for Machine Translation in the Americas.}\footnote{This paper won a ``Best Presentation Award'' at AMTA 2022.}
    
    \item \textit{\textbf{Translation Word-level Auto-Completion: What can we achieve out of the box?}} {\small \citep{Moslem2022-WLAC}} \\
    {\small \textbf{Yasmin Moslem}, Rejwanul Haque, and Andy Way. 2022.  In Proceedings of the Eighth Conference on Machine Translation. Abu Dhabi, UAE. Association for Computational Linguistics.}
\end{itemize}

\newpage
While the following publications still overlap in many aspects with my research topic, they are less relevant compared to the aforementioned publications; hence, they are only cited in this thesis as needed.

\begin{itemize}
    \item \textit{\textbf{Preparing an Endangered Language for the Digital Age: The Case of Judeo-Spanish}} {\small \citep{Oktem2022-Ladino}} \\
    {\small Alp \"Oktem, Rodolfo Zevallos, \textbf{Yasmin Moslem}, G\"une\c{s} \"Ozt\"urk, Karen \c{S}arhon. 2022. In Proceedings of the Workshop on Resources and Technologies for Indigenous, Endangered and Lesser-resourced Languages in Eurasia (EURALI), LREC 2022, pages 105–110, Marseille, France. European Language Resources Association (ELRA).}
    
    \item \textit{\textbf{Arabisc: Context-Sensitive Neural Spelling Checker}} {\small \citep{Moslem2020-Arabisc}} \\
    {\small \textbf{Yasmin Moslem}, Rejwanul Haque, and Andy Way. 2020. In Proceedings of the 6th Workshop on Natural Language Processing Techniques for Educational Applications, pages 11–19, Suzhou, China. Association for Computational Linguistics.}
    
    \item \textit{\textbf{Terminology-Aware Sentence Mining for NMT Domain Adaptation: ADAPT’s Submission to the Adap-MT 2020 English-to-Hindi AI Translation Shared Task}} {\small \citep{Haque2020-Terminology}} \\
    {\small Rejwanul Haque, \textbf{Yasmin Moslem}, and Andy Way. 2020. In Proceedings of the 17th International Conference on Natural Language Processing (ICON): Adap-MT 2020 Shared Task, pages 17–23, Patna, India. NLP Association of India (NLPAI).}
    
    \item \textit{\textbf{The ADAPT System Description for the STAPLE 2020 English-to-Portuguese Translation Task}} {\small \citep{Haque2020-STAPLE}} \\
    {\small Rejwanul Haque, \textbf{Yasmin Moslem}, and Andy Way. 2020. In Proceedings of the Fourth Workshop on Neural Generation and Translation: Simultaneous Translation and Paraphrasing for Language Education (STAPLE), pages 144–152, Online. Association for Computational Linguistics.}
\end{itemize}

\vspace{\fill}

\newpage
\section{Research Context}
\label{sec:research-context}

This section aims at reviewing the most related research work. It covers the following topics: adaptive and interactive MT, terminology-constrained MT, retrieval-augmented MT, retrieval-augmented LLMs, and in-context learning for LLMs and MT.

\subsection{Adaptive and Interactive  MT}

Adaptive MT usually refers to modifying the MT output at inference time to simulate the characteristics of the text that is currently translated. Moreover, in such an adaptive environment, the system is supposed to learn in real time from the edits implemented by the users. In other words, the MT system should adapt to reduce the likelihood that the error will be repeated in subsequent translations \citep{Richardson2007-MT,Farajian2017-AdaptiveMT,OBrien2022-PE}, which improves consistency and boosts productivity.  Interactive MT can be considered a type of adaptation, where an MT system can simultaneously adjust its output based on user input, such as typing part of the translation or selecting a word suggestion from a list of the most probable completions. In such an iterative prediction–correction process, every time the user corrects a word, the system reacts offering a new translation hypothesis, expected to be better than the previous one \citep{Langlais2000-TransType,Peris2017-INMT}.

State-of-the-art MT systems are usually Transformer-based encoder-decoder models \citep{Vaswani2017-attention}. Researchers investigated how to adjust the output of these models to adapt to certain domains at translation time. Such real-time adaptation might include fine-tuning a model on the fly on translation pairs similar to the current source segment, or manipulating the decoding step to adjust the output to match the characteristics of the source text and any other requirements such as terminology. There are several approaches that fall under adaptive and interactive MT, and I will try to cover popular methods that are relevant to my work.

To deal with a multi-domain NMT scenario, especially where the domain might not be known in advance, researchers proposed an unsupervised (on-the-fly) adaptation approach. Given a source input, the most similar translation pairs are extracted from the ``context'' dataset (TM)~\citep{Li2018-AdaptiveMT,Farajian2017-AdaptiveMT}. Then, the baseline MT model is fine-tuned with the retrieved pairs, which is then applied to translate the source. After a linguist edits the MT translation, the approved translations are added to the dataset/TM. It is also recommended to dedicate a ``context'' dataset to each client or project. Finally, the adapted model is reset to the original parameters. The same process is applied for each source segment.

In contrast to the aforementioned instance-based adaptation approach that utilised retrieved fuzzy matches to fine-tune the NMT model for each source segment independently, some researchers investigated real-time augmentation. In other words, they experimented with augmenting the source input with one or more fuzzy matches at inference time, without the need to repeatedly fine-tune the model for each segment \citep{Bulte2019-fuzzy,Xu2020-fuzzy}. Nevertheless, this approach usually works much better if the model is pre-trained to perform such a task. The main advantage of this approach over the instance-based adaptation approach is that it does not risk overfitting by fine-tuning the model on only a few segments. Moreover, as the model is already pre-trained, it does not necessarily need to be served on GPUs as it does not require further fine-tuning at inference time. Due to the relevance of such retrieval-augmented MT approaches to my work, I dedicate Section \ref{sec:retrieval-augmented-mt} to research on this area.

The concept of ``online learning'' in production environments has been the focus of many researchers, not only in NMT, but also in several fields of machine learning. However, according to \citet{Etchegoyhen2021-OnlineLearning}, this type of adaptation typically requires higher learning rates, which can affect the quality of the models over time. Alternatively, less aggressive online learning setups may preserve model stability, at the cost of reduced adaptation to user-generated corrections. Hence, in their work, they evaluated different online learning configurations over time, measuring their impact on user-generated samples, as well as separate in-domain and out-of-domain datasets. The results reported in their work indicate that mixed approaches combining online learning with periodical batch fine-tuning might be needed to balance the benefits of online learning with model stability.

When it comes to interactive MT, several approaches adopted the teacher forcing mode \citep{Williams1989-TeacherForcing} where ground truth previous tokens are fed into the decoder, instead of the predicted tokens y\textsubscript{i-1} as suggested by \citet{Bahdanau2015-JointlyLearn}. Then, the engine expands the next $N$ most likely words, and continues (auto-completes) the decoding for these $N$ hypotheses independently. \citet{Peris2019-OnlineLearning} explored the incremental update of NMT systems during the post-editing or interactive translation processes. Such modifications aim to incorporate the new knowledge from the edited sentences into the translation system. Updates to the model are performed on-the-fly as sentences are corrected via online learning techniques. They implemented an interactive, adaptive system, able to react to single-character interactions, hoping to reduce the human effort required for obtaining high-quality translations. As interactive prediction can be computing intensive, \citet{Wuebker2018-Personalized} demonstrated that a large proportion of model parameters can be frozen during adaptation with minimal or no reduction in translation quality, which significantly improves efficiency. They evaluated this technique for both batch and incremental adaptation across multiple data sets and language pairs.

\subsection{Terminology-Constrained MT}

Despite the impressive quality improvements yielded by NMT systems, controlling their translation output to adhere to user-provided terminology constraints remains an open challenge \citep{Hasler2018-TerminologyConstraintsMT}. The end-to-end nature of NMT removes many ways of manually guiding the translation process that were available in older paradigms \citep{Post2018-FastConstrainedDecoding}, which makes it very sensitive to domain shift \citep{Hu2019-LexiconInduction}. Hence, there have been several research efforts to boost NMT adherence to pre-approved terminology, either through adapting the decoding step only, or through training a terminology-aware NMT model.

Lexically constrained decoding is a modification to beam search that yields decoder outputs honouring user-supplied constraints. These constraints can be provided in the form of either positive constraints, which specify that certain token sequences \emph{must} be present in the output, or negative constraints, which specify token sequences that \emph{must not} be generated \citep{Hu2019-ImprovedConstrainedDecoding}. \citet{Hokamp2017-ConstrainedDecoding} proposed an approach to lexically constrained decoding using grid beam search, an algorithm which extends beam search to allow the inclusion of pre-specified lexical constraints. The algorithm can be used with any model which generates sequences token by token. Lexical constraints take the form of words or phrases that must be present in the output sequence. Their experiments on interactive-predictive translation and domain adaptation of NMT showed that the approach can provide large improvements in translation quality in interactive scenarios. Moreover, \citet{Hasler2018-TerminologyConstraintsMT} introduced an approach to constrained neural decoding which supports target-side constraints as well as constraints with corresponding aligned input text spans.

However, the aforementioned approaches have computational complexities. Hence, \citet{Post2018-FastConstrainedDecoding} presented an algorithm for lexically constrained decoding with less complexity, and higher efficiency. The algorithm was shipped as part of the Sockeye framework.\footnote{\url{https://github.com/awslabs/sockeye}} Likewise, the Fairseq framework\footnote{\url{https://github.com/facebookresearch/fairseq}} has implemented this approach along with the algorithm for faster decoding described by \citet{Hu2019-LexiconInduction}, who extended research in lexically constrained decoding to work with batching, leading to a five-fold improvement in throughput when working with positive constraints.

The aforementioned works have mainly proposed modifications to the decoding algorithm to constrain the output to include target terms at inference time. While effective, these constrained decoding methods add significant computational overhead to the inference step. In contrast, \citet{Dinu2019-TerminologyConstraintsTraining} approached this challenge by training an NMT system to learn how to use custom terminology when provided with the input, leading to efficiency gains comparable to constraint-free decoding. The authors used inline annotation of the target terms in the source segment plus source factor embeddings during training and inference. Later, \citet{Exel2020-TerminologyConstrainedMT} investigated variations of the approach proposed by \citet{Dinu2019-TerminologyConstraintsTraining} and compared them to constrained decoding. Similarly, \citet{Hu2019-LexiconInduction} proposed an unsupervised adaptation method which fine-tunes a pre-trained out-of-domain NMT model using a pseudo-in-domain corpus. Specifically, they performed lexicon induction to extract an in-domain lexicon, and construct a pseudo-parallel in-domain corpus by performing word-for-word back-translation of monolingual in-domain target sentences.

In the same context of teaching NMT models to use terminology, placeholders can be incorporated as part of a pre-processing step of the training data, and then training the system with this data that contains these special placeholders \citep{Crego2016-Placeholders}. A similar workflow is applied at inference time. Firstly, pre-processing replaces source terms with placeholders. Secondly, post-processing is applied over the NMT output to replace placeholders with target terms. \citet{Michon2020-Terminology} extended the approach to cover a wider variety of cases, and to control morphology. They used several placeholders indicating part-of-speech (POS) and morphological information, both in the source and target sides. For each source-target term pair, they encoded all possible inflections of the source and target word, labelled with inflection type. Similarly, \citet{Sun2022-Terminology-Prompt} proposed a prompt-based method that pre-trains an NMT model to adapt to terms augmented to the input at translation time.

\subsection{Retrieval-Augmented MT}
\label{sec:retrieval-augmented-mt}

In the realm of NMT, researchers discovered that it is possible to improve the performance of encoder-decoder NMT models by retrieving external knowledge at inference time. This can be achieved through diverse means, ranging from memorising the whole training data to retrieving similar translation pairs from relevant TMs and incorporating such knowledge at inference time. Retrieval can take different forms, ranging from sentence-based to document-based retrieval. On the one hand, utilisation of the retrieved information can be relatively explicit, e.g. by interpolating the distribution of the retrieved target tokens with the output distribution from the pre-trained MT model. On the other hand, the model can be provided with the retrieved segment of text along with the input source text, and hopefully it will try to modify its output translation accordingly.

Leveraging information retrieved from a TM is a simple yet powerful data augmentation approach for boosting MT performance. Researchers concatenated source segments with the targets of the retrieved fuzzy matches, and then trained an MT model on the augmented data \citep{Bulte2019-fuzzy}. Similarly, \citet{Xu2020-fuzzy} explored data augmentation methods for training an NMT system to make use of fuzzy matches. In particular, they simply provided the neural model with information from both source and target sides of the fuzzy matches. This augmentation step can be incorporated into both the training and inference stages to boost the NMT model quality as well as adherence to the preferred domain features. \citet{Pham2020-Priming} compared diverse approaches to augmentation of fuzzy matches and retrieval algorithms. While both \citet{Pham2020-Priming} and \citet{Xu2020-fuzzy} tagged related and unrelated words in the retrieved target tokens to avoid copying them to the new translation, \citet{Pham2020-Priming} modified the decoding process to make it more efficient. \citet{Zhang2018-TranslationPieces} proposed a method to improve MT of low-frequency words or phrases by retrieving similar sentence pairs, extracting ``translation pieces'', and finally rewarding the outputs that contain the retrieved translation pieces at decoding time.

In contrast to the majority of work that uses a bilingual corpus as TM and employs source-side similarity search for memory retrieval, \citet{Cai2021-MonolingualTM} proposed a framework that uses monolingual memory and performs learnable memory retrieval in a cross-lingual manner. Hence, instead of retrieving bilingual translation pairs based on the similarity with current source and retrieved source, they use cross-lingual retrieval to measure the similarity of the current source with monolingual data in the target language. Moreover, the ability to leverage monolingual data makes this approach effective in low-resource and domain adaptation scenarios. Different from previous works that make use of mutually similar but redundant TMs, \citet{Cheng2022-TM} proposed contrastively retrieving TMs that are holistically similar to the source sentence while individually contrastive to each other, providing maximal information gains.

\citet{Dabre2017-MultiSource} explored a simple solution to ``multi-source'' NMT which relies solely on preprocessing a multilingual corpus without modifying the model architecture or training procedure. They simply concatenated the source sentences to form a single long multi-source input sentence while keeping the target-side sentence as is and trained an NMT system using this preprocessed corpus. Hence, in institutions that maintain their proceedings in multiple languages, they can use, for example, two languages to generate a translation in a third language. The authors evaluated the method in low-resource as well as rich-resource settings and showed its effectiveness, demonstrating how the NMT system leverages multilingual information to improve the translation to a target language. \citet{Zhang2018-Context} introduced an additional context encoder on the source side that receives the previous two source sentences as context, encodes them and passes the context encodings to both the encoder and decoder, integrating them using additional multi-head attention mechanisms.

Previous approaches to leveraging TMs to improve translation generated by encoder-decoder NMT models require either a significant update of the model architecture and/or additional training efforts to make the models well-behaved when TMs are taken as additional input. In their work, \citet{Reheman2023-PromptingMT} presented a simple but effective method to introduce TMs into NMT systems in a prompting fashion. Specifically, they treat fuzzy matches as prompts to the NMT model at inference time, without changing the training process. Although this approach can work with strong encoder-decoder NMT models without further adaptation, \citet{Reinauer2023-MT-ICL} found that it is most effective when the model is fine-tuned towards this task by concatenating similar translations to the training data.


In their work, \citet{Khandelwal2021-kNN-MT} introduced $k$-nearest-neighbour machine translation ($k$NN-MT), and demonstrated that it can lead to significant performance boosts over standard NMT systems. $k$NN-MT predicts tokens with a nearest neighbour classifier over a large datastore of cached examples, using representations from an NMT model for similarity search. In other words, the translation is generated word-by-word, and at each time step, they found the most similar contexts in the datastore, and computed a distribution over the corresponding target tokens. This distribution was then interpolated with the output distribution from the pre-trained MT model. The authors demonstrated that memorising the training data improves MT generalisation, and allows a multilingual model to specialise. As a result, a single translation model can adapt to multiple domains by memorising domain-specific data, without any in-domain training. They interpolated a pre-trained autoregressive language model (an NMT decoder, in this case) with a $k$NN model, with no additional training. However, $k$NN-MT heavily relies on high-quality in-domain parallel corpora, limiting its capability on unsupervised domain adaptation, where in-domain parallel corpora are scarce or non-existent. Hence, \citet{Zheng2021-kNN-Adaptive} proposed a framework that directly uses in-domain monolingual sentences in the target language to construct an effective datastore for k-nearest-neighbour retrieval. To this end, they first introduced an autoencoder task based on the target language, and then inserted lightweight adapters into the original NMT model to map the token-level representation of this task to the ideal representation of the translation task. Their experiments on multi-domain datasets demonstrated that the proposed approach significantly improves translation accuracy with target-side monolingual data, while achieving comparable performance with back-translation \citep{Sennrich2016-BT,Poncelas2019-BT}.

Another downside of $k$NN-MT is its inefficient performance, since it uses the entire reference corpus as the datastore for the nearest neighbour search at inference time. This means each step for each beam in the beam search has to search over the entire reference corpus. Therefore, other researchers extended the previous work and proposed approaches to improve efficiency \citep{Meng2022-kNN-MT-Fast,Wang2022-kNN-MT-Efficiency}. In their work, \citet{Meng2022-kNN-MT-Fast} proposed Fast $k$NN-MT to address this issue. Fast $k$NN-MT constructs a significantly smaller datastore for the nearest neighbour search: for each word in a source sentence, Fast $k$NN-MT first selects its nearest token-level neighbours, which is limited to tokens that are the same as the query token. Then at each decoding step, in contrast to using the entire corpus as the datastore, the search space is limited to target tokens corresponding to the previously selected reference source tokens. This strategy avoids searching through the whole datastore for nearest neighbours and hence improves decoding efficiency without loss of performance. The authors suggested that their approach enables the practical use of $k$NN-MT systems in real-world MT applications. Similarly, \citet{Wang2022-kNN-MT-Efficiency} explored a more efficient $k$NN-MT implementation and proposed to use clustering for feature reduction to improve the retrieval efficiency, while retaining translation quality.

\subsection{Retrieval-Augmented LLMs}
\label{sec:retrieval-augmented-llms}

In recent years, several pre-trained LLMs have been made available to the research community, covering a wide range of linguistic tasks. Among the state-of-the-art LLMs are GPT-3 \citep{Brown2020-GPT-3}, GPT-J \citep{Wang2021-GPT-J}, GPT-NeoX \citep{Black2022-GPT-NeoX}, PaLM \citep{Chowdhery2022-PaLM}, BLOOM \citep{BLOOM2022}, GLM \citep{Zeng2022-GLM}, OPT \citep{Zhang2022-OPT}, Falcon \citep{Penedo2023-Falcon}, Llama \citep{Touvron2023-Llama1,Touvron2023-Llama2}, Mistral \citep{Jiang2023-Mistral}, and Phi-2 \citep{Li2023-Phi}.

According to \citet{Lewis2020-RetrievalLLM}, large pre-trained language models have been shown to store factual knowledge in their parameters, and achieve state-of-the-art results when fine-tuned for downstream NLP tasks. However, their ability to access and precisely manipulate knowledge is still limited, and hence on knowledge-intensive tasks, their performance lags behind task-specific architectures. Hence, the authors explored a general-purpose fine-tuning recipe for retrieval-augmented generation (RAG), models which combine pre-trained parametric and non-parametric memory for language generation. The authors introduced RAG models where the parametric memory is a pre-trained seq2seq model and the non-parametric memory is a dense vector index of Wikipedia, accessed with a pre-trained neural retriever. They found that RAG models generate more specific, diverse and factual language than a state-of-the-art parametric-only seq2seq baseline.
Similarly, augmenting language model pre-training with a knowledge-retriever allows it to capture knowledge in a more modular and interpretable way. In this context, \citet{Guu2020-RetrievalLLM} augmented language model pre-training with a latent knowledge retriever, which allows the model to retrieve and attend over documents from a large corpus such as Wikipedia, used during pre-training, fine-tuning and inference. They demonstrated the effectiveness of retrieval-augmented language model pre-training by fine-tuning on the challenging task of open-domain question answering. Even when there are multiple sources of knowledge, it is possible to use tools such as Toolformer \citep{Schick2023-Toolformer}, a model trained to decide which sources of knowledge (in this case, APIs) to call, when to call them, what arguments to pass, and how to best incorporate the results into future token prediction. The authors showed that their approach achieves substantially improved zero-shot performance across various downstream tasks, often competitive with much larger models, without sacrificing its core language modelling abilities.

\citet{Borgeaud2021-kNN-LM} demonstrated that autoregressive language models can be enhanced by conditioning on document chunks retrieved from a large corpus, based on local similarity with preceding tokens. Their Retrieval-Enhanced Transformer (RETRO) obtained comparable performance to the state-of-the-art LLMs despite using 25$\times$ fewer parameters. After fine-tuning, RETRO performance can translate to downstream knowledge-intensive tasks such as question answering.

When it comes to few-shot in-context learning with retrieval-augmented generation, LLMs have shown impressive few-shot results on a wide range of tasks. However, when knowledge is key for such results, as is the case for tasks such as question answering and fact checking, massive parameter counts to store knowledge seem to be needed. \citet{Izacard2022-AtlasRetrievalLLM} showed that retrieval-augmented LLMs excel at knowledge-intensive tasks without the need for as many parameters even in few-shot settings. In their work, they presented Atlas, a carefully designed and pre-trained retrieval-augmented language model able to learn knowledge-intensive tasks with very few training examples. They reported that Atlas outperforms a 540B parameters model by 3\% despite having 50x fewer parameters. Moreover, \citet{Shi2022-kNN-Prompt} further addressed whether retrieval-augmented LLMs achieve similar gains in few-shot and zero-shot end-task accuracy. They believed that the main challenge was to achieve coverage of the verbaliser tokens that define the different end-task class labels. To address this challenge, they introduced $k$NN-Prompt, a simple and effective $k$NN-LM with automatically expanded fuzzy verbalisers. In other words, they expanded a word like ``great'' to also include ``excellent'' and other task-specific synonyms for sentiment classification. In their experiments, $k$NN-Prompt was effective for domain adaptation with no further training.

In spite of the success of retrieval-augmented generation, LLMs can only afford fixed-sized inputs due to the input length limit, preventing them from utilising rich long-context information from past inputs. For instance, GPT-3 increases the input length from 1k tokens supported by GPT-2 to 2k tokens for capturing better long-range dependencies. However, this approach typically incurs computation-intensive training. To address this, \citet{Wang2023-MemoryLLM} proposed a framework, Language Models Augmented with Long-Term Memory (LongMem), which enables LLMs to memorise long history. They designed a novel decoupled network architecture, with the original backbone LLM frozen as a memory encoder and an adaptive residual side-network as a memory retriever and reader. Enhanced with memory-augmented adaptation training, LongMem can thus memorise long past context and use long-term memory for language modelling. Typically, LongMem can enlarge the long-form memory to 65k tokens (instead of the original 1k supported by GPT-2) and thus cache many-shot extra demonstration examples as long-form memory for in-context learning. It is worth noting that LongMem is inspired by the previous work ``Memorising Transformer'' (MemTRM) \citep{Wu2022-MemTRM}. The main difference is that MemTRM faces a memory staleness challenge during training due to its coupled memory design, which uses a single model for encoding memory and fusing memory for language modelling, while uses a decoupled memory module to address the issue of memory staleness. The results demonstrate that the proposed method is effective in helping language models to memorise and utilise long-form contents.

\subsection{In-Context Learning for LLMs and MT}

Recent research has shown that scaling up language models greatly improves task-agnostic, few-shot performance, sometimes even reaching competitiveness with prior state-of-the-art fine-tuning approaches. The idea was highlighted by \citet{Brown2020-GPT-3} who trained GPT-3, an autoregressive language model with 175 billion parameters, and tested its performance in the few-shot setting. For all tasks, GPT-3 was applied without any gradient updates or fine-tuning, with few-shot demonstrations specified purely via text interaction with the model. In few-shot settings, the model is given $K$ examples of the task at inference time. An example typically has a context and a desired completion (e.g. a source sentence and its translation), and then one final example of context, with the model expected to provide the completion. In their experiments, GPT-3 achieved strong performance on many NLP datasets, including question answering and translation. Figure \ref{fig:llms} shows examples of popular LLMs and illustrates how research in this area has evolved from scaling LLMs \citep{Brown2020-GPT-3,Chowdhery2022-PaLM} to building smaller and efficient LLMs with comparable performance \citep{Jiang2023-Mistral,Li2023-Phi,Penedo2023-Falcon,Touvron2023-Llama2}.

\begin{figure}[htb]
\includegraphics[width=1\textwidth]{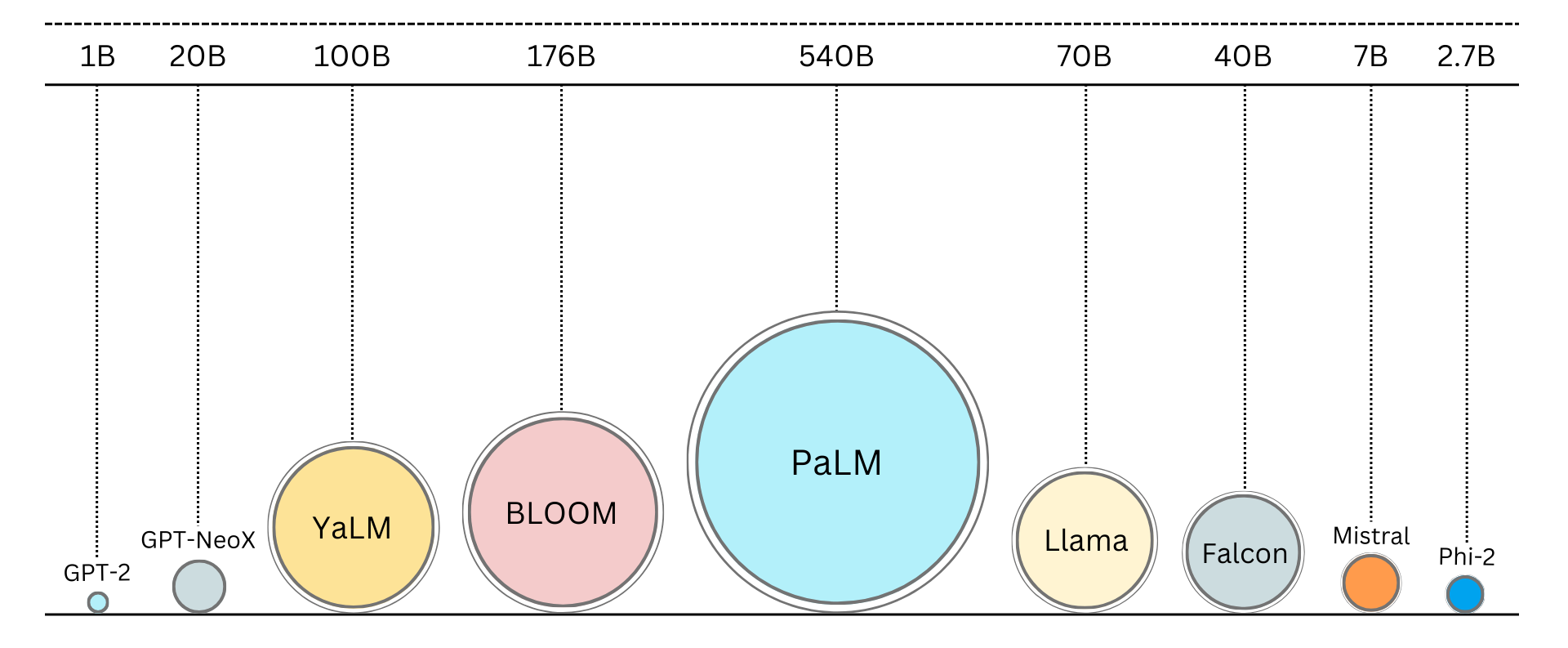}
\caption{Large language models moving from scaling to efficiency}
\label{fig:llms}
\end{figure}

The introduction of such a new paradigm opened up new horizons for researchers to explore and leverage the in-context learning capabilities of LLMs in diverse NLP tasks. \citet{Wang2021-LM4MT} demonstrated that a single language model (LM4MT) can achieve comparable performance with strong encoder-decoder NMT models on standard MT benchmarks. Similarly, LLMs that have been trained on multilingual but not parallel text exhibit a remarkable ability to translate between languages. \citet{Lin2022-LLM-MT} trained XGLM, a multilingual generative language model (up to 7.5 billion parameters), and presented a comprehensive study of multilingual zero-shot and in-context few-shot learning. They trained the models using a large-scale corpus of 500B tokens that comprises 30 diverse languages, over-sampling the less-resourced languages to render a more balanced language representation. They evaluated the models on multiple multilingual tasks, including translation. They conducted an in-depth analysis of different multilingual prompting approaches, showing in particular that strong few-shot learning performance across languages can be achieved via cross-lingual transfer through both templates and demonstration examples. Similarly, \citet{Vilar2023-PaLM-MT} conducted an in-depth study demonstrating the strong translation performance of PaLM among similarly trained LLMs. They investigated various strategies for choosing translation examples for few-shot prompting, concluding that example quality is the most important factor.

Researchers, including myself, explored the use of few-shot prompting strategies for translation, showing that the number, quality, and domain of the in-context examples significantly impact translation performance \citep{Agrawal2023-SelectionMT,Moslem2023-AdaptiveMT,Mu2023-MT-LLM-TM,Zhang2023-PromptingMT}. \citet{Garcia2023-FewShotMT} demonstrated that by incorporating examples of high-quality translation pairs at inference time, a Transformer decoder-only model trained solely with self-supervised learning is able to match specialised supervised state-of-the-art models. \citet{Sarti2023-LLM-MT} introduced an in-context learning approach to leverage attribute annotations and similar same-language or cross-lingual examples for better prompting quality. They demonstrated its effectiveness with multilingual LLMs for both formality-controlled and gender-controlled translation. Interestingly, such translation capabilities of LLMs are not only limited to commercial systems, but they are also available in open-source LLMs such as BLOOM \citep{Bawden2023-BLOOM-MT,Moslem2023-AdaptiveMT} and GLM \citep{Zhang2023-PromptingMT}.

In addition to real-time domain adaptation with fuzzy matches, my research \citep{Moslem2023-AdaptiveMT} was among the first efforts to study in-domain lexical and \mbox{terminology-constrained} translation with LLMs, and showed how incorporating domain-related information while prompting LLMs enhances in-context learning and improves the translation quality of domain-specific texts. Later, \citet{Ghazvininejad2023-DictionaryMT} proposed using bilingual dictionaries to provide control hints in the prompts, thereby enabling fine-grained phrase-level prompted control of the LLM. Moreover, they showed the effectiveness of their approach even for low-resource languages.

\citet{Schioppa2023-LLM-MT-CrossLingual} demonstrated that pre-training Large Language Models on a mixture of a self-supervised language modelling objective and the supervised MT objective, therefore including multilingual parallel data during pre-training, yields models with better in-context learning abilities. \citet{Chen2023-LLM-MT} and \citet{Liu2023-LLM-MT} proposed methods to enhance the instruction-following capability of LLMs by shifting the position of task instructions after the input sentences and adding a global instruction representation on the following input and response representations, respectively.

When it comes to low-resource languages, researchers evaluated the translation performance of LLMs for a wide range of languages and dialects, revealing that these models approach or exceed traditional MT model performance for some high-resource languages, but consistently lag for low-resource languages \citep{Kadaoui2023-LLM-MT-Arabic,Ojo2023-MT-LLM-African,Robinson2023-ChatGPT-MT}. This matches my research outcomes \citep{Moslem2023-AdaptiveMT}, while we observe that leveraging in-context learning via augmenting the translation prompt with similar translation pairs or even aligned phrases can significantly improve the quality of translation of low-resource languages. However, for very low-resource languages, fine-tuning the LLM might be required.

\section{Contribution}

With real-world production scenarios in mind, I designed and conducted the experiments in this research, building on the efforts of previous researchers to boost real-time adaptivity of MT systems. Exploring new horizons beyond traditional MT, I have taken a step forward towards leveraging LLMs for scenarios involving human interaction and continuous feedback, where the MT system is expected to concurrently adapt to such interactions.

In the light of the research questions and related work, the contribution of this research can be outlined as follows:

\begin{itemize}
    \item leveraging pretrained LLMs for domain-specific data augmentation, through generating bilingual in-domain synthetic data and using it to fine-tune an MT model, resulting in significant improvements in translation quality and adherence to pre-approved terminology \citep{Moslem2022-MT-LM,Moslem2023-Terminology}. The detailed experiments are in Chapter \ref{chapter:generation} and Chapter \ref{chapter:terminology}.
    \item exploring the gains that can be achieved through incorporating LLMs in real-time adaptive MT workflows, showing that LLMs can adapt to pre-approved in-domain translation pairs and terminology, while being solely used for translation, or while post-editing translations generated by specialised MT systems \citep{Moslem2023-AdaptiveMT,Moslem2023-AdaptiveMT-LLM-Finetuning,Moslem2023-Terminology}. The detailed experiments are in Chapter \ref{chapter:adaptive-MT}, Chapter \ref{chapter:finetuning}, and Chapter \ref{chapter:terminology}.
    \item investigating approaches to encouraging MT output to use pre-approved terms, which is usually referred to as terminology-constrained MT \citep{Moslem2023-AdaptiveMT,Moslem2022-MT-LM}, or to complete user-typed sequences (which can be terms or regular words) through either word-level auto-completion or predictive auto-suggestions \citep{Moslem2022-WLAC}. The detailed experiments are in Chapter~\ref{chapter:autosuggest}, Chapter~\ref{chapter:adaptive-MT} and Chapter~\ref{chapter:terminology}.
    
\end{itemize}

\end{large}

\begin{large}

\chapter{Domain-Specific Text Generation for Machine Translation}
\label{chapter:generation}

\vspace{-1\baselineskip}
{Yasmin Moslem, Rejwanul Haque, John Kelleher, and Andy Way}

\vspace{-12pt}
\singlespacing
\noindent {\footnotesize In Proceedings of the 15th biennial conference of the Association for Machine Translation in the Americas (AMTA 2022), Volume 1: Research Track, pages 14–30, Orlando, USA. Association for Machine Translation in the Americas.\footnote{Published at: \url{https://aclanthology.org/2022.amta-research.2/}}}
\smallskip

\onehalfspacing

\begin{abstract}
\smallskip
\nohyphens{
\normalsize
  Preservation of domain knowledge from the source to target is crucial in any translation workflow. It is common in the translation industry to receive highly specialised projects, where there is hardly any parallel in-domain data. In such scenarios where there is insufficient in-domain data to fine-tune Machine Translation (MT) models, producing translations that are consistent with the relevant context is challenging. In this work, we propose a novel approach to domain adaptation leveraging state-of-the-art pretrained language models (LMs) for \mbox{domain-specific} data augmentation for MT, simulating the domain characteristics of either (a) a small bilingual dataset, or (b) the monolingual source text to be translated. Combining this idea with back-translation, we can generate huge amounts of synthetic bilingual \mbox{in-domain} data for both use-cases. For our investigation, we use the \mbox{state-of-the-art} Transformer architecture. We employ mixed fine-tuning to train models that significantly improve translation of in-domain texts. More specifically, in both scenarios, our proposed methods achieve improvements of approximately \mbox{5-6 BLEU} and 2-3 BLEU, respectively, on the \mbox{Arabic-to-English} and \mbox{English-to-Arabic} language pairs. Furthermore, the outcome of human evaluation corroborates the automatic evaluation results.}
\end{abstract}

\newpage
\section{Context}
\enlargethispage{0.4\baselineskip}

The emergence of open-source large language models (LLMs) such as GPT-2 \citep{Radford2019-GPT-2} and BERT \citep{Devlin2019-BERT} opened the door for researchers to explore the capabilities of such models. At the time, the use of these models was limited to certain NLP tasks, and was conditioned by their architectures (e.g. encoder-only, decoder-only, or encoder-decoder models), which work for some tasks better than for others. In 2020, the release of GPT-3, an autoregressive LM with 175 billion parameters, was announced. The associated paper \citep{Brown2020-GPT-3} mentioned an attractive feature that the authors called ``in-context learning''. This refers to pattern recognition abilities that an LLM develops during pre-training and then uses at inference time to rapidly adapt to or recognise the desired task. The authors showed how to employ ``in-context learning'' in a range of tasks, including translation, question answering, natural language inference, and reasoning. For some of these tasks, the results reported in the paper were surprising, since the use of an autoregressive decoder-only model such as GPT-2 or GPT-3 was not the common practice for certain tasks. The authors noted that there was a consistent trend of quality improvement as the model scales, as well as a tendency for translation into English to be stronger than translation from English. When translating French, German, and Romanian into English, the reported scores for few-shot translation with GPT-3 were either on par with or better than the state-of-the-art WMT results for these language pairs. Similarly, the authors demonstrated the advantages of using ``in-context learning'' in other tasks. However, as GPT-3 was never open-sourced, this `magical power' was kept in a black box. As the scalability `space race' started to emerge, companies and institutions built larger and larger LMs, some of which were propriety such as  Gopher \citep{Rae2021-Gopher}, MT-NLG \citep{Smith2022-MT-NLG}, and PaLM \citep{Chowdhery2022-PaLM}, while others were open-sourced such as GPT-J \citep{Wang2021-GPT-J}, GPT-NeoX \citep{Black2022-GPT-NeoX}, BLOOM \citep{BLOOM2022}, mGPT \citep{Shliazhko2022-mGPT}, and OPT \citep{Zhang2022-OPT}. By the time of writing this paper, some of these models were publicly available, such as GPT-2, and GPT-J, while others were released later. I thought that we should try these models to improve MT domain adaptation, especially in scenarios of in-domain data scarcity, and the co-authors (my supervisors) encouraged the idea, which led to this research work. This paper won a Best Presentation Award at AMTA 2022.

\section{Introduction} 

Preservation of domain knowledge from the source to target is crucial in any translation workflow. \mbox{Domain} adaptation of MT systems on in-domain parallel texts has been an active area of research to handle this situation. Among popular contributions to the domain adaptation research, \citet{Luong2015-mt} proposed to adapt an already existing NMT system to a new domain, with further training on the in-domain data only. In an effort to avoid overfitting on the in-domain data, \citet{Chu2017-mixed-fine-tuning} employed the mixed fine-tuning approach, resuming training the baseline NMT model on a mix of in-domain and out-of-domain data. Other researchers suggested adding domain tags to either the source or target sentences of the in-domain data, to inform the NMT model about the domain during training and decoding \citep{Britz2017-domain-mixing,Kobus2017-domain-control,Stergiadis2021-domain-tag}.

In this sense, several research works on domain adaptation assume the availability of in-domain data. However, in-domain data scarcity is common in translation settings, due to the lack of specialised datasets and terminology, or inconsistency and inaccuracy of available in-domain translations. To tackle this problem, researchers have proposed diverse approaches, such as utilising large monolingual datasets through selecting instances related to a given test set, then automatically translating this source-synthetic corpus, and finally fine-tuning the general NMT system on this data \citep{Chinea-Rios2017-ln}. Similarly, some works have investigated retrieving similar translations (fuzzy matches) from bilingual datasets, and then applying on-the-fly domain adaptation through fine-tuning the baseline model at translation time \mbox{\citep{Farajian2017-AdaptiveMT}}, or integrating them into NMT training \citep{Bulte2019-fuzzy,Xu2020-fuzzy}.

While the aforementioned approaches prove to be helpful in certain scenarios of domain adaptation, we believe there is a need for further research in this area to address current challenges of in-domain data scarcity and synthetic data creation. Some approaches, such as on-the-fly domain adaptation, require using GPUs synchronously at translation time, which presents a challenge for some institutions due to the lack of resources. When it comes to mining monolingual or bilingual datasets for similar instances, in several domains a good similar sentence can be a mix of portions of multiple sentences. Besides, with the lack of in-house specialised translation memories, mining publicly available datasets can be an inefficient process.

In this work, we introduce a new approach to MT domain adaptation, leveraging state-of-the-art pre-trained language models (LMs) for domain-specific data augmentation. Our method can generate an unlimited number of in-domain sentences out of the box. 
Recently, there has been a considerable advancement in training large LMs \citep{Radford2019-GPT-2,Brown2020-GPT-3,Black2022-GPT-NeoX,Zhang2022-OPT}, not only for English, but also for diverse languages \citep{Antoun2021-ct,Zhang2021-CPM,Muller2022-Cedille}. More specifically, our current work exploits GPT-J \citep{Wang2021-GPT-J} and mGPT \citep{Shliazhko2022-mGPT} to generate texts from in-domain sentences. We investigate the \mbox{feasibility} of this domain-specific text generation technique when either no or limited bilingual in-domain dataset is available. Incorporating this approach in a process of bilingual in-domain synthetic data creation and then fine-tuning our baseline generic MT model on the new dataset (cf. Section \ref{sec:methods-amta}), we report significant improvements in the translation quality of the in-domain test set (cf. Section \ref{sec:results-amta}).

The rest of the paper is organised as follows. In Section \ref{sec:related-work-amta}, we discuss the related work in detail. Then, we present our methods in Section \ref{sec:methods-amta}. In Section \ref{sec:experimental-setup-amta}, we describe the experimental setup and present the results of our experiments in Section \ref{sec:results-amta}. Finally, we conclude the paper and discuss future work in Section \ref{sec:conclusion-amta}.

\section{Related Work}
\label{sec:related-work-amta}
\vspace{-1.5ex}

In recent years, several pre-trained large LMs have been made available to the research community, covering a wide range of linguistic tasks. Among the state-of-the-art LMs are GPT-2 \citep{Radford2019-GPT-2}, BERT \citep{Devlin2019-BERT}, RoBERTa \citep{Liu2019-RoBERTa}, XLNet \citep{Yang2019-aq}, GPT-3 \citep{Brown2020-GPT-3}, ELECTRA \citep{Clark2020-ELECTRA}, DeBERTa \citep{He2020-DeBERTa,He2021-DeBERTaV3}, T5 \citep{Raffel2020-T5}, Gopher \citep{Rae2021-Gopher}, GPT-J \citep{Wang2021-GPT-J}, GPT-NeoX \citep{Black2022-GPT-NeoX}, PaLM \citep{Chowdhery2022-PaLM}, Chinchilla \citep{Hoffmann2022-chinchilla}, ELMFOREST \citep{Li2022-ELMFOREST}, MT-NLG \citep{Smith2022-MT-NLG}, and OPT \citep{Zhang2022-OPT}. Some of these models are multilingual, such as BLOOM \citep{BLOOM2022}, AlexaTM \citep{FitzGerald2022-AlexaTM} and mGPT \citep{Shliazhko2022-mGPT}.

Using LMs for specialised domains has been explored by previous works for diverse tasks. Researchers explored the possibility to retrieve factual knowledge from LMs in various domains \citep{Petroni2019-kb,Sung2021-med-kb}. Similarly, \citet{Horawalavithana2022-chemistry} developed large-scale models of foundational scientific knowledge that can be effectively used to perform a wide range of in-domain and out-of-domain tasks.

LMs have been used in Unsupervised NMT \citep{Lample2019-xy,Chronopoulou2021-zr,Wang2021-LM4MT}. Large-scale pre-trained LMs have also been employed in a variety of MT tasks, to improve the robustness of MT models or their ability to work on domain texts \citep{Bawden2020-WMT-Biomedical,Specia2020-WMT-Robustness,Wenzek2021-WMT-Multilingual}.


Recently, \citet{Chang2021-sf} aimed at addressing the lack of training data for new application domains for data-to-text generation. They automatically augmented the data available for training by (a) generating new text samples by replacing specific values with alternative ones from the same category, (b) generating new text samples using \mbox{GPT-2}, and (c) proposing an automatic method for pairing the new text samples with data samples. Their approach boosted the performance of a standard seq2seq model by over 5 BLEU points. \citet{Sawai2021-yw} investigated the use of \mbox{GPT-2} for source-side data augmentation to improve the robustness of a generic pre-trained NMT model. They first fine-tuned the pre-trained model, BERT-fused \citep{Zhu2020-BERT-MT}, on authentic bilingual data. Then, they augmented the English source with data generated by \mbox{GPT-2}. Thereafter, they forward-translated the source-side English monolingual data with the fine-tuned version of BERT-fused. Finally, they fine-tuned the model on a combination of the authentic and synthetic data. While the reported results showed reasonable improvement (approx. 2.0 BLEU points) for the English-to-Japanese language direction, insignificant improvement (avg. 0.3 BLEU) was achieved for both English-to-German and English-to-Chinese language directions. The authors concluded that the result could be due to the relatively small amount of the original English-to-Japanese data compared to the other two language directions. We conjecture that more factors might have led to this result, including using forward-translation (rather than back-translation) of a huge amount of data, due to the noise it introduces for the decoder \citep{Haddow2022-hr}. In our current work, we try to be more specific about the task description, focussing on domain adaptation in the absence of enough in-domain data; utilising back-translation as an effective data augmentation technique \citep{Edunov2018-au,Caswell2019-el}; and giving more attention to data distribution through applying approaches like mixed fine-tuning and oversampling \citep{Chu2017-mixed-fine-tuning}.

Back-translation \citep{Sennrich2016-BT,Fadaee2018-Difficult-Words,Poncelas2019-BT} \mbox{corresponds} to the scenario where target-side monolingual data is translated using an MT system to give corresponding synthetic source sentences, the idea being that it is particularly beneficial for the MT decoder to see well-formed sentences \citep{Haddow2022-hr}. Back-translation has become a popular strategy among MT researchers, especially in low-resource scenarios \citep{Haque2021-Low-resource-MT}.
\citet{Burlot2018-Monolingual} performed a systematic study, which showed that forward-translation might lead to some improvements in translation quality, but not nearly as much as back-translation. \citet{Bogoychev2019-Translationese} concluded that forward-translation is more sensitive to the quality of the system used to produce synthetic data. Compared to back-translation, biases and errors in synthetic data are intuitively more problematic in forward-translation, since they directly affect the gold labels. The authors also reported that human evaluators favoured their back-translation systems over forward-translation systems, mostly in terms of fluency, while adequacy was largely the same across all of them, especially on the original translation direction. In their analysis, \citet{Edunov2018-au} showed that sampling or noisy synthetic data gives a much stronger training signal than data generated by beam or greedy search. \citet{Caswell2019-el} proposed a simpler alternative to noising techniques, consisting of tagging back-translated source sentences with an extra token. \citet{Hoang2018-rj} empirically showed that the quality of the back-translation system matters for synthetic corpus creation, and that NMT performance can be improved by iterative back-translation in both high-resource and low-resource scenarios.

When it comes to fine-tuning strategies for MT domain adaptation, researchers demonstrated that applying the right data distribution can significantly mitigate catastrophic forgetting of strong baselines in domain adaptation settings. \citet{Chu2017-mixed-fine-tuning} proposed the mixed fine-tuning method, whose training procedure is as follows: (a) train an NMT model on out-of-domain data until convergence, and (b) resume training the NMT model from the first step on a mix of in-domain and out-of-domain data (by oversampling the in-domain data) until convergence. According to the authors, mixed fine-tuning can address the overfitting problem of regular fine-tuning. In addition, mixed fine-tuning does not worsen the quality of out-of-domain translations, while regular fine-tuning does. Similarly, \citet{Hasler2021-domain} studied the problem in an adaptation setting where the goal is to preserve the existing system quality while incorporating data for domains that were not the focus of the original MT system. They found that they could improve over the performance trade-off offered by Elastic Weight Consolidation \citep{Kirkpatrick2017-qq} with a relatively simple data mixing strategy.

\vspace{-1.3ex}

\section{Methods}
\label{sec:methods-amta}
\vspace{-1ex}

In this work, we investigate two scenarios of in-domain data scarcity, and propose approaches to leverage pre-trained LMs for domain-specific data generation for MT \mbox{training}.

\subsection{Use Case 1: Limited bilingual in-domain data available}

 This is a common scenario where a specialised translation project is received, and although there is a large bilingual generic dataset and a small bilingual in-domain dataset (e.g. translation memory), the in-domain data is insufficient for fine-tuning a baseline model. From now on, we will refer to this use case as ``Setup\,1''. To handle this situation, we propose the following steps:
 
 \begin{enumerate}
     \setlength\itemsep{-0.1em}
     \item  We employ text generation with a large LM in the target language to augment the in-domain data. In this process, each target sentence in the in-domain dataset is used as a prompt to generate synthetic segments using the pre-trained language model. As expected, the generated text preserves the domain characteristics of the authentic in-domain data. This step enables us to have sufficient data in the target language.
    
     \item To obtain parallel source sentences, we back-translate the target-side synthetic sentences that were generated in the previous step.
     
     \item We apply mixed fine-tuning proposed by \citet{Chu2017-mixed-fine-tuning} to the baseline model. In other words, we continue training our baseline model on a mix of (a) the synthetic bilingual in-domain dataset we got from the two previous steps, and (b) a randomly sampled portion of the original generic dataset, with a data size ratio of 1:9, respectively. To apply oversampling, we employ the dataset weights feature in OpenNMT-tf\footnote{\url{https://github.com/OpenNMT/OpenNMT-tf}} \citep{Klein2020-OpenNMT}, with weights 0.9 and 0.1, respectively. Hence, the dataset weights are inversely proportional to the sizes of the two datasets.\footnote{This configuration creates a weighted dataset where examples are randomly sampled from the data files according to the provided weights. In simple words, it sequentially samples 9 examples from the smaller in-domain dataset, and 1 example from the larger generic dataset, and so on.} As the in-domain corpus is smaller than the generic corpus, oversampling allows the model to pay equal attention to both corpora. As a result of the mixed fine-tuning process, we obtained a new model that translates in-domain data significantly better than the baseline (cf. Section \ref{sec:results-amta}).\footnote{Inspired by \citet{Hasler2021-domain} who applied 20x oversampling, we experimented with a higher oversampling ratio. Increasing both the data size and weight degraded performance on the in-domain test set, compared to our applied 9x ratio, while increasing the weight only did not result in a significant improvement. We might investigate the effect of changing the oversampling ratio further in the future.}

     \item Although the new fine-tuned model can still adequately translate generic data, we noticed it can degrade performance by 1-2 BLEU points. Therefore, we experimented with \mbox{checkpoint} averaging \citep{Vaswani2017-attention} of the fine-tuned model with the baseline model to reduce \mbox{variability} between trainings and address rapid overfitting during fine-tuning \citep{Tran2021-oj}. This step helps regain the higher evaluation score of the baseline model on generic data, while retaining the improved score of the fine-tuned model on in-domain data.
 \end{enumerate}

\subsection{Use Case 2: Zero bilingual in-domain data available}

In this case, we assume that there is no bilingual in-domain data at all. There is only the source text that requires translation. From now on, we will refer to this use case as ``Setup\,2''.

The first step is to use the baseline MT model for forward-translation of the source text. The generated translation might not be perfect; however, it can still include useful information about the domain. This approach bootstraps some parallel data for a situation where there was none. Then, we follow the same four steps mentioned in the previous use case.

\section{Experimental Setup}
\label{sec:experimental-setup-amta}

\subsection{Datasets}

For training Arabic-to-English and English-to-Arabic generic models, we collect high-quality datasets from OPUS \citep{Tiedemann2012-OPUS}.
The breakdown of segment numbers in our datasets before and after filtering is shown in Table \ref{tab:datasets-generic}.
To ensure the quality of our datasets, we apply a multi-filtering process. First, we apply rule-based filtering to individual datasets, removing duplicates, source-copied segments, those with too long source/target (ratio 200\% and $>$ 200 words), and HTML tags. Then, we calculate the similarity between each source and target to semantically filter out segments with a similarity threshold lower than 0.45.
Finally, we concatenate the datasets and apply global filtering. For the development and test datasets, we randomly sampled 5000 segments each from the original dataset.\footnote{Our MT preparation scripts are publicly available at: \url{https://github.com/ymoslem/MT-Preparation}}

For in-domain NMT models, we use TICO-19 \citep{Anastasopoulos2020-TICO-19}, a dataset in the Public Health domain. After filtering, the dataset includes 3062 segments. Table \ref{tab:dataset-in-domain} shows the dataset details.
We split the TICO-19 dataset into a development dataset, with 1000 segments, and a test dataset which includes the rest, i.e. 2062 segments. The whole TICO-19 dataset is used for generating a large synthetic in-domain training dataset, as described in Section \ref{sec:data-lm}.

\subsection{Vocabulary}

To create our vocabulary, we first train SentencePiece unigram models \citep{Kudo2018-SentencePiece,Kudo2018-subword} for the source and target individually, to learn subword units from untokenised text.\footnote{In SentencePiece, we utilise the training options \url{--split_digits} to split all digits into separate pieces, and \url{--byte_fallback} to decompose unknown pieces into UTF-8 byte pieces to help avoid out-of-vocabulary tokens.} Then, we utilise this SentencePiece model to subword our dataset. We use a vocabulary size of 50,000. Subsequently, we convert the learned subword units into our final vocabulary in the format supported by OpenNMT-tf. Segments are automatically augmented with start and end tokens via \textit{source\_sequence\_controls} option.

\subsection{NMT Model Architecture}

Our baseline generic NMT models use the Transformer ``Big" architecture \citep{Vaswani2017-attention} as implemented in OpenNMT-tf, and relative position representations (Shaw et al., 2018) with a clipping distance k=20. The models consist of 6 layers with a model dimension of 1,024, split into 16 heads, and a feedforward dimension of 4,096.

\begin{table}[H]
\captionsetup{font=small,labelfont=small}
\centering
\begin{small}
\begin{tabular}{@{}llll@{}}
\toprule
                          &            & \multicolumn{2}{c}{\textbf{Filtering}} \\ \cmidrule(l){3-4} 
\textbf{Dataset} & \textbf{Raw}        & \multicolumn{1}{c}{\textbf{Rule-based}} & \multicolumn{1}{c}{\textbf{Semantic}} \\ \midrule
Bible                     & 62,195     & 47,699                 & 43,951        \\
ELRC\_2922                & 15,129     & 14,937                 & 14,850        \\
GlobalVoices              & 63,071     & 55,201                 & 51,220        \\
GNOME                     & 150        & 143                    & 134           \\
Infopankki                & 50,769     & 15,531                 & 14,635        \\
KDE4                      & 116,239    & 85,003                 & 68,180        \\
MultiUN                   & 9,759,125  & 7,807,811              & 7,508,443     \\
News-Commentary           & 97,384     & 80,744                 & 77,715        \\
OpenSubtitles             & 29,823,188 & 23,666,245             & 20,176,228    \\
Tatoeba                   & 27,905     & 27,649                 & 26,714        \\
Ubuntu                    & 5,978      & 5,617                  & 5,340         \\
UN                        & 74,067     & 63,074                 & 62,901        \\
UNPC                      & 20,044,478 & 15,696,210             & 15,441,996    \\
Wikimedia                 & 407,543    & 335,783                & 317,285       \\
Wikipedia                 & 151,136    & 117,859                & 116,940       \\ \midrule
\textbf{Total}   & \textbf{60,698,357} & \textbf{48,019,506}                     & \textbf{43,926,532}                   \\ \midrule
\textbf{Global Filtering} &            & \textbf{40,207,905}    &               \\ \bottomrule
\end{tabular}
\end{small}
\vspace{-0.5\baselineskip}
\caption{Generic datasets}
\label{tab:datasets-generic}
\end{table}

\vspace{-0.9\baselineskip}

\begin{table}[!ht]
\captionsetup{font=small,labelfont=small}
\centering
\begin{small}
\begin{tabular}{@{}p{2.5cm}p{1.4cm}llll@{}}
\toprule
                 &              & \multicolumn{2}{c}{\textbf{Filtering}}                            \\ \cmidrule(l){3-4} 
\textbf{Dataset} & \textbf{Raw} & \textbf{Rule-based}       & \multicolumn{1}{c}{\textbf{Semantic}} \\ \midrule
TICO-19          & 3,071        & \multicolumn{1}{l}{3,069} & 3,062                                 \\ \bottomrule
\end{tabular}
\end{small}
\vspace{-0.5\baselineskip}
\caption{In-domain dataset (Public Health)}
\label{tab:dataset-in-domain}
\end{table}

\subsection{Training}
The training takes place on 2x NVIDIA RTX A4000 GPUs, with a batch size of 2048 tokens per GPU, for an effective batch size of 25k tokens/step. The Arabic-to-English model is trained for 240k steps, while the English-to-Arabic model is trained for 105k steps. Early stopping is used after 3 evaluations with less than 0.01 BLEU improvement on the development dataset.

\subsection{Domain-Specific Data Generation with LMs}
\label{sec:data-lm}

For English, we use GPT-J \citep{Wang2021-GPT-J}, a Transformer-based language model with 6B trainable parameters.\footnote{\url{https://huggingface.co/EleutherAI/gpt-j-6B}} 
For Arabic, we use mGPT \citep{Shliazhko2022-mGPT}, a multilingual language model.\footnote{\url{https://huggingface.co/sberbank-ai/mGPT}}

To fit the models onto an NVIDIA RTX A4000 GPU (16 GB of GPU memory), the half-precision floating-point (float16) format is used.\footnote{In Hugging Face Transformers, we also set the option \url{low_cpu_mem_usage} to \url{True}.} We also use a batch size of 1.\footnote{It is worth mentioning though that for batch generation (i.e. $>$1), padding and attention masking should be used; note that left padding is required for GPT-like models.} For inference, we employ 50 Top-K sampling and 0.95 Top-p (nucleus) sampling \citep{Fan2018-zm,Holtzman2018-rf,Radford2019-GPT-2,Holtzman2020-nucleus}. The maximum length of the generated text is set to 300 tokens, and we return 5 sequences for each segment, to get multiple independently sampled outputs. Finally, we split the generated text into sentences.\footnote{Our scripts are available at: \url{https://github.com/ymoslem/MT-LM}}

As explained in Section \ref{sec:methods-amta}, we have two use-cases: (a) a small bilingual in-domain dataset is available; and (b) the source only is available, so we utilise forward-translation to generate the target side. After that, each target sentence of the in-domain dataset TICO-19 (i.e. the authentic target in the first case, or the MT-ed target in the second case) is fed to the LM as a prompt to generate synthetic in-domain segments. We use random seeds to generate multiple datasets, namely 2 for English and 3 for Arabic.\footnote{As two data generation runs for Arabic resulted in a less amount of data than for English, we increased the data size for Arabic by generating a third dataset (cf. Table \ref{tab:datasets-lm}).} We filter the concatenated datasets, by removing duplicates and cleaning lines with a wrong language, and those including only dashes or filenames. Table \ref{tab:datasets-lm} illustrates the numbers of in-domain synthetic segments generated by the LMs.

\begin{table}[H]
\captionsetup{font=small,labelfont=small}
\centering
\begin{scriptsize}
\begin{tabular}{@{}llllllllllll@{}}
\toprule
\textbf{}        & \textbf{} & \multicolumn{5}{c|}{\textbf{Setup\,1}}           & \multicolumn{5}{c}{\textbf{Setup\,2}}             \\ \cmidrule(l){3-12} 
\textbf{Language} &
  \textbf{LM} &
  \textbf{1st Run} &
  \textbf{2nd Run} &
  \textbf{3rd Run} &
  \textbf{Total} &
  \textbf{Filtered} &
  \textbf{1st Run} &
  \textbf{2nd Run} &
  \textbf{3rd Run} &
  \textbf{Total} &
  \textbf{Filtered} \\ \midrule
\textbf{English} & GPT-J     & 131,730 & 131,554 & N/A    & 263,284 & 242,469 & 137,705 & 138,702 & N/A     & 276,407 & 253,287 \\
\textbf{Arabic}  & mGPT      & 96,296  & 97,031  & 94,513 & 287,840 & 271,665 & 103,272 & 103,459 & 103,303 & 310,034 & 294,391 \\ \bottomrule
\end{tabular}
\end{scriptsize}
\vspace{-0.5\baselineskip}
\caption{Data generated by language models (LMs)}
\label{tab:datasets-lm}
\end{table}

\subsection{Back-Translation}

For back-translation, we use OPUS models,\footnote{\url{https://github.com/Helsinki-NLP/Tatoeba-Challenge/tree/master/models}} specifically the Transformer-Big versions. For efficiency purposes, we convert the models to the CTranslate2\footnote{\url{https://github.com/OpenNMT/CTranslate2}} format (INT8 quantisation). We use beam size 5. After back-translation, we run the same rule-based and semantic filtering on the generated dataset as we did for the original datasets. Table \ref{tab:datasets-bt} elaborates on the numbers.

\begin{table}[H]
\captionsetup{font=small,labelfont=small}
\centering
\begin{small}
\begin{tabular}{@{}lllllll@{}}
\toprule
\textbf{}        & \multicolumn{3}{c|}{\textbf{Setup\,1}}                & \multicolumn{3}{c}{\textbf{Setup\,2}}                 \\ \cmidrule(l){2-7} 
                 &            & \multicolumn{2}{c}{\textbf{Filtering}} &            & \multicolumn{2}{c}{\textbf{Filtering}} \\ \cmidrule(lr){3-4} \cmidrule(l){6-7} 
\textbf{Language} & \textbf{Translated} & \textbf{Rule-based} & \textbf{Semantic} & \textbf{Translated} & \textbf{Rule-based} & \textbf{Semantic} \\ \midrule
\textbf{English} & 242,469    & 240,329            & 239,931           & 253,287    & 251,357            & 250,317           \\
\textbf{Arabic}  & 271,665    & 271,645            & 270,743            & 294,391    & 294,234            & 293,252           \\ \bottomrule
\end{tabular}
\end{small}
\vspace{-0.6\baselineskip}
\caption{Back-translated datasets}
\label{tab:datasets-bt}
\end{table}

\subsection{Mixed Fine-tuning}

Following \citet{Chu2017-mixed-fine-tuning}, we employ the mixed fine-tuning approach. We randomly sample a portion from the generic data we used to train the baseline model, and use it during the fine-tuning step along with the in-domain dataset. Oversampling the in-domain data is a crucial step, as explained in Section \ref{sec:methods-amta}. We first train a baseline NMT model on out-of-domain data until convergence, and then continue training the NMT baseline model on a mix of in-domain and out-of-domain data (by oversampling the in-domain data) until convergence.

In most experiments, we fine-tuned the baseline for 5000 steps. However, for Setup\,2 of the English-to-Arabic language pair, we found that the best automatic evaluation scores were achieved with training for only 500 or 1000 steps. We believe that this might be due to the quality or distribution of the generated in-domain data compared to the original generic data. Although \citet{Chu2017-mixed-fine-tuning} observed that both regular fine-tuning and mixed fine-tuning tend to converge after 1 epoch of training, it seems there is no golden rule as to how many steps or epochs the baseline model should be fine-tuned on the mixed data. Depending on the size of data, we recommend conducting less-frequent evaluations on the development dataset during the fine-tuning process for finding out the best model checkpoint.

\section{Results}
\label{sec:results-amta}
\vspace{-1ex}

In this section, we elaborate on our automatic and human evaluations and discuss the results. As Table \ref{tab:evaluation-domain} shows, scores obtained from diverse automatic metics provide good correlation with the human evaluation. Moreover, the linguistic analysis (cf. Section \ref{sec:analysis}) supports these numerical results, and demonstrates how the models fine-tuned on synthetic in-domain data produce more accurate translations of the in-domain test set compared to the baseline model.


\subsection{Automatic Evaluation}

For automatic evaluation, we calculated spBLEU \citep{Papineni2002-BLEU,Goyal2022-spBLEU} which uses a \mbox{SentencePiece} tokeniser with 256,000 tokens and then the BLEU score is \mbox{computed} on the sub-worded text. spBLEU has been recently added to sacreBLEU v2.1.0.\footnote{\url{https://github.com/mjpost/sacrebleu}} \mbox{\citet{Goyal2022-spBLEU}} showed that spBLEU exhibits a strong correlation with the tokenisation-independent chrF++, yet has the advantage of keeping the familiarity of BLEU. To verify our results, we employed other evaluation metrics, namely the character-based metric chrF++ \citep{Popovic2017-chrF++}, and the word-based metric TER \citep{Snover2006-TER}, as implemented in sacreBLEU \citep{Post2018-sacreBLEU}. Furthermore, we integrated COMET\footnote{\url{https://github.com/Unbabel/COMET}} \citep{Rei2020-COMET} as a semantic evaluation metric, with the ``wmt20-comet-da'' model.

We experimented with averaging parameters across multiple model checkpoints \citep{Vaswani2017-attention}, to address bias towards recent training data \citep{Tran2021-oj}. Sometimes, averaging multiple checkpoints of a baseline model, or averaging a baseline model with a fine-tuned model could lead to extra improvements of the automatic and/or human evaluation of our \mbox{models}. Table \ref{tab:evaluation-domain} shows evaluation results on the in-domain test dataset, and Figure \ref{fig:evaluation} elaborates on all the automatic evaluation results, including the results for averaged models.

\begin{table}[H]
\captionsetup{font=small,labelfont=small}
\centering
\begin{small}
\begin{tabular}{@{}lllllll@{}}
\toprule
\textbf{Language} & \textbf{Model} & \textbf{spBLEU ↑} & \textbf{chrF++ ↑} & \textbf{TER ↓} & \textbf{COMET ↑} & \textbf{Human ↑} \\ \midrule
\multirow{3}{*}{\textbf{AR-EN}} & Baseline         & 44.57 & 66.68 & 46.67 & 65.78 & 87.0 \\
                                & Setup\,1 Mixed Fine-Tuning  & 49.79 & 70.54 & 43.32 & 71.89 & 93.5 \\
                                & Setup\,2 Mixed Fine-Tuning & 47.22 & 69.38 & 45.38 & 70.08 & 94.5 \\ \midrule
\multirow{3}{*}{\textbf{EN-AR}} & Baseline         & 36.15 & 58.3  & 58.29 & 57.5  & 87.0 \\
                                & Setup\,1 Mixed Fine-Tuning  & 42.38 & 62.52 & 53.99 & 67.48 & 90.0 \\
                                & Setup\,2 Mixed Fine-Tuning & 37.91 & 59.42 & 55.95 & 59.47 & 88.5 \\ \bottomrule
\end{tabular}
\end{small}
\vspace{-0.5\baselineskip}
\caption{Evaluation results on the in-domain test set, TICO-19}
\label{tab:evaluation-domain}
\end{table}

\subsection{Human Evaluation}

Since translation focusses mainly on word choice, syntax, and semantics, and how people perceive it, we decided to complement our evaluation process with human evaluation.

The evaluator was an Arabic native speaker and domain expert. We conducted a bilingual evaluation, providing the evaluator with both the original source sentences and translations generated by the MT models. The human test set contained 50 sentences, randomly extracted from the original test set, and verified as accepted translations. The evaluator was asked to assess the acceptability of each translation generated by our baselines and fine-tuned MT systems, using the scale proposed by \citet{Coughlin2003-MT-eval}, ranging from 1 to 4, and outlined as follows:

\vspace{-0.3\baselineskip}
\begin{small}
\begin{itemize}
    \setlength\itemsep{-0.3em}
    \item \textbf{4 = Ideal:} Not necessarily a perfect translation, but grammatically correct, with all information accurately transferred.
    \item \textbf{3 = Acceptable:}  Not perfect (stylistically or grammatically odd), but definitely comprehensible, AND with accurate transfer of all important information.
    \item \textbf{2 = Possibly Acceptable:}  Possibly comprehensible (given enough context and/or time to work it out); some information transferred accurately.
    \item \textbf{1 = Unacceptable:} Absolutely not comprehensible and/or little or no information is accurately \mbox{transferred}.
\end{itemize}
\end{small}

Human evaluation results on the in-domain dataset, TICO-19, are expressed in percentage points in the last column of Table \ref{tab:evaluation-domain}. In addition, Table \ref{tab:human-evaluation} elaborates on the results for all the systems, showing the mean value for each system on the 1-4 scale.\footnote{Sentence-level human evaluation is available at: \url{https://github.com/ymoslem/MT-LM}} The models fine-tuned on the domain-specific synthetic dataset achieve improvements on the in-domain test set, while retaining the baseline's quality on the generic holdout test set.

\begin{table}[H]
\captionsetup{font=footnotesize,labelfont=footnotesize}
\centering
\begin{tiny}
\begin{tabular}{@{}llcccccccc@{}}
\toprule
\textbf{Language} &  \textbf{Test Set} &   \textbf{BS} &  \textbf{BS-Avg8} &  \textbf{MixFT-1} &  \textbf{MixFT-1+BS} &  \textbf{MixFT-1+BS-Avg8} &  \textbf{MixFT-2} &  \textbf{MixFT-2+BS} &  \textbf{MixFT-2+BS-Avg8} \\
\midrule
   \multirow{2}{*}{\textbf{AR-EN}} & \textbf{Generic} & 3.84 & \textbf{3.90} & 3.84 & 3.88 & 3.88 & 3.84 & 3.84 & 3.84 \\
                         &  \textbf{TICO-19} & 3.48 & 3.62 & 3.74 & \textbf{3.82} & 3.80 & 3.78 & 3.72 & 3.74 \\
   \multirow{2}{*}{\textbf{EN-AR}} & \textbf{Generic} & \textbf{3.96} & 3.90 & 3.82 &  \textbf{3.96} & 3.90 & 3.94 & \textbf{3.96} & \textbf{3.96} \\
                         &  \textbf{TICO-19} & 3.48 & 3.50 & \textbf{3.60} & 3.54 & 3.52 & 3.54 & 3.56 & 3.54 \\
\bottomrule
\end{tabular}
\end{tiny}
\vspace{-0.5\baselineskip}
\caption{\nohyphens{Human evaluation of MT models for Arabic-to-English (AR-EN) and English-to-Arabic (EN-AR) language pairs, the baseline (BS), baseline averaged 8 checkpoints (BS-Avg8), mixed fine-tuning model (MixFT), averaging MixFT with BS (MixFT+BS), and averaging MixFT with BS-Avg8 (MixFT+BS-Avg8). MixFT-1 refers to Setup\,1 and MixFT-2 refers to Setup\,2.}}
\label{tab:human-evaluation}
\end{table}

\subsection{Linguistic Analysis}
\label{sec:analysis}

We observe that in several cases, the fine-tuned (\mbox{in-domain}) models generate more idiomatic translations or better capture the meaning in the Public Health context. Samples from the test dataset translated by the baseline model and in-domain models reflect these improvements.

Among Arabic-to-English examples, the phrase ``\begin{footnotesize}\<غير مسببة للأمراض في مضيفاتها المستودعة الطبيعية>\end{footnotesize}'' was translated as ``not pathogenic in their naturally \mbox{occurring} host'' by the baseline, and ``non-pathogenic in their natural reservoir hosts'' by both in-domain models. The former \mbox{translation} somehow conveys the meaning; however, the latter translation is more idiomatic in the \mbox{medical} context. The baseline system translated ``\begin{footnotesize}\<حمامات الولادة>\end{footnotesize}'' as ``maternity wards'' which is an incorrect translation, while the in-domain models in Setup\,1 and Setup\,2 produced more relevant translations as ``birthing pools'' and ``birth baths'', respectively. The baseline model translated ``\begin{footnotesize}\<مسحة بلعومية أنفية>\end{footnotesize}'' as ``a nasal laryngeal swab'' which is an inaccurate translation. In contrast, both in-domain models translated the term as ``a nasal nasopharyngeal swab'', which uses the \mbox{accurate} ``nasopharyngeal'' medical term. It can still be edited by removing the redundant ``nasal''; however, our evaluator gave it a higher score than the translation provided by the baseline. The term ``\begin{footnotesize}\<اختبارات مَصليَّة>\end{footnotesize}'' was translated as ``serum tests'' by the baseline, while it was translated as ``serological tests" by both in-domain models, which is more idiomatic.

Examining some of the English-to-Arabic translations, the baseline model mistranslated ``HCoVs'' as ``\begin{footnotesize}\<فيروسات نقص المناعة البشرية$/$متلازمة نقص المناعة المكتسب (الإيدز)>\end{footnotesize}'' (HIV/AIDS), as opposed to the in-domain models, which correctly translated it as ``\begin{footnotesize}\<فيروسات كورونا البشرية>\end{footnotesize}'' or just \mbox{``\begin{footnotesize}HCoV \<فيروسات>\end{footnotesize}''}. Interestingly, even for a simpler phrase like ``five times more cases", the \mbox{baseline} incorrectly translated it as ``\begin{footnotesize}\<خمس حالات>\end{footnotesize}'' (five cases), whilst the in-domain models correctly conveyed the meaning as ``\begin{footnotesize}\<خمسة أضعاف الحالات>\end{footnotesize}''.

There are also examples where one of the in-domain systems generated the correct translation while the other could not. For instance, both the baseline and Setup\,2 in-domain model translated ``If you do wear a mask'' as ``\begin{footnotesize}\<إذا كنت لا ارتداء قناع>\end{footnotesize}'', which is both syntactically and semantically incorrect. In contrast, the Setup\,1 in-domain model perfectly translated it as ``\begin{footnotesize}\<إذا كنت ترتدي قناعًا>\end{footnotesize}''. The baseline model translated the phrase ``passive antibody therapy'' as ``\begin{footnotesize}\<العلاج السلبي للأجسام المضادة>\end{footnotesize}'', which uses the preposition ``\begin{footnotesize}\<لـ>\end{footnotesize}'' (of) instead of ``\begin{footnotesize}\<بـ>\end{footnotesize}'' (with), missing the fact that in this context ``antibody'' is equivalent to \mbox{``antibody-based''} rather than being the issue to be treated. Similarly, the Setup\,2 in-domain model mistranslated it as ``\begin{footnotesize}\<العلاج المضاد السلبي>\end{footnotesize}'' while the Setup\,1 in-domain model accurately translated it as ``\begin{footnotesize}\<العلاج السلبي بالأجسام المضادة>\end{footnotesize}''.


Since some phrases can be expressed in multiple ways, we notice that sometimes the evaluator equally ranked different translations. This might explain why automatic metrics evaluate Arabic-to-English Setup\,1 higher than Setup\,2, whereas the human evaluation shows that the translation quality of both setups is comparable.

\section{Conclusion and Future Work}
\label{sec:conclusion-amta}

In this work, we propose two simple methods to utilise pre-trained language models for domain-specific data augmentation for NMT systems. We report significant improvements, supported by both automatic and human evaluation. The proposed techniques enable the generation of large amounts of data, simulating the characteristics of the specialised text to be translated, and facilitating the domain adaptation process.

For the Arabic-to-English language direction, human evaluation demonstrates that Setup\,2 is on par with Setup\,1 even though in the former we did not have any authentic bilingual in-domain data (cf. Section \ref{sec:methods-amta}). Nevertheless, the English-to-Arabic model in Setup\,2 has lower performance compared to the Setup\,1 model, although both setups outperform the baseline on the in-domain test set. We believe this might be due to the quality of synthetic data generated for Arabic, which is an interesting aspect to explore further.

In the future, we would like to investigate utilising terminology for domain-specific data generation, and experiment with employing the same proposed approaches for low-resource languages and multilingual settings.

\end{large}

\newpage

\newgeometry{bottom=20pt,top=50pt}

\centering \titlepage{\textbf{Figures: Automatic Evaluation}}
\vspace{-2\baselineskip}

\begin{figure}[H]
    \centering
    \subfloat[\tiny{AR-EN Setup\,1 - In-Domain Test Set (TICO-19)}\vspace{-2\baselineskip}]{
    \includegraphics[width=0.45\textwidth]{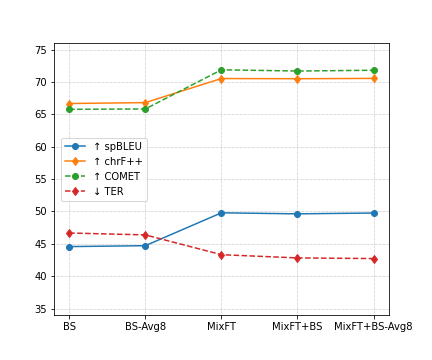}
    }
    \subfloat[\tiny{AR-EN Setup\,2 - In-Domain Test Set (TICO-19)}\vspace{-2\baselineskip}]{
    \includegraphics[width=0.45\textwidth]{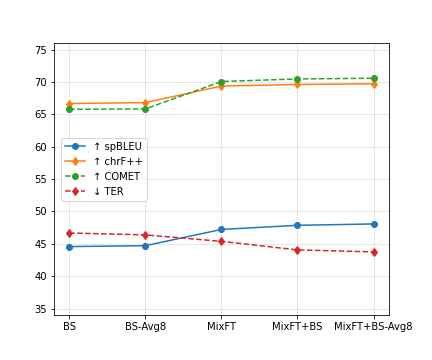}
    }%
    \vspace{-1.5\baselineskip}
    \subfloat[\tiny{AR-EN Setup\,1 - Generic Test Set}\vspace{-2\baselineskip}]{
    \includegraphics[width=0.45\textwidth]{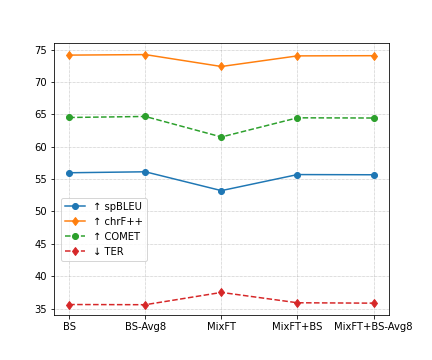}
    }
    \subfloat[\tiny{AR-EN Setup\,2 - Generic Test Set}\vspace{-2\baselineskip}]{
    \includegraphics[width=0.45\textwidth]{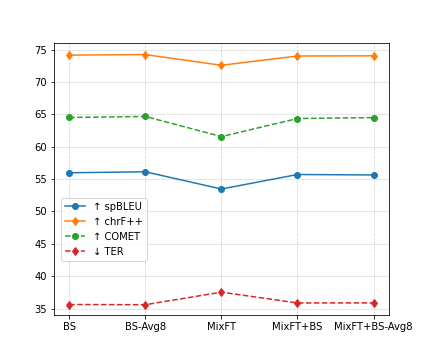}
    }
    \vspace{-0.5\baselineskip}
\end{figure}

\vspace{-1.2\baselineskip}
\centering\footnotesize{- - - - - - - - - - - - - - - - - - - - - - - - - - - - - - - - - - - - - - - - - - - - - - - - - - - - - - - - - - - - - - - - - - - - - - - - - - - - -}
\vspace{-2.2\baselineskip}

\begin{figure}[H]
\captionsetup{font=footnotesize,labelfont=footnotesize}
    \centering
    \subfloat[\tiny{EN-AR Setup\,1 - In-Domain Test Set (TICO-19)}\vspace{-2\baselineskip}]{
    \includegraphics[width=0.45\textwidth]{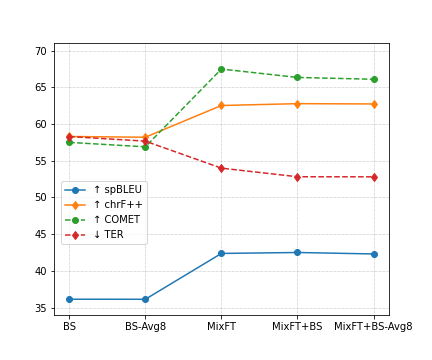}
    }
    \subfloat[\tiny{EN-AR Setup\,2 - In-Domain Test Set (TICO-19)}\vspace{-2\baselineskip}]{
    \includegraphics[width=0.45\textwidth]{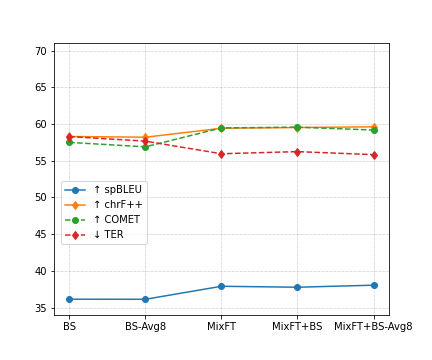}
    }%
    \vspace{-1.5\baselineskip}
    \subfloat[\tiny{EN-AR Setup\,1 - Generic Test Set}\vspace{-2\baselineskip}]{
    \includegraphics[width=0.45\textwidth]{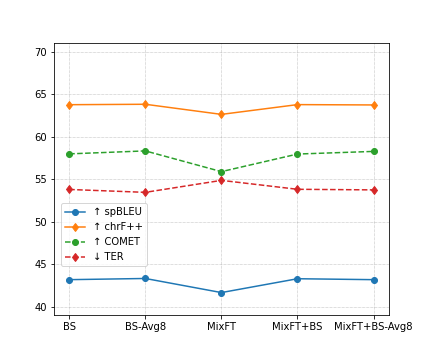}
    }
    \subfloat[\tiny{EN-AR Setup\,2 - Generic Test Set}\vspace{-2\baselineskip}]{
    \includegraphics[width=0.45\textwidth]{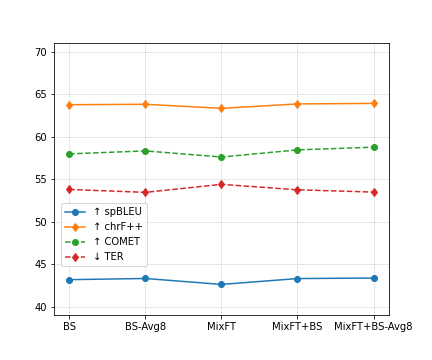}
    }
    \vspace{-0.8\baselineskip}
    \caption{\nohyphens{Performance comparison of 5 models for Arabic-to-English (AR-EN) and English-to-Arabic (EN-AR) language pairs, the baseline (BS), baseline averaged 8 checkpoints (BS-Avg8), mixed fine-tuning model (MixFT), averaging MixFT with BS (MixFT+BS), and averaging MixFT with BS-Avg8 (MixFT+BS-Avg8). The MixFT models fine-tuned on the domain-specific synthetic dataset achieve improvements on the in-domain test set (a,b \& e,f), while retaining the baselines quality on the generic test set (c,d \& g,h).}}
    \label{fig:evaluation}
\end{figure}

\restoregeometry

\restoregeometry

\addtocontents{toc}{\protect\newpage}

\begin{large}

\chapter{Translation Word-Level Auto-Completion}
\label{chapter:autosuggest}

\setcounter{page}{44}

\vspace{-1\baselineskip}
{Yasmin Moslem, Rejwanul Haque, John Kelleher, and Andy Way}

\vspace{-12pt}
\singlespacing
\noindent {\footnotesize In Proceedings of the Seventh Conference on Machine Translation (WMT 2022), pages 1176–1181, Abu Dhabi, United Arab Emirates. Association for Computational Linguistics.\footnote{Published at: \url{https://aclanthology.org/2022.wmt-1.119/}}}
\bigbreak

\onehalfspacing

\begin{abstract}
\smallskip
\nohyphens{
\normalsize
Research on Machine Translation (MT) has achieved important breakthroughs in several areas. While there is much more to be done in order to build on this success, we believe that the language industry needs better ways to take full advantage of current achievements. Due to a combination of factors, including time, resources, and skills, businesses tend to apply pragmatism into their AI workflows. Hence, they concentrate more on outcomes, e.g. delivery, shipping, releases, and features, and adopt high-level working production solutions, where possible. Among the features thought to be helpful for translators are sentence-level and word-level translation auto-suggestion and auto-completion. Suggesting alternatives can inspire translators and limit their need to refer to external resources, which hopefully boosts their productivity. This work describes our submissions to WMT's shared task on word-level auto-completion, for the Chinese-to-English, English-to-Chinese, German-to-English, and English-to-German language directions. We investigate the possibility of using pre-trained models and out-of-the-box features from available libraries. We employ random sampling to generate diverse alternatives, which reveals good results. Furthermore, we introduce our open-source API, based on CTranslate2, to serve translations, auto-suggestions, and auto-completions.
}
\end{abstract}

\newpage
\section{Context}

As large language models (LLMs) demonstrate a high level of generation diversity through sampling techniques, this reminds us that encoder-decoder MT models can also employ that autoaggressive feature at the decoder level. The research into interactive MT is not new. It was inspired by well-established techniques such as teacher forcing \citep{Williams1989-TeacherForcing}, where the ground truth previous tokens are fed into the decoder, instead of the predicted tokens y\textsubscript{i-1} as suggested by \citet{Bahdanau2015-JointlyLearn}, and then the model is expected to predict the next words. In this context, \citet{Langlais2000-TransType} proposed a system that watches over the user while typing a translation and repeatedly suggests completions for the text already entered. Later, several researchers studied this feature to improve the interactivity of encoder-decoder MT models. For example, \citet{Peris2017-INMT} proposed segment-based interactive MT; besides correcting a wrong word, the user can validate segments (word sequences) to be kept in future iterations. The system then offers alternative hypotheses that take into account the corrected word together with the validated segments. Such auto-suggestions help to adapt the translation to the desired style and terminology at inference time. Recently, researchers started to look into even prompting encoder-decoder models in a manner similar to prompting decoder-only LLMs \citep{Patel2023-Bidirectional-LM}. In 2022, WMT organised a shared task on MT word-level auto-completion. We found it a good opportunity to explore this direction that employs the autoregressive feature of the decoder of an MT model through random sampling techniques usually used in language modelling. The next sections elaborate on our submitted systems that employed random sampling to generate diverse alternatives at inference time, and achieved excellent results (1st and 2nd places in the shared task) based on both automatic and human evaluation.

\section{Introduction}
\label{sec:introduction-wlac}

Translation auto-suggestion and auto-completion are among the important features that can help translators better utilise MT systems. In a Computer-Aided Translation (CAT) environment, a translator can make use of the MT word auto-suggestion feature as follows:
\begin{itemize}
    \item typing a few words, or clicking a word in a proposed MT translation, a list of suggestions is displayed, as illustrated by Figure \ref{fig:autosuggest}.\footnote{The image is from our demo at: \url{https://www.machinetranslation.io/}}
    \item selecting one of the word suggestions from the list, the rest of the translation is modified accordingly.
\end{itemize}

\begin{figure*}[htp]
    \centering
    \includegraphics[width=15cm]{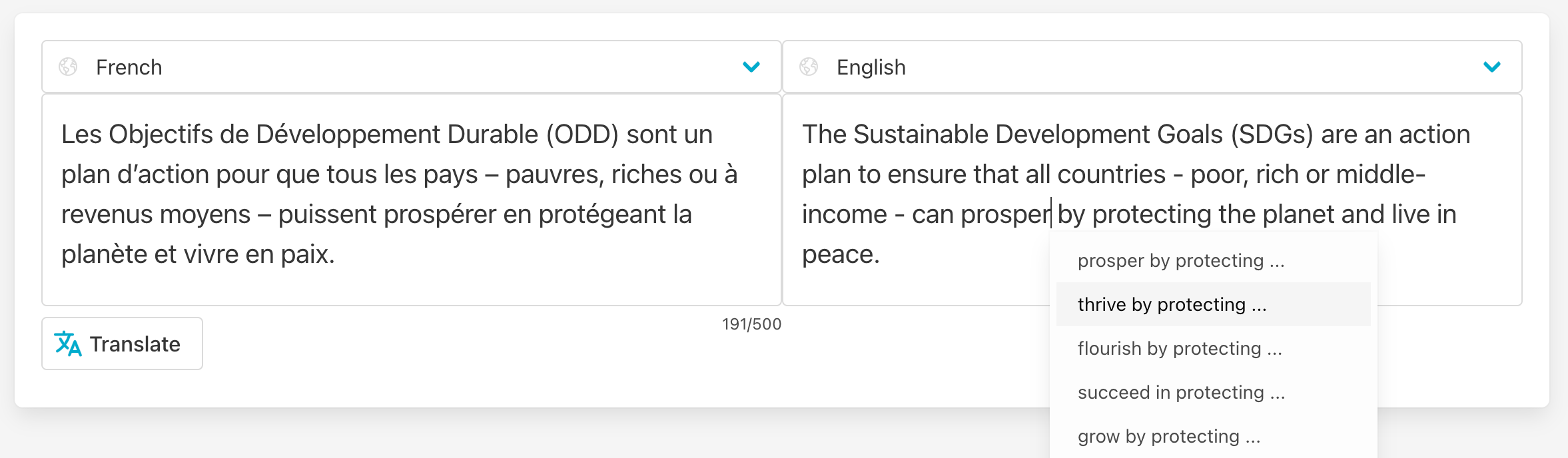}
    \caption{Auto-Suggest: Word Suggestions List}
    \label{fig:autosuggest}
\end{figure*}

The WMT's Word-Level AutoCompletion (WLAC) shared task addresses a more specific scenario, where the user types a few characters, and the system predicts and auto-completes the correct word, given the current context. The WLAC task even suggests that the context might be partial, and it can consist of preceding and/or following words. Given a source sequence $x$, typed character sequence $s$ and a context $c$, WLAC aims to predict a target word $w$ which is to be placed in the middle between the left context c\textsubscript{l} and right context c\textsubscript{r} to constitute a partial translation. Note that the last word of c\textsubscript{l}, the auto-completed word $w$, and the first word of c\textsubscript{r} are not necessary consecutive.

Previous work proposed diverse approaches, mostly to translation sentence-level auto-suggestion and auto-completion. In their work, \citet{Li2021-WLAC} proposed an approach to tackle the word-level auto-completion task. Given a tuple ($x$, $c$, $s$), the system decomposes the word autocompletion process into two parts: 1) model the distribution of the target word $w$ based on the source sequence $x$ and the translation context $c$; and 2) find the most possible word $w$ based on the distribution and human typed sequence $s$. Hence, they first use a single placeholder [MASK] to represent the unknown target word $w$, and use the representation of [MASK] learned from the word prediction model, based on BERT \citep{Devlin2019-BERT}, to predict it. Then, the predicted distribution of the masked token is used over the vocabulary to filter out invalid words, namely those that do not start with the human typed \mbox{sequence $s$}. Finally, they return the token with the highest probability over the new distribution.

Researchers in other natural language processing areas such as language modelling offered approaches to improve predictions of decoder-only autoregressive models, trained to predict the next word given the previous context. Among these approaches are top-K sampling and top-p (nucleus) sampling \citep{Fan2018-zm,Holtzman2018-rf,Radford2019-GPT-2,Holtzman2020-nucleus}. Since NMT inference depends on a decoder model, such approaches from language modelling can be employed. In particular, we investigate the use of top-K sampling during decoding to generate better word-level auto-completions.

\section{User Survey}
\label{sec:survey-wlac}

Previous work reported that a user can save over 60\% of the keystrokes needed to produce a translation in a word completion scenario \citep{Langlais2000-TransType}. Other researchers noted that post-editing is faster than MT auto-completion \citep{Koehn2009-CAT}, while MT auto-completion can yield higher quality translation when the baseline MT quality is high \citep{Green2014-zv}.

In a user survey we designed and distributed via social media networks, we asked participants whether they thought an MT word-level auto-suggestions feature could be helpful, and provided a simple definition and an illustrative image. If their answer was ``yes'', the respondent was asked to specify a reason. By the time of writing this work, we received 41 responses to our survey. While we do not believe this survey is enough to justify introducing an auto-suggestions feature into every MT system, it can be an indicator as to why some users think such a feature could be helpful. To answer the question, ``Which of the following best describes you?'' 46.3\% (19) of the respondents chose ``Translator/Linguist'', 31.7\% (13) selected ``NLP Engineer/Researcher'', and the rest 22\% (9) were other ``MT Users'', not included in the two aforementioned categories.

\begin{figure}[htp]
    \centering
    \includegraphics[width=7.9cm]{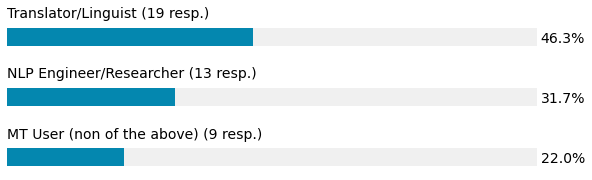}
    \caption{MT user categories}
    \label{fig:autosuggest-categories}
\end{figure}

Among the respondents to the survey, 90.2\% (37) answered ``Yes'' to the question ``In general, do you believe that a word-level auto-suggestions feature is helpful?'' Figure \ref{fig:autosuggest-why} shows the breakdown of answers to the question, ``Why do you believe that a word-level auto-suggestions feature can be helpful?'' taking into consideration those who answered ``No'' to the previous question.

Out of the 37 persons who believed a word-level auto-suggestions feature can be helpful,  40.5\% (15) of the respondents specified that it can give them some inspiration. This answer is specifically interesting as it is not constrained by time-saving benefits; hence, it focusses more on effectiveness rather than efficiency. The respondent that answered with ``Other'' mentioned that it allows them to look for alternative senses or phrasings, especially when they suspect the initial translation is bad, and referred to this as ``human in the loop''.

Respondents were allowed to give extra comments; among the notable comments were:
\begin{itemize}
\begin{normalsize}
    \item \emph{I think word-level suggestions can be a \mbox{useful} feature, particularly when the target language can have several translations of a single source word.}
    \item \emph{Word-level suggestions can be helpful, but sometimes you end up spending a lot of time figuring out if the MT suggestion is a valid translation in that context. So, I'm not really sure yet how I feel about it.}
    \item \emph{It's useful, as long as it's seen as a suggestion, and not inserted in the target where the translator is typing.}
\end{normalsize}
\end{itemize}

Among the respondents who chose ``For me, it is easier or faster than typing'', comments included:
\begin{itemize}
\begin{normalsize}
    \item \emph{Though most of the time; the suggestions are lousy.}
    \item \emph{I don’t think it gives me inspiration as I mostly need it for structures, not single words.}
    \item \emph{Auto-suggestion does not have to come from machine translation. History is much more useful.}
\end{normalsize}
\end{itemize}

The last comment above might be referring to the fact that in some CAT tools, auto-suggestions can also include glossary terms, and translation memory sub-segments, which encourages further research efforts to investigate methods to enhance leveraging and interaction between various translation resources in human-in-the-loop environments.

\bigbreak

\begin{figure}[htp]
\captionsetup{font=footnotesize,labelfont=footnotesize}
    \centering
    \includegraphics[width=7.9cm]{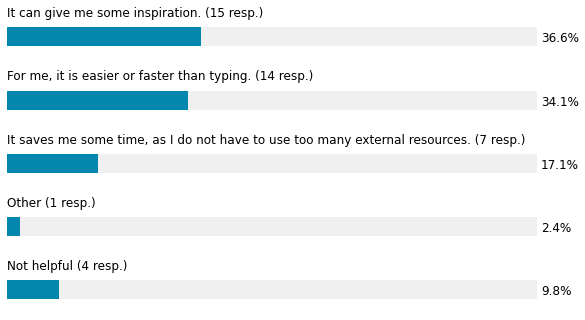}
    \caption{How translators and other MT users perceive word-level auto-suggestions}
    \label{fig:autosuggest-why}
\end{figure}

We hope this survey will inspire future user \mbox{studies} to look deeper into how diverse users of MT and CAT tools prefer to utilise certain features, such as auto-suggestions, and the value they seek. More aspects should be taken into consideration, such as language pairs, translation workflows, and user interfaces. This can help improve these features to better support linguists and other MT users and boost their productivity as well as translation quality.

\vspace{\fill}

\section{Experimental Setup}

\paragraph{Models}
We use OPUS pre-trained models\footnote{\url{https://github.com/Helsinki-NLP/Tatoeba-Challenge}} based on the Transformer architecture \citep{Vaswani2017-attention} for the Chinese-to-English, English-to-Chinese, German-to-English, and English-to-German language directions.

\paragraph{Tokenisers}
OPUS models depend on SentencePiece\footnote{\url{https://github.com/google/sentencepiece}} \citep{Kudo2018-SentencePiece} for tokenisation. Hence, we use their provided subword models in our pre-processing and post-processing processes. As OPUS's English-to-Chinese model requires defining the target dialect using a pre-specified token, we prepend \mbox{[``$>>$cmn\_Hans$<<$'']} to the list of tokens generated by SentencePiece. For word-level tokenisation, we use NLTK for English and German, and Jieba\footnote{\url{https://github.com/fxsjy/jieba}} for Chinese. This list of words can be used later to find the word that starts with the typed sequence.

\paragraph{Inference Engine}
We employ CTranslate2 \citep{Klein2020-OpenNMT} for sentence-level MT, as well as for translation auto-suggestions. To this end, we first convert OPUS models into the \mbox{CTranslate2} format. After that, we utilise a number of CTranslate2 decoding features, including ``alternatives at a position'' and ``auto-completion''.\footnote{\url{https://github.com/OpenNMT/CTranslate2/blob/master/docs/decoding.md}} The translation options $return\_alternatives$ and $num\_hypotheses$ are essential for all our experiments; the former should be set to $True$ while the latter determines the number of returned alternatives. These decoding options can be used with regular beam search, prefix-constrained \mbox{decoding}, and/or random sampling. If the decoding \mbox{option} $return\_alternatives$ is used along with $target\_prefix$, the provided target left \mbox{context} is fed into the decoder in the teacher forcing mode,\footnote{In \emph{teacher forcing} \citep{Williams1989-TeacherForcing}, ground truth previous tokens are fed into the decoder, instead of the predicted tokens y\textsubscript{i-1} as suggested by \citet{Bahdanau2015-JointlyLearn}} then the engine expands the next $N$ most likely words, and continues (auto-completes) the decoding for these $N$ hypotheses independently. The shared task investigates four context cases: \mbox{(a) empty} context, (b) right context only, (c) left context only, and (d) both the right and left contexts are provided. Hence, for all cases we returned multiple alternative translations, while for (c) and (d) we also returned another set of alternative auto-completions using the left context as a target prefix. In this sense, it is worth noting that we make use only of the left context, when available, and we do not use the right context at all, which we might investigate further in the future. To enhance the diversity of translations, especially for (a) and (b), we applied random sampling with the CTranslate2's decoding option $sampling\_topk$, with various sampling temperatures. Our experiments are further elaborated in Section \ref{sec:method} and Section \ref{sec:experiments}.

\paragraph{Pinyin}
The official Romanisation system for Standard Mandarin Chinese is called Pinyin. Since the task organisers used the pypinyin library\footnote{\url{https://github.com/mozillazg/python-pinyin}} to prepare the test files, we did too. OPUS English-to-Chinese models accept Chinese input, so we had to use the library to convert between the two writing systems. Since the conversion from Chinese characters to Pinyin is a lossy process and cannot be perfectly converted back, we keep a list of Chinese words resulted from tokenisation with Jieba to be able to map Pinyin tokens to Chinese tokens later.

\section{Method}
\label{sec:method}

We experimented with both beam search alternatives and random sampling, and found that the latter achieves better results. This could be due to the fact that alternatives generated from each beam are usually very similar, and lower beam values tend to generate translations of lower quality. This section elaborates on the actual methods we used for our submissions, while more details about the initial experiments that led us to these decisions are explained in \mbox{Section \ref{sec:experiments}}.

\begin{table}[ht]
\captionsetup{font=small,labelfont=small}
\centering
\begin{small}
\begin{tabular}{@{}llll@{}}
\toprule
Language               & Settings       & Accuracy    & Human  \\ \midrule
de-en                  & ST=1.0         & 0.61444 & 0.885  \\
                       & ST=1.3         & 0.60924 & 0.8875 \\ \midrule
\multirow{2}{*}{en-de} & ST=1.0         & 0.58942 & 0.6725 \\
                       & ST=1.3         & 0.58494 & 0.655  \\ \midrule
\multirow{4}{*}{zh-en} & ST=1.0 + detok & 0.50411 & 0.8675 \\
                       & ST=1.3 + detok & 0.50260 & 0.8675 \\
                       & ST=1.0         & 0.49348 & 0.86   \\
                       & ST=1.3         & 0.49062 & 0.87   \\ \midrule
\multirow{2}{*}{en-zh} & ST=1.0         & 0.31942 & 0.5775 \\
                       & ST=1.3         & 0.31935 & 0.5725 \\ \bottomrule
\end{tabular}
\end{small}
\caption{Evaluation results on the test datasets. Automatic evaluation uses the ``Accuracy'' metric. ``Human'' refers to human evaluation. Results obtained from sampling temperature (ST) 1.0 are slightly better than those with the value 1.3. When the source is Chinese, detokenisation (detok) resulted in slightly better scores.}
\label{tab:evaluation}
\end{table}

Random sampling is a decoding mode that randomly samples tokens from the model output distribution. In our experiments, we restrict the sampling to the top-10 candidates at each time-step. To obtain diverse generations from the MT model, we rely on randomness in the decoding method, in particular through top-K sampling that samples the next word from the top-K most probable choices \citep{Fan2018-zm, Holtzman2018-rf, Radford2019-GPT-2}, instead of aiming to decode text that maximises likelihood.

For each translation, we use the CTranlsate2 option \emph{return\_alternatives} to return 10 sequences, with 10 top-K sampling. If the entry has a left context starting with a capital letter, we use the prefix to constrain the decoding. In CTranslate2, combining \emph{target\_prefix} with the \emph{return\_alternatives} flag returns alternative sequences just after the prefix. We compose a list of alternatives with and without the prefix, and try to find the word starting with the typed sequence.\footnote{In a prefix-free target sequence, if multiple words start with the typed sequence, we return the first word. In practice, users could be prompted to choose from potential options.} If the word is not found, we repeat the same process for up to five runs. In each new run, random sampling can generate a new set of alternatives. Our experiments show that returning 20 sequences with 20 top-K sampling could lead to more correctly predicted words (cf. Table \ref{tab:experiments}); however, we had to consider the trade-off between quality and efficiency.\footnote{Our scripts are available at: \url{https://github.com/ymoslem/WLAC}}

Furthermore, we investigate increasing the randomness of the generation by using a value for sampling temperature between 1.0 and 1.3. For each run, a random value is generated in this range. The default sampling temperature in CTranslate2 is 1, which achieved relatively better results, as demonstrated in Table \ref{tab:evaluation}.\footnote{To measure the performance of the submitted systems, the organisers chose ``accuracy'' as the automatic evaluation metric and defined it as follows: $ACC$ $=$ $N_{match}$ $/$ $N_{all}$ where $N_{match}$ is the number of correct predicted words and $N_{all}$ is the number of \emph{all} test examples \citep{Casacuberta2022-2022}.}

\section{Other Experiments}
\label{sec:experiments}

This section elaborates on some initial experiments we conducted to decide what approach to use. The final approach we actually used in our submissions is explained in Section \ref{sec:method}.

We used 10,000 entries of a Chinese-to-English golden sample provided by the organisers to evaluate various experiments. For sentence translation, when there is no left context, we experimented with the following values:
\begin{itemize}
    \setlength\itemsep{-2pt}
    \item beam size 1, 5, and 10, without sampling
    \item beam size 1, with random sampling top-K 10, 20, and 50
\end{itemize}

Table \ref{tab:experiments} shows the results for these experiments, and demonstrates that random sampling achieves the best overall accuracy. Random sampling with beam size 1 reveals better results than mere beam size 1 and even beam sizes 5 and 10 without random sampling. Multiple runs of random sampling can result in more correctly predicted words.

\begin{table}[ht]
\captionsetup{font=small,labelfont=small}
\centering
\begin{small}
\begin{tabular}{@{}ccccc@{}}
\toprule
Beam Size & Sampling Top-K & Hypotheses & Accuracy & Runs \\ \midrule
1  & N/A & 10 & 0.6519 & 1 \\
5  & N/A & 10 & 0.6588 & 1 \\
10 & N/A & 10 & 0.6573 & 1 \\ \midrule
1  & 10 & 10  & 0.6918 & 1 \\
1  & 20 & 10  & 0.6907 & 1 \\
\textbf{1} & \textbf{20} & \textbf{20} & \textbf{0.7108} & \textbf{1} \\
1  & 50 & 10  & 0.6853 & 1 \\ \midrule
5  & N/A & 10  & 0.6588 & 5 \\
1  & 10 & 10  & 0.7165 & 5 \\
\textbf{1} & \textbf{20} & \textbf{20} & \textbf{0.7310} & \textbf{5} \\
\bottomrule
\end{tabular}
\end{small}
\caption{Results for the Chinese-to-English golden sample dataset (10,000 entries). Random sampling outperforms even higher beam sizes.}
\label{tab:experiments}
\end{table}

\section{Conclusion}

Random sampling is a decoding mode used for sequence generation. Instead of always selecting the most probable next word or token at each step, the model samples from the probability distribution over the vocabulary. In our experiments, we employed top-K sampling to obtain diverse generations from the MT model. In other words, the next word was sampled from the top-K most probable choices, given the typed context. We also investigated increasing the randomness of the generation using different values of sampling temperature. Our approach led to excellent results based on both automatic and human evaluation, across three language pairs.

\end{large}

\begin{large}

\chapter{Adaptive Machine Translation with Large Language Models}
\label{chapter:adaptive-MT}

\vspace{-1\baselineskip}
{Yasmin Moslem, Rejwanul Haque, John Kelleher, and Andy Way}

\vspace{-12pt}
\singlespacing
\noindent {\footnotesize In Proceedings of the 24th Annual Conference of the European Association for Machine Translation (EAMT 2023), Research: Technical, pages 227–237, Tampere, Finland. Association for Machine Translation in the Americas.\footnote{Published at: \url{https://aclanthology.org/2023.eamt-1.22/}}}
\bigbreak

\onehalfspacing

\begin{abstract}
\nohyphens{
\normalsize
Consistency is a key requirement of high-quality \mbox{translation}. It is especially important to adhere to pre-approved terminology and adapt to corrected translations in domain-specific projects. Machine translation (MT) has achieved significant progress in the area of domain adaptation. However, \mbox{real-time} adaptation remains challenging. Large-scale language models (LLMs) have recently shown interesting capabilities of in-context learning, where they learn to replicate certain input-output text generation patterns, without further fine-tuning. By feeding an LLM at inference time with a prompt that consists of a list of translation pairs, it can then simulate the domain and style characteristics. This work aims to investigate how we can utilise in-context learning to improve real-time adaptive MT. Our extensive experiments show promising results at translation time. For example, LLMs can adapt to a set of in-domain sentence pairs and/or terminology while translating a new sentence. We observe that the translation quality with few-shot in-context learning can surpass that of strong encoder-decoder MT systems, especially for high-resource languages. Moreover, we investigate whether we can combine MT from strong encoder-decoder models with fuzzy matches, which can further improve translation quality, especially for less supported languages. We conduct our experiments across five diverse language pairs, namely English-to-Arabic (EN-AR), English-to-Chinese (EN-ZH), English-to-French (EN-FR), English-to-Kinyarwanda (EN-RW), and English-to-Spanish (EN-ES).
}
\end{abstract}

\section{Context}

As explained in the previous chapters, the release of LLMs such as BLOOM \citep{BLOOM2022} (open-sourced), GPT-\{3.5,4\} \citep{Brown2020-GPT-3,Ouyang2022-InstructGPT} (via a commercial API), and PaLM \citep{Chowdhery2022-PaLM} (limited access) has paved the way for new approaches that utilise their in-context learning capabilities \citep{Dong2022-In-contextLearning}. By the end of 2022, many researchers started investigating the use of LLMs for all NLP areas. While this paper was a natural extension of my previous work, this research was among the earliest to conduct an extensive investigation into the employment of LLMs for MT in general and for adaptive MT in particular, and to compare the performance of a wide range of both open-source and commercial systems, across five language pairs. This paper first appeared as a preprint in January 2022, and then was peer-reviewed and published at EAMT 2022. By the time of writing this thesis, the paper is already well-cited in public literature. The following sections elaborate on the contribution of this work.

\section{Introduction}

Adaptive MT is a type of machine translation that utilises feedback from users to improve the quality of the translations over time. Feedback usually includes corrections to previous translations, terminology and style guides, as well as ratings of the quality of the translations. This can be particularly useful for domain-specific scenarios, where baseline MT systems may have insufficient relevant data to accurately translate certain terms or phrases. There are still several challenges to effectively incorporate user feedback into the translation process, especially at inference time. In this work, we use a relatively wide definition of adaptive MT to refer to learning from similar translations (fuzzy matches) found in approved translation memories (TMs) on the fly  \citep{Farajian2017-AdaptiveMT,Wuebker2018-Personalized,Peris2019-OnlineLearning,Etchegoyhen2021-OnlineLearning}, as well as real-time terminology-constrained MT \citep{Hokamp2017-ConstrainedDecoding,Post2018-FastConstrainedDecoding,Dinu2019-TerminologyConstraintsTraining,Michon2020-Terminology}.

Autoregressive decoder-only LLMs, such as GPT-3 \citep{Brown2020-GPT-3,Ouyang2022-InstructGPT}, BLOOM \citep{BLOOM2022}, PaLM \citep{Chowdhery2022-PaLM}, and Llama \citep{Touvron2023-Llama1,Touvron2023-Llama2} are trained to predict the next word given the previous context. During unsupervised pre-training, a language model develops a broad set of pattern recognition abilities. It then uses these abilities at inference time to rapidly recognise and adapt to the desired task. In their experiments, \citet{Brown2020-GPT-3} use the term ``in-context learning'' to describe a scenario where a pre-trained language model at inference time learns to replicate certain input-output text generation patterns without further fine-tuning. They show that autoregressive LLMs such as GPT-3 can perform well on diverse tasks, through zero-shot, one-shot, and few-shot in-context learning without weight updates. Instead of asking the model to directly perform a given task, the input can be augmented with relevant examples, which help the model adapt its output. The key idea of in-context learning is to learn from analogy. The model is expected to learn the pattern hidden in the demonstration and accordingly make better predictions \citep{Dong2022-In-contextLearning}.

Previous researchers investigated using neural language models for MT through few-shot in-context learning \citep{Vilar2023-PaLM-MT} and even in zero-shot settings \citep{Wang2021-LM4MT}. Other researchers proposed using LLMs for generating synthetic domain-specific data for MT domain adaptation \citep{Moslem2022-MT-LM}. Recently, researchers \citep{Agrawal2023-SelectionMT,Zhang2023-PromptingMT} confirmed the importance of in-context example selection for the quality of MT with LLMs.

\vspace{3ex}

\begin{figure}[htp]
\captionsetup{font=small,labelfont=small}
    \centering
    \includegraphics[width=12cm]{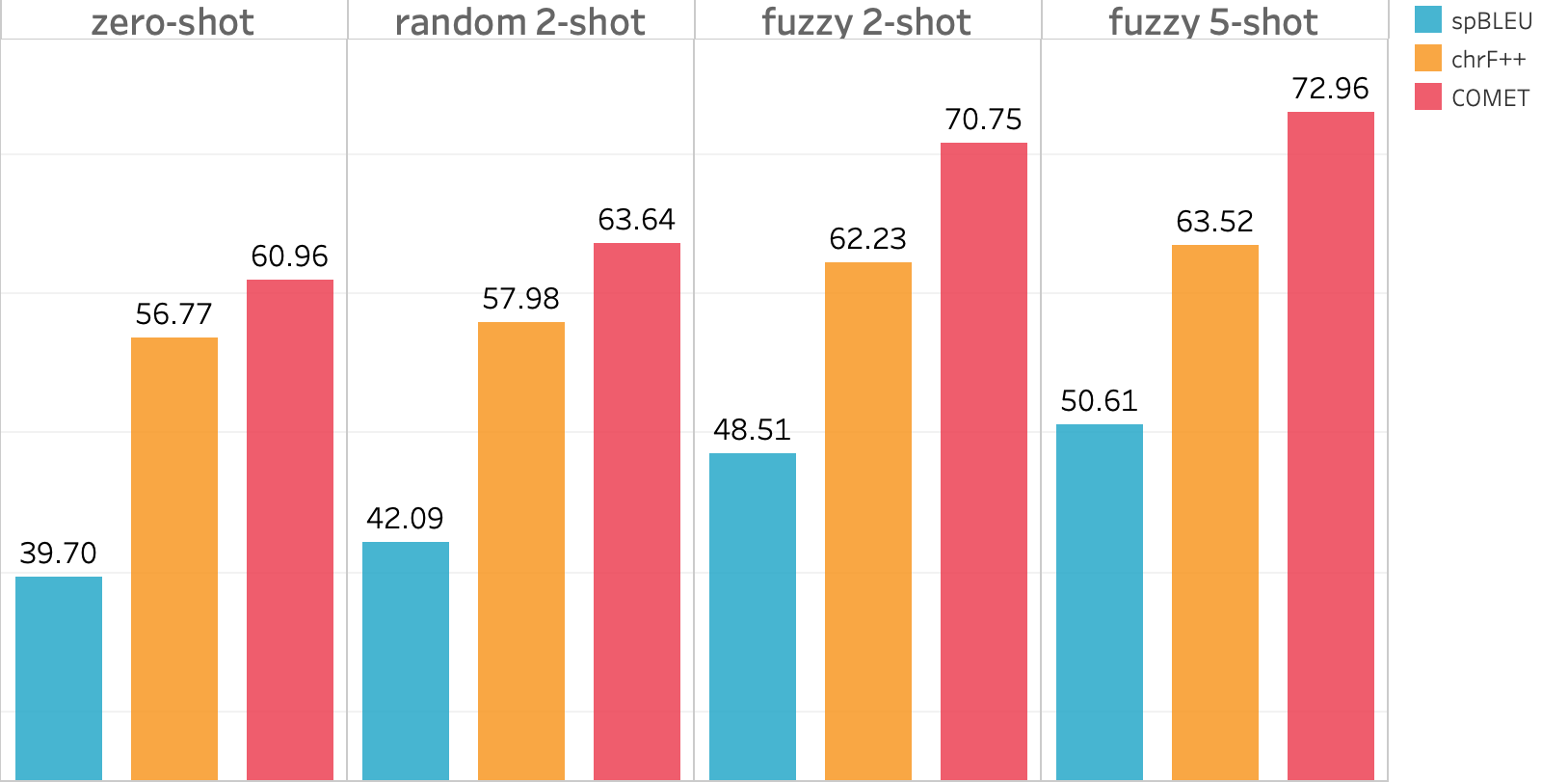}
    \caption{Evaluation results for GPT-3.5 zero-shot, and few-shot translation with random context or fuzzy matches. Average scores across EN-AR, EN-ES, EN-FR, and EN-ZH language pairs. While using a random context outperforms zero-shot translation, using fuzzy matches reveals the best results.}
    \label{fig:context-avg}
\end{figure}

The main contribution of this paper is investigating the capabilities of LLMs such as \mbox{GPT-3.5}, GPT-4 (including ChatGPT), and BLOOM for real-time adaptive MT through in-context learning. As illustrated by Figure \ref{fig:context-avg}, such LLMs can achieve better translation quality through adapting its output to adhere to the terminology and style used in previously approved translation pairs. In particular, we would like to understand the quality with which such models can perform the following tasks, without any further training:

\begin{itemize}
    \setlength\itemsep{-2pt}
    \item Adapting new translations to match the terminology and style of previously approved TM fuzzy matches, at inference time;
    \item Matching or outperforming the quality of translations generated by encoder-decoder MT models across a number of languages; 
    \item Fixing translations from stronger encoder-decoder MT systems using fuzzy matches, which is especially useful for low-resource languages; and
    \item Terminology-constrained MT, by first defining terminology in the relevant sentences or dataset, and then forcing new translations to use these terms.
\end{itemize}

\section{Experimental Setup}
\label{sec:setup}

In all our experiments, we use GPT-3.5 \emph{text-davinci-003} model via its official API.\footnote{\url{https://openai.com/api/}} For parameters, we use \mbox{\emph{top-p} 1}, with \emph{temperature} 0.3 for the three translation tasks, and 0 for the terminology extraction task.\footnote{To avoid over-generation, the option \emph{stop} can be set to [`\textbackslash{n}']. However, if a new line is generated by the model before the translation, this might result in not generating a translation. Alternatively, over-generation can be manually handled.} For the maximum length of tokens, we observe that French and Spanish tokens can be 3–4 times the number of English source words, while other languages can be longer. Hence, we roughly choose a length multiplier value, which we set to 8 for Arabic, 5 for Chinese and Kinyarwanda, and 4 for French and Spanish. We used batch requests with a batch size of 20 segments.\footnote{For higher values of few-shot translation into Arabic using \emph{text-davinci-003}, we had to decrease the batch size to avoid exceeding the tokens-per-minute limit.} Our scripts are publicly \mbox{available.}\footnote{\url{https://github.com/ymoslem/Adaptive-MT-LLM}}

 As we aim to simulate a document-level scenario where translators are required to adhere to a project's or client's TM, we use the domain-specific dataset, TICO-19 \citep{Anastasopoulos2020-TICO-19}, which includes 3070 unique segments. From now on, we will refer to it as the ``context dataset". We focus on a range of languages with diverse scripts and amounts of resources, namely English as the source language, and Arabic, Chinese, French, Kinyarwanda, and Spanish as the \mbox{target} languages.

\begin{table}[hp]
\captionsetup{font=small,labelfont=small}
\centering
\begin{small}
\begin{tabular}{@{}clcccc@{}}
\toprule
\multicolumn{1}{c}{\textbf{Lang}} & \textbf{Context} & \textbf{spBLEU ↑} & \textbf{chrF++ ↑} & \textbf{TER ↓} & \textbf{COMET ↑} \\ \midrule
\multirow{8}{*}{\textbf{EN-AR}} & zero-shot     & 27.6           & 48.36          & 70.6                                   & 41.28    \\
                                & random 2-shot & 28.94          & 49.35          & 70.55          & 43.32          \\
                                & fuzzy 1-shot   & 36.38          & 55.08          & 63.99          & 55.1           \\
                                & fuzzy 2-shot  & 38.41          & 56.57          & 62.31          & 57.36 \\
                                & fuzzy 3-shot  & 39.75          & 57.52 & 61.12          & 59.68          \\
                                & fuzzy 4-shot  & 40.84          & 58.27          & 60.39          & 62.16          \\
                                & fuzzy 5-shot  & 41.33          & 58.64          & 59.95          & 62.65 \\ 
                                & fuzzy 7-shot  & \textbf{41.81} & \textbf{59.1} & \textbf{59.38} & \textbf{64.01} \\ \midrule
\multirow{5}{*}{\textbf{EN-ES}} & zero-shot     & 53.91          & 72.61    & 36.86                                                & 84.0  \\
                                & random 2-shot & 54.78          & 73.12    & 36.09     
                                        & 85.25 \\
                                & fuzzy 2-shot  & 59.64          & 75.83          & 32.56          & 90.37          \\
                                & fuzzy 5-shot  & 61.24          & 76.73          & 31.32          & 91.51          \\
                                & fuzzy 10-shot & \textbf{61.77} & \textbf{77.05} & \textbf{30.9}  & \textbf{92.0}  \\ \midrule
\multirow{8}{*}{\textbf{EN-FR}} & zero-shot     & 44.87          & 65.29          & 50.34                                  & 58.67          \\
                                & random 2-shot & 45.91          & 65.4           & 49.92          & 57.6           \\
                                & fuzzy 1-shot  & 48.39          & 66.58          & 48.18          & 59.49          \\
                                & fuzzy 2-shot  & 49.79          & 67.41          & 46.79          & 61.38          \\
                                & fuzzy 3-shot  & 50.96          & 68.06          & 45.85          & 61.97          \\
                                & fuzzy 4-shot  & 51.89          & 68.5           & 44.94          & 62.7           \\
                                & fuzzy 5-shot  & 51.94          & 68.43          & 45.09          & 62.81          \\
                                & fuzzy 10-shot & \textbf{53.72} & \textbf{69.39} & \textbf{43.82} & \textbf{63.57} \\ \midrule
\multirow{5}{*}{\textbf{EN-RW}} & zero-shot     & 2.82           & 22.53          & 143.12                                         & N/A            \\
                                & random 2-shot & 3.8            & 25.19          & 129.88         & N/A            \\
                                & fuzzy 2-shot  & 12.23          & 36.66          & 105.54         & N/A            \\
                                & fuzzy 5-shot  & 14.96          & 39.84          & 100.11         & N/A            \\
                                & fuzzy 10-shot & \textbf{17.87} & \textbf{41.44} & \textbf{92.84} & N/A            \\ \midrule
\multirow{5}{*}{\textbf{EN-ZH}} & zero-shot  & 32.41    & 40.82          & 99.45                                           & 59.87          \\
                                & random 2-shot & 38.72  & 44.06      & 87.56  &   68.39            \\
                                & fuzzy 2-shot  & 46.18          & 49.12          & 69.0           & 73.9           \\
                                & fuzzy 5-shot  & 47.94          & 50.28          & 64.96          & 74.86          \\
                                & fuzzy 10-shot & \textbf{49.11} & \textbf{51.22} & \textbf{63.14} & \textbf{75.3}  \\ \bottomrule
\end{tabular}
\end{small}
\caption{Adaptive MT with fuzzy matches for GPT-3.5 few-shot in-context learning outperforms using random sentence pairs as context examples. Increasing the number of fuzzy matches can improve the translation quality further. The table shows consistent results for EN-AR, EN-ES, EN-FR, EN-RW, and EN-ZH language pairs.}
\label{tab:fuzzy-context}
\end{table}

\section{Adaptive MT with Fuzzy Matches}
\label{sec:adaptive-MT}

In translation environments, similar approved translated segments are usually referred to as ``fuzzy matches'', and are stored in parallel datasets, known as translation memories (TMs).\footnote{Segments stored in a TM can be smaller than a full sentence (e.g. a title) or larger. However, as most segments in a TM are supposed to be sentence pairs, we use the two words interchangeably throughout the paper.} Researchers have investigated the possibilities of improving MT quality and consistency with fuzzy matches \citep{Knowles2018-Fuzzy,Bulte2019-fuzzy,Xu2020-fuzzy}. Incorporating fuzzy matches into the MT process can help the system generate more accurate translations, and try to ensure adherence to pre-approved terminology and preferred style requirements.

In this set of experiments, we investigate the possibility of forcing the translation of a new sentence pair to adapt to fuzzy matches in the \mbox{context} dataset. To extract fuzzy matches, we use embedding similarity-based retrieval. Previous researchers have shown that approaches that depend on embeddings to retrieve fuzzy matches can outperform those that use Edit Distance \citep{Hosseini2020-DeepMatch,Pham2020-Priming}. To this end, we employ the paraphrase mining module from the Sentence-Transformers library \citep{Reimers2019-SentenceTransformers}. We use the \mbox{\emph{all-MiniLM-L6-v2}} model because of its high accuracy and efficiency.\footnote{\url{https://www.sbert.net/}} For each sentence, we retrieve up to $top\_k$ other sentences. We experiment with diverse values of \mbox{1 to 10} sentence(s) from the \mbox{context} dataset.\footnote{For Arabic, we could only integrate up to 7 matches (not 10 matches) because the tokeniser used by GPT-3.5 generates many more tokens for some Unicode languages, which can easily hit the max length of 4097 tokens. We observe that the issue has been alleviated by newer models.} \mbox{Table} \ref{tab:fuzzy-stats} elaborates on the statistics of fuzzy matches based on their similarity to the new source sentence in 2-shot and 5-shot scenarios.\footnote{While creating prompts, we arrange fuzzy matches in descending order, making higher matches closer to the segment to be translated. We experimented with reversing the order, and there was no significant difference in terms of translation quality.}

\begin{table}[H]
\captionsetup{font=small,labelfont=small}
\centering
\begin{footnotesize}
\begin{tabular}{@{}lllll@{}}
\toprule
\multicolumn{1}{c}{\multirow{2}{*}{\textbf{\begin{tabular}[c]{@{}c@{}}Similarity\\ Score\end{tabular}}}} &
  \multicolumn{4}{c}{\textbf{Segment Statistics}} \\ \cmidrule(l){2-5} 
\multicolumn{1}{c}{} &
  \multicolumn{2}{c|}{\textbf{fuzzy 2-shot}} &
  \multicolumn{2}{c}{\textbf{fuzzy 5-shot}} \\ \midrule
\textgreater{}90\% & 167   & 2.7\%  & 168   & 1.1\%  \\
89-80\%            & 751   & 12.2\% & 1,103 & 7.2\%  \\
79-70\%            & 1,593 & 25.9\% & 3,143 & 20.5\% \\
69-60\%            & 1,825 & 29.7\% & 4,661 & 30.4\% \\
\textless{}60\%    & 1,804 & 29.4\% & 6,275 & 40.9\% \\ \midrule
Total &
  \multicolumn{2}{l}{6,140 = 3,070*2} &
  \multicolumn{2}{l}{15,350 = 3,070*5} \\ \bottomrule
\end{tabular}
\end{footnotesize}
\caption{Numbers and percentages of segments based on their similarity to the new source segment, in the 2-shot and 5-shot experiments using fuzzy matches for in-context learning. The English source is used to calculate similarity across the 5 language pairs.}
\label{tab:fuzzy-stats}
\end{table}

The following illustrations show the difference between zero-shot and few-shot translation prompts. In the zero-shot prompt, only the source sentence and language names are provided, encouraging the model to generate the translation. The few-shot prompt incorporates translation examples to influence the style of the output.

\begin{multicols}{2}
\begin{tcolorbox}[enhanced,attach boxed title to top left={yshift=-3mm,yshifttext=-1mm,xshift=3mm},
  colback=blue!1!white,colframe=cyan!90!black,colbacktitle=cyan!90!black,boxrule=1pt,
  left=3pt, top=0pt,height=100pt, width=180pt,center,
  title=Prompt: EN-AR zero-shot translation,fonttitle=\small,
  boxed title style={size=small,colframe=cyan!90!black} ]
  \begin{scriptsize}
  \textmyfont{\emph{
    \begin{itemize}
    \setlength\itemsep{-1.5ex}
    \item[] English: $<$source\_segment$>$
    \item[] Arabic:
\end{itemize}
}}
\end{scriptsize}
\end{tcolorbox}

\begin{tcolorbox}[enhanced,attach boxed title to top left={yshift=-3mm,yshifttext=-1mm,xshift=3mm},
  colback=blue!1!white,colframe=cyan!90!black,colbacktitle=cyan!90!black,boxrule=1pt,
  left=3pt, top=0pt, height=100pt, width=180pt, center,
  title=Prompt: EN-AR two-shot translation,fonttitle=\small,
  boxed title style={size=small,colframe=cyan!90!black} ]
  \begin{scriptsize}
  \textmyfont{\emph{
    \begin{itemize}
    \setlength\itemsep{-1.5ex}
    \item[] English: $<$source\_fuzzy\_match\textsubscript{2}$>$
    \item[] Arabic: $<$target\_fuzzy\_match\textsubscript{2}$>$
    \item[] English: $<$source\_fuzzy\_match\textsubscript{1}$>$
    \item[] Arabic: $<$target\_fuzzy\_match\textsubscript{1}$>$
    \item[] English: $<$source\_segment$>$
    \item[] Arabic:
\end{itemize}
}}
\end{scriptsize}
\end{tcolorbox}
\end{multicols}

Results illustrated by Figure \ref{fig:context-avg} show that few-shot translation with \mbox{GPT-3.5} using fuzzy matches as context outperforms few-shot translation with random examples, although using random sentence pairs outperforms zero-shot translation. As demonstrated by Table \ref{tab:fuzzy-context}, across five language pairs, adding more fuzzy matches improves translation quality further. At some point, there might be diminishing returns of adding more similar sentences as their similarity score decreases. In other words, increasing the number of fuzzy matches from 2 sentences to 5 or 10 sentences incrementally improves translation quality, but with smaller quality gains.

\section{GPT-3 vs Encoder-Decoder MT Models}
\label{sec:compare-mt}

In this section, we aim to compare evaluation results we obtained from various MT encoder-decoder Transformer-based systems \citep{Vaswani2017-attention} with those from GPT-3.5. To this end, we translate our context dataset with a range of open-source and commercial MT models, including DeepL Translate API,\footnote{DeepL supports French, Spanish and Chinese, but not Arabic and Kinyarwanda.} Google Cloud Translation API, OPUS \citep{Tiedemann2020-Tatoeba},\footnote{We use OPUS models from the Tatoeba-Challenge, specifically the models augmented with back-translation, and trained with Transformer-Big.} and NLLB-200 \citep{NLLB2022}. We converted OPUS and NLLB models to the \mbox{CTranslate2} \citep{Klein2020-Efficient} format with int8 quantisation for efficiency. Inference parameters include \emph{beam\_size 4} and \emph{max\_batch\_size 2024}, on a GPU \emph{A100-SXM4-40GB} (Google Colab Pro). For tokenisation, we used SentencePiece \citep{Kudo2018-SentencePiece} with the source and target sub-word models provided for each OPUS model, and the multilingual model provided by NLLB for tokenisation.\footnote{$flores200\_sacrebleu\_tokenizer\_spm.model$ is used for both tokenisation for NLLB and also for spBLEU \citep{Goyal2022-spBLEU} in sacreBLEU.}

\vspace{2ex}

\begin{figure*}[htp]
\captionsetup{font=small,labelfont=small}
    \centering
    \includegraphics[width=15cm]{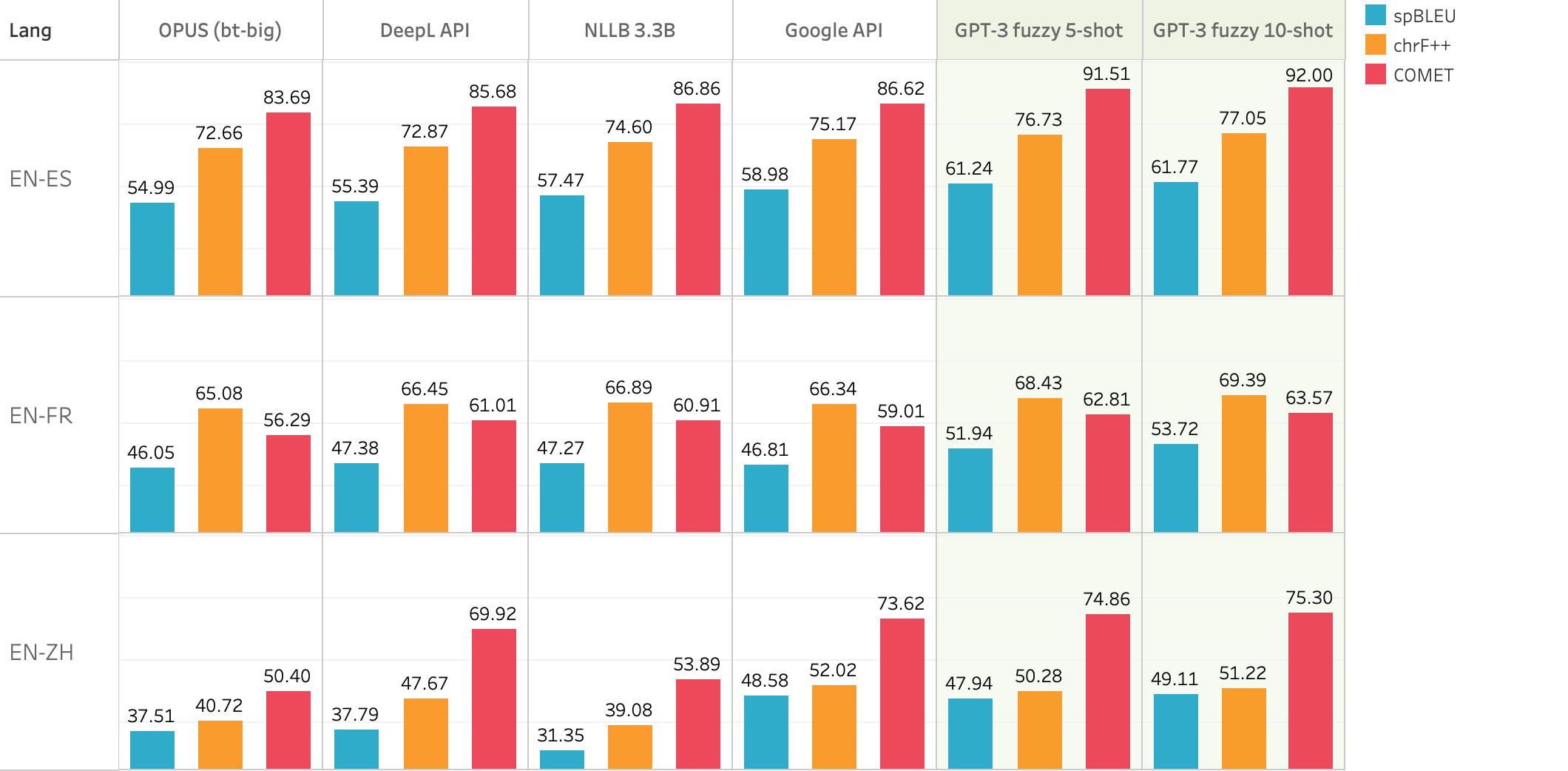}
    \caption{Evaluation results for GPT-3.5 few-shot translation with 5 or 10 fuzzy matches compared to encoder-decoder MT models (DeepL, Google, OPUS, and NLLB). Specifically, for EN-ES, EN-FR, and EN-ZH language pairs, few-shot translation with GPT-3.5 outperforms conventional systems.}
    \label{fig:compare-mt}
\end{figure*}

We observe that for high-resource languages, adaptive MT with fuzzy matches using \mbox{GPT-3.5} few-shot in-context learning (cf. Section \ref{sec:adaptive-MT}) can outperform strong encoder-decoder MT systems. For the English-to-French and English-to-Spanish language pairs, few-shot translation with \mbox{GPT-3.5} incorporating only 5 fuzzy matches outperforms strong encoder-decoder MT models, as demonstrated by Figure \ref{fig:compare-mt}. For English-to-Chinese translation, only when we used 10 fuzzy matches could we achieve better results. However, for English-to-Arabic and English-to-Kinyarwanda translations, results were not on par with the other three language pairs. The results are detailed in Table \ref{tab:compare-mt}.

Among the popular adaptive encoder-decoder MT systems is ModernMT.\footnote{\url{https://www.modernmt.com/}} Originally, the system adopted the instance-based adaptation approach proposed by \citet{Farajian2017-AdaptiveMT}. To control our experiments with ModernMT to match those with \mbox{GPT-3.5} few-shot translation, we created a new TM for each segment to include only the top-10 fuzzy matches for this segment. \mbox{Table}~\ref{tab:compare-mt} illustrates the evaluation results of \mbox{ModernMT} translation with and without a TM. In general, using a TM with ModernMT improves translation quality. Moreover, we observe that zero-shot translation performance (without a TM) of \mbox{ModernMT} outperforms \mbox{GPT-3.5} for the 4 supported language pairs. However, except for English-to-Arabic, few-shot translation with \mbox{GPT-3.5} using either 5 or 10 fuzzy matches outperforms the translation quality of ModernMT using a TM with 10 fuzzy matches per segment, for English-to-Chinese, English-to-French, and English-to-Spanish language pairs.

\section{Incorporating Encoder-Decoder MT}
\label{sec:fix-mt}

As we demonstrated in the previous section, encoder-decoder MT models have achieved high translation quality for several language pairs. \mbox{Nevertheless}, adaptive MT with LLM few-shot in-context learning can surpass such quality, especially for high-resource languages. In this section, we investigate whether we can utilise encoder-decoder MT models to further improve adaptive translation with \mbox{GPT-3.5}. In the next subsections, we study two scenarios:
\vspace*{-1pt}
\begin{itemize}
    \setlength\itemsep{-3pt}
    \item appending fuzzy matches with MT from an encoder-decoder model to enhance in-context learning.\footnote{This scenario can be considered an approach to ``automatic post-editing'' of MT generated by task-oriented models such as Google, OPUS and NLLB.}
    \item translating the source side of fuzzy matches, and using these MT translations for few-shot in-context learning along with the original translations.
\end{itemize}

\subsection{Fuzzy matches + new segment MT}
\label{sec:fuzzy-1-mt}

Incorporating a translation from an encoder-decoder MT model with fuzzy matches, we could achieve substantial improvements over the baseline MT performance. As illustrated by Table \ref{tab:fix-mt}, although OPUS English-to-Arabic translation quality outperforms \mbox{GPT-3.5} few-shot translation with 5 fuzzy matches, appending these fuzzy matches with OPUS translation outperforms both OPUS translation only and \mbox{GPT-3.5} translation with fuzzy matches only. Similarly, adding Google English-to-Chinese translation to 5 fuzzy matches outperforms both baselines. Even for the very low-resource English-to-Kinyarwanda language pair, we relatively notice a similar behaviour, using MT outputs of OPUS or NLLB models.

\enlargethispage{0.5\baselineskip}

However, we observe that if the translation with only fuzzy matches is significantly better than the encoder-decoder MT baseline, we may not achieve further gains. For example, the \mbox{GPT-3.5} translations with 5 fuzzy matches are already much better than the OPUS translation for English-to-French or Google translation for English-to-Spanish. That is why incorporating the MT output from OPUS or Google did not enhance the \mbox{GPT-3.5} translation quality for these language pairs.

\begin{table}[hp]
\captionsetup{font=scriptsize,labelfont=scriptsize}
\centering
\begin{scriptsize}
\begin{tabular}{@{}clcccc@{}}
\toprule
\textbf{Lang}    & \textbf{System}     & \textbf{spBLEU ↑} & \textbf{chrF++ ↑} & \textbf{TER ↓} & \textbf{COMET ↑} \\ \midrule
\multirow{10}{*}{\textbf{EN-AR}} & OPUS (bt-big) & 43.11    & 60.79    & 57.24    & 63.64      \\
                                 & NLLB 600M     & 35.66    & 54.6     & 62.07    & 54.53      \\
                                 & NLLB 1.2B     & 41.1     & 58.51    & 57.15    & 63.85      \\
                                 & NLLB 3.3B     & 43.42    & 60.11    & 55.58    & 66.8 \\
                                 & Google API    & 43.56    & 61.58    & 57.79    & 65.5    \\
                                 & ModernMT (no TM)    & 47.17    & 62.82    & 53.53    & 66.64   \\
                                 & ModernMT (TM) & \textbf{50.33} & \textbf{65.19}    & \textbf{50.19}    & \textbf{71.0}    \\
                                 & GPT-3 zero-shot     & 27.6     & 48.36    & 70.6     & 41.28    \\
                                 & GPT-3 fuzzy 5-shot  & 41.33    & 58.64    & 59.95    & 62.65    \\ 
                                 & GPT-3 fuzzy 7-shot  & 41.81 & 59.1   & 59.38    & 64.01 \\ \midrule
\multirow{10}{*}{\textbf{EN-ES}} & OPUS (bt-big) & 54.99    & 72.66    & 36.26    & 83.69      \\
                                 & NLLB 600M     & 53.31    & 72.19    & 37.13    & 83.09      \\
                                 & NLLB 1.2B     & 56.1     & 73.85    & 34.96    & 85.91      \\
                                 & NLLB 3.3B     & 57.47    & 74.6     & 33.99    & 86.86      \\
                                 & DeepL API     & 55.39    & 72.87    & 36.21    & 85.68      \\
                                 & Google API    & 58.98    & 75.17    & 32.46    & 86.62      \\
                                 & ModernMT (no TM)    & 57.09    &  74.2    &  34.27   &  85.53  \\
                                 & ModernMT (TM)       &  59.22   &  75.4    &  32.79   &  86.99  \\
                                 & GPT-3 zero-shot     & 53.91    & 72.61    & 36.86    & 84.0     \\
                                 & GPT-3 fuzzy 5-shot  & 61.24    & 76.73    & 31.32    & 91.51    \\
                                 & GPT-3 fuzzy 10-shot & \textbf{61.77} & \textbf{77.05} &
                                 \textbf{30.9}  & \textbf{92.0}   \\ \midrule
\multirow{11}{*}{\textbf{EN-FR}} & OPUS (bt-big)  & 46.05    & 65.08    & 49.8     & 56.29          \\
                                 & NLLB 600M     & 43.25    & 64.17    & 51.28    & 56.16          \\
                                 & NLLB 1.2B     & 46.3     & 66.25    & 48.68    & 59.76          \\
                                 & NLLB 3.3B     & 47.27    & 66.89    & 48.19    & 60.91          \\
                                 & DeepL API     & 47.38    & 66.45    & 48.47    & 61.01          \\
                                 & Google API    & 46.81    & 66.34    & 47.01    & 59.01          \\
                                 & ModernMT (no TM)  &   47.17  &   66.28  &   47.91  &   58.46    \\
                                 & ModernMT (TM)     &   49.24  &   67.41  &   46.17  &   59.84    \\
                                 & GPT-3 zero-shot     & 44.87    & 65.29    & 50.34    & 58.67    \\
                                 & GPT-3 fuzzy 5-shot  & 51.94    & 68.43    & 45.09    & 62.81    \\
                                 & GPT-3 fuzzy 10-shot & \textbf{53.72}    & \textbf{69.39}    & \textbf{43.82} & \textbf{63.57}   \\ \midrule
\multirow{8}{*}{\textbf{EN-RW}} & OPUS (Tatoeba 2021) & 1.38     & 15.32    & 153.58    & N/A      \\
                                 & OPUS (2020)   & 5.58     & 27.05    & 101.25   & N/A            \\
                                 & NLLB 600M     & 19.46    & 47.61    & 80.01    & N/A            \\
                                 & NLLB 1.2B     & 23.6     & 50.73    & 74.53    & N/A            \\
                                 & NLLB 3.3B     & \textbf{25.17} & \textbf{52.59} & \textbf{73.06} & N/A  \\
                                 & Google API    & 20.63    & 48.37    & 73.54    & N/A     \\
                                 & GPT-3 zero-shot     & 2.82     & 22.53      & 143.12    & N/A      \\
                                 & GPT-3 fuzzy 5-shot  & 14.96    & 39.84    & 100.11      & N/A     \\      
                                 & GPT-3 fuzzy 10-shot & 17.87    & 41.44    & 92.84 & N/A   \\ \midrule
\multirow{9}{*}{\textbf{EN-ZH}}  & OPUS (bt-big) & 37.51    & 40.72    & 121.49      & 50.4        \\
                                 & NLLB 600M     & 24.9     & 33.87    & 109.37      & 39.28       \\
                                 & NLLB 1.2B     & 29.02    & 37.45    & 110.22      & 50.05       \\
                                 & NLLB 3.3B     & 31.35    & 39.08    & 109.52      & 53.89       \\
                                 & DeepL API     & 37.79    & 47.67    & 100.83      & 69.92       \\
                                 & Google API    & 48.58    & \textbf{52.02} & 70.87    & 73.62    \\
                                 & ModernMT (no TM)    &  37.61   &  48.46   &  102.18  &  67.45   \\
                                 & ModernMT (TM)       &  39.85   &  50.95   &  101.53  &  69.64   \\
                                 & GPT-3 zero-shot     & 32.41    & 40.82    & 99.45    & 59.87    \\
                                 & GPT-3 fuzzy 5-shot  & 47.94    & 50.28    & 64.96    & 74.86    \\
                                 & GPT-3 fuzzy 10-shot & \textbf{49.11} & 51.22    & \textbf{63.14} & \textbf{75.3} \\ \bottomrule
\end{tabular}
\end{scriptsize}
\caption{Comparing GPT-3.5 few-shot translation using fuzzy matches with encoder-decoder MT systems, DeepL Translate API, Google Cloud Translation API, OPUS (Tatoeba-Challenge, with back-translation and Transformer-Big), and NLLB-200 (600M, 1.2B \& 3.3B parameters).}
\label{tab:compare-mt}
\end{table}

\subsection{Fuzzy matches + all segments MT}
\label{sec:fuzzy-all-mt}

In Section \ref{sec:fuzzy-1-mt}, we added MT of the new segment from an encoder-decoder model to fuzzy matches, which enhanced \mbox{GPT-3.5} in-context learning. In this experiment, we include MT for all fuzzy matches and also for the new source segment to be translated. For the English-to-Kinyarwanda and English-to-Spanish language pairs, it is not clear whether including MT for all in-context examples can significantly outperform including MT for only the new source segment to be translated. Again, this depends on the quality of the original MT and requires further investigation.

\section{Bilingual Terminology Extraction}
\label{sec:term-extract}

Terminology extraction is the task of automatically defining domain-specific terms in a dataset. Extracted terms are naturally used for building glossaries to help translators. Furthermore, it is possible to improve MT performance through finding sentences that include these terms and fine-tuning the system with them \citep{Hu2019-LexiconInduction,Haque2020-Terminology}. 

In this set of experiments, we ask \mbox{GPT-3.5} to extract 5 bilingual terms from each sentence pair in the context dataset. For parameters, we use temperature 0 and $top\_p$ 1.

\begin{table}[H]
\captionsetup{font=small,labelfont=small}
\centering
\begin{footnotesize}
\begin{tabular}{@{}ccccc@{}}
\toprule
\textbf{Lang}  & \textbf{Sentences} & \textbf{Terms} & \textbf{Correct} & \textbf{\%} \\ \midrule
\textbf{EN-AR} & 500                & 2,500          & 2,427            & 97.08         \\
\textbf{EN-ES} & 500                & 2,500          & 2,397            & 95.88          \\
\textbf{EN-FR} & 500                & 2,500          & 2,382            & 95.28           \\ \bottomrule
\end{tabular}
\end{footnotesize}
\caption{Human evaluation results for the terminology extraction task for English-to-Arabic (EN-AR), English-to-Spanish (EN-ES), and English-to-French (EN-FR) language pairs. The majority of the terms that GPT-3 extracted ($>$\,95\%) were accurate.}
\label{tab:term-extract}
\end{table}

Human evaluation was performed for Arabic, French,\footnote{We observe that the original English-to-French TICO-19 dataset includes several misaligned translation pairs. This can negatively affect the quality of tasks using such sentences. That is why it is important to filter parallel datasets to remove possible misalignments. The evaluation sample has been manually refined to include only well-aligned translation pairs. Automatic semantic filtering approaches can be applied to large datasets.} and Spanish. We provided the evaluators with a random sample of 500 sentences and their extracted terms. They were asked to use a 0-1 scale to determine whether each source and target term were equivalent, and whether the extracted terms were actually in the sentence pair (relevant inflexions are acceptable). In several cases where the evaluators marked the extracted term pair with 0, the model had made up either the source, target, or both; although it might be correct, it was not in the provided sentence pair. In other cases, the extracted term was partial, sometimes due to reaching the maximum length of tokens. Nevertheless, as Table \ref{tab:term-extract} illustrates, the majority of the terms in the provided sample were accurately extracted by the model.

\begin{table*}[hp]
\captionsetup{font=small,labelfont=small}
\centering
\begin{small}
\begin{tabular}{@{}clcccc@{}}
\toprule
\textbf{Lang} & \multicolumn{1}{c}{\textbf{System}} & \textbf{spBLEU ↑} & \textbf{chrF++ ↑} & \textbf{TER ↓} & \textbf{COMET ↑} \\ \midrule
\multirow{3}{*}{\textbf{EN-AR}} & MT (OPUS)                     & 43.11          & 60.79          & 57.24                                     & 63.64          \\
                                & GPT-3 fuzzy 5-shot            & 41.33          & 58.64          & 59.95     & 62.65          \\
                                & GPT-3 fuzzy 5-shot + 1-MT     & \textbf{45.9}  & \textbf{62.9}  & \textbf{55.14} & \textbf{67.74}   \\ \midrule
\multirow{7}{*}{\textbf{EN-ES}} & MT (Google)                   & 58.98          & 75.17          & 32.46                                     & 86.62          \\
                                & GPT-3 fuzzy 2-shot            & 59.64          & 75.83          & 32.56     & 90.37          \\
                                & GPT-3 fuzzy 2-shot + 1-MT     & 59.82          & 75.73          & \textbf{32.16} & 89.0           \\
                                & GPT-3 fuzzy 2-shot + all-MT   & \textbf{60.2}  & \textbf{76.06} & 32.32     & \textbf{92.0}  \\ \cmidrule(l){2-6} 
                                & GPT-3 fuzzy 5-shot            & \textbf{61.24} & \textbf{76.73} & \textbf{31.32} & 91.51          \\
                                & GPT-3 fuzzy 5-shot + 1-MT     & 60.49          & 76.16          & 31.49     & 89.55          \\
                                & GPT-3 fuzzy 5-shot + all-MT   & 61.1           & 76.52          & 31.8      & \textbf{92.07} \\ \midrule
\multirow{3}{*}{\textbf{EN-FR}} & MT (OPUS)                     & 46.05          & 65.08          & 49.8                                      & 56.29          \\
                                & GPT-3 fuzzy 5-shot            & \textbf{51.94} & \textbf{68.43} & \textbf{45.09} & \textbf{62.81} \\
                                & GPT-3 fuzzy 5-shot + 1-MT     & 47.95          & 66.72          & 48.34     & 59.69          \\ \midrule
\multirow{7}{*}{\textbf{EN-RW}} & MT \#1 (Google) & 20.63 & 48.37  & 73.54  & N/A     \\
                                & GPT-3 fuzzy 5-shot            & 14.96          & 39.84          & 100.11    & N/A            \\
                                & GPT-3 fuzzy 5-shot + 1-MT \#1 & 22.51 & \textbf{49.69} & \textbf{72.97} & N/A   \\ 
                                & GPT-3 fuzzy 5-shot + all-MT \#1 & \textbf{25.01} & 49.43 & 74.75 & N/A  \\
                                \cmidrule(l){2-6}
                                & MT \#2 (NLLB 3.3B)            & 25.17          & 52.59          & 73.06     & N/A            \\
                                & GPT-3 fuzzy 5-shot + 1-MT \#2 & 25.59 & 53.12 & \textbf{72.73} & N/A     \\
                                & GPT-3 fuzzy 5-shot + all-MT \#2 & \textbf{27.52} & \textbf{53.23} & 73.79 & N/A  \\ \midrule
\multirow{3}{*}{\textbf{EN-ZH}} & MT (Google)                   & 48.58          & 52.02          & 70.87                                     & 73.62          \\
                                & GPT-3 fuzzy 5-shot            & 47.94          & 50.28          & \textbf{64.96} & \textbf{74.86}          \\
                                & GPT-3 fuzzy 5-shot + 1-MT     & \textbf{49.45} & \textbf{52.4}  & 67.81     & 74.61 \\ \bottomrule
\end{tabular}
\end{small}

\caption{Combining fuzzy matches with high-quality MT from encoder-decoder systems can improve translation quality with GPT-3.5 few-shot in-context learning, especially for low-resource and medium-resource languages. 1-MT refers to appending fuzzy matches with the MT of the segment to be translated, while all-MT refers to additionally adding MT for each segment of the fuzzy matches along with its approved translation. For EN-AR and EN-RW improvements are clearer than for EN-ES, EN-FR and EN-ZH, potentially due to the limited support of EN-AR and EN-RW by GPT-3.5, which made them benefit more from incorporating MT from stronger encoder-decoder models.}
\label{tab:fix-mt}
\end{table*}


\section{Terminology-Constrained MT}
\label{sec:term-constrain}

As observed in Section \ref{sec:adaptive-MT}, adding more fuzzy matches enhances in-context learning and hence improves translation quality. However, early in a real-world translation project, we might not have so many fuzzy matches. By incorporating domain-specific terminology, the system can produce translations that are more accurate and consistent with the terminology used in that field. In this section, we investigate integrating terms in the process when there are $N$ fuzzy matches. For example, if we have only two fuzzy matches, we either extract terms from these similar sentences or from a \mbox{glossary}, and use those that match up to 5-gram phrases in the source sentence to be translated. In this work, we use the terminology extraction process elaborated in Section \ref{sec:term-extract}. Obviously, if a pre-approved glossary is available, it can be used instead. We investigate three scenarios:

\vspace*{-1.5ex}
\begin{itemize}
    \setlength\itemsep{-2pt}
    \item Few-shot translation with 2 fuzzy matches and their terms. As we do not have terms for the segment to be translated, we use terms from the 2 fuzzy matches if they are found in a set of n-grams (1-5) of the source segment to be translated. Integrating terms into two-shot prediction, i.e. using both terms and two fuzzy matches for in-context learning, \mbox{outperforms} using fuzzy matches only.

    \item We automatically compile a glossary including all terms from the dataset, with 2+ frequency, and up to 5-grams. If there are multiple targets for the same source, the term pair with the highest frequency is selected. Stop words and terms with empty source or target sides are excluded. The list is sorted by n-gram length, so terms with longer n-grams are prioritised.  As illustrated by Table \ref{tab:term-constrain}, integrating terms from a glossary outperforms adding terms from only two fuzzy matches, most likely due to the diversity that this option offers. In prompts (cf. Section \ref{a-prompts}), we use terms found in a set of n-grams (1-5) of the source segment to be translated. We experiment with adding maximum 5 terms and maximum 10 terms, which does not show a huge difference in performance; in some cases only a smaller number of terms is available in the glossary.

    \item Zero-shot translation, i.e. without any fuzzy matches. This is similar to the previous scenario, except that we only use terms from the glossary. In zero-shot prediction, adding terms from the glossary improves translation quality. As shown in Table \ref{tab:term-constrain}, improvements are significant across all 5 language pairs.
\end{itemize}

\begin{table*}[hp]
\captionsetup{font=small,labelfont=small}
\centering
\begin{small}
\begin{tabular}{@{}clcccc@{}}
\toprule
\textbf{Lang} & \textbf{GPT-3.5 Context}                                            & \textbf{spBLEU ↑} & \textbf{chrF++ ↑} & \textbf{TER ↓} & \textbf{COMET ↑} \\ \midrule
\multirow{6}{*}{\textbf{EN-AR}} & zero-shot     & 27.6                   & 48.36          &       70.6     &                                  41.28          \\
                                & zero-shot + max 5 terms (glossary)     & \textbf{35.38} & \textbf{54.53} & \textbf{65.36}  & \textbf{54.91} \\ \cmidrule(l){2-6} 
                                & fuzzy 2-shot                           & 38.41          & 56.57          & 62.31           & 57.36          \\
                                & fuzzy 2-shot + terms (fuzzy)           & 39.38          & 57.22          & 62.01           & 59.36          \\
                                & fuzzy 2-shot + max 5 terms (glossary)  & 41.27          & 58.84          & 60.09           & 62.17          \\
              & fuzzy 2-shot + max 10 terms (glossary)                     & \textbf{41.95}    & \textbf{59.34}    & \textbf{59.45} & \textbf{62.48}   \\ \midrule
\multirow{6}{*}{\textbf{EN-ES}} & zero-shot     & 53.91                  & 72.61          & 36.86          &                                  84.0           \\
                                & zero-shot + max 5 terms (glossary)     & \textbf{55.99} & \textbf{74.18} & \textbf{35.3}   & \textbf{87.21} \\ \cmidrule(l){2-6} 
                                & fuzzy 2-shot                           & 59.64          & 75.83          & 32.56           & 90.37          \\
                                & fuzzy 2-shot + terms (fuzzy)           & 59.66          & 75.91          & 32.53           & 90.04          \\
                                & fuzzy 2-shot + max 5 terms (glossary)  & 60.5           & 76.55          & \textbf{31.93}  & \textbf{91.05} \\
                                & fuzzy 2-shot + max 10 terms (glossary) & \textbf{60.54} & \textbf{76.58} & 32.02           & \textbf{91.05} \\ \midrule
\multirow{7}{*}{\textbf{EN-FR}} & zero-shot                              & 44.87          & 65.29          & 50.34           & 58.67          \\
                                & zero-shot + max 5 terms (glossary)     & \textbf{45.94} & \textbf{66.01} & \textbf{49.22}  & \textbf{59.78} \\ \cmidrule(l){2-6} 
                                & fuzzy 2-shot                           & 49.79          & 67.41          & 46.79           & 61.38          \\
                                & fuzzy 2-shot + terms (fuzzy)           & \textbf{50.58}          & \textbf{67.93} & 45.81           & 62.04          \\
                                & fuzzy 2-shot + max 3 terms (glossary)  & 50.46          & 67.69          & 46.22           & \textbf{68.94} \\
                                & fuzzy 2-shot + max 5 terms (glossary)  & 50.55 & 67.78         & \textbf{46.19}  & 60.24           \\
                                & fuzzy 2-shot + max 10 terms (glossary) & 49.64          & 66.86          & 47.34           & 58.57          \\ \midrule
\multirow{7}{*}{\textbf{EN-RW}} & zero-shot     & 2.82                   & 22.53          & 143.12         &                                  N/A            \\
                                & zero-shot + max 5 terms (glossary)     & \textbf{7.26}  & \textbf{30.83} & \textbf{115.44} & N/A            \\ \cmidrule(l){2-6} 
                                & fuzzy 2-shot                           & 12.23          & 36.66          & 105.54          & N/A            \\
                                & fuzzy 2-shot + terms (fuzzy)           & 12.43          & 36.48          & 102.22          & N/A            \\
                                & fuzzy 2-shot + max 5 terms (glossary)  & 15.34          & 39.96          & 96.09           & N/A            \\
                                & fuzzy 2-shot + max 10 terms (glossary) & \textbf{15.49}    & \textbf{40.53}    & \textbf{96.0}  & N/A          \\
                            \midrule
\multirow{7}{*}{\textbf{EN-ZH}} & zero-shot                              & 32.41          & 40.82          & 99.45           & 59.87          \\
                                & zero-shot + max 5 terms (glossary)     & 36.31          & 44.72          & 96.45           & 68.6           \\
              & \multicolumn{1}{l}{zero-shot + max 10 terms (glossary)}    & \textbf{36.64}    & \textbf{45.06}    & \textbf{96.24} & \textbf{68.94}   \\ \cmidrule(l){2-6} 
                                & fuzzy 2-shot                           & 46.18          & 49.12          & 69.0            & \textbf{73.9}  \\
                                & fuzzy 2-shot + terms (fuzzy)           & 46.16          & 49.11          & \textbf{68.79}  & 73.41          \\
                                & fuzzy 2-shot + max 5 terms (glossary)  & \textbf{46.6}  & \textbf{49.51} & 69.46           & 73.88          \\
                                & fuzzy 2-shot + max 10 terms (glossary) & 46.31          & 49.25          & 69.39           & 73.57          \\ \bottomrule
\end{tabular}
\end{small}
\caption{Terminology-constrained MT with GPT 3.5 outperforms both zero-shot and 2-shot translation with fuzzy matches, although gains are much higher for zero-shot translation. For zero-shot translation, we experimented with adding terms from a glossary. For 2-shot translation with fuzzy matches, we compared adding terms from these 2 fuzzy matches to adding terms from a glossary. The latter revealed better results.}
\label{tab:term-constrain}
\end{table*}

We conducted human evaluation for English-to-Arabic, English-to-French, and English-to-Spanish terminology-constrained MT, to see to what extent the model adheres to the required terms, and how this affects the overall translation quality. The evaluators are professional linguists in the respective languages. We provided the evaluators with 4 sets of 100 randomly selected sentence pairs (zero-shot, zero-shot with glossary terms, fuzzy two-shot, and fuzzy two-shot with glossary terms). They were asked to evaluate the sentence-level translation quality on a 1-4 scale \citep{Coughlin2003-MT-eval} and the usage of each provided term in the translation on a 0-1 scale, as elaborated by \mbox{Table \ref{tab:human-eval}.}

\begin{table}[H]
\captionsetup{font=small,labelfont=small}
\centering
\begin{small}
\begin{tabular}{@{}clcc@{}}
\toprule
\multicolumn{1}{l}{\textbf{\textbf{Lang}}} & \multicolumn{1}{c}{\textbf{GPT-3 Context}} & \textbf{\textbf{Human Eval. ↑}} & \textbf{\textbf{Terms ↑}} \\ \midrule
\multirow{5}{*}{\textbf{EN-AR}} & Zero-shot              & 2.80          & 0.67          \\
                                & Zero-shot + glossary terms      & \textbf{3.19} & \textbf{0.94} \\ \cmidrule(l){2-4}
                                & Fuzzy two-shot         & 2.89          & 0.80          \\
                                & Fuzzy two-shot + glossary terms & \textbf{3.03} & \textbf{0.94} \\ \midrule
\multirow{4}{*}{\textbf{EN-ES}} & Zero-shot              & 3.76          & 0.87          \\
                                & Zero-shot + glossary terms      & \textbf{3.93} & \textbf{0.96} \\ \cmidrule(l){2-4} 
                                & Fuzzy two-shot         & 3.77          & 0.89          \\
                                & Fuzzy two-shot + glossary terms & \textbf{3.84} & \textbf{0.97} \\ \midrule
\multirow{4}{*}{\textbf{EN-FR}} & Zero-shot              & 3.55          & 0.89          \\
                                & Zero-shot + glossary terms      & \textbf{3.64} & \textbf{0.97} \\ \cmidrule(l){2-4} 
                                & Fuzzy two-shot         & 3.50          & 0.91          \\
                                & Fuzzy two-shot + glossary terms & \textbf{3.55} & \textbf{0.92} \\ \bottomrule
\end{tabular}
\end{small}
\caption{Human evaluation of terminology-constrained MT, for EN-AR, EN-ES, and EN-FR. The results cover zero-shot and two-shot translation without and with (maximum 5) glossary terms. The column ``Human Eval." refers to the average evaluation score on a 1-4 scale. The column ``Terms" refers to the average number of terms that the model has successfully transferred into the translation on a 0-1 scale.}
\label{tab:human-eval}
\end{table}

According to the evaluators, for Arabic, French and Spanish, terminology-constrained MT successfully transferred the provided glossary terms into the target more often than zero-shot and few-shot translation without terminology incorporation. In several cases, forcing glossary terms to be used could help improve the overall translation quality; however, sometimes it was detrimental to grammatical accuracy. Although we provided the model with longer terms before shorter ones, contradictory terms can hurt translation quality. Hence, it might be better to exclude shorter terms if they overlap with longer ones.\footnote{For example, ``New York Times" can be transferred without translation into the target, while ``New York" might be translated. If the model is provided with both terms while it is actually supposed to use the former, this can cause confusion.} In production workflows, linguists can be provided with translation alternatives with and without fuzzy matches and/or terminology to be able to use the best translation. Alternatively, automatic quality estimation can be conducted to select the best translation.

Among interesting observations that human evaluation reveals is that in few-shot translation with fuzzy matches (even \emph{without} terms), the number of successfully used terms is more than those in zero-shot translation. This can help enhance \mbox{consistency} with approved translations. Moreover, incorporating glossary terms in a zero-shot prompt can result in quality gains comparable to those of few-shot translation with fuzzy matches.

\section{ChatGPT}

At the time of writing this paper, OpenAI has released new conversational models, publicly \mbox{referred} to as ChatGPT. This range of models includes: GPT-3.5 Turbo and GPT-4. In this section, we briefly investigate the translation capabilities of these models compared to GPT-3.5 Davinci. Generally, we observe that both of the new models solve some tokenisation issues, especially for non-Latin languages such as \mbox{Arabic}. While \emph{gpt-3.5-turbo} is more efficient than \emph{text-davinci-003}, it shows comparable quality for both zero-shot and few-shot translation (with fuzzy matches). The newest model \emph{gpt-4} provides better zero-shot translation quality, while the quality of few-shot translation is relatively similar to that of the two other models. Table \ref{tab:chatgpt} demonstrates the results.

\vspace{\fill}

\vspace*{-4pt}
\begin{table}[hp]
\captionsetup{font=small,labelfont=small}
\centering
\begin{small}
\begin{tabular}{@{}ccccccc@{}}
\toprule
\textbf{Lang} &
  \textbf{Model} &
  \textbf{Context} &
  \textbf{spBLEU ↑} &
  \textbf{chrF++ ↑} &
  \textbf{TER ↓} &
  \textbf{COMET ↑} \\ \midrule
\multirow{6}{*}{\textbf{EN-AR}} &
  GPT-3.5 Davinci &
  \multirow{3}{*}{0-shot} &
  27.6 &
  48.36 &
  70.6 &
  41.28 \\
 &
  GPT-3.5 Turbo &
   &
  38.06 &
  56.35 &
  61.34 &
  62.68 \\
 &
  GPT-4 &
   &
  \textbf{40.29} &
  \textbf{57.86} &
  \textbf{59.55} &
  \textbf{64.25} \\ \cmidrule(l){2-7} 
 &
  GPT-3.5 Davinci &
  \multirow{3}{*}{2-shot} &
  38.41 &
  56.57 &
  62.31 &
  57.36 \\
 &
  GPT-3.5 Turbo &
   &
  46.04 &
  62.18 &
  55.03 &
  73.35 \\
 &
  GPT-4 &
   &
  \textbf{47.52} &
  \textbf{63.28} &
  \textbf{53.04} &
  \textbf{73.7} \\ \midrule
\multirow{6}{*}{\textbf{EN-ES}} &
  GPT-3.5 Davinci &
  \multirow{3}{*}{0-shot} &
  53.91 &
  72.61 &
  36.86 &
  84.0 \\
 &
  GPT-3.5 Turbo &
   &
  52.91 &
  70.87 &
  38.86 &
  82.28 \\
 &
  GPT-4 &
   &
  \textbf{56.93} &
  \textbf{74.41} &
  \textbf{34.35} &
  \textbf{87.89} \\ \cmidrule(l){2-7} 
 &
  GPT-3.5 Davinci &
  \multirow{3}{*}{2-shot} &
  59.64 &
  75.83 &
  32.56 &
  90.37 \\
 &
  GPT-3.5 Turbo &
   &
  \textbf{60.35} &
  \textbf{76.51} &
  32.05 &
  91.57 \\
 &
  GPT-4 &
   &
  60.16 &
  \textbf{76.51} &
  \textbf{31.77} &
  \textbf{91.86} \\ \midrule
\multirow{6}{*}{\textbf{EN-FR}} &
  GPT-3.5 Davinci &
  \multirow{3}{*}{0-shot} &
  44.87 &
  65.29 &
  50.34 &
  58.67 \\
 &
  GPT-3.5 Turbo &
   &
  46.85 &
  66.75 &
  48.31 &
  61.34 \\
 &
  GPT-4 &
   &
  \textbf{47.39} &
  \textbf{67.14} &
  \textbf{48.03} &
  \textbf{61.93} \\ \cmidrule(l){2-7} 
 &
  GPT-3.5 Davinci &
  \multirow{3}{*}{2-shot} &
  49.79 &
  67.41 &
  46.79 &
  61.38 \\
 &
  GPT-3.5 Turbo &
   &
  \textbf{49.88} &
  68.33 &
  46.27 &
  63.62 \\
 &
  GPT-4 &
   &
  49.75 &
  \textbf{68.38} &
  \textbf{45.97} &
  \textbf{64.04} \\ \midrule
\multicolumn{1}{l}{\multirow{6}{*}{\textbf{EN-RW}}} &
  GPT-3.5 Davinci &
  \multirow{3}{*}{0-shot} &
  2.82 &
  22.53 &
  143.12 &
  N/A \\
\multicolumn{1}{l}{} &
  GPT-3.5 Turbo &
   &
  5.31 &
  29.77 &
  114.34 &
  N/A \\
\multicolumn{1}{l}{} &
  GPT-4 &
   &
  \textbf{8.95} &
  \textbf{35.28} &
  \textbf{93.15} &
  N/A \\ \cmidrule(l){2-7} 
\multicolumn{1}{l}{} &
  GPT-3.5 Davinci &
  \multirow{3}{*}{2-shot} &
  12.23 &
  36.66 &
  105.54 &
  N/A \\
\multicolumn{1}{l}{} &
  GPT-3.5 Turbo &
   &
  12.49 &
  39.37 &
  105.51 &
  N/A \\
\multicolumn{1}{l}{} &
  GPT-4 &
   &
  \textbf{16.78} &
  \textbf{44.21} &
  \textbf{83.31} &
  N/A \\ \midrule
\multicolumn{1}{l}{\multirow{6}{*}{\textbf{EN-ZH}}} & GPT-3.5 Davinci & \multirow{3}{*}{0-shot} & 32.41 & 40.82 & 99.45 & 59.87 \\
\multicolumn{1}{l}{} &
  GPT-3.5 Turbo &
   &
  36.83 &
  45.77 &
  99.83 &
  69.13 \\
\multicolumn{1}{l}{} &
  GPT-4 &
   &
  \textbf{37.65} &
  \textbf{47.02} &
  \textbf{99.37} &
  \textbf{70.75} \\ \cmidrule(l){2-7} 
\multicolumn{1}{l}{} &
  GPT-3.5 Davinci &
  \multirow{3}{*}{2-shot} &
  46.18 &
  49.12 &
  \textbf{69.0} &
  73.9 \\
\multicolumn{1}{l}{} &
  GPT-3.5 Turbo &
   &
  \textbf{45.95} &
  49.79 &
  74.53 &
  74.63 \\
\multicolumn{1}{l}{} &
  GPT-4 &
   &
  45.37 &
  \textbf{50.26} &
  79.29 &
  \textbf{74.9} \\ \bottomrule
\end{tabular}
\end{small}
\caption{Comparing GPT-3.5 \emph{text-davinci-003} to ChatGPT models \emph{gpt-3.5-turbo} and \emph{gpt-4} for zero-shot and few-shot translation with 2 fuzzy matches}
\label{tab:chatgpt}
\end{table}


\section{BLOOM and BLOOMZ}

In this section, we compare GPT-3.5 to open-source multilingual models, namely BLOOM \citep{BLOOM2022} and BLOOMZ \citep{Muennighoff2022-BLOOMZ-mT0}. While BLOOM is a general-purpose LLM, BLOOMZ belongs to a family of models capable of following human instructions in a zero-shot manner.

We use BLOOM and BLOOMZ via the Hugging Face's Inference API.\footnote{\url{https://huggingface.co/inference-api}} As mentioned in Section \ref{sec:setup}, recommended (sampling) parameters for translation with \mbox{GPT-3.5} are top-p 1 and temperature up to 0.3. For BLOOM, the same parameters are not good for translation.\footnote{Using lower sampling values of top-p and temperature such as 0.9 and 0.1, respectively, can generate good outputs. However, greedy search shows better translation performance.} We found that ``greedy search" achieves better results for BLOOM, which are reported in Table \ref{tab:bloom}. We use a batch size of 1, and set the $max\_new\_tokens$ parameter to be double the number of words of the source sentence if it is less than 250, the maximum number of new tokens allowed by BLOOM's API; otherwise, we set it to 250 tokens. For comparison purposes, we use the same values for BLOOMZ.\footnote{BLOOMZ is trained to generate the required output only; however, using BLOOM, we had to truncate over-generated text outputs, excluding anything generated in a new line.}

When providing each system with two fuzzy matches, generally GPT-3.5 outperforms both BLOOM and BLOOMZ for most language pairs, except English-to-Arabic translation. The English-to-French translation quality of BLOOM and \mbox{GPT-3.5} is comparable.

\vspace{\fill}

\begin{table}[H]
\captionsetup{font=small,labelfont=small}
\centering
\begin{small}
\begin{tabular}{@{}clcccc@{}}
\toprule
\textbf{Lang}                   & \multicolumn{1}{c}{\textbf{System}} & \textbf{spBLEU ↑} & \textbf{chrF++ ↑} & \textbf{TER ↓} & \textbf{COMET ↑} \\ \midrule
\multirow{3}{*}{\textbf{EN-AR}} & BLOOM fuzzy 2-shot  & \textbf{43.19} & \textbf{59.48}   & \textbf{57.58} &                                  \textbf{67.36}   \\
                                & BLOOMZ fuzzy 2-shot & 36.29          & 53.33            & 66.86       &
                                58.4           \\
                                & GPT-3 fuzzy 2-shot & 38.41          & 56.57          & 62.31          & 57.36          \\ \cmidrule(l){2-6} 
\multirow{3}{*}{\textbf{EN-ES}} & BLOOM fuzzy 2-shot & 57.67          & 74.25          & 34.86          &                                       86.48          \\
                                & BLOOMZ fuzzy 2-shot & 53.07          & 70.44         & 40.45          & 81.38         \\
                                & GPT-3 fuzzy 2-shot  & \textbf{59.64} & \textbf{75.83} & \textbf{32.56} & \textbf{90.37} \\ \cmidrule(l){2-6} 
\multirow{3}{*}{\textbf{EN-FR}} & BLOOM fuzzy 2-shot  & \textbf{50.52} & 66.81         & \textbf{46.45} &                                     55.74            \\
                                & BLOOMZ fuzzy 2-shot & 45.1           & 62.73         & 51.69           &
                                47.49             \\
                                & GPT-3 fuzzy 2-shot & 49.79          & \textbf{67.41} & 46.79          & \textbf{61.38} \\ \cmidrule(l){2-6} 
\multirow{3}{*}{\textbf{EN-RW}} & BLOOM fuzzy 2-shot & 10.95          & 31.87          & 91.07 & N/A                                   \\
                                & BLOOMZ fuzzy 2-shot & \textbf{12.26}          & 35.44          & \textbf{88.36}          & N/A \\
                                & GPT-3 fuzzy 2-shot & 12.23          & \textbf{36.66} & 105.54         & N/A            \\ \cmidrule(l){2-6} 
\multirow{3}{*}{\textbf{EN-ZH}} & BLOOM fuzzy 2-shot & 40.62          & 40.62          & 75.24          &                                     66.23          \\
                                & BLOOMZ fuzzy 2-shot & 34.82         & 38.23          & 80.03          & 59.92          \\
                                & GPT-3 fuzzy 2-shot & \textbf{46.18} & \textbf{49.12} & \textbf{69.0}  & \textbf{73.9}  \\ \bottomrule
\end{tabular}
\end{small}
\caption{Comparing GPT-3.5 to BLOOM and BLOOMZ for few-shot translation with 2 fuzzy matches}
\label{tab:bloom}
\end{table}

\section{Conclusion}

In this work, we conducted several experiments to assess the performance of GPT-3.5 across multiple translation tasks, namely adaptive MT using fuzzy matches (cf. Section \ref{sec:adaptive-MT}), MT post-editing (cf. Section \ref{sec:fix-mt}), terminology extraction \mbox{(cf. Section \ref{sec:term-extract})}, and terminology-constrained MT (cf. Section \ref{sec:term-constrain}). Moreover, we compared its translation quality with strong encoder-decoder MT systems. Generally speaking, results obtained from these experiments are very promising. While some high-resource languages such as English-to-French, English-to-Spanish and even English-to-Chinese show excellent results, other languages have lower support either because they are low-resource languages such as English-to-Kinyarwanda or because of \mbox{issues} in the \mbox{GPT-3.5} tokeniser such as English-to-Arabic. Nevertheless, when we used GPT-3.5 for MT post-editing of the English-to-Arabic translation obtained from OPUS, the quality significantly surpassed that obtained from both OPUS and Google Translation API. This means that different pipelines can be adopted in production for different language pairs, based on the level of support of these languages by an LLM.

Furthermore, we briefly compared \mbox{GPT-3.5} translation quality with open-source LLMs such as BLOOM and BLOOMZ. In the future, we would like to expand our experiments with open-source LLMs to cover more aspects.

For adaptive MT with fuzzy matches, it would be interesting to investigate \emph{dynamic} few-shot example selection. For instance, instead of selecting 5 fuzzy matches for all sentences, only high-quality fuzzy matches up to a certain similarity score are used. Similarly, when incorporating glossary terms or MT outputs from other systems, only those with certain quality characteristics are utilised. This can potentially enhance performance gains.

For terminology extraction, we would like to try ``phrases" instead of ``terms". This would generate longer strings. We would like to see the effect of using such longer phrases, especially for low-resource languages.

This work mainly aims at understanding the quality and level of support that LLMs can achieve (out of the box) for a range of translation tasks across diverse language pairs. In the future, we might consider starting with fine-tuning the model, and then conducting similar experiments.\footnote{Chapter \ref{chapter:finetuning} demonstrates preliminary experiments of fine-tuning Mistral 7B for the purpose of adaptive MT.} This can be especially beneficial for low-resource languages and rare domains, and can help enhance quality and efficiency.

\end{large}

\newpage
\section{Prompts}
\label{a-prompts}

These are some examples of the prompts we used for our experiments.

\begin{multicols}{2}
\vspace{-4ex}
\subsubsection{Zero-shot Translation}

\begin{tcolorbox}[enhanced,attach boxed title to top left={yshift=-3mm,yshifttext=-1mm,xshift=3mm},
  colback=blue!1!white,colframe=cyan!90!black,colbacktitle=cyan!90!black,boxrule=1pt,
  left=3pt, top=0pt, width=180pt,center,
  title=Prompt: EN-AR zero-shot translation,fonttitle=\tiny,
  boxed title style={size=small,colframe=cyan!90!black} ]
  \begin{tiny}
  \textmyfont{\emph{
    \begin{itemize}
    \setlength\itemsep{-2ex}
    \item[] English: $<$source\_segment$>$
    \item[] Arabic:
\end{itemize}
}}
\end{tiny}
\end{tcolorbox}

\vspace{-4ex}
\subsubsection{\RaggedRight Adaptive MT with Fuzzy Matches}

\begin{tcolorbox}[enhanced,attach boxed title to top left={yshift=-3mm,yshifttext=-1mm,xshift=3mm},
  colback=blue!1!white,colframe=cyan!90!black,colbacktitle=cyan!90!black,boxrule=1pt,
  left=3pt, top=0pt, width=180pt,center,
  title=Prompt: EN-AR two-shot translation,fonttitle=\tiny,
  boxed title style={size=small,colframe=cyan!90!black} ]
  \begin{tiny}
  \textmyfont{\emph{
    \begin{itemize}
    \setlength\itemsep{-2ex}
    \item[] English: $<$source\_fuzzy\_match\textsubscript{2}$>$
    \item[] Arabic: $<$target\_fuzzy\_match\textsubscript{2}$>$
    \item[] English: $<$source\_fuzzy\_match\textsubscript{1}$>$
    \item[] Arabic: $<$target\_fuzzy\_match\textsubscript{1}$>$
    \item[] English: $<$source\_segment$>$
    \item[] Arabic:
\end{itemize}
}}
\end{tiny}
\end{tcolorbox}

\vspace{-4ex}
\subsubsection{MT Post-editing}

\begin{tcolorbox}[enhanced,attach boxed title to top left={yshift=-3mm,yshifttext=-1mm,xshift=3mm},
  colback=blue!1!white,colframe=cyan!90!black,colbacktitle=cyan!90!black,boxrule=1pt,
  left=3pt, top=0pt, width=180pt,center,
  title=Prompt: EN-ZH two-shot + 1-MT,fonttitle=\tiny,
  boxed title style={size=small,colframe=cyan!90!black} ]
  \begin{tiny}
  \textmyfont{\emph{
    \begin{itemize}
    \setlength\itemsep{-2ex}
    \item[] English: $<$source\_fuzzy\_match\textsubscript{2}$>$
    \item[] Chinese: $<$target\_fuzzy\_match\textsubscript{2}$>$
    \item[] English: $<$source\_fuzzy\_match\textsubscript{1}$>$
    \item[] Chinese: $<$target\_fuzzy\_match\textsubscript{1}$>$
    \item[] English: $<$source\_segment$>$
    \item[] MT: $<$mt\_segment$>$
    \item[] Chinese:
\end{itemize}
}}
\end{tiny}
\end{tcolorbox}


\begin{tcolorbox}[enhanced,attach boxed title to top left={yshift=-3mm,yshifttext=-1mm,xshift=3mm},
  colback=blue!1!white,colframe=cyan!90!black,colbacktitle=cyan!90!black,boxrule=1pt,
  left=3pt, top=0pt, width=180pt,center,
  title=Prompt: EN-ZH two-shot + all-MT,fonttitle=\tiny,
  boxed title style={size=small,colframe=cyan!90!black} ]
  \begin{tiny}
  \textmyfont{\emph{
    \begin{itemize}
    \setlength\itemsep{-2ex}
    \item[] English: $<$source\_fuzzy\_match\textsubscript{2}$>$
    \item[] MT: $<$mt\_fuzzy\_match\textsubscript{2}$>$
    \item[] Chinese: $<$target\_fuzzy\_match\textsubscript{2}$>$
    \item[] English: $<$source\_fuzzy\_match\textsubscript{1}$>$
    \item[] MT: $<$mt\_fuzzy\_match\textsubscript{1}$>$
    \item[] Chinese: $<$target\_fuzzy\_match\textsubscript{1}$>$
    \item[] English: $<$source\_segment$>$
    \item[] MT: $<$mt\_segment$>$
    \item[] Chinese:
\end{itemize}
}}
\end{tiny}
\end{tcolorbox}

\vspace{-4ex}
\subsubsection{Terminology Extraction}

\begin{tcolorbox}[enhanced,attach boxed title to top left={yshift=-3mm,yshifttext=-1mm,xshift=3mm},
  colback=blue!1!white,colframe=cyan!90!black,colbacktitle=cyan!90!black,boxrule=1pt,
  left=-3pt, top=0pt, width=180pt,center,
  title=Prompt: terminology extraction,fonttitle=\tiny,
  boxed title style={size=small,colframe=cyan!90!black} ]
  \begin{tiny}
  \textmyfont{\emph{
    \begin{itemize}
    \setlength\itemsep{-2ex}
    \item[] $<$source\_lang$>$: $<$source\_sentence$>$
    \item[] $<$target\_lang$>$: $<$target\_sentence$>$
    \item[]
    \item[] Extract $<$number$>$ terms from the above sentence pair. Type each $<$source\_lang$>$ term and its $<$target\_lang$>$ equivalent in one line, separated by '$<$separator$>$'.
    \item[]
    \item[] 1.
\end{itemize}
}}
\end{tiny}
\end{tcolorbox}

\vspace{-4ex}
\subsubsection{Terminology-constrained MT}

\begin{tcolorbox}[enhanced,attach boxed title to top left={yshift=-3mm,yshifttext=-1mm,xshift=3mm},
  colback=blue!1!white,colframe=cyan!90!black,colbacktitle=cyan!90!black,boxrule=1pt,
  title=Prompt: EN-ES zero-shot + glossary terms,fonttitle=\tiny,
  left=-3pt, top=0pt, width=180pt,center,
  boxed title style={size=small,colframe=cyan!90!black} ]
  \begin{tiny}
  \textmyfont{\emph{
  \begin{itemize}
    \setlength\itemsep{-2ex}
    \item[] Terms: $<$src\_term\textsubscript{1}$>$ $=$ $<$tgt\_term\textsubscript{1}$>$ - $<$src\_term\textsubscript{2}$>$ $=$ $<$tgt\_term\textsubscript{2}$>$ ... $<$src\_term\textsubscript{5}$>$ $=$ $<$tgt\_term\textsubscript{5}$>$
    \item[] English: $<$source\_segment$>$
    \item[] Spanish:
\end{itemize}
}}
\end{tiny}
\end{tcolorbox}


\begin{tcolorbox}[enhanced,attach boxed title to top left={yshift=-3mm,yshifttext=-1mm,xshift=3mm},
  colback=blue!1!white,colframe=cyan!90!black,colbacktitle=cyan!90!black,boxrule=1pt,
  title=Prompt: EN-ES two-shot + fuzzy terms,fonttitle=\tiny,
  left=3pt, top=0pt, width=180pt,center,
  boxed title style={size=small,colframe=cyan!90!black} ]
  \begin{tiny}
  \textmyfont{\emph{
  \begin{itemize}
    \setlength\itemsep{-2ex}
    \item[] Terms: $<$terms\_fuzzy\_match\textsubscript{2}$>$
    \item[] English: $<$source\_fuzzy\_match\textsubscript{2}$>$
    \item[] Spanish: $<$target\_fuzzy\_match\textsubscript{2}$>$
    \item[] Terms: $<$terms\_fuzzy\_match\textsubscript{1}$>$
    \item[] English: $<$source\_fuzzy\_match\textsubscript{1}$>$
    \item[] Spanish: $<$target\_fuzzy\_match\textsubscript{1}$>$
    \item[] Terms: $<$terms\_from\_fuzzy\_matches\textsubscript{1+2}$>$
    \item[] English: $<$source\_segment$>$
    \item[] Spanish:
\end{itemize}
}}
\end{tiny}
\end{tcolorbox}


\begin{tcolorbox}[enhanced,attach boxed title to top left={yshift=-3mm,yshifttext=-1mm,xshift=3mm},
  colback=blue!1!white,colframe=cyan!90!black,colbacktitle=cyan!90!black,boxrule=1pt,
  title=Prompt: EN-ES two-shot + glossary terms,fonttitle=\tiny,
  left=3pt, top=0pt, width=180pt,center,
  boxed title style={size=small,colframe=cyan!90!black} ]
  \begin{tiny}
  \textmyfont{\emph{
  \begin{itemize}
\setlength\itemsep{-2ex}
    \item[] Terms: $<$terms\_fuzzy\_match\textsubscript{2}$>$
    \item[] English: $<$source\_fuzzy\_match\textsubscript{2}$>$
    \item[] Spanish: $<$target\_fuzzy\_match\textsubscript{2}$>$
    \item[] Terms: $<$terms\_fuzzy\_match\textsubscript{1}$>$
    \item[] English: $<$source\_fuzzy\_match\textsubscript{1}$>$
    \item[] Spanish: $<$target\_fuzzy\_match\textsubscript{1}$>$
    \item[] Terms: $<$terms\_from\_glossary$>$
    \item[] English: $<$source\_segment$>$
    \item[] Spanish:
\end{itemize}
}}
\end{tiny}
\end{tcolorbox}

\end{multicols}

\vspace{\fill}

\begin{large}

\chapter{Fine-tuning Large Language Models for Adaptive Machine Translation}
\label{chapter:finetuning}

This chapter\footnote{More details can be found in a preprint paper at: \url{https://arxiv.org/abs/2312.12740}} demonstrates the setup and results of experiments with fine-tuning an LLM, namely Mistral 7B \citep{Jiang2023-Mistral}, for ``adaptive" MT. Hence, the model is not only fine-tuned for regular (zero-shot) translation, but also for adaptation to one fuzzy match (one-shot) at translation time \citep{Moslem2023-AdaptiveMT-LLM-Finetuning}. The experiments were conducted for Spanish-to-English medical adaptive MT.
The code used for these experiments is publicly available.\footnote{GitHub repository: \url{https://github.com/ymoslem/Adaptive-MT-LLM-Fine-tuning}}

In Table \ref{tab:mistral-finetuning}, the last two rows show the results for fine-tuning Mistral 7B. The rest of the results are for baselines, i.e. without fine-tuning. As illustrated, fine-tuning has led to quality improvements in terms of both zero-shot and one-shot translation. The fine-tuned version of Mistral outperforms its own baseline (i.e. without fine-tuning) for both zero-shot and one-shot translation. Zero-shot translation quality of the fine-tuned Mistral outperforms ChatGPT ``gpt-3.5-turbo", while one-shot translation quality of the fine-tuned Mistral is on par with that of ChatGPT. Zero-shot translation of the fine-tuned Mistral is on par with NLLB 3.3B, while one-shot translation quality of the fine-tuned Mistral outperforms that of NLLB 3.3B. To conclude, fine-tuning an efficient LLM like Mistral 7B helps to produce a high-quality zero-shot translation comparable to that of MT task-oriented models such as NLLB 3.3B, while achieving adaptive gains of one-shot translation on par with commercial LLMs such as ChatGPT ``gpt-3.5-turbo".

\begin{table}[htp]
    \captionsetup{font=footnotesize,labelfont=footnotesize}
    \centering
    \begin{small}
    \begin{tabular}{ccccccc}
    \toprule
    \textbf{Lang} & \textbf{Model} & \textbf{Context} & \textbf{BLEU ↑} & \textbf{chrF++ ↑} & \textbf{TER ↓} & \textbf{COMET ↑} \\ \midrule 
    \multirow{8}{*}{\textbf{ES-EN}} & \multirow{2}{*}{NLLB 3.3B} & Source only (zero-shot) & \underline{47.02} & 68.82 & 43.43 & 66.46 \\
     &  & + Fuzzy (one-shot) & 47.42 & 68.77 & 45.26 & 64.57 \\ \cmidrule{2-7}
     & ChatGPT & Source only (zero-shot) & 44.65 & 68.36 & 44.28 & 74.48 \\
     & {``\footnotesize gpt-3.5-turbo"} & + Fuzzy (one-shot) & 48.34 & 70.54 & 40.80 & \textbf{80.25}* \\ \cmidrule{2-7}
     & \multirow{2}{*}{Mistral 7B} & Source only (zero-shot) & 42.88 & 66.03 & 46.54 & 69.56 \\
     &  & + Fuzzy (one-shot) & 47.35 & 69.25 & 42.53 & 76.37 \\ \cmidrule{2-7}
     & Mistral 7B & Source only (zero-shot) & 46.71 & \underline{69.55} & \underline{41.81} & \underline{77.44} \\
     & ``\textbf{Fine-tuned}" & + Fuzzy (one-shot) & \textbf{49.69} & \textbf{70.89} & \textbf{40.08} & \textbf{79.62} \\ \bottomrule
    \end{tabular}
    \end{small}
    \caption{Comparing adaptive MT with NLLB-200 3.3B, ChatGPT, and Mistral 7B (before and after fine-tuning). Our fine-tuned Mistral demonstrates quality gains for both zero-shot translation and one-shot adaptive MT.}
    \label{tab:mistral-finetuning}
\end{table}

\section{Data}
\label{sec:finetune-data}

In this experiment, fine-tuning uses a mix of 10,000 segments with zero-shot prompts and 10,000 segments with one-shot prompts. The whole dataset was split into 19,000 segments for training the model and 1,000 randomly selected segments for validation while training. Fuzzy matches are extracted from a ``context dataset" including 50,000 translation pairs. The test dataset includes 10,000 sentences, and it has its own unique context dataset, which consists of 50,000 unique translation pairs. Figure \ref{fig:finetune-prompts} shows examples of zero-shot and one-shot prompts. The retrieval process of fuzzy matches is detailed in Section \ref{sec:finetune-information-retrieval}.

\vspace{2ex}

\begin{figure}[H]
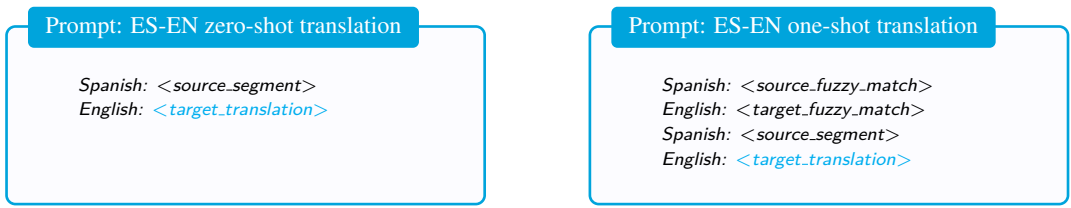

\captionsetup{font=footnotesize,labelfont=footnotesize}
\begin{multicols}{2}
\begin{tcolorbox}[enhanced,attach boxed title to top left={yshift=-3mm,yshifttext=-1mm,xshift=3mm},
  colback=blue!1!white,colframe=cyan!90!black,colbacktitle=cyan!90!black,boxrule=1pt,
  left=3pt, top=0pt,height=68pt, width=180pt,center,
  title=Prompt: ES-EN zero-shot translation,fonttitle=\small,
  boxed title style={size=small,colframe=cyan!90!black} ]
  \begin{scriptsize}
  \textmyfont{\emph{
    \begin{itemize}
    \setlength\itemsep{-1.5ex}
    \item[] Spanish: $<$source\_segment$>$
    \item[] English: {\color{cyan}$<$target\_translation$>$}
\end{itemize}
}}
\end{scriptsize}
\end{tcolorbox}

\begin{tcolorbox}[enhanced,attach boxed title to top left={yshift=-3mm,yshifttext=-1mm,xshift=3mm},
  colback=blue!1!white,colframe=cyan!90!black,colbacktitle=cyan!90!black,boxrule=1pt,
  left=3pt, top=0pt, height=68pt, width=180pt, center,
  title=Prompt: ES-EN one-shot translation,fonttitle=\small,
  boxed title style={size=small,colframe=cyan!90!black} ]
  \begin{scriptsize}
  \textmyfont{\emph{
    \begin{itemize}
    \setlength\itemsep{-1.5ex}
    \item[] Spanish: $<$source\_fuzzy\_match$>$
    \item[] English: $<$target\_fuzzy\_match$>$
    \item[] Spanish: $<$source\_segment$>$
    \item[] English: {\color{cyan}$<$target\_translation$>$}
\end{itemize}
}}
\end{scriptsize}
\end{tcolorbox}
\end{multicols}
\vspace{-1ex}
\caption{Zero-shot and one-shot prompts used for fine-tuning Mistral}
\label{fig:finetune-prompts}
\end{figure}

Originally, we mixed Spanish-to-English medical datasets from OPUS \citep{Tiedemann2012-OPUS}, namely ELRC \citep{ELRC2019}, EMEA \citep{EMA2012}, SciELO \citep{Soares2018-SciELO}, and TICO-19 \citep{Anastasopoulos2020-TICO-19} datasets. Then we filtered\footnote{Filtering scripts are available at: \url{https://github.com/ymoslem/MT-Preparation}} the resulted dataset to exclude duplicates and too long segments.\footnote{As NLLB supports a maximum token length of 512 \textit{tokens}, we exclude any segment whose source or target text is longer than 70 \textit{words} to also take into account the one-shot case that will augment another segment to the original one. As the context window of Mistral is much larger (8K tokens), it is theoretically possible to translate longer segments.}
The whole dataset includes 1,868,505 segments before filtering, and 922,343 segments after filtering.\footnote{We observe that almost two-thirds of the EMEA dataset are duplicates.} However, we used only part of it for this preliminary experiment. In the future, we would like to increase the size of the training data and compare the performance. Nevertheless, achieving these results (cf. Table \ref{tab:mistral-finetuning}) with such a small dataset (cf. Section \ref{sec:finetune-data}) shows how promising this approach is.

\section{Information Retrieval}
\label{sec:finetune-information-retrieval}

For indexing and retrieval of fuzzy matches, we use Sentence-Transformers \citep{Reimers2019-SentenceTransformers} and \textit{Faiss} \citep{Johnson2019-Faiss}, with a multilingual model to generate the embeddings for the datasets and later to extract fuzzy matches through semantic search.

\paragraph{Embedding:} To encode all the translation segments into embeddings, we employ a \textit{multilingual} model, namely Microsoft's \textit{``Multilingual-MiniLM-L12-H384"} \citep{Wang2020-MS-MiniLM}. These embeddings will be used later for both indexing and retrieval. The step of generating embeddings can be implemented with the Sentence-Transformers library\footnote{\url{https://www.sbert.net/}} \citep{Reimers2019-SentenceTransformers}.

\paragraph{Indexing:} For indexing, we use Faiss\footnote{\url{https://github.com/facebookresearch/faiss}} \citep{Johnson2019-Faiss}, a library for efficient similarity search and clustering of dense vectors. We train an \emph{IndexIVFFlat} index, which uses \emph{IndexFlatL2} as a quantiser.\footnote{\url{https://github.com/facebookresearch/faiss/wiki/Faster-search}}
The embedding size is 384, which is the same as the embedding size of model used.
For the number of clusters at indexing time, the \emph{nlist} parameter was set to 4096, while the number of clusters to explore at search time \textit{nprobe} was set to search for nearest neighbours in 32 clusters.\footnote{According to Faiss' guidelines for choosing an index, the number of clusters is recommended to be between \emph{4*sqrt(N)} to \emph{16*sqrt(N)}, where \emph{N} is the size of the dataset. Obviously, this would depend on the available computational resources.}

\paragraph{Semantic Search:} This step computes the cosine similarity between the query and all the documents in the corpus based on their embeddings, and retrieves the top $k$ matching entries. In this case, our query is each source segment, and the corpus is the unique ``context dataset" (cf. Section \ref{sec:finetune-data}) leveraged to extract fuzzy matches.


\section{Fine-tuning}
\label{sec:finetune-finetuning}

We used QLoRA \citep{Hu2021-LoRA,Dettmers2023-QLoRA} for efficient fine-tuning with 4bit quantisation, with Hugging Face Transformers.\footnote{\url{https://github.com/huggingface/transformers}} Fine-tuning was for only one epoch, which revealed better results than fine-tuning for 4 epochs. The configuration of quantisation through \mbox{\textit{BitsAndBytes}} includes: \textit{load\_in\_4bit=True}, \textit{bnb\_4bit\_quant\_type=``nf4"}, \textit{bnb\_4bit\_use\_double\_quant=True}, and \textit{bnb\_4bit\_compute\_dtype=torch.bfloat16}. LoRA configuration was set via the PEFT library\footnote{\url{https://github.com/huggingface/peft}} as follows: the dimension of the low-rank matrices \textit{r=64}, the scaling factor for the weight matrices \textit{lora\_alpha=16}, dropout probability of the LoRA layers \textit{lora\_dropout=0.1}, and without training the bias parameters for better performance \textit{bias=``none"}.
Training arguments include: batch size for training and evaluation 32 examples, \textit{warmup\_steps=0.03}, \textit{learning\_rate=2e-3}, \textit{lr\_scheduler\_type=``constant"}, and \textit{bf16=True}.
Both training and inference utilise Google Colab Pro+ with one GPU \textit{NVIDIA A100-SXM4-40GB}.

\section{Inference}

For inference (translation), we experimented with a number of models including NLLB-200 \citep{NLLB2022} whose architecture is encoder-decoder Transformer as well as ChatGPT \citep{Brown2020-GPT-3,Ouyang2022-InstructGPT} and Mistral 7B, which are autoregressive decoder-only Transformer-based LLMs. Mistral 7B was used both without fine-tuning and after fine-tuning on a mix of zero-shot and one-shot translation prompts.

\paragraph{Mistral 7B:} We converted both the baseline and our fine-tuned models of Mistral 7B to the CTranslate2 \footnote{\url{https://github.com/OpenNMT/CTranslate2}} \citep{Klein2020-Efficient} format (with 8int quantisation) for more efficiency. We employed greedy search by setting \textit{sampling\_topk=1} and added the new line \verb|\n| character to \textit{end\_token} to avoid overgeneration. Mistral through CTranslate2 translates the zero-shot test dataset that includes 10,000 sentences in 2–3 minutes (approx. 80 segments/second). The time almost doubles for the one-shot test dataset. Figure \ref{fig:finetune-prompts} illustrates the prompts used for both inference and fine-tuning. Two versions of Mistral were tested, the baseline model without fine-tuning, and the model we fine-tuned on a mix of zero-shot and one-shot translation prompts. Table \ref{tab:mistral-finetuning} shows translation evaluation results of both models. Fine-tuning Mistral on a mix of zero-shot and one-shot prompts improved both its regular translation quality (i.e. when only the source text is available) and adaptive translation quality (in this case, when one fuzzy match is provided) at inference time.

\paragraph{ChatGPT:} The model used is ``gpt-3.5-turbo'' with \textit{temperature=0.3} and \textit{top\_p=1}. Requests were sent in batches of 20 segments, and the \textit{max\_tokens} argument was set as the largest number of words per source segment in a batch multiplied by 4, which is a rough number that can be increased or decreased based on the language and model. When the source text is augmented with one fuzzy match, the translation quality is improved by several points across all the automatic evaluation metrics. However, it is worth noting that although batch processing was employed, there is no guarantee of the generation time with ChatGPT, which can range from several minutes to a couple of hours. In this sense, we observe that Mistral 7B is much more efficient.

\paragraph{NLLB-200:} In this set of experiments, the NLLB-200 model was used as is, i.e. without fine-tuning. The first two rows of Table \ref{tab:mistral-finetuning} show the evaluation scores of using NLLB 3.3B for translation without a fuzzy match (zero-shot) and with a fuzzy match (one-shot). The same test dataset and its unique context dataset were used; however, fuzzy matching augmentation was done differently to match the architecture of the NLLB model. Each source sentence was augmented with its best fuzzy match, and the two sentences were separated by the language code of the source language (in this case \textit{``spa\_Latn"}). As NLLB was mainly pre-trained on sentences, we had to add an extra token that usually comes at the beginning of sentences, such as a bullet point (``{\tiny •}") after the language code between the two source sentences. Hence, the target fuzzy match was fed to the model as a prefix augmented by the target language code (in this case \textit{``eng\_Latn"}) and the extra token. In this sense, the model was encouraged to complete the translation through teacher-forcing \citep{Williams1989-TeacherForcing}, i.e. using the ground truth as input, instead of the model output. In other words, the model is not required to translate the target fuzzy match, but rather to use the provided translation as is to guide the translation of the new untranslated source sentence. Translation arguments include: \textit{batch\_type=``tokens"}, \textit{max\_batch\_size=2024}, \textit{beam\_size=2}, \textit{min\_decoding\_length=2}, and \textit{max\_decoding\_length=512}. Although there is a marginal improvement given the BLEU score, the performance degrades according to the other reported automatic evaluation metrics, chrF++, TER and COMET. The fact that NLLB was trained to translate individual sentences rather than a series of sentences or full documents could be the main reason for this. In the future, we would like to experiment with fine-tuning NLLB with fuzzy matching augmentation and compare the results.

\section{Conclusion and Future Work}

In this chapter, we showed how fine-tuning a general purpose LLM such as Mistral 7B can improve its in-context learning ability, especially for real-time adaptive MT (cf. Table \ref{tab:mistral-finetuning}). Moreover, such translation quality gains were achievable through fine-tuning using a relatively small dataset (20,000 segments). Incorporating a mix of zero-shot and one-shot prompts in the training data helps improve both regular zero-shot translation, and one-shot translation that incorporates a fuzzy match. It is worth noting that Mistral 7B is much more efficient than ChatGPT, which is an added benefit in production scenarios.

In the future, we would like to experiment with other domains and language pairs, including low-resource languages. As currently there are several multilingual LLMs such as BLOOM (46 languages) \citep{BLOOM2022}, Falcon (EN-DE-ES-FR) \citep{Penedo2023-Falcon}, larger versions of Mistral/Mixtral (EN-DE-ES-FR-IT) \citep{Jiang2023-Mistral}, Jais (AR-EN) \citep{Sengupta2023-Jais}, Baichuan (ZH) \citep{Yang2023-Baichuan}, and Qwen (ZH) \citep{Bai2023-Qwen}, it can be insightful to apply the same approach with these models. Furthermore, as we experimented with NLLB-200 without fine-tuning, we would like to experiment with fine-tuning for fair comparison. While we fine-tuned Mistral on a small dataset, it is recommended to experiment with fine-tuning on more data, especially from the same domain. It can also be helpful to incorporate different types of prompts, such as few-shot prompts, and terminology-based prompts.

\end{large}

\begin{large}

\chapter{Domain Terminology \mbox{Integration} into Machine Translation:
\mbox{Leveraging} Large Language \mbox{Models}}
\label{chapter:terminology}

\vspace{-1\baselineskip}
{\small Yasmin Moslem, Gianfranco Romani, Mahdi Molaei, Rejwanul Haque, John Kelleher, and Andy Way}

\vspace{-15pt}
\singlespacing
\noindent {\footnotesize In Proceedings of the Eighth Conference on Machine Translation (WMT 2023), pages 902–911, Sentosa, Singapore. Association for Computational Linguistics.\footnote{Published at: \url{https://aclanthology.org/2023.wmt-1.82/}}}
\bigbreak

\onehalfspacing

\begin{abstract}
\smallskip
\nohyphens{
\normalsize
This paper discusses the methods that we used for our submissions to the WMT 2023 \mbox{Terminology} Shared Task for German-to-English~(DE-EN), English-to-Czech~(EN-CS), and Chinese-to-English~(ZH-EN) language pairs. The task aims to advance machine \mbox{translation} (MT) by challenging participants to develop systems that accurately translate technical terms, ultimately enhancing communication and understanding in specialised domains. To this end, we conduct experiments that utilise large language models (LLMs) for two purposes: generating synthetic bilingual terminology-based data, and post-editing translations generated by an MT model through incorporating pre-approved terms.~Our system employs a four-step process:~(i) using an LLM to generate bilingual synthetic data based on the provided terminology, (ii) fine-tuning a generic encoder-decoder MT model, with a mix of the terminology-based synthetic data generated in the first step and a randomly sampled portion of the original generic training data, (iii) generating translations with the fine-tuned MT model, and (iv) finally, leveraging an LLM for terminology-constrained automatic post-editing of the translations that do not include the required terms. The results demonstrate the effectiveness of our proposed approach in improving the integration of pre-approved terms into translations. The number of terms incorporated into the translations of the blind dataset increases from an average of 36.67\% with the generic model to an average of 72.88\% by the end of the process. In other words, successful utilisation of terms nearly doubles across the three language pairs.
}
\end{abstract}

\section{Context}

This work builds on our two previous papers for AMTA 2022 and EAMT 2023. It investigates techniques to enhance adherence to pre-approved terminology in translation, including terminology-based data generation and MT automatic post-editing with LLMs. Section \ref{sec:intro-terminology} elaborates on the research context of this paper.

\section{Introduction}
\label{sec:intro-terminology}

The primary goal of the WMT 2023 Terminology Shared Task is to evaluate the ability of MT systems to accurately translate technical terminology.\footnote{For the test dataset of the Chinese-to-English language pair, the organisers used the BWB corpus, which comprises texts extracted from novels. The test dataset of the English-to-Czech language pair consists of NLP paper abstracts, while the test dataset of the German-to-English language pair consists of medical paper abstracts \citep{Semenov2023-WMT23-Terminology}.} The task aims to assess the extent to which MT models can utilise additional information regarding the translation of terminology. The shared task requires the participants to provide three translations, one without terms and the others with two individual sets of terms.

There have been several advancements in the area of MT domain adaptation, where an MT model is expected to follow the style and terminology of a certain domain or client \citep{Chu2017-mixed-fine-tuning,Kobus2017-domain-control}. Moreover, some researchers give special focus to terminology while training and fine-tuning MT systems \citep{Dinu2019-TerminologyConstraintsTraining,Hu2019-LexiconInduction,Haque2020-Terminology,Michon2020-Terminology,Nayak2023-Instance}. However, forcing an MT model to adhere to certain terminology at inference time is among the most challenging aspects of MT. Hence, several researchers have investigated approaches to terminology-constrained decoding at translation time \citep{Hokamp2017-ConstrainedDecoding,Hasler2018-TerminologyConstraintsMT,Post2018-FastConstrainedDecoding,Hu2019-ImprovedConstrainedDecoding,Exel2020-TerminologyConstrainedMT}. The goal is to ensure that the MT system can accommodate unseen terminology while retaining translation accuracy and fluency.

\newpage

Recently, since the emergence of advanced LLMs such as GPT-3~\citep{Brown2020-GPT-3}, BLOOM~\citep{BLOOM2022}, PaLM~\citep{Chowdhery2022-PaLM}, Falcon \citep{Penedo2023-Falcon}, Llama~2 \citep{Touvron2023-Llama2}, Mistral \citep{Jiang2023-Mistral}, and Jais \citep{Sengupta2023-Jais} to mention just a few, researchers have been exploring the capabilities of these models for a number of tasks including MT~\citep{Bawden2023-BLOOM-MT,Hendy2023-LLM-MT,Jiao2023-LLM-MT,Moslem2023-AdaptiveMT,Vilar2023-PaLM-MT}. Some work investigates whether it is possible to employ the in-context learning feature of LLMs for adaptive MT with fuzzy matches \citep{Agrawal2023-SelectionMT,Moslem2023-AdaptiveMT}, and terminology-constrained MT using a pre-defined glossary~\citep{Moslem2023-AdaptiveMT} or even a dictionary~\citep{Ghazvininejad2023-DictionaryMT}. They found the approach is generally effective in increasing the number of terms used in the translation, even for low-resource languages.

We highlight our key contributions with the systems that we submitted for the WMT 2023 Terminology Shared Task as follows:

\begin{itemize}
    \item \textbf{LLMs for domain-specific data augmentation:} In our previous work \citep{Moslem2022-MT-LM}, we employed LLMs, namely GPT-J \citep{Wang2021-GPT-J} and mGPT \citep{Shliazhko2022-mGPT}, to generate domain-specific datasets based on the target sentences in a small authentic dataset, then generated the source sentences with back-translation \citep{Sennrich2016-BT,Poncelas2019-BT}, and finally fine-tuned an encoder-decoder MT model on this data. In this work, we take a couple of steps forward by instructing an LLM, namely ChatGPT \citep{Brown2020-GPT-3,Ouyang2022-InstructGPT}, to generate terminology-based bilingual synthetic data. In other words, the LLM will generate both the source and target sides of translation pairs, making sure the pre-approved target terms provided by the organisers are used in the translations.
    
    \item \textbf{LLMs for terminology-constrained MT and MT post-editing:} In our previous work, we utilised an LLM for translation and provided it with a list of terms to support in-context learning, which improved adherence to the required terminology at inference time \citep{Moslem2023-AdaptiveMT}. We also investigated whether we could use an LLM for post-editing MT generated by other systems. In this work, we prompt ChatGPT to insert missing terms into translations generated by an encoder-decoder MT system. In other words, if some of the translations generated by a fine-tuned MT model still do not include the terms provided by the organisers, we feed these translations into an LLM, namely ChatGPT, instructing it to incorporate these terms while using the same translation.
\end{itemize}

\section{Method}

In our submissions to the WMT 2023 Terminology Shared Task, we followed these steps:

\begin{enumerate}[(i)]
    \item Generate bilingual synthetic data based on the pre-approved terms, using an LLM, namely ChatGPT.
    \item Fine-tune a generic model, OPUS \citep{Tiedemann2020-OPUS-MT}, on a mix of the terminology-based synthetic data generated in (i) and a randomly sampled portion of the original generic training data.
    \item Generate translations of the dev, test, and blind datasets provided by the organisers with the fine-tuned model from (ii).
    \item Apply terminology-constrained automatic post-editing using ChatGPT to incorporate missing terms into translations that do not yet include the required terminology.
\end{enumerate}

\subsection{Synthetic Data Generation}
\label{sec:data-generation}

We used ChatGPT ``gpt-3.5-turbo''\footnote{The model ``gpt-3.5-turbo'' is a relatively efficient and cost-effective option, so we wanted to understand the quality we can achieve with it.} to generate bilingual sentence pairs, using the terms provided by the organisers. So, given a target term, the model was asked to generate multiple translation pairs, including both the source (e.g. German) and the target (e.g. English). For parameters of ChatGPT's API, we used \emph{top\_p} 1 and \emph{temperature} values 0 and 0.3 to generate diverse outputs.

\singlespacing
\begin{tcolorbox}[title={Example prompt: Terminology-based generation}, colbacktitle=gray, fonttitle=\footnotesize, boxrule=0.2pt, right=7pt, left=3pt]
    \begin{footnotesize}
    Please use the ``Federal Ministry of Science'' to generate just 20 numbered sentences in German-English in one Python dictionary format.
    \end{footnotesize}
\end{tcolorbox}
\bigbreak

\onehalfspacing

To filter the generated data, we first removed duplicate sentences from the whole dataset, based on both the source and target. Then, we applied language detection of both sides of the data using \textit{fastText}\footnote{\url{https://fasttext.cc/docs/en/language-identification.html}} and \textit{pycld2}\footnote{\url{https://github.com/aboSamoor/pycld2}} libraries to ensure that the generated sentences were in our desired languages. We excluded any sentences whose scores were below a certain threshold, namely 0.9 for \textit{fastText} and 90 for \textit{pycld2}.

The filtering step removed less than 1\% of the generated data. However, due to computational resource and time limitations, we could not use all the generated data. Table \ref{tab:generated-data} reports the number of generated, filtered, and used translation pairs.

Initially, we only had the development and test datasets, so we used them for the German-to-English language pair. Later, when the organisers released the blind dataset, we used the development, test and blind datasets for the Chinese-to-English and English-to-Czech language pairs.

\begin{table}[htp]
\captionsetup{font=footnotesize,labelfont=footnotesize}
\centering
\begin{small}
\begin{tabular}{@{}llll@{}}
\toprule
\multicolumn{1}{c}{\textbf{Lang}} & \multicolumn{1}{c}{\textbf{Raw}} & \multicolumn{1}{c}{\textbf{Filtered}} & \multicolumn{1}{c}{\textbf{Used}} \\ \midrule
\textbf{DE-EN} & 124,215 & 104,318 & 68,265 \\
\textbf{EN-CS} & 187,471 & 103,797 & 64,218 \\
\textbf{ZH-EN} & 90,538 & 72,695 & 49,001 \\ \bottomrule
\end{tabular}
\end{small}
\caption{Terminology-based bilingual data generated by \mbox{ChatGPT} for fine-tuning the OPUS model}
\label{tab:generated-data}
\end{table}

To assess the quality of the bilingual data generated by ChatGPT, we computed cross-entropy scores \citep{Moore2010-Scoring} of the synthetic translation pairs based on the strong encoder-decoder MT model, NLLB-200 3.3B \citep{NLLB2022}. For scoring, we used CTranslate2\footnote{\url{https://github.com/OpenNMT/CTranslate2}} \citep{Klein2020-Efficient} \emph{score\_batch()} method with the parameters \emph{batch\_type} ``tokens'' and \emph{max\_batch\_size} 2024. We scored each synthetic translation pair generated by ChatGPT, and then calculated the average score for the whole dataset. Computing dual cross-entropy scores according to two inverse translation models trained on clean data is an effective method to evaluate data quality \citep{Junczys-Dowmunt2018-Scoring}. Hence, we computed the scores of both directions of each language pair according to the multilingual MT model NLLB-200 3.3B because both directions are generated by ChatGPT. To produce a baseline for translation quality, we generated the translations of the same datasets using NLLB-200 3.3B for each language direction with \emph{beam\_size} 4, and then scored these translations with the same model. As the scores are in the form of negative log probabilities, we converted them to their exponential equivalents for readability, which are reported in Table \ref{tab:scoring}. It is normal that the model NLLB-200 generates higher scores for its own translations; however, we wanted to know to what extent such scores are comparable to those of ChatGPT's synthetic translation pairs. According to the scores, the German$\leftrightarrow$English language pair had the most comparable quality, followed by Czech$\leftrightarrow$English, and Chinese $\leftrightarrow$English language pairs. 

Among the approaches that can be employed for assessing the quality of synthetic bilingual data is semantic similarity between the two sides of each translation pair (e.g. with mUSE \citep{Yang2020-mUSE}). However, the scoring approach that we previously described and used achieves a similar goal while comparing the quality of the synthetic bilingual data to the translation quality of a strong MT baseline model, namely NLLB-200 3.3B.

\smallskip
\begin{table}[htp]
    \captionsetup{font=footnotesize,labelfont=footnotesize}
    \centering
    \begin{small}
    \begin{tabular}{cccc}
     \textbf{Lang}    &  \textbf{ChatGPT}    &  \textbf{NLLB}  & \textbf{Diff.} \\
     \toprule
     \textbf{DE-EN}    &  0.59    &  0.68   & 0.09 \\
     \textbf{EN-DE}    &  0.56    &  0.64   & 0.08 \\
     \textbf{Avg.}    &  0.58    &  0.66    & 0.08 \\ \midrule
     \textbf{CS-EN}    &  0.58    &  0.70    & 0.12 \\
     \textbf{EN-CS}    &  0.49    &  0.58    & 0.09 \\
     \textbf{Avg.}    &  0.54    &  0.64  & 0.10 \\ \midrule
     \textbf{ZH-EN}    &  0.39    &  0.56   & 0.17 \\
     \textbf{EN-ZH}    &  0.09    &  0.34   & 0.25 \\
     \textbf{Avg.}    &  0.24    &  0.45    & 0.21 \\ \bottomrule
    \end{tabular}
    \end{small}
    \caption{Scores of translation pairs generated by ChatGPT based on the NLLB-200 3.3B model}
    \label{tab:scoring}
\end{table}

\subsection{Fine-tuning}
\label{sec:fine-tuning}

Using the term-based synthetic bilingual data generated in the previous step, we fine-tuned encoder-decoder Transformer-based MT models \citep{Vaswani2017-attention}. In particular, we fine-tuned OPUS MT models, with Hugging Face Transformers.\footnote{\url{https://github.com/huggingface/transformers}} We applied mixed fine-tuning \citep{Chu2017-mixed-fine-tuning}; in other words, we fine-tuned the baseline model with a mix of the terminology-based synthetic data generated from the previous step (cf. Section \ref{sec:data-generation}) and a randomly sampled portion of the original generic data used to train the OPUS baseline model. The numbers of segments taken from the OPUS generic data are as follows: CS: 372,928, DE: 419,881, ZH: 462,780. We over-sampled the synthetic terminology-based data to make it the same size as the used portion of generic data. The fine-tuning parameters are as follows: \emph{train} = 0.9, \emph{val} = 0.1, \emph{batch\_size} = 32, \emph{learning\_rate} = 2e-5, \emph{accumulate\_gradient} = 4, \emph{weight\_decay} = 0.01, \emph{num\_train\_epochs} = 1, \emph{max\_input\_length} = 256, \emph{max\_target\_length} = 256. Finally, we used the fine-tuned model to generate translations for the development, test, and blind sets.

At first glance, the fine-tuning step might look redundant if the LLM can achieve the same translation quality directly, either via zero-shot translation or few-shot in-context learning \citep{Moslem2023-AdaptiveMT}. However, domain-specific or terminology-based knowledge distillation \citep{Treviso2023-Efficient-NLP} from a massive LLM to a compact task-oriented MT model can help boost efficiency at inference time while enhancing domain adaptation and terminology adherence. Obviously, when authentic in-domain data is available, it can be used for fine-tuning instead of synthetic data for domain adaptation of the MT model. In production workflows, only segments that do not meet specific quality criteria are passed to either human or automatic post-editing. Hence, deployment of a model fine-tuned on in-domain data can reduce the number of translations that need post-editing.

\subsection{Terminology-constrained Automatic Post-Editing}
\label{sec:term-ape}

For the shared task, the organisers provided two term sets for each source sentence in the test and blind datasets, and expected the participants to generate two translations that use one term set each. In this step of terminology-constrained automatic post-editing, we aim to refine the translations generated by an MT system by inserting the required terminology. To this end, we checked the translations generated by the fine-tuned model from the previous step (cf. Section \ref{sec:fine-tuning}). For each term set provided for the sentence, if the translation does not include all the terms, we ran this step of terminology insertion into the translation.

This step involves instructing ChatGPT to post-edit the translation by making sure it includes all the terms without changing the rest of the translation. For the API's parameters, we used \emph{top\_p} 1 and \emph{temperature} values 0 and 0.2, and then chose the generation that fixed more terms.

\begin{tcolorbox}[title={Example prompt: Terminology-constrained post-editing}, colbacktitle=gray, fonttitle=\footnotesize, boxrule=0.2pt, right=7pt, left=3pt]
    \begin{scriptsize}
    \begin{itemize}
    \itemindent=-13pt
    \setlength\itemsep{-0.5pt}
    \item[] In the following $<$tgt\_lang$>$ translation, use the $<$tgt\_term$>$ to translate the $<$src\_lang$>$ term $<$src\_term$>$, and the...\footnotemark \item[] Leave everything else the same.\textbackslash n\textbackslash n
    \item[] 
    \item[] $<$src\_lang$>$: $<$src\_segment$>$\textbackslash n
    \item[] $<$tgt\_lang$>$: $<$tgt\_segment$>$
    \end{itemize}
    \end{scriptsize}
\end{tcolorbox}

\footnotetext{We can add more terms, if needed.}

\section{Evaluation}

To assess the effectiveness of our process, we conducted two types of evaluation: (i) term-level evaluation in order to measure the level of adherence to the required terminology, and (ii) sentence-level evaluation in order to see whether the process \mbox{affected} the quality of the overall translation.

\subsection{Term-level Evaluation}

In Tables \ref{tab:test-used-terms} and \ref{tab:blind-used-terms}, we report the number of terms used in the translations of the test and blind datasets, respectively, in respect to the two term sets provided by the organisers. The results show the effectiveness of our proposed process, increasing the integration of the required terms in the final translations of the blind dataset from an average of 36.67\% with the baseline generic model to an average of 72.88\% after the LLM-based post-editing, across the three language pairs. Interestingly, prompting an LLM to integrate the required terms into the translations generated by a fine-tuned encoder-decoder MT model was more effective than solely using the fine-tuned model.

\subsection{Sentence-level Evaluation}

After the end of the submission phase, the organisers released the references for the participants to conduct automatic evaluation. The main purpose of this sentence-based evaluation process is to determine whether terminology integration affected the overall quality of translation. In general, as demonstrated in Table \ref{tab:blind-used-terms} and Table \ref{tab:mt-automatic-evaluation}, this terminology-constrained automatic post-editing step significantly increased the inclusion of the necessary terms into the final translation while improving translation quality across the three language pairs.

\bigbreak

\begin{table}[H]
\captionsetup{font=footnotesize,labelfont=footnotesize}
\centering
\begin{small}
\begin{tabular}{ccccccc}
\toprule
\textbf{Lang} & \textbf{System} &   \textbf{Total [1]} &  \textbf{Used [1]} &  \textbf{Total [2]} &  \textbf{Used [2]}  & \textbf{Avg~\%}    \\
\midrule
\multirow{3}{*}{\textbf{DE-EN}}  &    Baseline &           432 &          291 &           317 &          168  &  60.18\\
                        &    Fine-tuned &           432 &          302 &           317 &          165  &  60.98 \\
                        &    Term APE &           432 &          \textbf{397} &           317 &          \textbf{239}  &  \textbf{83.65} \\
                        \midrule
\multirow{3}{*}{\textbf{EN-CS}}  &    Baseline &           550 &          221 &           313 &          139  &  42.30\\
                        &    Fine-tuned &           550 &          135 &           313 &          108  &  29.53 \\
                        &    Term APE &           550 &          \textbf{466} &           313 &          \textbf{283}  &  \textbf{87.57} \\
                        \midrule
\multirow{3}{*}{\textbf{ZH-EN}}  &    Baseline &          1779 &          498 &          1938 &          491  &  26.66\\
                        &    Fine-tuned &          1779 &          854 &          1938 &          570  &  38.71 \\
                        &    Term APE &          1779 &         \textbf{1137} &          1938 &          \textbf{886}  &  \textbf{54.81} \\
                        \midrule
\multirow{3}{*}{\textbf{Avg.~\%}}& Baseline  & & & & &  43.05 \\
                        & Fine-tuned  & & & & & 43.07 \\
                        & Term APE  & & & & & \textbf{75.34} \\ \bottomrule
\end{tabular}
\end{small}
\caption{For the test dataset, the number of terms used in the translations from the first term set~[1] and the second term set~[2]. According to the results, terminology-constrained automatic post-editing (``Term APE'') using ChatGPT achieved the best adoption of the required terminology.}
\label{tab:test-used-terms}
\end{table}
\label{tab:test}

\begin{table}[H]
\captionsetup{font=footnotesize,labelfont=footnotesize}
\centering
\begin{small}
\begin{tabular}{ccccccc}
\toprule
\textbf{Lang} & \textbf{System} &   \textbf{Total [1]} &  \textbf{Used [1]} &  \textbf{Total [2]} &  \textbf{Used [2]}  & \textbf{Avg~\%}    \\
\midrule
\multirow{3}{*}{\textbf{DE-EN}}  &    Baseline  &         11357 &         4120 &         11202 &         4623  &  38.77\\
                        &    Fine-tuned &         11357 &         4130 &         11202 &         4621  &  38.81\\
                        &    Term APE &         11357 &         \textbf{6257} &         11202 &         \textbf{5893}  &  \textbf{53.85}\\
      \midrule
\multirow{3}{*}{\textbf{EN-CS}}  &    Baseline  &         10626 &         3964 &         10563 &         5122  &  42.90\\
                        &    Fine-tuned &         10626 &         3397 &         10563 &         4412  &  36.87\\
                        &    Term APE &         10626 &         \textbf{8727} &         10563 &         \textbf{8681}  &  \textbf{82.16}\\
      \midrule
\multirow{3}{*}{\textbf{ZH-EN}}  &    Baseline  &          2892 &         1375 &          2908 &          265  &  28.33\\
                        &    Fine-tuned &          2892 &         1422 &          2908 &          970  &  41.26\\
                        &    Term APE &          2892 &         \textbf{2471} &          2908 &         \textbf{2322}  &  \textbf{82.65}\\
\midrule
\multirow{3}{*}{\textbf{Avg.~\%}}&    Baseline  &          &         &          &         &  36.67\\
                        & Fine-tuned  & & & & &38.98\\
                        & Term APE& & & & &\textbf{72.88}\\ \bottomrule
\end{tabular}
\end{small}
\caption{For the blind dataset, the number of terms used in the translations from the first term set~[1] and the second term set~[2]. According to the results, terminology-based automatic post-editing (``Term APE'') using ChatGPT achieved the best adoption of the required terminology.}
\label{tab:blind-used-terms}
\end{table}

For the automatic evaluation of each MT system, we used the BLEU \citep{Papineni2002-BLEU}, chrF++ \citep{Popovic2017-chrF++}, and COMET \citep{Rei2020-COMET} metrics. Since many of the Chinese-to-English segments in the blind dataset did not have two term sets, we evaluated only those that had two term sets (1629 segments out of 2640 segments). We observe that the evaluation scores of the Chinese-to-English translation task are much lower than those of the two other language pairs. This can be due to the literary nature of the blind dataset extracted from Chinese novels, which might be difficult for both the MT model and automatic evaluation metrics.

\begin{table}[H]
\captionsetup{font=footnotesize,labelfont=footnotesize}
\centering
\begin{small}
\begin{tabular}{ccccccc}
\toprule
\textbf{Lang}   &   \textbf{Count}     &  \textbf{System}      &  \textbf{BLEU}     & \textbf{ chrF++}   &   \textbf{COMET}  \\
\toprule
\multirow{5}{*}{\textbf{DE-EN}} &  \multirow{5}{*}{2963} &  Baseline    &  19.81    &  48.04    &   21.81  \\
                                &        &  Fine-tuned  &  19.27    &  47.75    &   21.51   \\
\cmidrule{3-6}
                                &  &    Term APE [1]    & 32.36     & 60.84     &   40.25   \\
                                &  &    Term APE [2]    & 27.84     & 56.84     &   33.20   \\
                                &   &   Term APE Avg.  &  \textbf{30.10}    &  \textbf{58.84}    &   \textbf{36.73}   \\
\midrule
\multirow{5}{*}{\textbf{EN-CS}}     &   \multirow{5}{*}{3005} &  Baseline   &  29.13  &  53.11  & 50.90   \\
                                    &   &  Fine-tuned         &  24.54      &  49.14  &  33.78    \\
\cmidrule{3-6}
                                    &  &  Term APE [1]  &  45.65    &  67.36    &   79.84    \\
                                    &   &  Term APE [2] &  37.88    &  61.19   &   63.64    \\
                                    &  &  Term APE Avg. & \textbf{41.77}    & \textbf{64.28}    &   \textbf{71.74}   \\
\midrule
\multirow{5}{*}{\textbf{ZH-EN}}  &   \multirow{5}{*}{1629}   &  Baseline  &  6.95 &  27.95 & -50.90 \\
                                 &  &  Fine-tuned     & 7.76  & 29.26     &   -38.83 \\
\cmidrule{3-6}                   &   &  Term APE [1]  &  9.56 &  32.80    &   -18.96  \\
                                 &   &  Term APE [2]  &  11.93    &  35.30    &   -13.51     \\
                                 &  &  Term APE Avg. & \textbf{10.75}     & \textbf{34.05}     & \textbf{-16.24}     \\
\bottomrule
\end{tabular}
\end{small}
\caption{Automatic evaluation of the overall translation quality across the three language pairs based on the blind dataset. The ``Baseline'' refers to the OPUS model without fine-tuning, while ``Fine-tuned'' refers to the model after domain adaptation with the bilingual terminology-based synthetic data generated by an LLM. Finally, the three last rows for each language pair refer to using ChatGPT for terminology-constrained automatic post-editing (``Term APE'') of the MT output generated by the fine-tuned model. In other words, ``Term APE [1]'' indicates the results when the first term set was used to prompt ChatGPT to integrate terms of this set into the translation generated by the fine-tuned model, while ``Term APE [2]'' refers to using the second term set. Finally, ``Term APE Avg.'' is the average of ``Term APE [1]'' and ``Term APE [2]'' for each language pair. Terminology-constrained automatic post-editing with ChatGPT achieves the best results across the three language pairs in terms of the overall translation quality. As reported in Table \ref{tab:blind-used-terms}, the number of terms integrated after the automatic post-editing step also increased.}
\label{tab:mt-automatic-evaluation}
\end{table}

Moreover, it is worth noting that we used the English term while generating bilingual synthetic data (cf. Section \ref{sec:data-generation}) for the three language pairs. However, English is the target language for both Chinese-to-English and German-to-English language directions, while it is the source language for the English-to-Czech language direction. This can explain the performance degradation after the fine-tuning step in the English-to-Czech language direction (cf. Tables \ref{tab:blind-used-terms} and \ref{tab:mt-automatic-evaluation}). In other words, it is recommended in the step of bilingual synthetic data generation to either use the target term or both the source and target terms while prompting the LLM to generate translation pairs.

As explained in Section \ref{sec:term-ape}, our final step of terminology-constrained automatic post-editing involves instructing an LLM to insert terms that were missing from the output of the fine-tuned model. This significantly increased term usage across all the Chinese-to-English, English-to-Czech, and German-to-English language pairs (cf. Table \ref{tab:blind-used-terms}). Furthermore, as demonstrated in Table \ref{tab:mt-automatic-evaluation}, this step had no detrimental effects on translation quality. In fact, integrating the necessary terms into the translation using ChatGPT improved translation quality according to our automatic evaluation.

\section{Conclusion and Future Work}

In this work, we showed that applying a multistep process of mixed fine-tuning on terminology-based synthetic bilingual data and then terminology-constrained automatic post-editing with an LLM can increase the adherence to the pre-approved terms in the generated translations. By the end of the process, the use of the required terms has increased in the translations of the blind dataset across the three language pairs from an average of 36.67\% with the baseline generic model to an average of 72.88\% after instructing an LLM to integrate the required terms into the translations.

Due to the task restrictions, we had to fine-tune OPUS models only. We would like to experiment with fine-tuning NLLB models, and probably the new SeamlessM4T \citep{Barrault2023-SeamlessM4T}, Mistral \citep{Jiang2023-Mistral}, and MADLAD-400 models \citep{Kudugunta2023-MADLAD}, on the same data and compare the output quality. In our experiments, we employed ChatGPT ``gpt-3.5-turbo'' for both terminology-based synthetic data generation and terminology-constrained automatic post-editing, as it is a relatively efficient and cost-effective option. In the future, we would like to repeat the same experiments with GPT-4 in order to assess the benefit of using a stronger language model on overall performance. We observe that BLOOM can be used as an alternative LLM for data generation; however, one-shot generation might work better than zero-shot generation. In this case, the prompt can consist of a term, a bilingual sentence pair, and then another term. Interestingly, the model will predict a new translation pair including the second term. While certain open-source models such as Llama~2 and Falcon might be employed for the terminology-constrained automatic post-editing step for certain languages, we suspect that they will need fine-tuning before being reliably usable for most languages.

In future work, we will carry out a deeper analysis\footnote{It is recommended to ask professional linguists to analyse both the synthetic data and output translations, assess the quality, and report common linguistic characteristics.} of the generated synthetic data together with the outputs of the fine-tuned models in order to understand how the properties of the synthetic data affect the fine-tuning results. It is important also to test the same approach for other languages, especially low-resource language pairs.

Moreover, it would be interesting to exclude the fine-tuning step and assess the overall translation quality after LLM-based post-editing. It is possible that domain adaptation through fine-tuning the baseline MT model either on authentic or synthetic data would still be beneficial. It can lead to domain-specific improvements in the overall translation quality that may not be achievable by the baseline model or the terminology-constrained post-editing step. Again, deploying a model fine-tuned on in-domain data into production can enhance terminology adherence in initial translations. As there is no need to send the translations that already include the pre-approved terms to the LLM for terminology-constrained post-editing, this can reduce the number of translations that require post-editing. Such an efficient workflow can allow us to save resources, and minimise latency at inference time. Similarly, there are potential advantages of employing an LLM for post-editing rather than for direct translation. Instead of solely relying on the translation quality of the LLM, quality estimation can be performed to select the best MT model in general or for the current source text segment. \mbox{Ultimately,} only segments that do not meet quality criteria are then passed to the LLM for post-editing.

\end{large}

\begin{large}

\chapter{Conclusions and Future Work}
\label{chapter:conclusion}

The previous chapters explained how I addressed the two main research questions and elaborated on the findings of my research. This chapter concludes the accomplished progress and potential future work.

\section{Conclusions}

In the first research question, I examined how to leverage language modelling techniques in general and LLMs in particular to improve translation features that involve human interaction and continuous feedback, such as adaptive MT, terminology-constrained MT, domain-aware automatic post-editing, auto-suggestion and auto-completion. In the second research question, I addressed a common scenario in the translation industry, namely receiving highly specialised projects, where there is hardly any parallel in-domain data. In such scenarios where there is insufficient in-domain data to fine-tune MT models, producing translations that are consistent with the relevant context is challenging. One way to address this question is through domain-specific text generation with LLMs and then fine-tuning an MT model using the generated bilingual in-domain synthetic data. Furthermore, this question overlaps with the first question, since real-time adaptive MT with LLMs can serve as an alternative approach to address in-domain data scarcity.

In Chapter \ref{chapter:generation}, I proposed a novel approach to domain adaptation leveraging state-of-the-art pretrained LLMs for \mbox{domain-specific} data augmentation for MT, simulating the domain characteristics of either (a) a small bilingual dataset, or (b) the monolingual source text to be translated. Combining this idea with back-translation, I was able to generate huge amounts of synthetic bilingual \mbox{in-domain} data for both use-cases. For this investigation, I used the \mbox{state-of-the-art} Transformer architecture. I employed mixed fine-tuning to train models that significantly improve translation of in-domain texts. In this work, I suggested potential future work ideas, such as terminology-aware text generation, as well as low-resource and multilingual text generation for domain-specific MT, some of which I have already addressed in the next papers/chapters.

In Chapter \ref{chapter:autosuggest}, I described our submissions to WMT 2022 Shared Task on Word-level Auto-completion, for the Chinese-to-English, English-to-Chinese,
German-to-English, and English-to-German language directions. I discussed how utilising random sampling to generate diverse alternatives can reveal good results. Moreover, the survey I conducted shows that suggesting alternatives can inspire translators and limit their need to refer to external resources, which hopefully boosts their productivity. In the future, I intend to expand this work by investigating whether domain adaptation and/or simulated annealing can improve translation suggestions. In a user survey I conducted (cf. Chapter 3), 90.2\% of the participants stated that they believed that word-level auto-completion was helpful. This can be considered an indicator that users appreciate adaptivity and interactivity features in MT system.

Chapter \ref{chapter:adaptive-MT} emphasises the significance of consistency for high-quality translation, especially in domain-specific projects. It addresses the challenge of real-time adaptation through leveraging LLMs. In-context learning involves replicating text generation patterns without additional fine-tuning. Inspired by this idea, I explored the use of in-context learning to improve real-time adaptive MT. I investigated whether prompting LLMs with similar translations (fuzzy matches) and/or terminology enables them to simulate domain and style characteristics. Experiments strongly suggest that LLMs can adapt to domain-specific translation pairs and terminology, and that the quality of adaptive translations can surpass that of strong encoder-decoder MT systems, especially for high-resource languages.

In Chapter \ref{chapter:terminology}, I described the systems submitted to the WMT 2023 Terminology Shared Task for German-to-English, English-to-Czech, and Chinese-to-English language pairs. The aim of the task was to improve MT by developing systems that can accurately translate technical terms. The methods used involve leveraging LLMs to generate synthetic bilingual terminology-based data and to post-edit translations though inserting missing terminology. The process consists of four steps: generating terminology-based synthetic data with an LLM, fine-tuning a generic encoder-decoder MT model, generating translations with the fine-tuned model, and finally using an LLM to fix the translations that do not include the required terms. The results indicate that the use of pre-approved terms in the translations doubles across the three language pairs, which is very encouraging. 

Revisiting my research questions, this research was able to answer both of the two main research questions and their sub-questions. For the first research question, this work explored a number of scenarios to employ language models to improve the quality of adaptive MT at inference time, and achieved performance gains based on both human and automatic evaluation. Chapter \ref{chapter:autosuggest} demonstrated that by only utilising the autoregressive property of NMT models through teacher forcing, the performance of word-level auto-completion can surpass other approaches, without using any external models.\footnote{Our systems achieved the first and second places at WMT 2023 Shared Task on Word-level Auto-completion.} Chapter \ref{chapter:adaptive-MT} showed that LLM in-context learning can enhance real-time adaptive MT, leveraging user (translator) feedback and modifications. These models (e.g. ChatGPT, BLOOM, Mistral) can adapt to approved in-domain translation pairs and/or terminology while translating  new texts. Moreover, Chapter \ref{chapter:terminology} demonstrated that LLMs can be prompted for terminology-constrained automatic post-editing of translations generated by MT models, which doubled the integration of the pre-approved terminology into the final translations. In other words, instead of inputting a raw source sentence to translate, providing more linguistic information improves translation quality. As for the second research question, this work demonstrated that pre-trained LLMs can be used for in-domain data augmentation to improve domain-specific MT models. In-domain texts or terminology can be fed to an LLM to generate more data, simulating the domain characteristics of the original text. As demonstrated by Chapter \ref{chapter:generation} and Chapter \ref{chapter:terminology}, if there is only a small in-domain dataset (e.g. 1000 sentences) or a glossary, LLMs can be used to generate large domain-specific datasets (e.g. 200,000 sentences). Then, the generic model can be fine-tuned on this synthetic data, which improves translation quality of in-domain texts.

This research is beneficial for translators and MT providers alike. For translators, it aims to provide them with better flexibility and consistency while translating. Boosting MT real-time adaptation can be considered one method of enhancing the user-friendliness of translation environments. Once a linguist edits a translation, the system should comply with the approved modifications in subsequent translations. For MT providers, adaptability is essential for serving a wider range of customers from diverse domains. While some MT providers offer the feature of fine-tuning on customised data, giving their clients the ability to adapt an MT system out of the box without any further fine-tuning (or on top of fine-tuning) is an added benefit. Language service providers seek to achieve the consistency required by their clients efficiently. However, new clients might not have sufficient TMs if any, and access to TMs owned by other clients is restricted by ethics and laws of intellectual property. Hence, I believe that the ability to employ both in-domain synthetic data for fine-tuning, and in-context learning for real-time adaptivity can help improve the productivity and satisfaction of both linguists and clients.

\section{Future Work}

The design and execution of these experiments were guided by a commitment to making the research findings highly applicable and beneficial to real-world production scenarios. For instance, in Chapter \ref{chapter:adaptive-MT}, I tried to address a number of common use cases, such as real-time sentence-level translation adaptivity and terminology adherence. Moreover, throughout my research, I covered a wide range of languages with diverse scripts and availability of data resources, exploring the opportunities and limitations associated with these languages. While Chapter \ref{chapter:adaptive-MT} addresses using LLMs for both translation and real-time adaptation to fuzzy matches and terminology, Chapter \ref{chapter:terminology} investigates using LLMs for terminology-constrained automatic post-editing of translations generated by other MT systems. As shown in Chapter \ref{chapter:adaptive-MT}, when the original translation quality of an LLM is much lower than that produced by an (encoder-decoder) MT model for a language pair, starting from this strong baseline and trying to improve it can reveal better results (cf. Section \ref{sec:fix-mt}). Nevertheless, this depends on the language and its level of support by the LLM. As stated in Chapter \ref{chapter:adaptive-MT} and Chapter \ref{chapter:terminology}, real-time adaptive MT is not a replacement for fine-tuning. Hence, starting with fine-tuning, when possible, should be more efficient as the original translation quality of a model will be better, which minimises the need for post-editing. As serving multiple models in production can be challenging, the optimal scenario would be to fine-tune an LLM on both zero-shot and few-shot translation data. As demonstrated in Chapter \ref{chapter:finetuning}, this can help improve not only its ability to produce high-quality zero-shot translation, but also to improve its in-context learning capability to adapt to a range of translation-related instructions.

In the future, I would like to further focus on relevant topics, elaborated in the following sections.

\subsection{MT adaptation for multilingual and low-resource settings}

I aim to investigate the aforementioned approaches in multilingual and/or low-resource settings. During my PhD, I worked with more than 20 language pairs. In respect of my research questions, I already experimented with some low-resource languages, such as Czech and Kinyarwanda. Moreover, I co-authored a couple of works on MT for low-resource languages. For example, in our work \citep{Haque2020-Terminology}, we applied domain adaptation to an English-to-Hindi MT system. To this end, we used terminology to select relevant target sentences for the AI domain, generated the sources with back-translation, and finally applied mixed fine-tuning of the generic model, which improved the translation quality for the AI domain while retaining the translation quality of generic texts. Similarly, in our work \citep{Oktem2022-Ladino}, we built NMT systems for a very low-resource language, Ladino, utilising diverse data augmentation approaches, employing rule-based MT and back-translation. We collected data in English-to-Spanish and Turkish-to-Spanish language pairs, since Spanish shares a wide range of linguistic characteristics with Ladino. Then, we used a rule-based system to translate Spanish to Ladino. This process resulted in English-to-Synthetic-Ladino and Turkish-to-Synthetic-Ladino datasets, which I used later for building NMT models. Furthermore, I have experience with building a multilingual MT model\footnote{Notes on multilingual MT: \url{https://blog.machinetranslation.io/multilingual-nmt/}} for 10 low-resource Indic languages.\footnote{MT demo: \url{https://www.machinetranslation.io/}} Multilingual models can exploit the similarity between languages and particularly benefit
low-resource languages \citep{Imankulova2019-MT-Multi-Low-resource,Liu2020-mBART,Gala2023-IndicTrans2}.

Text generation for low-resource languages can be challenging, since some major LLMs lack decent support for such languages \citep{Moslem2023-AdaptiveMT,Robinson2023-ChatGPT-MT}. As the quality of synthetic data generated affects the quality of the MT model fine-tuned on such data, applying the same approach I used in my research \citep{Moslem2022-MT-LM} might not work out of the box. Among the research directions to consider is to start by fine-tuning the LLM on monolingual data in the low-resource target language or a mix of monolingual data in both the source and target languages. Then the fine-tuned model can be used for text generation. Nevertheless, it is always important to evaluate the generated synthetic data through both human linguistic analysis and automatic quality estimation before using it for fine-tuning.

Similarly, real-time adaptive MT for low-resource languages might benefit from the availability of target-side monolingual data. Hence, instead of feeding the model with similar sentence pairs, monolingual sentences with higher similarity to the source can be used to adapt the translation. This would require multilingual embeddings to calculate similarity between monolingual data in the target language and the input text in the source language. Moreover, I recommend experimenting with extracting chunks of short bilingual phrases of a low-resource language, and then prompting an LLM to use these phrases either during translation or text generation. Augmenting NMT systems with phrases can be useful in diverse scenarios \citep{Gupta2021-Phrases,Ghazvininejad2023-DictionaryMT,Puduppully2023-DecoMT-LLM}.

\subsection{Domain-specific word-level autosuggestions and autocompletion}

In Chapter \ref{chapter:autosuggest}, I concluded that random sampling can improve word-level auto-completion using generic NMT models. In the future, I hope to investigate the effect of incorporating domain adaptation into this process. Potentially, fine-tuning the baseline model on in-domain data can achieve better performance on the test dataset of the same domain, especially when combined with random sampling.

\subsection{Analysis of critical errors and text generation for difficult-to-translate words}

The core of this idea is first defining critical errors in translation of in-domain texts (e.g. in the medical domain) and then fixing these errors. In their work, \citet{Fadaee2018-Difficult-Words} introduced variations of sampling strategies targeting difficult-to-predict words using prediction losses and frequencies of words. In addition, they also target the contexts of difficult words and sample sentences that are similar in context. They proposed a two-stage approach, identifying difficult words and sampling with the objective of increasing occurrences of these words, and identifying contexts where these words are difficult to predict and sample sentences similar to the difficult contexts. With targeted sampling of sentences for back-translation, their approach achieved improvements of up to 1.7 BLEU points over back-translation using random sampling. Similarly, \citet{Huck2019-MT-OOV} started by translating the new text and finding out-of-vocabulary words (OOVs). Then, they translated them using bilingual word embeddings. Afterwards, they used the 5-best target words as queries to mine target sentences, and back-translated these sentences, forcing the back-translation of each of the five proposed target OOVs to be the original source OOV. Finally, they used this synthetic data to fine-tune the system. As a result, the translation of OOVs was significantly improved. To boost the quality of target-to-source word alignments of attention-based models, researchers proposed guided alignment training \citep{Chen2016-MT-Align,Garg2019-MT-Align}.

In this sense, we can improve the robustness of the MT model by finding such difficult words, and then using them or their full sentences for text generation, or for prompting LLMs through in-context learning. The idea is similar to my previous work \citep{Moslem2023-AdaptiveMT,Moslem2023-Terminology}, where I used pre-approved terminology to enhance the translation quality. However, it goes beyond terminology to consider other error categories. For example, these problematic words can be named entities (e.g. person, location, organisation, product) in the medical domain, or untranslatables (e.g. websites, email addresses, or phone numbers of hospitals in a certain country). Words or phrases that are difficult to translate can be automatically defined as in \citet{Fadaee2018-Difficult-Words}, or using human annotation of a sample dataset. They can also be extracted from localisation comments that are originally authored by software developers to provide rules or hints for localisers. While the idea can be employed to improve MT models through leveraging LLMs, the same concept can be applied to knowledge distillation from a larger teacher LLM to a smaller student LLM to enhance efficiency.

\subsection{Improving random sampling via simulated annealing}

For word-level auto-completion and to complement our previous work \citep{Moslem2022-WLAC}, I will investigate using simulated annealing. In the beginning, we will allow the search process to explore less-probable options, and it will not always choose the token with the highest probability. Towards the end of the translation, the model will usually choose the option with the highest probability. Sampling temperature and potentially top-k/top-p parameters can control this. This idea can also be investigated to improve the quality of text generation using pre-trained language models \citep{Moslem2022-MT-LM}. 

\subsection{Domain-Specific Efficient Fine-tuning of LLMs}

Several researchers have demonstrated the abilities of LLMs for diverse translation tasks \citep{Akter2023-Gemini,Bawden2023-BLOOM-MT,Cheng2023-MT-LLM-SCALE,He2023-MT-LLM-Human,Hendy2023-LLM-MT,Hoang2023-MT-LLM-Ensemble,Jiao2023-LLM-MT,Koneru2023-MT-LLM-MTPE,Kumar2023-MT-LLM-CTQScorer,Moslem2023-AdaptiveMT,Peng2023-MT-LLM,Sia2023-LLM-MT,Tan2023-MT-LLM,Vilar2023-PaLM-MT,Wang2023-DocumentMT,Zhu2023-LLM-MT}. Nevertheless, fine-tuning LLMs specifically for MT tasks can help improve translation quality and efficiency \citep{Moslem2023-AdaptiveMT-LLM-Finetuning}. So far, training LLMs to work better with MT mostly involves either: (i) pre-training an LLM to work with zero-shot MT \citep{Li2023-MT-LLM-MT2,Schioppa2023-LLM-MT-CrossLingual}; or (ii) fine-tuning LLMs with the purpose of improving zero-shot capabilities \citep{Sia2022-LLM-MT,Alves2023-LLM-MT-Finetuning,Iyer2023-LLMs-Disambiguation,Jiao2023-Parrot,Li2023-MT-LLM-Multilingual-Finetuning,Xu2023-Llama-Finetuning,Yang2023-BigTranslate,Zhang2023-MT-Efficient-Finetuning}. Moreover, there are several works that investigate pre-training or fine-tuning encoder-decoder MT models for adaptive MT (cf. Section \ref{sec:research-context}: \textit{Research Context}), and there is at least one work that compares this with using in-context learning of LLMs for adaptive MT \citep{Reinauer2023-MT-ICL}. However, there is still a need for research that instead investigates fine-tuning available open-source models to enhance their in-context learning ability for real-time adaptive MT and compares this to current approaches. To this end, these models can be fine-tuned to perform better at in-context learning scenarios, where special prompt templates incorporate in-domain sentences, phrases, or terminology. In Chapter \ref{chapter:finetuning}, I demonstrated that with a relatively small dataset that incorporates a mix of zero-shot and one-shot prompts, fine-tuning significantly enhances Mistral's in-context learning ability, especially for real-time adaptive MT \citep{Moslem2023-AdaptiveMT-LLM-Finetuning}. This direction can improve both translation quality and efficiency, especially as fewer examples might be required for in-context learning.  Moreover, both inference and fine-tuning of the largest LLMs can be challenging. Hence, knowledge distillation approaches that transfer robust in-context learning features from stronger teachers to smaller students for the purpose of adaptive and interactive MT should be investigated.

Accordingly, I can start by fine-tuning an LLM for zero-shot, one-shot, two-shot translation, and then build on this. The model used can be at least one encoder-decoder model, e.g. NLLB-200 \citep{NLLB2022}, and one decoder-only model, e.g. BLOOM \citep{BLOOM2022}. Furthermore, I can compare other models like Llama \citep{Touvron2023-Llama2}, Falcon (EN-DE-ES-FR) \citep{Penedo2023-Falcon}, Mistral/Mixtral (EN-DE-ES-FR-IT) \citep{Jiang2023-Mistral}, Jais (AR-EN) \citep{Sengupta2023-Jais}, Baichuan (ZH) \citep{Yang2023-Baichuan}, and/or Qwen (ZH) \citep{Bai2023-Qwen}. I am interested in conducting an in-depth investigation into fine-tuning LLMs to enhance their domain-specific in-context learning capability, and improve their ability to work better in adaptive MT scenarios. To this end, I would like to conduct a range of experiments, including: instruction fine-tuning on the in-domain data only; instruction mixed fine-tuning on the in-domain data and a randomly sampled portion of the original generic data, over-sampling the in-domain data; and investigating diverse scenarios of boosting in-context learning for adaptive MT, ranging from zero-shot and few-shot translation to integration of terminology and in-domain monolingual data.



The aforementioned experiments should employ efficient fine-tuning approaches \citep{Wan2023-EfficientLLMs} such as QLoRA \citep{Hu2021-LoRA,Dettmers2023-QLoRA}. I would like to apply the experiments to a range of diverse languages, including high-resource, medium-resource, and low-resource languages, as well as different domains. Preliminary experiments in this direction demonstrate promising results (cf. Chapter \ref{chapter:finetuning}).

Among the benefits of fine-tuning open-source LLMs are efficient self-hosting. In other words, those who would like to serve their own LLMs for privacy reasons can utilise an open-source model, and efficiently achieve quality gains comparable to those of strong commercial LLMs. Moreover, so far, for very high-resource languages, LLMs can be used independently. However, for other languages, a hybrid approach using both conventional MT models and LLMs leads to better results, which means we have to deploy/use two models at translation time. If fine-tuning a small ``standalone'' LLM is possible for both regular (zero-shot) and adaptive (one-shot or few-shot) translation, this would be much more efficient. In addition, researchers can definitely build on this direction rather than having to rely on closed models. Open-source research can lead to more interpretability as we know better what is going on in the background.


\subsection{Cross-Lingual Retrieval-Augmented Generation}
\label{sec:cross-lingual-retrieval}

Retrieval-augmented MT or cross-lingual generation is crucial in diverse scenarios. For example, legal companies receive huge amounts of documents in one language while they are operating in another language. In such case, they want a quick way to retrieve information in their working language. However, translating them manually is time-consuming and expensive. It can lead to inaccuracies and delays due to the complexity and specificity of legal terminology. Similarly, refugees and immigrants need to retrieve information, e.g. about health and education, in the hosting countries. However, they might require mass information from different countries, and they need to receive this information in their native languages.

Advances in technology have made it possible to use MT for this purpose to quickly translate large volumes of text, making it easier for legal professionals to access the information they need. Due to the specificity and importance of legal documents, these translations often still need to be reviewed or edited by human translators. However, if information retrieval is successfully used, only the documents that matter most to the case at hand can be professionally translated.

Employing information retrieval approaches such as semantic search entails that users would receive results that refer them to the most probable documents. Such an outcome can suffice if the user is seeking information in the style provided by search engines. Nowadays, users might prefer retrieving information in a more comprehensive way that probably summarises the whole document, relevant portions of the document, or even multiple documents, in a cross-lingual manner. 

In this future work, I aim to understand the effect of using multilingual information retrieval approaches in combination with LLMs to retrieve information based on legal or medical queries. In other words, I would like to investigate the effect of initially employing MT before retrieving information with semantic search versus merely using multilingual embeddings, while the final output is generated by a multilingual LLM. Analysing the quality of the generated output by the LLM based on the retrieved information can help understand whether explicit MT is still required, or whether the use of cross-lingual semantic search based on multilingual embedding is sufficient while using a multilingual LLM to generate the final output.

\vspace*{\fill}

\end{large}

\break
\renewcommand{\thepage}{}
\pagestyle{empty}
\bibliographystyle{hapalike}
\bibliography{paperpile,software}

\begin{thebibliography}{}

\bibitem[Agrawal et~al., 2023]{Agrawal2023-SelectionMT}
Sweta Agrawal, Chunting Zhou, Mike Lewis, Luke Zettlemoyer, and Marjan Ghazvininejad (2023).
\newblock {In-context Examples Selection for Machine Translation}.
\newblock In {\em {Findings of the Association for Computational Linguistics: ACL 2023}}, pages 8857--8873, Toronto, Canada. Association for Computational Linguistics.

\bibitem[Akter et~al., 2023]{Akter2023-Gemini}
Syeda~Nahida Akter, Zichun Yu, Aashiq Muhamed, Tianyue Ou, Alex B{\"a}uerle, {\'A}ngel~Alexander Cabrera, Krish Dholakia, Chenyan Xiong, and Graham Neubig (2023).
\newblock {An in-depth look at Gemini's language abilities}.
\newblock \emph{arXiv preprint arXiv:2312.11444 [cs.CL]}.

\bibitem[Alves et~al., 2023]{Alves2023-LLM-MT-Finetuning}
Duarte Alves, Nuno Guerreiro, Jo{\~a}o Alves, Jos{\'e} Pombal, Ricardo Rei, Jos{\'e} de~Souza, Pierre Colombo, and Andre Martins (2023).
\newblock {Steering Large Language Models for Machine Translation with Finetuning and In-Context Learning}.
\newblock In {\em {Findings of the Association for Computational Linguistics: EMNLP 2023}}, pages 11127--11148, Singapore. Association for Computational Linguistics.

\bibitem[Anastasopoulos et~al., 2020]{Anastasopoulos2020-TICO-19}
Antonios Anastasopoulos, Alessandro Cattelan, Zi-Yi Dou, Marcello Federico, Christian Federmann, Dmitriy Genzel, Franscisco Guzm{\'a}n, Junjie Hu, Macduff Hughes, Philipp Koehn, Rosie Lazar, Will Lewis, Graham Neubig, Mengmeng Niu, Alp {\"O}ktem, Eric Paquin, Grace Tang, and Sylwia Tur (2020).
\newblock {{TICO}-19: the Translation Initiative for {CO}vid-19}.
\newblock In {\em {Proceedings of the 1st Workshop on {NLP} for {COVID}-19 (Part 2) at {EMNLP} 2020}}, Online. Association for Computational Linguistics.

\bibitem[Antoun et~al., 2021]{Antoun2021-ct}
Wissam Antoun, Fady Baly, and Hazem Hajj (2021).
\newblock {{A}ra{GPT}2: Pre-Trained Transformer for {A}rabic Language Generation}.
\newblock In {\em {Proceedings of the Sixth Arabic Natural Language Processing Workshop}}, pages 196--207, Kyiv, Ukraine (Virtual). Association for Computational Linguistics.

\bibitem[Axelrod et~al., 2011]{Axelrod2011-Data}
Amittai Axelrod, Xiaodong He, and Jianfeng Gao (2011).
\newblock {Domain Adaptation via Pseudo In-Domain Data Selection}.
\newblock In {\em {Proceedings of the 2011 Conference on Empirical Methods in Natural Language Processing}}, pages 355--362, Edinburgh, Scotland, UK. Association for Computational Linguistics.

\bibitem[Bahdanau et~al., 2015]{Bahdanau2015-JointlyLearn}
Dzmitry Bahdanau, Kyunghyun Cho, and Yoshua Bengio (2015).
\newblock {Neural Machine Translation by Jointly Learning to Align and Translate}.
\newblock In {\em {Proceedings of the 3rd International Conference on Learning Representations, ICLR 2015}}, San Diego, CA, USA.

\bibitem[Bai et~al., 2023]{Bai2023-Qwen}
Jinze Bai, Shuai Bai, Yunfei Chu, Zeyu Cui, Kai Dang, Xiaodong Deng, Yang Fan, Wenbin Ge, Yu Han, Fei Huang, Binyuan Hui, Luo Ji, Mei Li, Junyang Lin, Runji Lin, Dayiheng Liu, Gao Liu, Chengqiang Lu, Keming Lu, Jianxin Ma, Rui Men, Xingzhang Ren, Xuancheng Ren, Chuanqi Tan, Sinan Tan, Jianhong Tu, Peng Wang, Shijie Wang, Wei Wang, Shengguang Wu, Benfeng Xu, Jin Xu, An Yang, Hao Yang, Jian Yang, Shusheng Yang, Yang Yao, Bowen Yu, Hongyi Yuan, Zheng Yuan, Jianwei Zhang, Xingxuan Zhang, Yichang Zhang, Zhenru Zhang, Chang Zhou, Jingren Zhou, Xiaohuan Zhou, and Tianhang Zhu (2023).
\newblock {Qwen Technical Report}.
\newblock \emph{arXiv preprint arXiv:2309.16609 [cs.CL]}.

\bibitem[Barrault et~al., 2023]{Barrault2023-SeamlessM4T}
Loic Barrault, Andy Chung, David Dale, Ning~Dong (ai), Paul-Ambroise Duquenne, Hady Elsahar, Hongyu Gong, Kevin Heffernan, John Hoffman, Christopher Klaiber, Peng-Jen Chen, Daniel Licht, Jean Maillard, Alice Rakotoarison, Kaushik~Ram Sadagopan, Guillaume Wenzek, Abinesh Ramakrishnan, Alexandre Mourachko, Amanda Kallet, Ann Lee, Anna Sun, Bapi Akula, Benjamin Peloquin, Bernie Huang, Bokai Yu, Brian Ellis, Can Balioglu, Carleigh Wood, Changhan Wang, Christophe Ropers, Cynthia Gao, Daniel~Li (fair), Elahe Kalbassi, Ethan Ye, Gabriel~Mejia Gonzalez, Hirofumi Inaguma, Holger Schwenk, Igor Tufanov, Ilia Kulikov, Janice Lam, Jeff~Wang (pm Ai), Juan Pino, Justin Haaheim, Justine Kao, Prangthip Hasanti, Kevin Tran, Maha Elbayad, Marta~R Costa-jussa, Mohamed Ramadan, Naji El~Hachem, Onur {\c C}elebi, Paco Guzm{\'a}n, Paden Tomasello, Pengwei Li, Pierre Andrews, Ruslan Mavlyutov, Russ Howes, Safiyyah Saleem, Skyler Wang, Somya Jain, Sravya Popuri, Tuan Tran, Vish Vogeti, Xutai Ma, and Yilin Yang (2023).
\newblock {SeamlessM4T---Massively Multilingual \& Multimodal Machine Translation}.
\newblock Meta AI.
\newblock {\em Meta AI}.

\bibitem[Bawden et~al., 2020]{Bawden2020-WMT-Biomedical}
Rachel Bawden, Giorgio~Maria Di~Nunzio, Cristian Grozea, Inigo Jauregi~Unanue, Antonio Jimeno~Yepes, Nancy Mah, David Martinez, Aur{\'e}lie N{\'e}v{\'e}ol, Mariana Neves, Maite Oronoz, Olatz Perez-de Vi{\~n}aspre, Massimo Piccardi, Roland Roller, Amy Siu, Philippe Thomas, Federica Vezzani, Maika Vicente~Navarro, Dina Wiemann, and Lana Yeganova (2020).
\newblock {Findings of the WMT 2020 biomedical translation shared task: Basque, Italian and Russian as new additional languages}.
\newblock In {\em {Proceedings of the Fifth Conference on Machine Translation}}, pages 660--687, Online. Association for Computational Linguistics.

\bibitem[Bawden and Yvon, 2023]{Bawden2023-BLOOM-MT}
Rachel Bawden and Fran{\c c}ois Yvon (2023).
\newblock {Investigating the Translation Performance of a Large Multilingual Language Model: the Case of {BLOOM}}.
\newblock In {\em {Proceedings of the 24th Annual Conference of the European Association for Machine Translation}}, pages 157--170, Tampere, Finland. European Association for Machine Translation.

\bibitem[Berzins et~al., 2019]{ELRC2019}
Aivars Berzins, Khalid Choukri, Maria Giagkou, Andrea L{\"o}sch, H{\'e}l{\`e}ne Mazo, Stelios Piperidis, Micka{\"e}l Rigault, Eileen Schnur, Lilli Smal, Josef van Genabith, and Andrejs Vasiljevs (2019).
\newblock {\em {ELRC White Paper: Sustainable Language Data Sharing to Support Language Equality in Multilingual Europe: why Language Data Matters}}.
\newblock European Language Resource Coordination \& OVD Verlag.

\bibitem[Black et~al., 2022]{Black2022-GPT-NeoX}
Sidney Black, Stella Biderman, Eric Hallahan, Quentin Anthony, Leo Gao, Laurence Golding, Horace He, Connor Leahy, Kyle McDonell, Jason Phang, Michael Pieler, Usvsn~Sai Prashanth, Shivanshu Purohit, Laria Reynolds, Jonathan Tow, Ben Wang, and Samuel Weinbach (2022).
\newblock {{GPT}-{N}eo{X}-20{B}: An Open-Source Autoregressive Language Model}.
\newblock In {\em {Proceedings of BigScience Episode \#5 -- Workshop on Challenges \& Perspectives in Creating Large Language Models}}, pages 95--136, Stroudsburg, PA, USA. Association for Computational Linguistics.

\bibitem[Bogoychev and Sennrich, 2019]{Bogoychev2019-Translationese}
Nikolay Bogoychev and Rico Sennrich (2019).
\newblock {Domain, Translationese and Noise in Synthetic Data for Neural Machine Translation}.
\newblock \emph{arXiv preprint arXiv:1911.03362 [cs.CL]}.

\bibitem[Borgeaud et~al., 2021]{Borgeaud2021-kNN-LM}
Sebastian Borgeaud, Arthur Mensch, Jordan Hoffmann, Trevor Cai, Eliza Rutherford, Katie Millican, George van~den Driessche, Jean-Baptiste Lespiau, Bogdan Damoc, Aidan Clark, Diego de~Las~Casas, Aurelia Guy, Jacob Menick, Roman Ring, Tom Hennigan, Saffron Huang, Loren Maggiore, Chris Jones, Albin Cassirer, Andy Brock, Michela Paganini, Geoffrey Irving, Oriol Vinyals, Simon Osindero, Karen Simonyan, Jack~W Rae, Erich Elsen, and Laurent Sifre (2021).
\newblock {Improving language models by retrieving from trillions of tokens}.
\newblock \emph{arXiv preprint arXiv:2112.04426 [cs.CL]}.

\bibitem[Britz et~al., 2017]{Britz2017-domain-mixing}
Denny Britz, Quoc Le, and Reid Pryzant (2017).
\newblock {Effective Domain Mixing for Neural Machine Translation}.
\newblock In {\em {Proceedings of the Second Conference on Machine Translation}}, pages 118--126, Copenhagen, Denmark. Association for Computational Linguistics.

\bibitem[Brown et~al., 2020]{Brown2020-GPT-3}
Tom~B Brown, Benjamin Mann, Nick Ryder, Melanie Subbiah, Jared Kaplan, Prafulla Dhariwal, Arvind Neelakantan, Pranav Shyam, Girish Sastry, Amanda Askell, Sandhini Agarwal, Ariel Herbert-Voss, Gretchen Krueger, Tom Henighan, Rewon Child, Aditya Ramesh, Daniel~M Ziegler, Jeffrey Wu, Clemens Winter, Christopher Hesse, Mark Chen, Eric Sigler, Mateusz Litwin, Scott Gray, Benjamin Chess, Jack Clark, Christopher Berner, Sam McCandlish, Alec Radford, Ilya Sutskever, and Dario Amodei (2020).
\newblock {Language Models are Few-Shot Learners}.
\newblock In {\em {Advances in Neural Information Processing Systems (NeurIPS 2020)}}, volume~33, pages 1877--1901, Virtual. Curran Associates, Inc.

\bibitem[Bulte and Tezcan, 2019]{Bulte2019-fuzzy}
Bram Bulte and Arda Tezcan (2019).
\newblock {Neural Fuzzy Repair: Integrating Fuzzy Matches into Neural Machine Translation}.
\newblock In {\em {Proceedings of the 57th Annual Meeting of the Association for Computational Linguistics}}, pages 1800--1809, Florence, Italy. Association for Computational Linguistics.

\bibitem[Burlot and Yvon, 2018]{Burlot2018-Monolingual}
Franck Burlot and Fran{\c c}ois Yvon (2018).
\newblock {Using Monolingual Data in Neural Machine Translation: a Systematic Study}.
\newblock In {\em {Proceedings of the Third Conference on Machine Translation: Research Papers}}, pages 144--155, Brussels, Belgium. Association for Computational Linguistics.

\bibitem[Cai et~al., 2021]{Cai2021-MonolingualTM}
Deng Cai, Yan Wang, Huayang Li, Wai Lam, and Lemao Liu (2021).
\newblock {Neural Machine Translation with Monolingual Translation Memory}.
\newblock In {\em {Proceedings of the 59th Annual Meeting of the Association for Computational Linguistics and the 11th International Joint Conference on Natural Language Processing (Volume 1: Long Papers)}}, pages 7307--7318, Online. Association for Computational Linguistics.

\bibitem[Casacuberta et~al., 2022]{Casacuberta2022-2022}
Francisco Casacuberta, George Foster, Guoping Huang, Philipp Koehn, Geza Kovacs, Lemao Liu, Shuming Shi, Taro Watanabe, and Chengqing Zong (2022).
\newblock {Findings of the Word-Level {A}uto{C}ompletion Shared Task in {WMT} 2022}.
\newblock In {\em {Proceedings of the Seventh Conference on Machine Translation (WMT)}}, pages 812--820, Abu Dhabi, United Arab Emirates (Hybrid). Association for Computational Linguistics.

\bibitem[Caswell et~al., 2019]{Caswell2019-el}
Isaac Caswell, Ciprian Chelba, and David Grangier (2019).
\newblock {Tagged Back-Translation}.
\newblock In {\em {Proceedings of the Fourth Conference on Machine Translation (Volume 1: Research Papers)}}, pages 53--63, Florence, Italy. Association for Computational Linguistics.

\bibitem[Chang et~al., 2021]{Chang2021-sf}
Ernie Chang, Xiaoyu Shen, Dawei Zhu, Vera Demberg, and Hui Su (2021).
\newblock {Neural Data-to-Text Generation with {LM}-based Text Augmentation}.
\newblock In {\em {Proceedings of the 16th Conference of the European Chapter of the Association for Computational Linguistics: Main Volume}}, pages 758--768, Online. Association for Computational Linguistics.

\bibitem[Chen et~al., 2016]{Chen2016-MT-Align}
Wenhu Chen, Evgeny Matusov, Shahram Khadivi, and Jan-Thorsten Peter (2016).
\newblock {Guided Alignment Training for Topic-Aware Neural Machine Translation}.
\newblock In {\em {Conferences of the Association for Machine Translation in the Americas: MT Researchers' Track}}, pages 121--134, Austin, TX, USA. The Association for Machine Translation in the Americas.

\bibitem[Chen et~al., 2023]{Chen2023-LLM-MT}
Yijie Chen, Yijin Liu, Fandong Meng, Yufeng Chen, Jinan Xu, and Jie Zhou (2023).
\newblock {Improving translation faithfulness of large Language Models via augmenting instructions}.
\newblock \emph{arXiv preprint arXiv:2308.12674 [cs.CL]}.

\bibitem[Cheng et~al., 2022]{Cheng2022-TM}
Xin Cheng, Shen Gao, Lemao Liu, Dongyan Zhao, and Rui Yan (2022).
\newblock {Neural Machine Translation with Contrastive Translation Memories}.
\newblock In {\em {Proceedings of the 2022 Conference on Empirical Methods in Natural Language Processing}}, pages 3591--3601, Abu Dhabi, United Arab Emirates. Association for Computational Linguistics.

\bibitem[Cheng et~al., 2023]{Cheng2023-MT-LLM-SCALE}
Xin Cheng, Xun Wang, Tao Ge, Si-Qing Chen, Furu Wei, Dongyan Zhao, and Rui Yan (2023).
\newblock {SCALE: Synergized Collaboration of Asymmetric Language Translation Engines}.
\newblock \emph{arXiv preprint arXiv:2309.17061 [cs.CL]}.

\bibitem[Chinea-R{\'\i}os et~al., 2017]{Chinea-Rios2017-ln}
Mara Chinea-R{\'\i}os, {\'A}lvaro Peris, and Francisco Casacuberta (2017).
\newblock {Adapting Neural Machine Translation with Parallel Synthetic Data}.
\newblock In {\em {Proceedings of the Second Conference on Machine Translation}}, pages 138--147, Copenhagen, Denmark. Association for Computational Linguistics.

\bibitem[Chowdhery et~al., 2022]{Chowdhery2022-PaLM}
Aakanksha Chowdhery, Sharan Narang, Jacob Devlin, Maarten Bosma, Gaurav Mishra, Adam Roberts, Paul Barham, Hyung~Won Chung, Charles Sutton, Sebastian Gehrmann, Parker Schuh, Kensen Shi, Sasha Tsvyashchenko, Joshua Maynez, Abhishek Rao, Parker Barnes, Yi Tay, Noam Shazeer, Vinodkumar Prabhakaran, Emily Reif, Nan Du, Ben Hutchinson, Reiner Pope, James Bradbury, Jacob Austin, Michael Isard, Guy Gur-Ari, Pengcheng Yin, Toju Duke, Anselm Levskaya, Sanjay Ghemawat, Sunipa Dev, Henryk Michalewski, Xavier Garcia, Vedant Misra, Kevin Robinson, Liam Fedus, Denny Zhou, Daphne Ippolito, David Luan, Hyeontaek Lim, Barret Zoph, Alexander Spiridonov, Ryan Sepassi, David Dohan, Shivani Agrawal, Mark Omernick, Andrew~M Dai, Thanumalayan~Sankaranarayana Pillai, Marie Pellat, Aitor Lewkowycz, Erica Moreira, Rewon Child, Oleksandr Polozov, Katherine Lee, Zongwei Zhou, Xuezhi Wang, Brennan Saeta, Mark Diaz, Orhan Firat, Michele Catasta, Jason Wei, Kathy Meier-Hellstern, Douglas Eck, Jeff Dean, Slav Petrov, and Noah Fiedel (2022).
\newblock {PaLM: Scaling Language Modeling with Pathways}.
\newblock \emph{arXiv preprint arXiv:2204.02311 [cs.CL]}.

\bibitem[Chronopoulou et~al., 2021]{Chronopoulou2021-zr}
Alexandra Chronopoulou, Dario Stojanovski, and Alexander Fraser (2021).
\newblock {Improving the Lexical Ability of Pretrained Language Models for Unsupervised Neural Machine Translation}.
\newblock In {\em {Proceedings of the 2021 Conference of the North American Chapter of the Association for Computational Linguistics: Human Language Technologies}}, pages 173--180, Online. Association for Computational Linguistics.

\bibitem[Chu et~al., 2017]{Chu2017-mixed-fine-tuning}
Chenhui Chu, Raj Dabre, and Sadao Kurohashi (2017).
\newblock {An Empirical Comparison of Domain Adaptation Methods for Neural Machine Translation}.
\newblock In {\em {Proceedings of the 55th Annual Meeting of the Association for Computational Linguistics (Volume 2: Short Papers)}}, pages 385--391, Vancouver, Canada. Association for Computational Linguistics.

\bibitem[Clark et~al., 2020]{Clark2020-ELECTRA}
Kevin Clark, Minh-Thang Luong, Quoc~V Le, and Christopher~D Manning (2020).
\newblock {ELECTRA: Pre-training Text Encoders as Discriminators Rather Than Generators}.
\newblock In {\em {Proceedings of the 8th International Conference on Learning Representations (ICLR)}}, Virtual.

\bibitem[Costa-juss{\`a} et~al., 2022]{NLLB2022}
Marta~R Costa-juss{\`a}, James Cross, Onur {\c C}elebi, Maha Elbayad, Kenneth Heafield, Kevin Heffernan, Elahe Kalbassi, Janice Lam, Daniel Licht, Jean Maillard, Anna Sun, Skyler Wang, Guillaume Wenzek, Al Youngblood, Bapi Akula, Loic Barrault, Gabriel~Mejia Gonzalez, Prangthip Hansanti, John Hoffman, Semarley Jarrett, Kaushik~Ram Sadagopan, Dirk Rowe, Shannon Spruit, Chau Tran, Pierre Andrews, Necip~Fazil Ayan, Shruti Bhosale, Sergey Edunov, Angela Fan, Cynthia Gao, Vedanuj Goswami, Francisco Guzm{\'a}n, Philipp Koehn, Alexandre Mourachko, Christophe Ropers, Safiyyah Saleem, Holger Schwenk, and Jeff Wang (2022).
\newblock {No Language Left Behind: Scaling human-centered machine translation}.
\newblock \emph{arXiv preprint arXiv:2207.04672 [cs.CL]}.

\bibitem[Coughlin, 2003]{Coughlin2003-MT-eval}
Deborah Coughlin (2003).
\newblock {Correlating automated and human assessments of machine translation quality}.
\newblock In {\em {Proceedings of Machine Translation Summit IX: Papers}}, New Orleans, USA.

\bibitem[Crego et~al., 2016]{Crego2016-Placeholders}
Josep Crego, Jungi Kim, Guillaume Klein, Anabel Rebollo, Kathy Yang, Jean Senellart, Egor Akhanov, Patrice Brunelle, Aurelien Coquard, Yongchao Deng, Satoshi Enoue, Chiyo Geiss, Joshua Johanson, Ardas Khalsa, Raoum Khiari, Byeongil Ko, Catherine Kobus, Jean Lorieux, Leidiana Martins, Dang-Chuan Nguyen, Alexandra Priori, Thomas Riccardi, Natalia Segal, Christophe Servan, Cyril Tiquet, Bo Wang, Jin Yang, Dakun Zhang, Jing Zhou, and Peter Zoldan (2016).
\newblock {SYSTRAN's Pure Neural Machine Translation Systems}.
\newblock \emph{arXiv preprint arXiv:1610.05540 [cs.CL]}.

\bibitem[Dabre et~al., 2017]{Dabre2017-MultiSource}
Raj Dabre, Fabien Cromieres, and Sadao Kurohashi (2017).
\newblock {Enabling Multi-Source Neural Machine Translation By Concatenating Source Sentences In Multiple Languages}.
\newblock In {\em {Proceedings of Machine Translation Summit XVI: Research Track}}, pages 96--107, Nagoya Japan.

\bibitem[Dettmers et~al., 2023]{Dettmers2023-QLoRA}
Tim Dettmers, Artidoro Pagnoni, Ari Holtzman, and Luke Zettlemoyer (2023).
\newblock {QLoRA: Efficient Finetuning of Quantized LLMs}.
\newblock \emph{arXiv preprint arXiv:2305.14314 [cs.LG]}.

\bibitem[Devlin et~al., 2019]{Devlin2019-BERT}
Jacob Devlin, Ming-Wei Chang, Kenton Lee, and Kristina Toutanova (2019).
\newblock {{BERT}: Pre-training of Deep Bidirectional Transformers for Language Understanding}.
\newblock In {\em {Proceedings of the 2019 Conference of the North {A}merican Chapter of the Association for Computational Linguistics: Human Language Technologies, Volume 1 (Long and Short Papers)}}, pages 4171--4186, Minneapolis, Minnesota. Association for Computational Linguistics.

\bibitem[Dinu et~al., 2019]{Dinu2019-TerminologyConstraintsTraining}
Georgiana Dinu, Prashant Mathur, Marcello Federico, and Yaser Al-Onaizan (2019).
\newblock {Training Neural Machine Translation to Apply Terminology Constraints}.
\newblock In {\em {Proceedings of the 57th Annual Meeting of the Association for Computational Linguistics}}, pages 3063--3068, Florence, Italy. Association for Computational Linguistics.

\bibitem[Dong et~al., 2022]{Dong2022-In-contextLearning}
Qingxiu Dong, Lei Li, Damai Dai, Ce Zheng, Zhiyong Wu, Baobao Chang, Xu Sun, Jingjing Xu, Lei Li, and Zhifang Sui (2022).
\newblock {A Survey on In-context Learning}.
\newblock \emph{arXiv preprint arXiv:2301.00234 [cs.CL]}.

\bibitem[Edunov et~al., 2018]{Edunov2018-au}
Sergey Edunov, Myle Ott, Michael Auli, and David Grangier (2018).
\newblock {Understanding Back-Translation at Scale}.
\newblock In {\em {Proceedings of the 2018 Conference on Empirical Methods in Natural Language Processing}}, pages 489--500, Brussels, Belgium. Association for Computational Linguistics.

\bibitem[{EMA}, 2012]{EMA2012}
{EMA} (2012).
\newblock {European public assessment reports}.
\newblock The European Medicines Agency (EMA).

\bibitem[Etchegoyhen et~al., 2021]{Etchegoyhen2021-OnlineLearning}
Thierry Etchegoyhen, David Ponce, Harritxu Gete, and Victor Ruiz (2021).
\newblock {Online Learning over Time in Adaptive Neural Machine Translation}.
\newblock In {\em {Proceedings of the International Conference on Recent Advances in Natural Language Processing (RANLP 2021)}}, pages 411--420, Held Online. INCOMA Ltd.

\bibitem[Exel et~al., 2020]{Exel2020-TerminologyConstrainedMT}
Miriam Exel, Bianka Buschbeck, Lauritz Brandt, and Simona Doneva (2020).
\newblock {Terminology-Constrained Neural Machine Translation at {SAP}}.
\newblock In {\em {Proceedings of the 22nd Annual Conference of the European Association for Machine Translation}}, pages 271--280, Lisboa, Portugal. European Association for Machine Translation.

\bibitem[Fadaee and Monz, 2018]{Fadaee2018-Difficult-Words}
Marzieh Fadaee and Christof Monz (2018).
\newblock {Back-Translation Sampling by Targeting Difficult Words in Neural Machine Translation}.
\newblock In {\em {Proceedings of the 2018 Conference on Empirical Methods in Natural Language Processing}}, pages 436--446, Brussels, Belgium. Association for Computational Linguistics.

\bibitem[Fan et~al., 2018]{Fan2018-zm}
Angela Fan, Mike Lewis, and Yann Dauphin (2018).
\newblock {Hierarchical Neural Story Generation}.
\newblock In {\em {Proceedings of the 56th Annual Meeting of the Association for Computational Linguistics (Volume 1: Long Papers)}}, pages 889--898, Melbourne, Australia. Association for Computational Linguistics.

\bibitem[Farajian et~al., 2017]{Farajian2017-AdaptiveMT}
M~Amin Farajian, Marco Turchi, Matteo Negri, and Marcello Federico (2017).
\newblock {Multi-Domain Neural Machine Translation through Unsupervised Adaptation}.
\newblock In {\em {Proceedings of the Second Conference on Machine Translation}}, pages 127--137, Copenhagen, Denmark. Association for Computational Linguistics.

\bibitem[FitzGerald et~al., 2022]{FitzGerald2022-AlexaTM}
Jack FitzGerald, Shankar Ananthakrishnan, Konstantine Arkoudas, Davide Bernardi, Abhishek Bhagia, Claudio Delli~Bovi, Jin Cao, Rakesh Chada, Amit Chauhan, Luoxin Chen, Anurag Dwarakanath, Satyam Dwivedi, Turan Gojayev, Karthik Gopalakrishnan, Thomas Gueudre, Dilek Hakkani-Tur, Wael Hamza, Jonathan~J H{\"u}ser, Kevin~Martin Jose, Haidar Khan, Beiye Liu, Jianhua Lu, Alessandro Manzotti, Pradeep Natarajan, Karolina Owczarzak, Gokmen Oz, Enrico Palumbo, Charith Peris, Chandana~Satya Prakash, Stephen Rawls, Andy Rosenbaum, Anjali Shenoy, Saleh Soltan, Mukund~Harakere Sridhar, Lizhen Tan, Fabian Triefenbach, Pan Wei, Haiyang Yu, Shuai Zheng, Gokhan Tur, and Prem Natarajan (2022).
\newblock {Alexa Teacher Model: Pretraining and Distilling Multi-Billion-Parameter Encoders for Natural Language Understanding Systems}.
\newblock In {\em {Proceedings of the 28th ACM SIGKDD Conference on Knowledge Discovery and Data Mining}}, KDD '22, pages 2893--2902, New York, NY, USA. Association for Computing Machinery.

\bibitem[Gala et~al., 2023]{Gala2023-IndicTrans2}
Jay Gala, Pranjal~A Chitale, A~K Raghavan, Varun Gumma, Sumanth Doddapaneni, Kumar~M Aswanth, Janki~Atul Nawale, Anupama Sujatha, Ratish Puduppully, Vivek Raghavan, Pratyush Kumar, Mitesh~M Khapra, Raj Dabre, and Anoop Kunchukuttan (2023).
\newblock {IndicTrans2: Towards High-Quality and Accessible Machine Translation Models for all 22 Scheduled Indian Languages}.
\newblock Transactions on Machine Learning Research.
\newblock {\em Transactions on Machine Learning Research}.

\bibitem[Garcia et~al., 2023]{Garcia2023-FewShotMT}
Xavier Garcia, Yamini Bansal, Colin Cherry, George Foster, Maxim Krikun, Fangxiaoyu Feng, Melvin Johnson, and Orhan Firat (2023).
\newblock {The unreasonable effectiveness of few-shot learning for machine translation}.
\newblock \emph{arXiv preprint arXiv:2302.01398 [cs.CL]}.

\bibitem[Garg et~al., 2019]{Garg2019-MT-Align}
Sarthak Garg, Stephan Peitz, Udhyakumar Nallasamy, and Matthias Paulik (2019).
\newblock {Jointly Learning to Align and Translate with Transformer Models}.
\newblock \emph{arXiv preprint arXiv:1909.02074 [cs.CL]}.

\bibitem[Ghazvininejad et~al., 2023]{Ghazvininejad2023-DictionaryMT}
Marjan Ghazvininejad, Hila Gonen, and Luke Zettlemoyer (2023).
\newblock {Dictionary-based Phrase-level Prompting of Large Language Models for Machine Translation}.
\newblock \emph{arXiv preprint arXiv:2302.07856 [cs.CL]}.

\bibitem[Goyal et~al., 2022]{Goyal2022-spBLEU}
Naman Goyal, Cynthia Gao, Vishrav Chaudhary, Peng-Jen Chen, Guillaume Wenzek, Da Ju, Sanjana Krishnan, Marc'aurelio Ranzato, Francisco Guzm{\'a}n, and Angela Fan (2022).
\newblock {The Flores-101 evaluation benchmark for low-resource and multilingual machine translation}.
\newblock Trans. Assoc. Comput. Linguist.
\newblock 10:522--538, {\em Trans. Assoc. Comput. Linguist.}, 10:522--538.

\bibitem[Green et~al., 2014]{Green2014-zv}
Spence Green, Sida~I Wang, Jason Chuang, Jeffrey Heer, Sebastian Schuster, and Christopher~D Manning (2014).
\newblock {Human Effort and Machine Learnability in Computer Aided Translation}.
\newblock In {\em {Proceedings of the 2014 Conference on Empirical Methods in Natural Language Processing ({EMNLP})}}, pages 1225--1236, Doha, Qatar. Association for Computational Linguistics.

\bibitem[Grootendorst, 2020]{Grootendorst2020-KeyBERT}
Maarten Grootendorst (2020).
\newblock Key{BERT}: Minimal keyword extraction with bert.
\newblock GitHub.
\newblock \emph{doi:10.5281/zenodo.4461265}.

\bibitem[Gupta et~al., 2021]{Gupta2021-Phrases}
Kamal~Kumar Gupta, Sukanta Sen, Rejwanul Haque, Asif Ekbal, Pushpak Bhattacharyya, and Andy Way (2021).
\newblock {Augmenting training data with syntactic phrasal-segments in low-resource neural machine translation}.
\newblock Machine Translation.
\newblock 35(4):661--685, {\em Machine Translation}, 35(4):661--685.

\bibitem[Guu et~al., 2020]{Guu2020-RetrievalLLM}
Kelvin Guu, Kenton Lee, Zora Tung, Panupong Pasupat, and Ming-Wei Chang (2020).
\newblock {REALM: Retrieval-Augmented Language Model Pre-Training}.
\newblock \emph{arXiv preprint arXiv:2002.08909 [cs.CL]}.

\bibitem[Haddow et~al., 2022]{Haddow2022-hr}
Barry Haddow, Rachel Bawden, Antonio Valerio~Miceli Barone, Jind{\v r}ich Helcl, and Alexandra Birch (2022).
\newblock {Survey of Low-Resource Machine Translation}.
\newblock Computational Linguistics.
\newblock 06:1--67, {\em Computational Linguistics}, 06:1--67.

\bibitem[Haddow and Koehn, 2012]{Haddow2012-Data}
Barry Haddow and Philipp Koehn (2012).
\newblock {Analysing the Effect of Out-of-Domain Data on {SMT} Systems}.
\newblock In {\em {Proceedings of the Seventh Workshop on Statistical Machine Translation}}, pages 422--432, Montr{\'e}al, Canada. Association for Computational Linguistics.

\bibitem[Haque et~al., 2021]{Haque2021-Low-resource-MT}
Rejwanul Haque, Chao-Hong Liu, and Andy Way (2021).
\newblock {Recent Advances of Low-resource Neural Machine Translation}.
\newblock Machine Translation.
\newblock 35(4):451--474, {\em Machine Translation}, 35(4):451--474.

\bibitem[Haque et~al., 2020a]{Haque2020-Terminology}
Rejwanul Haque, Yasmin Moslem, and Andy Way (2020a).
\newblock {Terminology-Aware Sentence Mining for {NMT} Domain Adaptation: {ADAPT}{'}s Submission to the Adap-{MT} 2020 {E}nglish-to-{H}indi {AI} Translation Shared Task}.
\newblock In {\em {Proceedings of the 17th International Conference on Natural Language Processing (ICON): Adap-MT 2020 Shared Task}}, pages 17--23, Patna, India. NLP Association of India (NLPAI).

\bibitem[Haque et~al., 2020b]{Haque2020-STAPLE}
Rejwanul Haque, Yasmin Moslem, and Andy Way (2020b).
\newblock {The {ADAPT} System Description for the {STAPLE} 2020 {E}nglish-to-{P}ortuguese Translation Task}.
\newblock In {\em {Proceedings of the Fourth Workshop on Neural Generation and Translation}}, pages 144--152, Online. Association for Computational Linguistics.

\bibitem[Hasler et~al., 2018]{Hasler2018-TerminologyConstraintsMT}
Eva Hasler, Adri{\`a} de~Gispert, Gonzalo Iglesias, and Bill Byrne (2018).
\newblock {Neural Machine Translation Decoding with Terminology Constraints}.
\newblock In {\em {Proceedings of the 2018 Conference of the North {A}merican Chapter of the Association for Computational Linguistics: Human Language Technologies, Volume 2 (Short Papers)}}, pages 506--512, New Orleans, Louisiana. Association for Computational Linguistics.

\bibitem[Hasler et~al., 2021]{Hasler2021-domain}
Eva Hasler, Tobias Domhan, Jonay Trenous, Ke Tran, Bill Byrne, and Felix Hieber (2021).
\newblock {Improving the Quality Trade-Off for Neural Machine Translation Multi-Domain Adaptation}.
\newblock In {\em {Proceedings of the 2021 Conference on Empirical Methods in Natural Language Processing}}, pages 8470--8477, Online and Punta Cana, Dominican Republic. Association for Computational Linguistics.

\bibitem[He et~al., 2021a]{He2021-DeBERTaV3}
Pengcheng He, Jianfeng Gao, and Weizhu Chen (2021a).
\newblock {DeBERTaV3: Improving DeBERTa using ELECTRA-Style Pre-Training with Gradient-Disentangled Embedding Sharing}.
\newblock \emph{arXiv preprint arXiv:2111.09543 [cs.CL]}.

\bibitem[He et~al., 2021b]{He2020-DeBERTa}
Pengcheng He, Xiaodong Liu, Jianfeng Gao, and Weizhu Chen (2021b).
\newblock {DeBERTa: Decoding-enhanced BERT with Disentangled Attention}.
\newblock In {\em {Proceedings of the 9th International Conference on Learning Representations (ICLR)}}, Virtual.

\bibitem[He et~al., 2023]{He2023-MT-LLM-Human}
Zhiwei He, Tian Liang, Wenxiang Jiao, Zhuosheng Zhang, Yujiu Yang, Rui Wang, Zhaopeng Tu, Shuming Shi, and Xing Wang (2023).
\newblock {Exploring Human-Like Translation Strategy with Large Language Models}.
\newblock \emph{arXiv preprint arXiv:2305.04118 [cs.CL]}.

\bibitem[Hendy et~al., 2023]{Hendy2023-LLM-MT}
Amr Hendy, Mohamed Abdelrehim, Amr Sharaf, Vikas Raunak, Mohamed Gabr, Hitokazu Matsushita, Young~Jin Kim, Mohamed Afify, and Hany~Hassan Awadalla (2023).
\newblock {How Good Are GPT Models at Machine Translation? A Comprehensive Evaluation}.
\newblock \emph{arXiv preprint arXiv:2302.09210 [cs.CL]}.

\bibitem[Hoang et~al., 2023]{Hoang2023-MT-LLM-Ensemble}
Hieu Hoang, Huda Khayrallah, and Marcin Junczys-Dowmunt (2023).
\newblock {On-the-Fly Fusion of Large Language Models and Machine Translation}.
\newblock \emph{arXiv preprint arXiv:2311.08306 [cs.CL]}.

\bibitem[Hoang et~al., 2018]{Hoang2018-rj}
Vu~Cong~Duy Hoang, Philipp Koehn, Gholamreza Haffari, and Trevor Cohn (2018).
\newblock {Iterative Back-Translation for Neural Machine Translation}.
\newblock In {\em {Proceedings of the 2nd Workshop on Neural Machine Translation and Generation}}, pages 18--24, Melbourne, Australia. Association for Computational Linguistics.

\bibitem[Hoffmann et~al., 2022]{Hoffmann2022-chinchilla}
Jordan Hoffmann, Sebastian Borgeaud, Arthur Mensch, Elena Buchatskaya, Trevor Cai, Eliza Rutherford, Diego de~Las~Casas, Lisa~Anne Hendricks, Johannes Welbl, Aidan Clark, Tom Hennigan, Eric Noland, Katie Millican, George van~den Driessche, Bogdan Damoc, Aurelia Guy, Simon Osindero, Karen Simonyan, Erich Elsen, Jack~W Rae, Oriol Vinyals, and Laurent Sifre (2022).
\newblock {Training Compute-Optimal Large Language Models}.
\newblock \emph{arXiv preprint arXiv:2203.15556 [cs.CL]}.

\bibitem[Hokamp and Liu, 2017]{Hokamp2017-ConstrainedDecoding}
Chris Hokamp and Qun Liu (2017).
\newblock {Lexically Constrained Decoding for Sequence Generation Using Grid Beam Search}.
\newblock In {\em {Proceedings of the 55th Annual Meeting of the Association for Computational Linguistics (Volume 1: Long Papers)}}, pages 1535--1546, Vancouver, Canada. Association for Computational Linguistics.

\bibitem[Holtzman et~al., 2020]{Holtzman2020-nucleus}
Ari Holtzman, Jan Buys, Li Du, Maxwell Forbes, and Yejin Choi (2020).
\newblock {The Curious Case of Neural Text Degeneration}.
\newblock In {\em {Proceedings of the 8th International Conference on Learning Representations (ICLR)}}, Virtual.

\bibitem[Holtzman et~al., 2018]{Holtzman2018-rf}
Ari Holtzman, Jan Buys, Maxwell Forbes, Antoine Bosselut, David Golub, and Yejin Choi (2018).
\newblock {Learning to Write with Cooperative Discriminators}.
\newblock In {\em {Proceedings of the 56th Annual Meeting of the Association for Computational Linguistics (Volume 1: Long Papers)}}, pages 1638--1649, Melbourne, Australia. Association for Computational Linguistics.

\bibitem[Horawalavithana et~al., 2022]{Horawalavithana2022-chemistry}
Sameera Horawalavithana, Ellyn Ayton, Shivam Sharma, Scott Howland, Megha Subramanian, Scott Vasquez, Robin Cosbey, Maria Glenski, and Svitlana Volkova (2022).
\newblock {Foundation Models of Scientific Knowledge for Chemistry: Opportunities, Challenges and Lessons Learned}.
\newblock In {\em {Proceedings of BigScience Episode \#5 -- Workshop on Challenges \& Perspectives in Creating Large Language Models}}, pages 160--172, virtual+Dublin. Association for Computational Linguistics.

\bibitem[Hosseini et~al., 2020]{Hosseini2020-DeepMatch}
Kasra Hosseini, Federico Nanni, and Mariona Coll~Ardanuy (2020).
\newblock {{D}eezy{M}atch: A Flexible Deep Learning Approach to Fuzzy String Matching}.
\newblock In {\em {Proceedings of the 2020 Conference on Empirical Methods in Natural Language Processing: System Demonstrations}}, pages 62--69, Online. Association for Computational Linguistics.

\bibitem[Hu et~al., 2021]{Hu2021-LoRA}
Edward~J Hu, Yelong Shen, Phillip Wallis, Zeyuan Allen-Zhu, Yuanzhi Li, Shean Wang, Lu Wang, and Weizhu Chen (2021).
\newblock {LoRA: Low-Rank Adaptation of Large Language Models}.
\newblock \emph{arXiv preprint arXiv:2106.09685 [cs.CL]}.

\bibitem[Hu et~al., 2019a]{Hu2019-LexiconInduction}
Junjie Hu, Mengzhou Xia, Graham Neubig, and Jaime Carbonell (2019a).
\newblock {Domain Adaptation of Neural Machine Translation by Lexicon Induction}.
\newblock In {\em {Proceedings of the 57th Annual Meeting of the Association for Computational Linguistics}}, pages 2989--3001, Florence, Italy. Association for Computational Linguistics.

\bibitem[Hu et~al., 2019b]{Hu2019-ImprovedConstrainedDecoding}
J~Edward Hu, Huda Khayrallah, Ryan Culkin, Patrick Xia, Tongfei Chen, Matt Post, and Benjamin Van~Durme (2019b).
\newblock {Improved Lexically Constrained Decoding for Translation and Monolingual Rewriting}.
\newblock In {\em {Proceedings of the 2019 Conference of the North {A}merican Chapter of the Association for Computational Linguistics: Human Language Technologies, Volume 1 (Long and Short Papers)}}, pages 839--850, Minneapolis, Minnesota. Association for Computational Linguistics.

\bibitem[Huck et~al., 2019]{Huck2019-MT-OOV}
Matthias Huck, Viktor Hangya, and Alexander Fraser (2019).
\newblock {Better {OOV} Translation with Bilingual Terminology Mining}.
\newblock In {\em {Proceedings of the 57th Annual Meeting of the Association for Computational Linguistics}}, pages 5809--5815, Florence, Italy. Association for Computational Linguistics.

\bibitem[Imankulova et~al., 2019]{Imankulova2019-MT-Multi-Low-resource}
Aizhan Imankulova, Raj Dabre, Atsushi Fujita, and Kenji Imamura (2019).
\newblock {Exploiting Out-of-Domain Parallel Data through Multilingual Transfer Learning for Low-Resource Neural Machine Translation}.
\newblock pages 128--139, Dublin, Ireland. European Association for Machine Translation.

\bibitem[Iyer et~al., 2023]{Iyer2023-LLMs-Disambiguation}
Vivek Iyer, Pinzhen Chen, and Alexandra Birch (2023).
\newblock {Towards Effective Disambiguation for Machine Translation with Large Language Models}.
\newblock In {\em {Proceedings of the Eighth Conference on Machine Translation}}, pages 482--495, Singapore. Association for Computational Linguistics.

\bibitem[Izacard et~al., 2022]{Izacard2022-AtlasRetrievalLLM}
Gautier Izacard, Patrick Lewis, Maria Lomeli, Lucas Hosseini, Fabio Petroni, Timo Schick, Jane Dwivedi-Yu, Armand Joulin, Sebastian Riedel, and Edouard Grave (2022).
\newblock {Atlas: Few-shot Learning with Retrieval Augmented Language Models}.
\newblock \emph{arXiv preprint arXiv:2208.03299 [cs.CL]}.

\bibitem[Jiang et~al., 2023]{Jiang2023-Mistral}
Albert~Q Jiang, Alexandre Sablayrolles, Arthur Mensch, Chris Bamford, Devendra~Singh Chaplot, Diego de~las Casas, Florian Bressand, Gianna Lengyel, Guillaume Lample, Lucile Saulnier, L{\'e}lio~Renard Lavaud, Marie-Anne Lachaux, Pierre Stock, Teven Le~Scao, Thibaut Lavril, Thomas Wang, Timoth{\'e}e Lacroix, and William El~Sayed (2023).
\newblock {Mistral 7B}.
\newblock \emph{arXiv preprint arXiv:2310.06825 [cs.CL]}.

\bibitem[Jiao et~al., 2023a]{Jiao2023-Parrot}
Wenxiang Jiao, Jen-Tse Huang, Wenxuan Wang, Xing Wang, Shuming Shi, and Zhaopeng Tu (2023a).
\newblock {ParroT: Translating During Chat Using Large Language Models}.
\newblock \emph{arXiv preprint arXiv:2304.02426 [cs.CL]}.

\bibitem[Jiao et~al., 2023b]{Jiao2023-LLM-MT}
Wenxiang Jiao, Wenxuan Wang, Jen-Tse Huang, Xing Wang, and Zhaopeng Tu (2023b).
\newblock {Is ChatGPT A Good Translator? Yes With GPT-4 As The Engine}.
\newblock \emph{arXiv preprint arXiv:2301.08745 [cs.CL]}.

\bibitem[Johnson et~al., 2019]{Johnson2019-Faiss}
Jeff Johnson, Matthijs Douze, and Herv{\'e} J{\'e}gou (2019).
\newblock {Billion-Scale Similarity Search with GPUs}.
\newblock IEEE Transactions on Big Data.
\newblock 7(3):535--547, {\em IEEE Transactions on Big Data}, 7(3):535--547.

\bibitem[Junczys-Dowmunt, 2018]{Junczys-Dowmunt2018-Scoring}
Marcin Junczys-Dowmunt (2018).
\newblock {Dual Conditional Cross-Entropy Filtering of Noisy Parallel Corpora}.
\newblock In {\em {Proceedings of the Third Conference on Machine Translation: Shared Task Papers}}, pages 888--895, Belgium, Brussels. Association for Computational Linguistics.

\bibitem[Kadaoui et~al., 2023]{Kadaoui2023-LLM-MT-Arabic}
Karima Kadaoui, Samar~M Magdy, Abdul Waheed, Md~Tawkat~Islam Khondaker, Ahmed~Oumar El-Shangiti, El~Moatez~Billah Nagoudi, and Muhammad Abdul-Mageed (2023).
\newblock {TARJAMAT: Evaluation of Bard and ChatGPT on Machine Translation of Ten Arabic Varieties}.
\newblock In {\em {The First Arabic Natural Language Processing Conference (ArabicNLP 2023)}}, Sentosa, Singapore.

\bibitem[Khandelwal et~al., 2021]{Khandelwal2021-kNN-MT}
Urvashi Khandelwal, Angela Fan, Dan Jurafsky, Luke Zettlemoyer, and Mike Lewis (2021).
\newblock {Nearest Neighbor Machine Translation}.
\newblock In {\em {Proceedings of the 9th International Conference on Learning Representations (ICLR)}}, Virtual.

\bibitem[Kirkpatrick et~al., 2017]{Kirkpatrick2017-qq}
James Kirkpatrick, Razvan Pascanu, Neil Rabinowitz, Joel Veness, Guillaume Desjardins, Andrei~A Rusu, Kieran Milan, John Quan, Tiago Ramalho, Agnieszka Grabska-Barwinska, Demis Hassabis, Claudia Clopath, Dharshan Kumaran, and Raia Hadsell (2017).
\newblock {Overcoming catastrophic forgetting in neural networks}.
\newblock Proc. Natl. Acad. Sci. U. S. A.
\newblock 114(13):3521--3526, {\em Proc. Natl. Acad. Sci. U. S. A.}, 114(13):3521--3526.

\bibitem[Klein et~al., 2020a]{Klein2020-OpenNMT}
Guillaume Klein, Fran{\c c}ois Hernandez, Vincent Nguyen, and Jean Senellart (2020a).
\newblock {The {O}pen{NMT} Neural Machine Translation Toolkit: 2020 Edition}.
\newblock In {\em {Proceedings of the 14th Conference of the Association for Machine Translation in the Americas (Volume 1: Research Track)}}, pages 102--109, Virtual. Association for Machine Translation in the Americas.

\bibitem[Klein et~al., 2020b]{Klein2020-Efficient}
Guillaume Klein, Dakun Zhang, Cl{\'e}ment Chouteau, Josep Crego, and Jean Senellart (2020b).
\newblock {Efficient and high-quality neural machine translation with {OpenNMT}}.
\newblock In {\em {Proceedings of the Fourth Workshop on Neural Generation and Translation}}, pages 211--217, Stroudsburg, PA, USA. Association for Computational Linguistics.

\bibitem[Knowles et~al., 2018]{Knowles2018-Fuzzy}
Rebecca Knowles, John Ortega, and Philipp Koehn (2018).
\newblock {A Comparison of Machine Translation Paradigms for Use in Black-Box Fuzzy-Match Repair}.
\newblock In {\em {Proceedings of the {AMTA} 2018 Workshop on Translation Quality Estimation and Automatic Post-Editing}}, pages 249--255, Boston, MA. Association for Machine Translation in the Americas.

\bibitem[Kobus et~al., 2017]{Kobus2017-domain-control}
Catherine Kobus, Josep Crego, and Jean Senellart (2017).
\newblock {Domain Control for Neural Machine Translation}.
\newblock In {\em {Proceedings of Recent Advances in Natural Language Processing}}, pages 372--378, Varna, Bulgaria.

\bibitem[Koehn, 2009]{Koehn2009-CAT}
Philipp Koehn (2009).
\newblock {A process study of computer-aided translation}.
\newblock Mach. Transl.
\newblock 23(4):241--263, {\em Mach. Transl.}, 23(4):241--263.

\bibitem[Koehn and Knowles, 2017]{Koehn2017-qj}
Philipp Koehn and Rebecca Knowles (2017).
\newblock {Six Challenges for Neural Machine Translation}.
\newblock In {\em {Proceedings of the First Workshop on Neural Machine Translation}}, pages 28--39, Vancouver. Association for Computational Linguistics.

\bibitem[Koneru et~al., 2023]{Koneru2023-MT-LLM-MTPE}
Sai Koneru, Miriam Exel, Matthias Huck, and Jan Niehues (2023).
\newblock {Contextual Refinement of Translations: Large Language Models for Sentence and Document-Level Post-Editing}.
\newblock \emph{arXiv preprint arXiv:2310.14855 [cs.CL]}.

\bibitem[Kudo, 2018]{Kudo2018-subword}
Taku Kudo (2018).
\newblock {Subword Regularization: Improving Neural Network Translation Models with Multiple Subword Candidates}.
\newblock In {\em {Proceedings of the 56th Annual Meeting of the Association for Computational Linguistics (Volume 1: Long Papers)}}, pages 66--75, Melbourne, Australia. Association for Computational Linguistics.

\bibitem[Kudo and Richardson, 2018]{Kudo2018-SentencePiece}
Taku Kudo and John Richardson (2018).
\newblock {{S}entence{P}iece: A simple and language independent subword tokenizer and detokenizer for Neural Text Processing}.
\newblock In {\em {Proceedings of the 2018 Conference on Empirical Methods in Natural Language Processing: System Demonstrations}}, pages 66--71, Brussels, Belgium. Association for Computational Linguistics.

\bibitem[Kudugunta et~al., 2023]{Kudugunta2023-MADLAD}
Sneha Kudugunta, Isaac Caswell, Biao Zhang, Xavier Garcia, Christopher~A Choquette-Choo, Katherine Lee, Derrick Xin, Aditya Kusupati, Romi Stella, Ankur Bapna, and Orhan Firat (2023).
\newblock {MADLAD-400: A Multilingual And Document-Level Large Audited Dataset}.
\newblock \emph{arXiv preprint arXiv:2309.04662 [cs.CL]}.

\bibitem[Kumar et~al., 2023]{Kumar2023-MT-LLM-CTQScorer}
Aswanth Kumar, Ratish Puduppully, Raj Dabre, and Anoop Kunchukuttan (2023).
\newblock {{CTQS}corer: Combining Multiple Features for In-context Example Selection for Machine Translation}.
\newblock In {\em {Findings of the Association for Computational Linguistics: EMNLP 2023}}, pages 7736--7752, Singapore. Association for Computational Linguistics.

\bibitem[Lample and Conneau, 2019]{Lample2019-xy}
Guillaume Lample and Alexis Conneau (2019).
\newblock {Cross-lingual Language Model Pretraining}.
\newblock In {\em {Advances in Neural Information Processing Systems (NeurIPS 2019)}}, volume~32, Vancouver, Canada. Curran Associates, Inc.

\bibitem[Langlais et~al., 2000]{Langlais2000-TransType}
Philippe Langlais, George Foster, and Guy Lapalme (2000).
\newblock {{T}rans{T}ype: a Computer-Aided Translation Typing System}.
\newblock In {\em {{ANLP}-{NAACL} 2000 Workshop: Embedded Machine Translation Systems}}.

\bibitem[Le~Scao et~al., 2022]{BLOOM2022}
Teven Le~Scao, Angela Fan, Christopher Akiki, Ellie Pavlick, Suzana Ili{\'c}, Daniel Hesslow, Roman Castagn{\'e}, Alexandra~Sasha Luccioni, Fran{\c c}ois Yvon, Matthias Gall{\'e}, Jonathan Tow, Alexander~M Rush, Stella Biderman, Albert Webson, Pawan~Sasanka Ammanamanchi, Thomas Wang, Beno{\^\i}t Sagot, Niklas Muennighoff, Albert~Villanova del Moral, Olatunji Ruwase, Rachel Bawden, Stas Bekman, Angelina McMillan-Major, Iz Beltagy, Huu Nguyen, Lucile Saulnier, Samson Tan, Pedro~Ortiz Suarez, Victor Sanh, Hugo Lauren{\c c}on, Yacine Jernite, Julien Launay, Margaret Mitchell, Colin Raffel, Aaron Gokaslan, Adi Simhi, Aitor Soroa, Alham~Fikri Aji, Amit Alfassy, Anna Rogers, Ariel~Kreisberg Nitzav, Canwen Xu, Chenghao Mou, Chris Emezue, Christopher Klamm, Colin Leong, Daniel van Strien, David~Ifeoluwa Adelani, Dragomir Radev, Eduardo~Gonz{\'a}lez Ponferrada, Efrat Levkovizh, Ethan Kim, Eyal~Bar Natan, Francesco De~Toni, G{\'e}rard Dupont, Germ{\'a}n Kruszewski, Giada Pistilli, Hady Elsahar, Hamza Benyamina, Hieu
  Tran, Ian Yu, Idris Abdulmumin, Isaac Johnson, Itziar Gonzalez-Dios, Javier de~la Rosa, Jenny Chim, Jesse Dodge, Jian Zhu, Jonathan Chang, J{\"o}rg Frohberg, Joseph Tobing, Joydeep Bhattacharjee, Khalid Almubarak, Kimbo Chen, Kyle Lo, Leandro Von~Werra, Leon Weber, Long Phan, Loubna Ben~allal, Ludovic Tanguy, Manan Dey, Manuel~Romero Mu{\~n}oz, Maraim Masoud, Mar{\'\i}a Grandury, Mario {\v S}a{\v s}ko, Max Huang, Maximin Coavoux, Mayank Singh, Mike Tian-Jian Jiang, Minh~Chien Vu, Mohammad~A Jauhar, Mustafa Ghaleb, Nishant Subramani, Nora Kassner, Nurulaqilla Khamis, Olivier Nguyen, Omar Espejel, Ona de~Gibert, Paulo Villegas, Peter Henderson, Pierre Colombo, Priscilla Amuok, Quentin Lhoest, Rheza Harliman, Rishi Bommasani, Roberto~Luis L{\'o}pez, Rui Ribeiro, Salomey Osei, Sampo Pyysalo, Sebastian Nagel, Shamik Bose, Shamsuddeen~Hassan Muhammad, Shanya Sharma, Shayne Longpre, Somaieh Nikpoor, Stanislav Silberberg, Suhas Pai, Sydney Zink, Tiago~Timponi Torrent, Timo Schick, Tristan Thrush, Valentin Danchev,
  Vassilina Nikoulina, Veronika Laippala, Violette Lepercq, Vrinda Prabhu, Zaid Alyafeai, Zeerak Talat, Arun Raja, Benjamin Heinzerling, Chenglei Si, Davut~Emre Ta{\c s}ar, Elizabeth Salesky, Sabrina~J Mielke, Wilson~Y Lee, Abheesht Sharma, Andrea Santilli, Antoine Chaffin, Arnaud Stiegler, Debajyoti Datta, Eliza Szczechla, Gunjan Chhablani, Han Wang, Harshit Pandey, Hendrik Strobelt, Jason~Alan Fries, Jos Rozen, Leo Gao, Lintang Sutawika, M Saiful~Bari, Maged~S Al-shaibani, Matteo Manica, Nihal Nayak, Ryan Teehan, Samuel Albanie, Sheng Shen, Srulik Ben-David, Stephen~H Bach, Taewoon Kim, Tali Bers, Thibault Fevry, Trishala Neeraj, Urmish Thakker, Vikas Raunak, Xiangru Tang, Zheng-Xin Yong, Zhiqing Sun, Shaked Brody, Yallow Uri, Hadar Tojarieh, Adam Roberts, Hyung~Won Chung, Jaesung Tae, Jason Phang, {Ofir Press}, Conglong Li, Deepak Narayanan, Hatim Bourfoune, Jared Casper, Jeff Rasley, Max Ryabinin, Mayank Mishra, Minjia Zhang, Mohammad Shoeybi, Myriam Peyrounette, Nicolas Patry, Nouamane Tazi, Omar
  Sanseviero, Patrick von Platen, Pierre Cornette, Pierre~Fran{\c c}ois Lavall{\'e}e, R{\'e}mi Lacroix, Samyam Rajbhandari, Sanchit Gandhi, Shaden Smith, St{\'e}phane Requena, Suraj Patil, Tim Dettmers, Ahmed Baruwa, Amanpreet Singh, Anastasia Cheveleva, Anne-Laure Ligozat, Arjun Subramonian, Aur{\'e}lie N{\'e}v{\'e}ol, Charles Lovering, Dan Garrette, Deepak Tunuguntla, Ehud Reiter, Ekaterina Taktasheva, Ekaterina Voloshina, Eli Bogdanov, Genta~Indra Winata, Hailey Schoelkopf, Jan-Christoph Kalo, Jekaterina Novikova, Jessica~Zosa Forde, Jordan Clive, Jungo Kasai, Ken Kawamura, Liam Hazan, Marine Carpuat, Miruna Clinciu, Najoung Kim, Newton Cheng, Oleg Serikov, Omer Antverg, Oskar van~der Wal, Rui Zhang, Ruochen Zhang, Sebastian Gehrmann, Shachar Mirkin, Shani Pais, Tatiana Shavrina, Thomas Scialom, Tian Yun, Tomasz Limisiewicz, Verena Rieser, Vitaly Protasov, Vladislav Mikhailov, Yada Pruksachatkun, Yonatan Belinkov, Zachary Bamberger, Zden{\v e}k Kasner, Alice Rueda, Amanda Pestana, Amir Feizpour, Ammar
  Khan, Amy Faranak, Ana Santos, Anthony Hevia, Antigona Unldreaj, Arash Aghagol, Arezoo Abdollahi, Aycha Tammour, Azadeh HajiHosseini, Bahareh Behroozi, Benjamin Ajibade, Bharat Saxena, Carlos~Mu{\~n}oz Ferrandis, Daniel McDuff, Danish Contractor, David Lansky, Davis David, Douwe Kiela, Duong~A Nguyen, Edward Tan, Emi Baylor, Ezinwanne Ozoani, Fatima Mirza, Frankline Ononiwu, Habib Rezanejad, Hessie Jones, Indrani Bhattacharya, Irene Solaiman, Irina Sedenko, Isar Nejadgholi, Jesse Passmore, Josh Seltzer, Julio~Bonis Sanz, Livia Dutra, Mairon Samagaio, Maraim Elbadri, Margot Mieskes, Marissa Gerchick, Martha Akinlolu, Michael McKenna, Mike Qiu, Muhammed Ghauri, Mykola Burynok, Nafis Abrar, Nazneen Rajani, Nour Elkott, Nour Fahmy, Olanrewaju Samuel, Ran An, Rasmus Kromann, Ryan Hao, Samira Alizadeh, Sarmad Shubber, Silas Wang, Sourav Roy, Sylvain Viguier, Thanh Le, Tobi Oyebade, Trieu Le, Yoyo Yang, Zach Nguyen, Abhinav~Ramesh Kashyap, Alfredo Palasciano, Alison Callahan, Anima Shukla, Antonio
  Miranda-Escalada, Ayush Singh, Benjamin Beilharz, Bo Wang, Caio Brito, Chenxi Zhou, Chirag Jain, Chuxin Xu, Cl{\'e}mentine Fourrier, Daniel~Le{\'o}n Peri{\~n}{\'a}n, Daniel Molano, Dian Yu, Enrique Manjavacas, Fabio Barth, Florian Fuhrimann, Gabriel Altay, Giyaseddin Bayrak, Gully Burns, Helena~U Vrabec, Imane Bello, Ishani Dash, Jihyun Kang, John Giorgi, Jonas Golde, Jose~David Posada, Karthik~Rangasai Sivaraman, Lokesh Bulchandani, Lu Liu, Luisa Shinzato, Madeleine~Hahn de~Bykhovetz, Maiko Takeuchi, Marc P{\`a}mies, Maria~A Castillo, Marianna Nezhurina, Mario S{\"a}nger, Matthias Samwald, Michael Cullan, Michael Weinberg, Michiel De~Wolf, Mina Mihaljcic, Minna Liu, Moritz Freidank, Myungsun Kang, Natasha Seelam, Nathan Dahlberg, Nicholas~Michio Broad, Nikolaus Muellner, Pascale Fung, Patrick Haller, Ramya Chandrasekhar, Renata Eisenberg, Robert Martin, Rodrigo Canalli, Rosaline Su, Ruisi Su, Samuel Cahyawijaya, Samuele Garda, Shlok~S Deshmukh, Shubhanshu Mishra, Sid Kiblawi, Simon Ott, Sinee
  Sang-aroonsiri, Srishti Kumar, Stefan Schweter, Sushil Bharati, Tanmay Laud, Th{\'e}o Gigant, Tomoya Kainuma, Wojciech Kusa, Yanis Labrak, Yash~Shailesh Bajaj, Yash Venkatraman, Yifan Xu, Yingxin Xu, Yu Xu, Zhe Tan, Zhongli Xie, Zifan Ye, Mathilde Bras, Younes Belkada, and Thomas Wolf (2022).
\newblock {BLOOM: A 176B-Parameter Open-Access Multilingual Language Model}.
\newblock \emph{arXiv preprint arXiv:2211.05100 [cs.CL]}.

\bibitem[Lewis et~al., 2020]{Lewis2020-RetrievalLLM}
Patrick Lewis, Ethan Perez, Aleksandra Piktus, Fabio Petroni, Vladimir Karpukhin, Naman Goyal, Heinrich K{\"u}ttler, Mike Lewis, Wen-Tau Yih, Tim Rockt{\"a}schel, Sebastian Riedel, and Douwe Kiela (2020).
\newblock {Retrieval-Augmented Generation for Knowledge-Intensive NLP Tasks}.
\newblock \emph{arXiv preprint arXiv:2005.11401 [cs.CL]}.

\bibitem[Li et~al., 2023a]{Li2023-MT-LLM-MT2}
Chunyou Li, Mingtong Liu, Hongxiao Zhang, Yufeng Chen, Jinan Xu, and Ming Zhou (2023a).
\newblock {{MT}2: Towards a Multi-Task Machine Translation Model with Translation-Specific In-Context Learning}.
\newblock In {\em {Proceedings of the 2023 Conference on Empirical Methods in Natural Language Processing}}, pages 8616--8627, Singapore. Association for Computational Linguistics.

\bibitem[Li et~al., 2021]{Li2021-WLAC}
Huayang Li, Lemao Liu, Guoping Huang, and Shuming Shi (2021).
\newblock {{GWLAN}: General Word-Level {A}utocompletio{N} for Computer-Aided Translation}.
\newblock In {\em {Proceedings of the 59th Annual Meeting of the Association for Computational Linguistics and the 11th International Joint Conference on Natural Language Processing (Volume 1: Long Papers)}}, pages 4792--4802, Online. Association for Computational Linguistics.

\bibitem[Li et~al., 2023b]{Li2023-MT-LLM-Multilingual-Finetuning}
Jiahuan Li, Hao Zhou, Shujian Huang, Shanbo Cheng, and Jiajun Chen (2023b).
\newblock {Eliciting the Translation Ability of Large Language Models via Multilingual Finetuning with Translation Instructions}.
\newblock \emph{arXiv preprint arXiv:2305.15083 [cs.CL]}.

\bibitem[Li et~al., 2022]{Li2022-ELMFOREST}
Margaret Li, Suchin Gururangan, Tim Dettmers, Mike Lewis, Tim Althoff, Noah~A Smith, and Luke Zettlemoyer (2022).
\newblock {Branch-Train-Merge: Embarrassingly Parallel Training of Expert Language Models}.
\newblock \emph{arXiv preprint arXiv:2208.03306 [cs.CL]}.

\bibitem[Li et~al., 2018]{Li2018-AdaptiveMT}
Xiaoqing Li, Jiajun Zhang, and Chengqing Zong (2018).
\newblock {One Sentence One Model for Neural Machine Translation}.
\newblock In {\em {Proceedings of the Eleventh International Conference on Language Resources and Evaluation ({LREC} 2018)}}, Miyazaki, Japan. European Language Resources Association (ELRA).

\bibitem[Li et~al., 2023c]{Li2023-Phi}
Yuanzhi Li, S{\'e}bastien Bubeck, Ronen Eldan, Allie Del~Giorno, Suriya Gunasekar, and Yin~Tat Lee (2023c).
\newblock {Textbooks Are All You Need II: phi-1.5 technical report}.
\newblock \emph{arXiv preprint arXiv:2309.05463 [cs.CL]}.

\bibitem[Lin et~al., 2020]{Lin2020-CommonGen}
Bill~Yuchen Lin, Wangchunshu Zhou, Ming Shen, Pei Zhou, Chandra Bhagavatula, Yejin Choi, and Xiang Ren (2020).
\newblock {{C}ommon{G}en: A Constrained Text Generation Challenge for Generative Commonsense Reasoning}.
\newblock In {\em {Findings of the Association for Computational Linguistics: EMNLP 2020}}, pages 1823--1840, Online. Association for Computational Linguistics.

\bibitem[Lin et~al., 2022]{Lin2022-LLM-MT}
Xi~Victoria Lin, Todor Mihaylov, Mikel Artetxe, Tianlu Wang, Shuohui Chen, Daniel Simig, Myle Ott, Naman Goyal, Shruti Bhosale, Jingfei Du, Ramakanth Pasunuru, Sam Shleifer, Punit~Singh Koura, Vishrav Chaudhary, Brian O'Horo, Jeff Wang, Luke Zettlemoyer, Zornitsa Kozareva, Mona Diab, Veselin Stoyanov, and Xian Li (2022).
\newblock {Few-shot Learning with Multilingual Generative Language Models}.
\newblock In {\em {Proceedings of the 2022 Conference on Empirical Methods in Natural Language Processing}}, pages 9019--9052, Abu Dhabi, United Arab Emirates. Association for Computational Linguistics.

\bibitem[Liu et~al., 2020]{Liu2020-mBART}
Yinhan Liu, Jiatao Gu, Naman Goyal, Xian Li, Sergey Edunov, Marjan Ghazvininejad, Mike Lewis, and Luke Zettlemoyer (2020).
\newblock {Multilingual Denoising Pre-training for Neural Machine Translation}.
\newblock Transactions of the Association for Computational Linguistics.
\newblock 8:726--742, {\em Transactions of the Association for Computational Linguistics}, 8:726--742.

\bibitem[Liu et~al., 2019]{Liu2019-RoBERTa}
Yinhan Liu, Myle Ott, Naman Goyal, Jingfei Du, Mandar Joshi, Danqi Chen, Omer Levy, Mike Lewis, Luke Zettlemoyer, and Veselin Stoyanov (2019).
\newblock {RoBERTa: A Robustly Optimized BERT Pretraining Approach}.
\newblock \emph{arXiv preprint arXiv:1907.11692 [cs.CL]}.

\bibitem[Liu et~al., 2023]{Liu2023-LLM-MT}
Yijin Liu, Xianfeng Zeng, Fandong Meng, and Jie Zhou (2023).
\newblock {Instruction Position Matters in Sequence Generation with Large Language Models}.
\newblock \emph{arXiv preprint arXiv:2308.12097 [cs.CL]}.

\bibitem[Luong and Manning, 2015]{Luong2015-mt}
Minh-Thang Luong and Christopher Manning (2015).
\newblock {{S}tanford neural machine translation systems for spoken language domains}.
\newblock In {\em {Proceedings of the 12th International Workshop on Spoken Language Translation: Evaluation Campaign}}, pages 76--79, Da Nang, Vietnam.

\bibitem[Martins et~al., 2023]{Martins2023-kNN-MT}
Pedro~Henrique Martins, Jo{\~a}o Alves, T{\^a}nia Vaz, Madalena Gon{\c c}alves, Beatriz Silva, Marianna Buchicchio, Jos{\'e} G~C de~Souza, and Andr{\'e} F~T Martins (2023).
\newblock {Empirical Assessment of k{NN}-{MT} for Real-World Translation Scenarios}.
\newblock In {\em {Proceedings of the 24th Annual Conference of the European Association for Machine Translation}}, pages 115--124, Tampere, Finland. European Association for Machine Translation.

\bibitem[Meng et~al., 2022]{Meng2022-kNN-MT-Fast}
Yuxian Meng, Xiaoya Li, Xiayu Zheng, Fei Wu, Xiaofei Sun, Tianwei Zhang, and Jiwei Li (2022).
\newblock {Fast Nearest Neighbor Machine Translation}.
\newblock In {\em {Findings of the Association for Computational Linguistics: ACL 2022}}, pages 555--565, Dublin, Ireland. Association for Computational Linguistics.

\bibitem[Michon et~al., 2020]{Michon2020-Terminology}
Elise Michon, Josep Crego, and Jean Senellart (2020).
\newblock {Integrating Domain Terminology into Neural Machine Translation}.
\newblock In {\em {Proceedings of the 28th International Conference on Computational Linguistics}}, pages 3925--3937, Barcelona, Spain (Online). International Committee on Computational Linguistics.

\bibitem[Mondragón, 2021]{Mondragon2021-TM2TB}
Luis Mondragón (2021).
\newblock {TM2TB}: Bilingual term extraction and matching with spa{C}y and sentence transformers.
\newblock GitHub.
\newblock \emph{doi:""}.

\bibitem[Moore and Lewis, 2010]{Moore2010-Scoring}
Robert~C Moore and William Lewis (2010).
\newblock {Intelligent Selection of Language Model Training Data}.
\newblock In {\em {Proceedings of the {ACL} 2010 Conference Short Papers}}, pages 220--224, Uppsala, Sweden. Association for Computational Linguistics.

\bibitem[Moslem et~al., 2022a]{Moslem2022-MT-LM}
Yasmin Moslem, Rejwanul Haque, John Kelleher, and Andy Way (2022a).
\newblock {Domain-Specific Text Generation for Machine Translation}.
\newblock In {\em {Proceedings of the 15th biennial conference of the Association for Machine Translation in the Americas (Volume 1: Research Track)}}, pages 14--30, Orlando, USA. Association for Machine Translation in the Americas.

\bibitem[Moslem et~al., 2023a]{Moslem2023-AdaptiveMT}
Yasmin Moslem, Rejwanul Haque, John~D Kelleher, and Andy Way (2023a).
\newblock {Adaptive Machine Translation with Large Language Models}.
\newblock In {\em {Proceedings of the 24th Annual Conference of the European Association for Machine Translation}}, pages 227--237, Tampere, Finland. European Association for Machine Translation.

\bibitem[Moslem et~al., 2020]{Moslem2020-Arabisc}
Yasmin Moslem, Rejwanul Haque, and Andy Way (2020).
\newblock {Arabisc: Context-Sensitive Neural Spelling Checker}.
\newblock In {\em {Proceedings of the 6th Workshop on Natural Language Processing Techniques for Educational Applications}}, pages 11--19, Suzhou, China. Association for Computational Linguistics.

\bibitem[Moslem et~al., 2022b]{Moslem2022-WLAC}
Yasmin Moslem, Rejwanul Haque, and Andy Way (2022b).
\newblock {Translation Word-Level Auto-Completion: What Can We Achieve Out of the Box?}
\newblock In {\em {Proceedings of the Seventh Conference on Machine Translation (WMT)}}, pages 1176--1181, Abu Dhabi, United Arab Emirates (Hybrid). Association for Computational Linguistics.

\bibitem[Moslem et~al., 2023b]{Moslem2023-AdaptiveMT-LLM-Finetuning}
Yasmin Moslem, Rejwanul Haque, and Andy Way (2023b).
\newblock {Fine-tuning Large Language Models for Adaptive Machine Translation}.
\newblock \emph{arXiv preprint arXiv:2312.12740 [cs.CL]}.

\bibitem[Moslem et~al., 2023c]{Moslem2023-Terminology}
Yasmin Moslem, Gianfranco Romani, Mahdi Molaei, John~D Kelleher, Rejwanul Haque, and Andy Way (2023c).
\newblock {Domain Terminology Integration into Machine Translation: Leveraging Large Language Models}.
\newblock In {\em {Proceedings of the Eighth Conference on Machine Translation}}, pages 902--911, Singapore. Association for Computational Linguistics.

\bibitem[Mu et~al., 2023]{Mu2023-MT-LLM-TM}
Yongyu Mu, Abudurexiti Reheman, Zhiquan Cao, Yuchun Fan, Bei Li, Yinqiao Li, Tong Xiao, Chunliang Zhang, and Jingbo Zhu (2023).
\newblock {Augmenting Large Language Model Translators via Translation Memories}.
\newblock In {\em {Findings of the Association for Computational Linguistics: ACL 2023}}, pages 10287--10299, Toronto, Canada. Association for Computational Linguistics.

\bibitem[Muennighoff et~al., 2022]{Muennighoff2022-BLOOMZ-mT0}
Niklas Muennighoff, Thomas Wang, Lintang Sutawika, Adam Roberts, Stella Biderman, Teven Le~Scao, M Saiful~Bari, Sheng Shen, Zheng-Xin Yong, Hailey Schoelkopf, Xiangru Tang, Dragomir Radev, Alham~Fikri Aji, Khalid Almubarak, Samuel Albanie, Zaid Alyafeai, Albert Webson, Edward Raff, and Colin Raffel (2022).
\newblock {Crosslingual Generalization through Multitask Finetuning}.
\newblock \emph{arXiv preprint arXiv:2211.01786 [cs.CL]}.

\bibitem[M{\"u}ller and Laurent, 2022]{Muller2022-Cedille}
Martin M{\"u}ller and Florian Laurent (2022).
\newblock {Cedille: A large autoregressive French language model}.
\newblock \emph{arXiv preprint arXiv:2202.03371 [cs.CL]}.

\bibitem[Nayak et~al., 2023]{Nayak2023-Instance}
Prashanth Nayak, Rejwanul Haque, John~D Kelleher, and Andy Way (2023).
\newblock {Instance-Based Domain Adaptation for Improving Terminology Translation}.
\newblock In {\em {Proceedings of Machine Translation Summit XIX: Research Track}}, pages 222--231, Macau SAR, China. Association for Machine Translation in the Americas.

\bibitem[Neves et~al., 2022]{Neves2022-ClinSpEn}
Mariana Neves, Antonio Jimeno~Yepes, Amy Siu, Roland Roller, Philippe Thomas, Maika Vicente~Navarro, Lana Yeganova, Dina Wiemann, Giorgio~Maria Di~Nunzio, Federica Vezzani, Christel Gerardin, Rachel Bawden, Darryl~Johan Estrada, Salvador Lima-lopez, Eulalia Farre-maduel, Martin Krallinger, Cristian Grozea, and Aurelie Neveol (2022).
\newblock {Findings of the {WMT} 2022 Biomedical Translation Shared Task: Monolingual Clinical Case Reports}.
\newblock In {\em {Proceedings of the Seventh Conference on Machine Translation (WMT)}}, pages 694--723, Abu Dhabi, United Arab Emirates (Hybrid). Association for Computational Linguistics.

\bibitem[O'Brien, 2022]{OBrien2022-PE}
Sharon O'Brien (2022).
\newblock {How to deal with errors in machine translation: Post-editing}.
\newblock In {\em {Machine translation for everyone}}, pages 105--120. Language Science Press.

\bibitem[Ojo and Ogueji, 2023]{Ojo2023-MT-LLM-African}
Jessica Ojo and Kelechi Ogueji (2023).
\newblock {How Good are Commercial Large Language Models on African Languages?}
\newblock In {\em {Proceedings of the 4th Workshop on African Natural Language Processing, AfricaNLP@ICLR 2023}}, Kigali, Rwanda.

\bibitem[{\"O}ktem et~al., 2022]{Oktem2022-Ladino}
Alp {\"O}ktem, Rodolfo Zevallos, Yasmin Moslem, {\"O}zg{\"u}r~G{\"u}ne{\c s} {\"O}zt{\"u}rk, and Karen Gerson~{\c S}arhon (2022).
\newblock {Preparing an endangered language for the digital age: The Case of {J}udeo-{S}panish}.
\newblock In {\em {Proceedings of the Workshop on Resources and Technologies for Indigenous, Endangered and Lesser-resourced Languages in Eurasia within the 13th Language Resources and Evaluation Conference}}, pages 105--110, Marseille, France. European Language Resources Association.

\bibitem[{OpenAI}, 2023]{OpenAI2023-GPT-4}
{OpenAI} (2023).
\newblock {GPT-4 Technical Report}.
\newblock \emph{arXiv preprint arXiv:2303.08774 [cs.CL]}.

\bibitem[Ouyang et~al., 2022]{Ouyang2022-InstructGPT}
Long Ouyang, Jeff Wu, Xu Jiang, Diogo Almeida, Carroll~L Wainwright, Pamela Mishkin, Chong Zhang, Sandhini Agarwal, Katarina Slama, Alex Ray, John Schulman, Jacob Hilton, Fraser Kelton, Luke Miller, Maddie Simens, Amanda Askell, Peter Welinder, Paul Christiano, Jan Leike, and Ryan Lowe (2022).
\newblock {Training language models to follow instructions with human feedback}.
\newblock In {\em {Advances in Neural Information Processing Systems (NeurIPS 2022)}}, pages 27730--27744, New Orleans, Louisiana, USA. Curran Associates, Inc.

\bibitem[Papineni et~al., 2002]{Papineni2002-BLEU}
Kishore Papineni, Salim Roukos, Todd Ward, and Wei-Jing Zhu (2002).
\newblock {{B}leu: a Method for Automatic Evaluation of Machine Translation}.
\newblock In {\em {Proceedings of the 40th Annual Meeting of the Association for Computational Linguistics}}, pages 311--318, Philadelphia, Pennsylvania, USA. Association for Computational Linguistics.

\bibitem[Patel et~al., 2023]{Patel2023-Bidirectional-LM}
Ajay Patel, Bryan Li, Mohammad~Sadegh Rasooli, Noah Constant, Colin Raffel, and Chris Callison-Burch (2023).
\newblock {Bidirectional Language Models Are Also Few-shot Learners}.
\newblock In {\em {Proceedings of the Eleventh International Conference on Learning Representations}}.

\bibitem[Penedo et~al., 2023]{Penedo2023-Falcon}
Guilherme Penedo, Quentin Malartic, Daniel Hesslow, Ruxandra Cojocaru, Alessandro Cappelli, Hamza Alobeidli, Baptiste Pannier, Ebtesam Almazrouei, and Julien Launay (2023).
\newblock {The RefinedWeb Dataset for Falcon LLM: Outperforming Curated Corpora with Web Data, and Web Data Only}.
\newblock \emph{arXiv preprint arXiv:2306.01116 [cs.CL]}.

\bibitem[Peng et~al., 2023]{Peng2023-MT-LLM}
Keqin Peng, Liang Ding, Qihuang Zhong, Li Shen, Xuebo Liu, Min Zhang, Yuanxin Ouyang, and Dacheng Tao (2023).
\newblock {Towards Making the Most of {C}hat{GPT} for Machine Translation}.
\newblock In {\em {Findings of the Association for Computational Linguistics: EMNLP 2023}}, pages 5622--5633, Singapore. Association for Computational Linguistics.

\bibitem[Peris and Casacuberta, 2019]{Peris2019-OnlineLearning}
{\'A}lvaro Peris and Francisco Casacuberta (2019).
\newblock {Online learning for effort reduction in interactive neural machine translation}.
\newblock Computer Speech \& Language.
\newblock 58:98--126, {\em Computer Speech \& Language}, 58:98--126.

\bibitem[Peris et~al., 2017]{Peris2017-INMT}
{\'A}lvaro Peris, Miguel Domingo, and Francisco Casacuberta (2017).
\newblock {Interactive neural machine translation}.
\newblock Computer Speech \& Language.
\newblock 45:201--220, {\em Computer Speech \& Language}, 45:201--220.

\bibitem[Petroni et~al., 2019]{Petroni2019-kb}
Fabio Petroni, Tim Rockt{\"a}schel, Sebastian Riedel, Patrick Lewis, Anton Bakhtin, Yuxiang Wu, and Alexander Miller (2019).
\newblock {Language Models as Knowledge Bases?}
\newblock In {\em {Proceedings of the 2019 Conference on Empirical Methods in Natural Language Processing and the 9th International Joint Conference on Natural Language Processing (EMNLP-IJCNLP)}}, pages 2463--2473, Hong Kong, China. Association for Computational Linguistics.

\bibitem[Pham et~al., 2020]{Pham2020-Priming}
Minh~Quang Pham, Jitao Xu, Josep Crego, Fran{\c c}ois Yvon, and Jean Senellart (2020).
\newblock {Priming Neural Machine Translation}.
\newblock In {\em {Proceedings of the Fifth Conference on Machine Translation}}, pages 516--527, Online. Association for Computational Linguistics.

\bibitem[Poncelas et~al., 2019]{Poncelas2019-BT}
Alberto Poncelas, Gideon Maillette de~Buy Wenniger, and Andy Way (2019).
\newblock {Adaptation of Machine Translation Models with Back-Translated Data Using Transductive Data Selection Methods}.
\newblock In {\em {Proceedings of the 20th International Conference on Computational Linguistics and Intelligent Text Processing CICLing 2019: Computational Linguistics and Intelligent Text Processing}}, pages 567--579, La Rochelle, France. Springer Nature Switzerland.

\bibitem[Popovi{\'c}, 2017]{Popovic2017-chrF++}
Maja Popovi{\'c} (2017).
\newblock {chr{F}++: words helping character n-grams}.
\newblock In {\em {Proceedings of the Second Conference on Machine Translation}}, pages 612--618, Copenhagen, Denmark. Association for Computational Linguistics.

\bibitem[Popovi{\'c} et~al., 2023]{Popovic2023-Reproducibility}
Maja Popovi{\'c}, Mohammad Arvan, Natalie Parde, and Anya Belz (2023).
\newblock {Exploring Variation of Results from Different Experimental Conditions}.
\newblock In {\em {Findings of the Association for Computational Linguistics: ACL 2023}}, pages 2746--2757, Toronto, Canada. Association for Computational Linguistics.

\bibitem[Post, 2018]{Post2018-sacreBLEU}
Matt Post (2018).
\newblock {A Call for Clarity in Reporting {BLEU} Scores}.
\newblock In {\em {Proceedings of the Third Conference on Machine Translation: Research Papers}}, pages 186--191, Brussels, Belgium. Association for Computational Linguistics.

\bibitem[Post and Vilar, 2018]{Post2018-FastConstrainedDecoding}
Matt Post and David Vilar (2018).
\newblock {Fast Lexically Constrained Decoding with Dynamic Beam Allocation for Neural Machine Translation}.
\newblock In {\em {Proceedings of the 2018 Conference of the North {A}merican Chapter of the Association for Computational Linguistics: Human Language Technologies, Volume 1 (Long Papers)}}, pages 1314--1324, New Orleans, Louisiana. Association for Computational Linguistics.

\bibitem[Puduppully et~al., 2023]{Puduppully2023-DecoMT-LLM}
Ratish Puduppully, Anoop Kunchukuttan, Raj Dabre, Ai~Ti Aw, and Nancy Chen (2023).
\newblock {{D}eco{MT}: Decomposed Prompting for Machine Translation Between Related Languages using Large Language Models}.
\newblock In {\em {Proceedings of the 2023 Conference on Empirical Methods in Natural Language Processing}}, pages 4586--4602, Singapore. Association for Computational Linguistics.

\bibitem[Radford et~al., 2019]{Radford2019-GPT-2}
Alec Radford, Jeff Wu, Rewon Child, D Luan, Dario Amodei, and Ilya Sutskever (2019).
\newblock {Language Models are Unsupervised Multitask Learners}.
\newblock Technical report, OpenAi.

\bibitem[Rae et~al., 2021]{Rae2021-Gopher}
Jack~W Rae, Sebastian Borgeaud, Trevor Cai, Katie Millican, Jordan Hoffmann, Francis Song, John Aslanides, Sarah Henderson, Roman Ring, Susannah Young, Eliza Rutherford, Tom Hennigan, Jacob Menick, Albin Cassirer, Richard Powell, George van~den Driessche, Lisa~Anne Hendricks, Maribeth Rauh, Po-Sen Huang, Amelia Glaese, Johannes Welbl, Sumanth Dathathri, Saffron Huang, Jonathan Uesato, John Mellor, Irina Higgins, Antonia Creswell, Nat McAleese, Amy Wu, Erich Elsen, Siddhant Jayakumar, Elena Buchatskaya, David Budden, Esme Sutherland, Karen Simonyan, Michela Paganini, Laurent Sifre, Lena Martens, Xiang~Lorraine Li, Adhiguna Kuncoro, Aida Nematzadeh, Elena Gribovskaya, Domenic Donato, Angeliki Lazaridou, Arthur Mensch, Jean-Baptiste Lespiau, Maria Tsimpoukelli, Nikolai Grigorev, Doug Fritz, Thibault Sottiaux, Mantas Pajarskas, Toby Pohlen, Zhitao Gong, Daniel Toyama, Cyprien de~Masson~d'Autume, Yujia Li, Tayfun Terzi, Vladimir Mikulik, Igor Babuschkin, Aidan Clark, Diego de~Las~Casas, Aurelia Guy, Chris Jones,
  James Bradbury, Matthew Johnson, Blake Hechtman, Laura Weidinger, Iason Gabriel, William Isaac, Ed Lockhart, Simon Osindero, Laura Rimell, Chris Dyer, Oriol Vinyals, Kareem Ayoub, Jeff Stanway, Lorrayne Bennett, Demis Hassabis, Koray Kavukcuoglu, and Geoffrey Irving (2021).
\newblock {Scaling Language Models: Methods, Analysis \& Insights from Training Gopher}.
\newblock \emph{arXiv preprint arXiv:2112.11446 [cs.CL]}.

\bibitem[Raffel et~al., 2020]{Raffel2020-T5}
Colin Raffel, Noam Shazeer, Adam Roberts, Katherine Lee, Sharan Narang, Michael Matena, Yanqi Zhou, Wei Li, and Peter~J Liu (2020).
\newblock {Exploring the Limits of Transfer Learning with a Unified Text-to-Text Transformer}.
\newblock Journal of Machine Learning Research.
\newblock 21(140):1--67, {\em Journal of Machine Learning Research}, 21(140):1--67.

\bibitem[Reheman et~al., 2023]{Reheman2023-PromptingMT}
Abudurexiti Reheman, Tao Zhou, Yingfeng Luo, Di Yang, Tong Xiao, and Jingbo Zhu (2023).
\newblock {Prompting Neural Machine Translation with Translation Memories}.
\newblock AAAI.
\newblock 37(11):13519--13527, {\em AAAI}, 37(11):13519--13527.

\bibitem[Rei et~al., 2020]{Rei2020-COMET}
Ricardo Rei, Craig Stewart, Ana~C Farinha, and Alon Lavie (2020).
\newblock {{COMET}: A Neural Framework for {MT} Evaluation}.
\newblock In {\em {Proceedings of the 2020 Conference on Empirical Methods in Natural Language Processing (EMNLP)}}, pages 2685--2702, Online. Association for Computational Linguistics.

\bibitem[Reimers and Gurevych, 2019]{Reimers2019-SentenceTransformers}
Nils Reimers and Iryna Gurevych (2019).
\newblock {Sentence-{BERT}: Sentence Embeddings using {S}iamese {BERT}-Networks}.
\newblock In {\em {Proceedings of the 2019 Conference on Empirical Methods in Natural Language Processing and the 9th International Joint Conference on Natural Language Processing (EMNLP-IJCNLP)}}, pages 3982--3992, Hong Kong, China. Association for Computational Linguistics.

\bibitem[Reinauer et~al., 2023]{Reinauer2023-MT-ICL}
Raphael Reinauer, Patrick Simianer, Kaden Uhlig, Johannes E~M Mosig, and Joern Wuebker (2023).
\newblock {Neural Machine Translation Models Can Learn to be Few-shot Learners}.
\newblock \emph{arXiv preprint arXiv:2309.08590 [cs.CL]}.

\bibitem[Richardson and Rashid, 2007]{Richardson2007-MT}
Stephen~D Richardson and Richard~F Rashid (2007).
\newblock {Adaptive Machine Translation}.
\newblock US Patent No. US7295963B2 (Microsoft). United States Patent and Trademark Office.

\bibitem[Robinson et~al., 2023]{Robinson2023-ChatGPT-MT}
Nathaniel Robinson, Perez Ogayo, David~R Mortensen, and Graham Neubig (2023).
\newblock {{C}hat{GPT} {MT}: Competitive for High- (but Not Low-) Resource Languages}.
\newblock In {\em {Proceedings of the Eighth Conference on Machine Translation}}, pages 392--418, Singapore. Association for Computational Linguistics.

\bibitem[Sarti et~al., 2023]{Sarti2023-LLM-MT}
Gabriele Sarti, Phu~Mon Htut, Xing Niu, Benjamin Hsu, Anna Currey, Georgiana Dinu, and Maria Nadejde (2023).
\newblock {{RAMP}: Retrieval and Attribute-Marking Enhanced Prompting for Attribute-Controlled Translation}.
\newblock In {\em {Proceedings of the 61st Annual Meeting of the Association for Computational Linguistics (Volume 2: Short Papers)}}, pages 1476--1490, Toronto, Canada. Association for Computational Linguistics.

\bibitem[Saunders, 2022]{Saunders2022-Survey}
Danielle Saunders (2022).
\newblock {Domain Adaptation and Multi-Domain Adaptation for Neural Machine Translation: A Survey}.
\newblock JAIR.
\newblock 75:351--424, {\em JAIR}, 75:351--424.

\bibitem[Sawai et~al., 2021]{Sawai2021-yw}
Ranto Sawai, Incheon Paik, and Ayato Kuwana (2021).
\newblock {Sentence Augmentation for Language Translation Using GPT-2}.
\newblock Electronics.
\newblock 10(24):3082, {\em Electronics}, 10(24):3082.

\bibitem[Schick et~al., 2023]{Schick2023-Toolformer}
Timo Schick, Jane Dwivedi-Yu, Roberto Dess{\`\i}, Roberta Raileanu, Maria Lomeli, Luke Zettlemoyer, Nicola Cancedda, and Thomas Scialom (2023).
\newblock {Toolformer: Language Models Can Teach Themselves to Use Tools}.
\newblock \emph{arXiv preprint arXiv:2302.04761 [cs.CL]}.

\bibitem[Schioppa et~al., 2023]{Schioppa2023-LLM-MT-CrossLingual}
Andrea Schioppa, Xavier Garcia, and Orhan Firat (2023).
\newblock {Cross-Lingual Supervision improves Large Language Models Pre-training}.
\newblock \emph{arXiv preprint arXiv:2305.11778 [cs.CL]}.

\bibitem[Semenov et~al., 2023]{Semenov2023-WMT23-Terminology}
Kirill Semenov, Vil{\'e}m Zouhar, Tom Kocmi, Dongdong Zhang, Wangchunshu Zhou, A Yuchen, Eleanor Jiang, Charles University, Eth Z{\"u}rich, M Microsoft, A Aiwaves, and Chinese English (2023).
\newblock {Findings of the WMT 2023 shared task on machine translation with terminologies}.
\newblock In {\em {Proceedings of the Seventh Conference on Machine Translation (WMT)}}, Sentosa, Singapore.

\bibitem[Sengupta et~al., 2023]{Sengupta2023-Jais}
Neha Sengupta, Sunil~Kumar Sahu, Bokang Jia, Satheesh Katipomu, Haonan Li, Fajri Koto, Osama~Mohammed Afzal, Samta Kamboj, Onkar Pandit, Rahul Pal, Lalit Pradhan, Zain~Muhammad Mujahid, Massa Baali, Alham~Fikri Aji, Zhengzhong Liu, Andy Hock, Andrew Feldman, Jonathan Lee, Andrew Jackson, Preslav Nakov, Timothy Baldwin, and Eric Xing (2023).
\newblock {Jais and Jais-chat: Arabic-Centric Foundation and Instruction-Tuned Open Generative Large Language Models}.
\newblock \emph{arXiv preprint arXiv:2308.16149 [cs.CL]}.

\bibitem[Sennrich et~al., 2016]{Sennrich2016-BT}
Rico Sennrich, Barry Haddow, and Alexandra Birch (2016).
\newblock {Improving Neural Machine Translation Models with Monolingual Data}.
\newblock In {\em {Proceedings of the 54th Annual Meeting of the Association for Computational Linguistics (Volume 1: Long Papers)}}, pages 86--96, Berlin, Germany. Association for Computational Linguistics.

\bibitem[Shi et~al., 2022]{Shi2022-kNN-Prompt}
Weijia Shi, Julian Michael, Suchin Gururangan, and Luke Zettlemoyer (2022).
\newblock {kNN-Prompt: Nearest Neighbor Zero-Shot Inference}.
\newblock \emph{arXiv preprint arXiv:2205.13792 [cs.CL]}.

\bibitem[Shliazhko et~al., 2022]{Shliazhko2022-mGPT}
Oleh Shliazhko, Alena Fenogenova, Maria Tikhonova, Vladislav Mikhailov, Anastasia Kozlova, and Tatiana Shavrina (2022).
\newblock {mGPT: Few-Shot Learners Go Multilingual}.
\newblock Transactions of the Association for Computational Linguistics (TACL).
\newblock {\em Transactions of the Association for Computational Linguistics (TACL)}.
\newblock \emph{arXiv preprint arXiv:2204.07580 [cs.CL]}.

\bibitem[Sia and Duh, 2022]{Sia2022-LLM-MT}
Suzanna Sia and Kevin Duh (2022).
\newblock {Prefix Embeddings for In-context Machine Translation}.
\newblock In {\em {Proceedings of the 15th biennial conference of the Association for Machine Translation in the Americas (Volume 1: Research Track)}}, pages 45--57, Orlando, USA. Association for Machine Translation in the Americas.

\bibitem[Sia and Duh, 2023]{Sia2023-LLM-MT}
Suzanna Sia and Kevin Duh (2023).
\newblock {In-context Learning as Maintaining Coherency: A Study of On-the-fly Machine Translation Using Large Language Models}.
\newblock In {\em {Proceedings of Machine Translation Summit XIX, Vol. 1: Research Track}}, pages 173--185, Macau SAR, China. Asia-Pacific Association for Machine Translation.

\bibitem[Smith et~al., 2022]{Smith2022-MT-NLG}
Shaden Smith, Mostofa Patwary, Brandon Norick, Patrick LeGresley, Samyam Rajbhandari, Jared Casper, Zhun Liu, Shrimai Prabhumoye, George Zerveas, Vijay Korthikanti, Elton Zhang, Rewon Child, Reza~Yazdani Aminabadi, Julie Bernauer, Xia Song, Mohammad Shoeybi, Yuxiong He, Michael Houston, Saurabh Tiwary, and Bryan Catanzaro (2022).
\newblock {Using DeepSpeed and Megatron to Train Megatron-Turing NLG 530B, A Large-Scale Generative Language Model}.
\newblock \emph{arXiv preprint arXiv:2201.11990 [cs.CL]}.

\bibitem[Snover et~al., 2006]{Snover2006-TER}
Matthew Snover, Bonnie Dorr, Rich Schwartz, Linnea Micciulla, and John Makhoul (2006).
\newblock {A Study of Translation Edit Rate with Targeted Human Annotation}.
\newblock In {\em {Proceedings of the 7th Conference of the Association for Machine Translation in the Americas: Technical Papers}}, pages 223--231, Cambridge, Massachusetts, USA. Association for Machine Translation in the Americas.

\bibitem[Soares et~al., 2018]{Soares2018-SciELO}
Felipe Soares, Viviane Moreira, and Karin Becker (2018).
\newblock {A Large Parallel Corpus of Full-Text Scientific Articles}.
\newblock In {\em {Proceedings of the Eleventh International Conference on Language Resources and Evaluation ({LREC} 2018)}}, Miyazaki, Japan. European Language Resources Association (ELRA).

\bibitem[Specia et~al., 2020]{Specia2020-WMT-Robustness}
Lucia Specia, Zhenhao Li, Juan Pino, Vishrav Chaudhary, Francisco Guzm{\'a}n, Graham Neubig, Nadir Durrani, Yonatan Belinkov, Philipp Koehn, Hassan Sajjad, Paul Michel, and Xian Li (2020).
\newblock {Findings of the {WMT} 2020 Shared Task on Machine Translation Robustness}.
\newblock In {\em {Proceedings of the Fifth Conference on Machine Translation}}, pages 76--91, Online. Association for Computational Linguistics.

\bibitem[Stergiadis et~al., 2021]{Stergiadis2021-domain-tag}
Emmanouil Stergiadis, Satendra Kumar, Fedor Kovalev, and Pavel Levin (2021).
\newblock {Multi-Domain Adaptation in Neural Machine Translation Through Multidimensional Tagging}.
\newblock In {\em {Proceedings of Machine Translation Summit XVIII: Users and Providers Track}}, pages 396--420, Virtual. Association for Machine Translation in the Americas.

\bibitem[Sun et~al., 2022]{Sun2022-Terminology-Prompt}
Zewei Sun, Qingnan Jiang, Shujian Huang, Jun Cao, Shanbo Cheng, and Mingxuan Wang (2022).
\newblock {Zero-shot Domain Adaptation for Neural Machine Translation with Retrieved Phrase-level Prompts}.
\newblock \emph{arXiv preprint arXiv:2209.11409 [cs.CL]}.

\bibitem[Sung et~al., 2021]{Sung2021-med-kb}
Mujeen Sung, Jinhyuk Lee, Sean Yi, Minji Jeon, Sungdong Kim, and Jaewoo Kang (2021).
\newblock {Can Language Models be Biomedical Knowledge Bases?}
\newblock In {\em {Proceedings of the 2021 Conference on Empirical Methods in Natural Language Processing}}, pages 4723--4734, Online and Punta Cana, Dominican Republic. Association for Computational Linguistics.

\bibitem[Tan et~al., 2023]{Tan2023-MT-LLM}
Weiting Tan, Haoran Xu, Lingfeng Shen, Shuyue~Stella Li, Kenton Murray, Philipp Koehn, Benjamin Van~Durme, and Yunmo Chen (2023).
\newblock {Narrowing the Gap between Zero- and Few-shot Machine Translation by Matching Styles}.
\newblock \emph{arXiv preprint arXiv:2311.02310 [cs.CL]}.

\bibitem[Tiedemann, 2012]{Tiedemann2012-OPUS}
J{\"o}rg Tiedemann (2012).
\newblock {Parallel Data, Tools and Interfaces in {OPUS}}.
\newblock In {\em {Proceedings of the Eighth International Conference on Language Resources and Evaluation ({LREC}'12)}}, pages 2214--2218, Istanbul, Turkey. European Language Resources Association (ELRA).

\bibitem[Tiedemann, 2020]{Tiedemann2020-Tatoeba}
J{\"o}rg Tiedemann (2020).
\newblock {The Tatoeba Translation Challenge {--} Realistic Data Sets for Low Resource and Multilingual {MT}}.
\newblock In {\em {Proceedings of the Fifth Conference on Machine Translation}}, pages 1174--1182, Online. Association for Computational Linguistics.

\bibitem[Tiedemann and Thottingal, 2020]{Tiedemann2020-OPUS-MT}
J{\"o}rg Tiedemann and Santhosh Thottingal (2020).
\newblock {{OPUS-MT} --- {B}uilding open translation services for the {W}orld}.
\newblock In {\em {Proceedings of the 22nd Annual Conferenec of the European Association for Machine Translation (EAMT)}}, Lisbon, Portugal.

\bibitem[Touvron et~al., 2023a]{Touvron2023-Llama1}
Hugo Touvron, Thibaut Lavril, Gautier Izacard, Xavier Martinet, Marie-Anne Lachaux, Timoth{\'e}e Lacroix, Baptiste Rozi{\`e}re, Naman Goyal, Eric Hambro, Faisal Azhar, Aurelien Rodriguez, Armand Joulin, Edouard Grave, and Guillaume Lample (2023a).
\newblock {LLaMA: Open and Efficient Foundation Language Models}.
\newblock \emph{arXiv preprint arXiv:2302.13971 [cs.CL]}.

\bibitem[Touvron et~al., 2023b]{Touvron2023-Llama2}
Hugo Touvron, Louis Martin, Kevin Stone, Peter Albert, Amjad Almahairi, Yasmine Babaei, Nikolay Bashlykov, Soumya Batra, Prajjwal Bhargava, Shruti Bhosale, Dan Bikel, Lukas Blecher, Cristian~Canton Ferrer, Moya Chen, Guillem Cucurull, David Esiobu, Jude Fernandes, Jeremy Fu, Wenyin Fu, Brian Fuller, Cynthia Gao, Vedanuj Goswami, Naman Goyal, Anthony Hartshorn, Saghar Hosseini, Rui Hou, Hakan Inan, Marcin Kardas, Viktor Kerkez, Madian Khabsa, Isabel Kloumann, Artem Korenev, Punit~Singh Koura, Marie-Anne Lachaux, Thibaut Lavril, Jenya Lee, Diana Liskovich, Yinghai Lu, Yuning Mao, Xavier Martinet, Todor Mihaylov, Pushkar Mishra, Igor Molybog, Yixin Nie, Andrew Poulton, Jeremy Reizenstein, Rashi Rungta, Kalyan Saladi, Alan Schelten, Ruan Silva, Eric~Michael Smith, Ranjan Subramanian, Xiaoqing~Ellen Tan, Binh Tang, Ross Taylor, Adina Williams, Jian~Xiang Kuan, Puxin Xu, Zheng Yan, Iliyan Zarov, Yuchen Zhang, Angela Fan, Melanie Kambadur, Sharan Narang, Aurelien Rodriguez, Robert Stojnic, Sergey Edunov, and Thomas
  Scialom (2023b).
\newblock {Llama 2: Open Foundation and Fine-Tuned Chat Models}.
\newblock \emph{arXiv preprint arXiv:2307.09288 [cs.CL]}.

\bibitem[Tran et~al., 2021]{Tran2021-oj}
Chau Tran, Shruti Bhosale, James Cross, Philipp Koehn, Sergey Edunov, and Angela Fan (2021).
\newblock {{F}acebook {AI}{'}s {WMT}21 News Translation Task Submission}.
\newblock In {\em {Proceedings of the Sixth Conference on Machine Translation}}, pages 205--215, Online. Association for Computational Linguistics.

\bibitem[Trask et~al., 2015]{Trask2015-sense2vec}
Andrew Trask, Phil Michalak, and John Liu (2015).
\newblock {sense2vec - A Fast and Accurate Method for Word Sense Disambiguation In Neural Word Embeddings}.
\newblock \emph{arXiv preprint arXiv:1511.06388 [cs.CL]}.

\bibitem[Treviso et~al., 2023]{Treviso2023-Efficient-NLP}
Marcos Treviso, Ji-Ung Lee, Tianchu Ji, Betty van Aken, Qingqing Cao, Manuel~R Ciosici, Michael Hassid, Kenneth Heafield, Sara Hooker, Colin Raffel, Pedro~H Martins, Andr{\'e} F~T Martins, Jessica~Zosa Forde, Peter Milder, Edwin Simpson, Noam Slonim, Jesse Dodge, Emma Strubell, Niranjan Balasubramanian, Leon Derczynski, Iryna Gurevych, and Roy Schwartz (2023).
\newblock {Efficient methods for natural language processing: A survey}.
\newblock Transactions of the Association for Computational Linguistics.
\newblock 11:826--860, {\em Transactions of the Association for Computational Linguistics}, 11:826--860.

\bibitem[Vaswani et~al., 2017]{Vaswani2017-attention}
Ashish Vaswani, Noam Shazeer, Niki Parmar, Jakob Uszkoreit, Llion Jones, Aidan~N Gomez, Lukasz Kaiser, and Illia Polosukhin (2017).
\newblock {Attention Is All You Need}.
\newblock In {\em {Advances in Neural Information Processing Systems (NIPS 2017)}}, volume~30. Curran Associates, Inc.

\bibitem[Vilar et~al., 2023]{Vilar2023-PaLM-MT}
David Vilar, Markus Freitag, Colin Cherry, Jiaming Luo, Viresh Ratnakar, and George Foster (2023).
\newblock {Prompting {P}a{LM} for Translation: Assessing Strategies and Performance}.
\newblock In {\em {Proceedings of the 61st Annual Meeting of the Association for Computational Linguistics (Volume 1: Long Papers)}}, pages 15406--15427, Toronto, Canada. Association for Computational Linguistics.

\bibitem[Wan et~al., 2023]{Wan2023-EfficientLLMs}
Zhongwei Wan, Xin Wang, Che Liu, Samiul Alam, Yu Zheng, Zhongnan Qu, Shen Yan, Yi Zhu, Quanlu Zhang, Mosharaf Chowdhury, and Mi Zhang (2023).
\newblock {Efficient Large Language Models: A Survey}.
\newblock \emph{arXiv preprint arXiv:2312.03863 [cs.CL]}.

\bibitem[Wang and Komatsuzaki, 2021]{Wang2021-GPT-J}
Ben Wang and Aran Komatsuzaki (2021).
\newblock {GPT-J-6B: A 6 Billion Parameter Autoregressive Language Model}.
\newblock Github (mesh-transformer-jax).

\bibitem[Wang et~al., 2022]{Wang2022-kNN-MT-Efficiency}
Dexin Wang, Kai Fan, Boxing Chen, and Deyi Xiong (2022).
\newblock {Efficient Cluster-Based $k$-Nearest-Neighbor Machine Translation}.
\newblock In {\em {Proceedings of the 60th Annual Meeting of the Association for Computational Linguistics (Volume 1: Long Papers)}}, pages 2175--2187, Dublin, Ireland. Association for Computational Linguistics.

\bibitem[Wang et~al., 2023a]{Wang2023-DocumentMT}
Longyue Wang, Chenyang Lyu, Tianbo Ji, Zhirui Zhang, Dian Yu, Shuming Shi, and Zhaopeng Tu (2023a).
\newblock {Document-Level Machine Translation with Large Language Models}.
\newblock \emph{arXiv preprint arXiv:2304.02210 [cs.CL]}.

\bibitem[Wang et~al., 2021]{Wang2021-LM4MT}
Shuo Wang, Zhaopeng Tu, Zhixing Tan, Wenxuan Wang, Maosong Sun, and Yang Liu (2021).
\newblock {Language Models are Good Translators}.
\newblock ArXiv.
\newblock {\em ArXiv}.

\bibitem[Wang et~al., 2023b]{Wang2023-MemoryLLM}
Weizhi Wang, Li Dong, Hao Cheng, Xiaodong Liu, Xifeng Yan, Jianfeng Gao, and Furu Wei (2023b).
\newblock {Augmenting Language Models with Long-Term Memory}.
\newblock In {\em {Advances in Neural Information Processing Systems (NeurIPS 2023)}}, New Orleans, Louisiana, USA.

\bibitem[Wang et~al., 2020]{Wang2020-MS-MiniLM}
Wenhui Wang, Furu Wei, Li Dong, Hangbo Bao, Nan Yang, and Ming Zhou (2020).
\newblock {MINILM: deep self-attention distillation for task-agnostic compression of pre-trained transformers}.
\newblock In {\em {Proceedings of the 34th International Conference on Neural Information Processing Systems}}, number Article 485 in NIPS'20, pages 5776--5788, Red Hook, NY, USA. Curran Associates Inc.

\bibitem[Wenzek et~al., 2021]{Wenzek2021-WMT-Multilingual}
Guillaume Wenzek, Vishrav Chaudhary, Angela Fan, Sahir Gomez, Naman Goyal, Somya Jain, Douwe Kiela, Tristan Thrush, and Francisco Guzm{\'a}n (2021).
\newblock {Findings of the {WMT} 2021 Shared Task on Large-Scale Multilingual Machine Translation}.
\newblock In {\em {Proceedings of the Sixth Conference on Machine Translation}}, pages 89--99, Online. Association for Computational Linguistics.

\bibitem[Williams and Zipser, 1989]{Williams1989-TeacherForcing}
Ronald~J Williams and David Zipser (1989).
\newblock {A Learning Algorithm for Continually Running Fully Recurrent Neural Networks}.
\newblock Neural Comput.
\newblock 1(2):270--280, {\em Neural Comput.}, 1(2):270--280.

\bibitem[Wu et~al., 2022]{Wu2022-MemTRM}
Yuhuai Wu, Markus~Norman Rabe, Delesley Hutchins, and Christian Szegedy (2022).
\newblock {Memorizing Transformers}.
\newblock In {\em {Proceedings of the Tenth International Conference on Learning Representations}}, Virtual.

\bibitem[Wuebker et~al., 2018]{Wuebker2018-Personalized}
Joern Wuebker, Patrick Simianer, and John DeNero (2018).
\newblock {Compact Personalized Models for Neural Machine Translation}.
\newblock In {\em {Proceedings of the 2018 Conference on Empirical Methods in Natural Language Processing}}, pages 881--886, Brussels, Belgium. Association for Computational Linguistics.

\bibitem[Xu et~al., 2023]{Xu2023-Llama-Finetuning}
Haoran Xu, Young~Jin Kim, Amr Sharaf, and Hany~Hassan Awadalla (2023).
\newblock {A Paradigm Shift in Machine Translation: Boosting Translation Performance of Large Language Models}.
\newblock \emph{arXiv preprint arXiv:2309.11674 [cs.CL]}.

\bibitem[Xu et~al., 2020]{Xu2020-fuzzy}
Jitao Xu, Josep Crego, and Jean Senellart (2020).
\newblock {Boosting Neural Machine Translation with Similar Translations}.
\newblock In {\em {Proceedings of the 58th Annual Meeting of the Association for Computational Linguistics}}, pages 1580--1590, Online. Association for Computational Linguistics.

\bibitem[Yang et~al., 2023a]{Yang2023-Baichuan}
Aiyuan Yang, Bin Xiao, Bingning Wang, Borong Zhang, Ce Bian, Chao Yin, Chenxu Lv, Da Pan, Dian Wang, Dong Yan, Fan Yang, Fei Deng, Feng Wang, Feng Liu, Guangwei Ai, Guosheng Dong, Haizhou Zhao, Hang Xu, Haoze Sun, Hongda Zhang, Hui Liu, Jiaming Ji, Jian Xie, Juntao Dai, Kun Fang, Lei Su, Liang Song, Lifeng Liu, Liyun Ru, Luyao Ma, Mang Wang, Mickel Liu, Mingan Lin, Nuolan Nie, Peidong Guo, Ruiyang Sun, Tao Zhang, Tianpeng Li, Tianyu Li, Wei Cheng, Weipeng Chen, Xiangrong Zeng, Xiaochuan Wang, Xiaoxi Chen, Xin Men, Xin Yu, Xuehai Pan, Yanjun Shen, Yiding Wang, Yiyu Li, Youxin Jiang, Yuchen Gao, Yupeng Zhang, Zenan Zhou, and Zhiying Wu (2023a).
\newblock {Baichuan 2: Open Large-scale Language Models}.
\newblock \emph{arXiv preprint arXiv:2309.10305 [cs.CL]}.

\bibitem[Yang et~al., 2023b]{Yang2023-BigTranslate}
Wen Yang, Chong Li, Jiajun Zhang, and Chengqing Zong (2023b).
\newblock {BigTranslate: Augmenting Large Language Models with Multilingual Translation Capability over 100 Languages}.
\newblock \emph{arXiv preprint arXiv:2305.18098 [cs.CL]}.

\bibitem[Yang et~al., 2020]{Yang2020-mUSE}
Yinfei Yang, Daniel Cer, Amin Ahmad, Mandy Guo, Jax Law, Noah Constant, Gustavo Hernandez~Abrego, Steve Yuan, Chris Tar, Yun-Hsuan Sung, Brian Strope, and Ray Kurzweil (2020).
\newblock {Multilingual Universal Sentence Encoder for Semantic Retrieval}.
\newblock In {\em {Proceedings of the 58th Annual Meeting of the Association for Computational Linguistics: System Demonstrations}}, pages 87--94, Online. Association for Computational Linguistics.

\bibitem[Yang et~al., 2019]{Yang2019-aq}
Zhilin Yang, Zihang Dai, Yiming Yang, Jaime Carbonell, Ruslan Salakhutdinov, and Quoc~V Le (2019).
\newblock {XLNet: Generalized Autoregressive Pretraining for Language Understanding}.
\newblock In {\em {Advances in Neural Information Processing Systems (NeurIPS 2019)}}, volume~32, Vancouver, Canada. Curran Associates, Inc.

\bibitem[Zeng et~al., 2022]{Zeng2022-GLM}
Aohan Zeng, Xiao Liu, Zhengxiao Du, Zihan Wang, Hanyu Lai, Ming Ding, Zhuoyi Yang, Yifan Xu, Wendi Zheng, Xiao Xia, Weng~Lam Tam, Zixuan Ma, Yufei Xue, Jidong Zhai, Wenguang Chen, Peng Zhang, Yuxiao Dong, and Jie Tang (2022).
\newblock {GLM-130B: An Open Bilingual Pre-trained Model}.
\newblock \emph{arXiv preprint arXiv:2210.02414 [cs.CL]}.

\bibitem[Zhang et~al., 2023a]{Zhang2023-PromptingMT}
Biao Zhang, Barry Haddow, and Alexandra Birch (2023a).
\newblock {Prompting large language model for machine translation: A case study}.
\newblock In {\em {Proceedings of the 40 th International Conference on Machine Learning}}, Honolulu, Hawaii, USA.

\bibitem[Zhang et~al., 2018a]{Zhang2018-Context}
Jiacheng Zhang, Huanbo Luan, Maosong Sun, Feifei Zhai, Jingfang Xu, Min Zhang, and Yang Liu (2018a).
\newblock {Improving the Transformer Translation Model with Document-Level Context}.
\newblock In {\em {Proceedings of the 2018 Conference on Empirical Methods in Natural Language Processing}}, pages 533--542, Brussels, Belgium. Association for Computational Linguistics.

\bibitem[Zhang et~al., 2018b]{Zhang2018-TranslationPieces}
Jingyi Zhang, Masao Utiyama, Eiichro Sumita, Graham Neubig, and Satoshi Nakamura (2018b).
\newblock {Guiding Neural Machine Translation with Retrieved Translation Pieces}.
\newblock In {\em {Proceedings of the 2018 Conference of the North {A}merican Chapter of the Association for Computational Linguistics: Human Language Technologies, Volume 1 (Long Papers)}}, pages 1325--1335, New Orleans, Louisiana. Association for Computational Linguistics.

\bibitem[Zhang et~al., 2022]{Zhang2022-OPT}
Susan Zhang, Stephen Roller, Naman Goyal, Mikel Artetxe, Moya Chen, Shuohui Chen, Christopher Dewan, Mona Diab, Xian Li, Xi~Victoria Lin, Todor Mihaylov, Myle Ott, Sam Shleifer, Kurt Shuster, Daniel Simig, Punit~Singh Koura, Anjali Sridhar, Tianlu Wang, and Luke Zettlemoyer (2022).
\newblock {OPT: Open Pre-trained Transformer Language Models}.
\newblock \emph{arXiv preprint arXiv:2205.01068 [cs.CL]}.

\bibitem[Zhang et~al., 2023b]{Zhang2023-MT-Efficient-Finetuning}
Xuan Zhang, Navid Rajabi, Kevin Duh, and Philipp Koehn (2023b).
\newblock {Machine Translation with Large Language Models: Prompting, Few-shot Learning, and Fine-tuning with {QL}o{RA}}.
\newblock In {\em {Proceedings of the Eighth Conference on Machine Translation}}, pages 468--481, Singapore. Association for Computational Linguistics.

\bibitem[Zhang et~al., 2020]{Zhang2020-POINTER}
Yizhe Zhang, Guoyin Wang, Chunyuan Li, Zhe Gan, Chris Brockett, and Bill Dolan (2020).
\newblock {{POINTER}: Constrained Progressive Text Generation via Insertion-based Generative Pre-training}.
\newblock In {\em {Proceedings of the 2020 Conference on Empirical Methods in Natural Language Processing (EMNLP)}}, pages 8649--8670, Online. Association for Computational Linguistics.

\bibitem[Zhang et~al., 2021]{Zhang2021-CPM}
Zhengyan Zhang, Xu Han, Hao Zhou, Pei Ke, Yuxian Gu, Deming Ye, Yujia Qin, Yusheng Su, Haozhe Ji, Jian Guan, Fanchao Qi, Xiaozhi Wang, Yanan Zheng, Guoyang Zeng, Huanqi Cao, Shengqi Chen, Daixuan Li, Zhenbo Sun, Zhiyuan Liu, Minlie Huang, Wentao Han, Jie Tang, Juanzi Li, Xiaoyan Zhu, and Maosong Sun (2021).
\newblock {CPM: A large-scale generative Chinese Pre-trained language model}.
\newblock AI Open.
\newblock 2:93--99, {\em AI Open}, 2:93--99.

\bibitem[Zheng et~al., 2021]{Zheng2021-kNN-Adaptive}
Xin Zheng, Zhirui Zhang, Shujian Huang, Boxing Chen, Jun Xie, Weihua Luo, and Jiajun Chen (2021).
\newblock {Non-Parametric Unsupervised Domain Adaptation for Neural Machine Translation}.
\newblock In {\em {Findings of the Association for Computational Linguistics: EMNLP 2021}}, pages 4234--4241, Punta Cana, Dominican Republic. Association for Computational Linguistics.

\bibitem[Zhu et~al., 2020]{Zhu2020-BERT-MT}
Jinhua Zhu, Yingce Xia, Lijun Wu, Di He, Tao Qin, Wengang Zhou, Houqiang Li, and Tie-Yan Liu (2020).
\newblock {Incorporating BERT into Neural Machine Translation}.
\newblock In {\em {Proceedings of the 8th International Conference on Learning Representations, {ICLR} 2020}}, Virtual.

\bibitem[Zhu et~al., 2023]{Zhu2023-LLM-MT}
Wenhao Zhu, Hongyi Liu, Qingxiu Dong, Jingjing Xu, Shujian Huang, Lingpeng Kong, Jiajun Chen, and Lei Li (2023).
\newblock {Multilingual Machine Translation with Large Language Models: Empirical Results and Analysis}.
\newblock \emph{arXiv preprint arXiv:2304.04675 [cs.CL]}.

\end{thebibliography}

\end{document}